\documentclass{article}

\usepackage[dandb, final]{neurips_2025}
\usepackage{fancyhdr}

\usepackage{booktabs}
\usepackage{longtable}
\usepackage{hhline}
\usepackage{varwidth}
\usepackage[x11names, table]{xcolor}
\definecolor{celadon}{rgb}{0.67, 0.88, 0.69}

\usepackage[utf8]{inputenc} 
\usepackage[T1]{fontenc}    
\usepackage{hyperref}       
\usepackage{url}            
\usepackage{booktabs}       
\usepackage{amsfonts}       
\usepackage{nicefrac}       
\usepackage{booktabs}       
\usepackage{tabularx}       
\usepackage{amsmath}
\usepackage{graphicx}
\usepackage{microtype}      
\hypersetup{
    colorlinks,
    linkcolor={red!50!black},
    citecolor={blue!50!black},
    urlcolor={blue!80!black}
}
\usepackage{tcolorbox}      
\usepackage{scalerel}
\usepackage{xspace}
\usepackage{verbatim}
\usepackage{tabularx} 
\usepackage[export]{adjustbox} 
\usepackage{makecell}   
\usepackage{arydshln}
\usepackage{tabularx}
\usepackage{marvosym}
\usepackage{longtable}

\newcommand{\mptext}[2]{%
  Our annotation guide is inspired by the work of \citet{shah-etal-2023-trillion}. We modify their annotation guide further, keeping in mind the changing global macroeconomic dynamics. Also, we have two additional categories (\texttt{Forward Looking} or \texttt{Not Forward Looking} and \texttt{Certain} or \texttt{Uncertain}) of labels other than monetary policy stance.
  
  \par
  The annotation guide for monetary policy stance is summarized in Table~\ref{tb:#1_mp_stance_guide}. It is built by dividing each target sentence into #2 defined categories: %
}

\newcommand{\usubsection}[1]{%
  \refstepcounter{subsection}
  \subsection*{\centering\LARGE #1}
  \addcontentsline{toc}{subsection}{\protect\numberline{\thesubsection}#1}
}
\newcommand{\fwcertaintytext}[1]{%
  The annotation guide for whether the sentence has \texttt{Forward Looking} information or not is summarized in the Table~\ref{tb:#1_forward_looking_guide} and whether the committee is talking about the sentence with \texttt{Certainty} attest to it or not in Table~\ref{tb:#1_certainty_guide}.%
}

\newcommand{\SD}{\texttt{Stance Detection}\xspace}
\newcommand{\TC}{\texttt{Temporal Classification}\xspace}
\newcommand{\CE}{\texttt{Uncertainty Estimation}\xspace}
\newcommand{\forward}{\texttt{Forward Looking}\xspace}
\newcommand{\notforward}{\texttt{Not Forward Looking}\xspace}
\newcommand{\certain}{\texttt{Certain}\xspace}
\newcommand{\uncertain}{\texttt{Uncertain}\xspace}
\newcommand{\hawk}{\texttt{Hawkish}\xspace}
\newcommand{\dov}{\texttt{Dovish}\xspace}
\newcommand{\neut}{\texttt{Neutral}\xspace}
\newcommand{\irr}{\texttt{Irrelevant}\xspace}

\newcommand{\mptitle}[1]{%
    #1 Annotation Guide%
}

\newcommand{\fwtitle}[1]{%
    #1 Annotation Guide to identify whether a sentence has forward-looking information or not.%
}

\newcommand{\certaintytitle}[1]{%
    #1 Annotation Guide to identify whether a sentence echoes certainty or not.%
}

\definecolor{ForestGreen}{rgb}{0.13, 0.55, 0.13}
\newcommand{\gcheck}{\textcolor{ForestGreen}{\checkmark}}

\usepackage[colorinlistoftodos]{todonotes}

\newcommand{\bbx}[1]{
  \setlength{\fboxsep}{0pt}
  \setlength{\fboxrule}{0.05pt}
  \fbox{\includegraphics{#1}}%
}

\newcommand{\simplebbx}[1]{%
  \includegraphics[height=1em]{#1}%
}

\newcommand{\finbert}{\scalerel*{\includegraphics{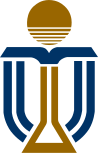}}{\textrm{C}}\xspace}
\newcommand{\MB}{\scalerel*{\includegraphics{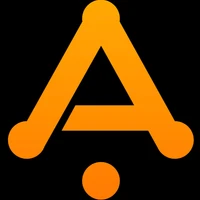}}{\textrm{C}}\xspace}
\newcommand{\FinMA}{\scalerel*{\includegraphics{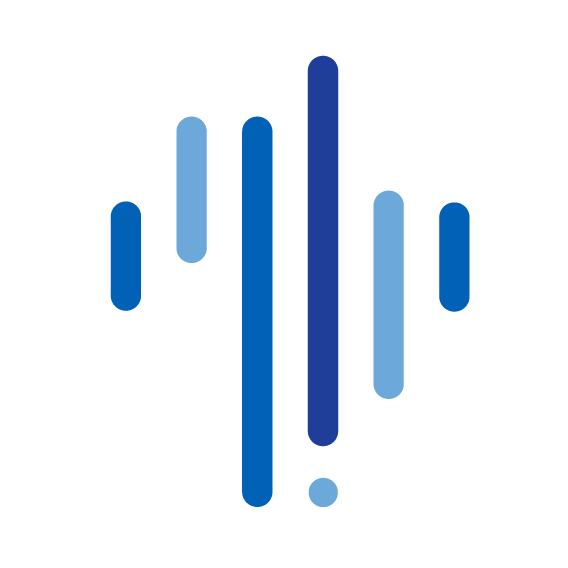}}{\textrm{C}}\xspace}
\newcommand{\openai}{\scalerel*{\includegraphics{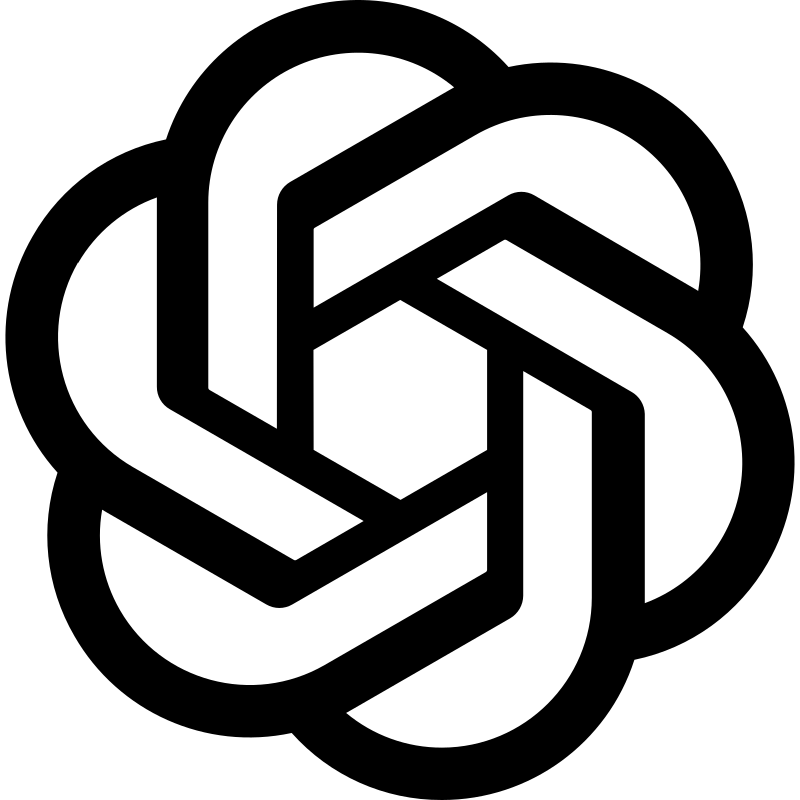}}{\textrm{C}}\xspace}
\newcommand{\google}{\scalerel*{\includegraphics{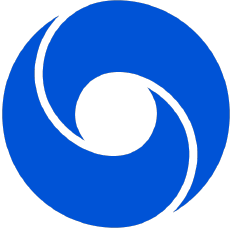}}{\textrm{C}}\xspace}
\newcommand{\meta}{\scalebox{0.7}{\scalerel*{\includegraphics{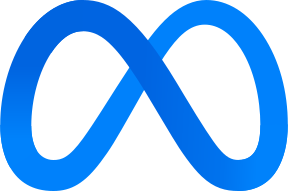}}{\textrm{C}}}\xspace}

\newcommand{\qwen}{\scalebox{1}{\scalerel*{\includegraphics{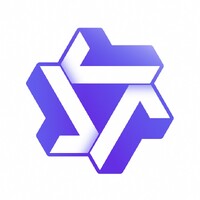}}{\textrm{C}}}\xspace}
\newcommand{\deepseek}{\scalebox{0.7}{\scalerel*{\includegraphics{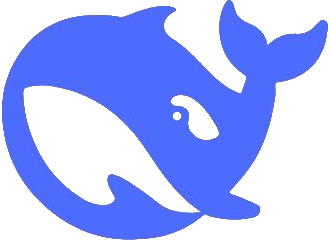}}{\textrm{C}}}\xspace}
\newcommand{\USA}{\scalerel*{\bbx{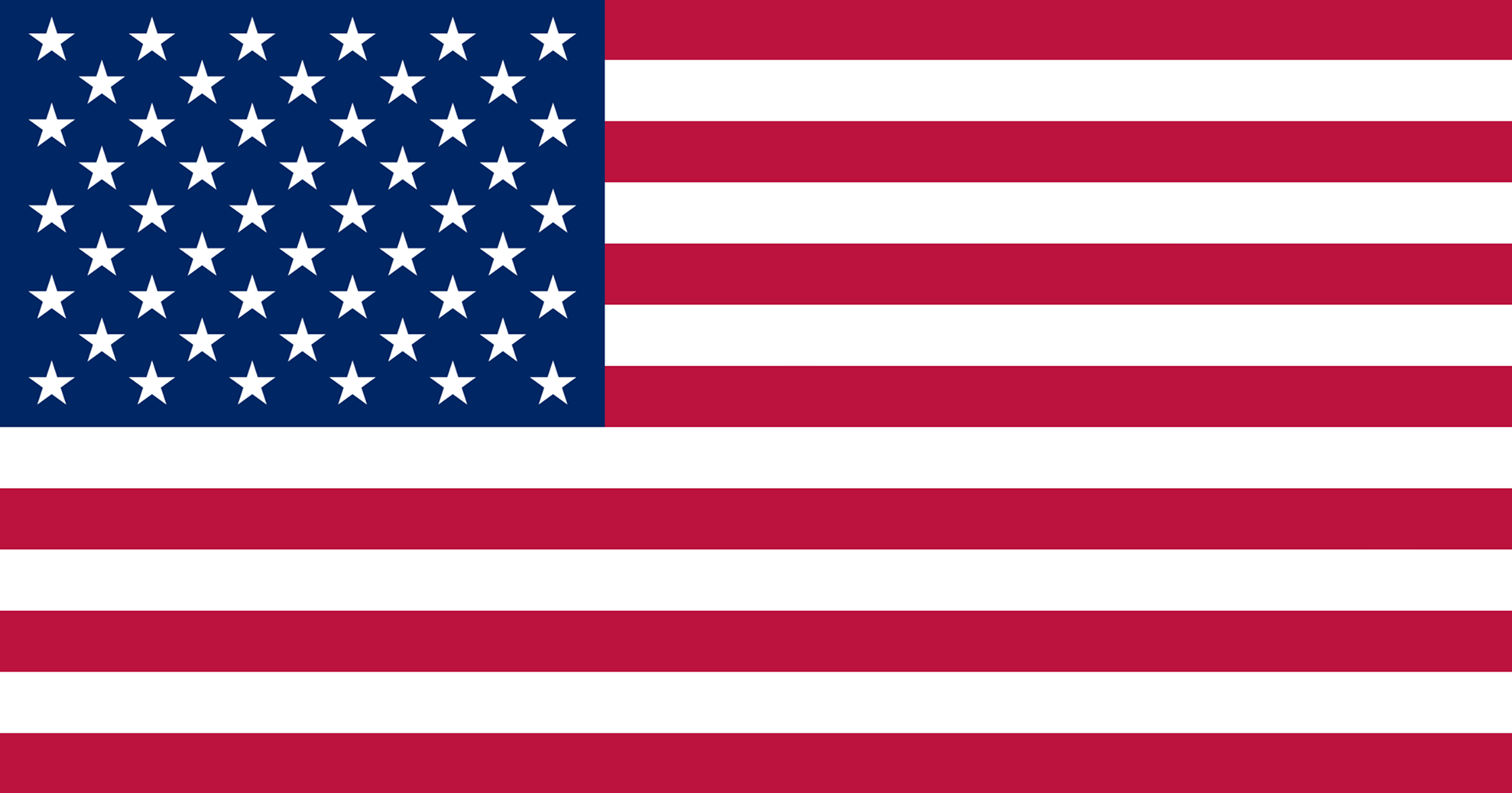}}{\textrm{C}}\xspace}
\newcommand{\AUS}{\scalerel*{\bbx{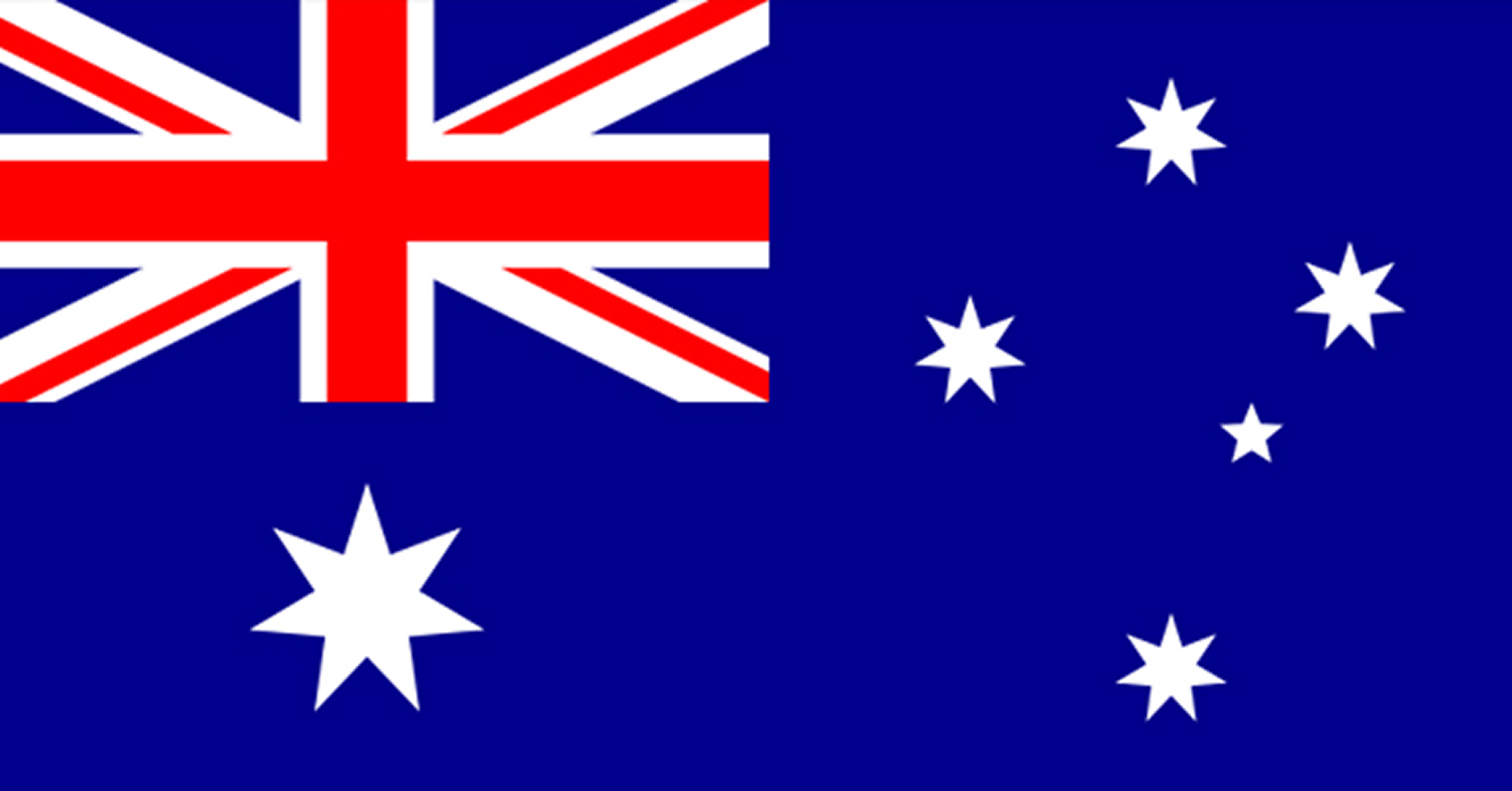}}{\textrm{C}}\xspace}
\newcommand{\CHILE}{\scalerel*{\bbx{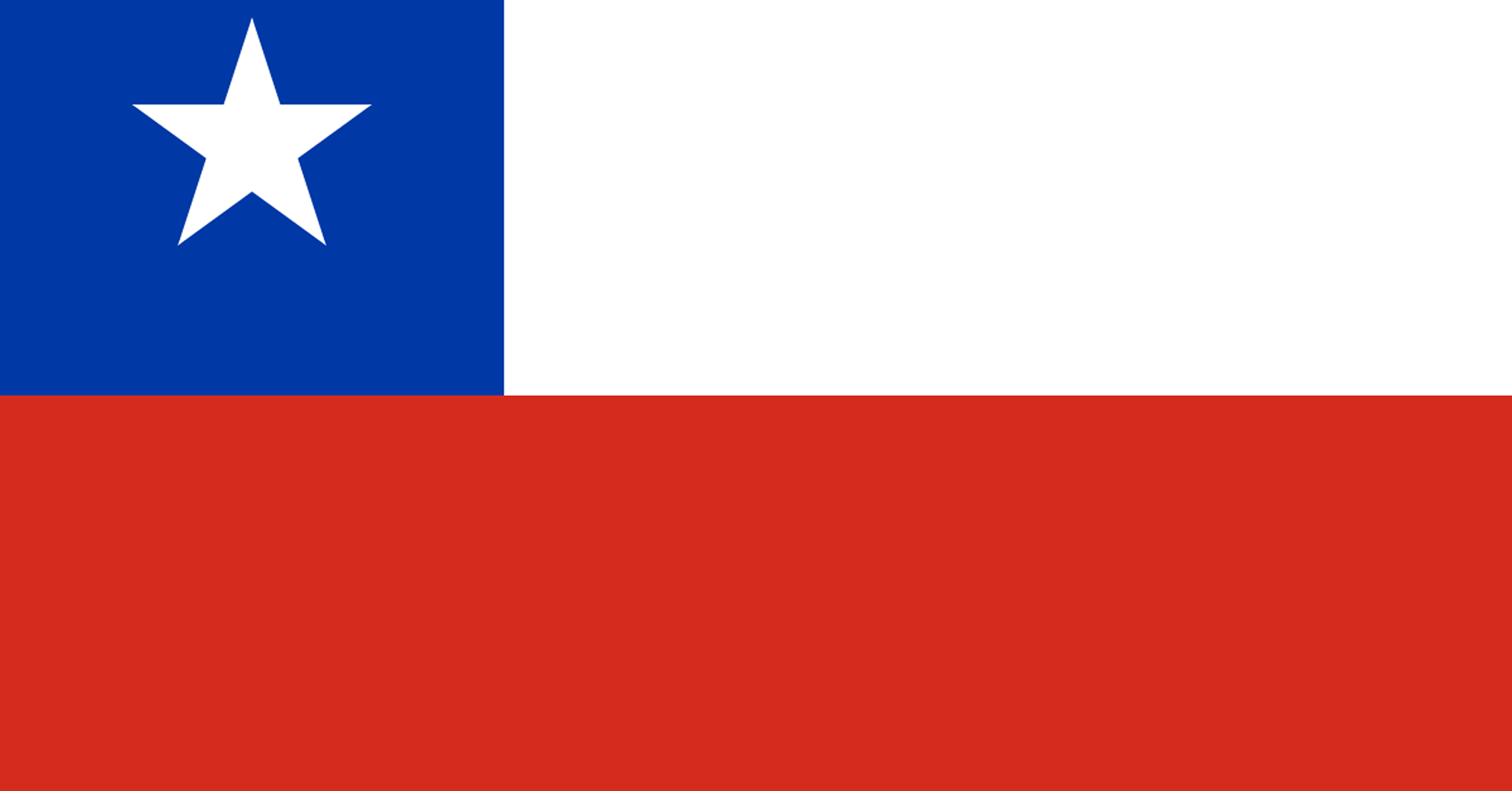}}{\textrm{C}}\xspace}
\newcommand{\BRAZIL}{\scalerel*{\bbx{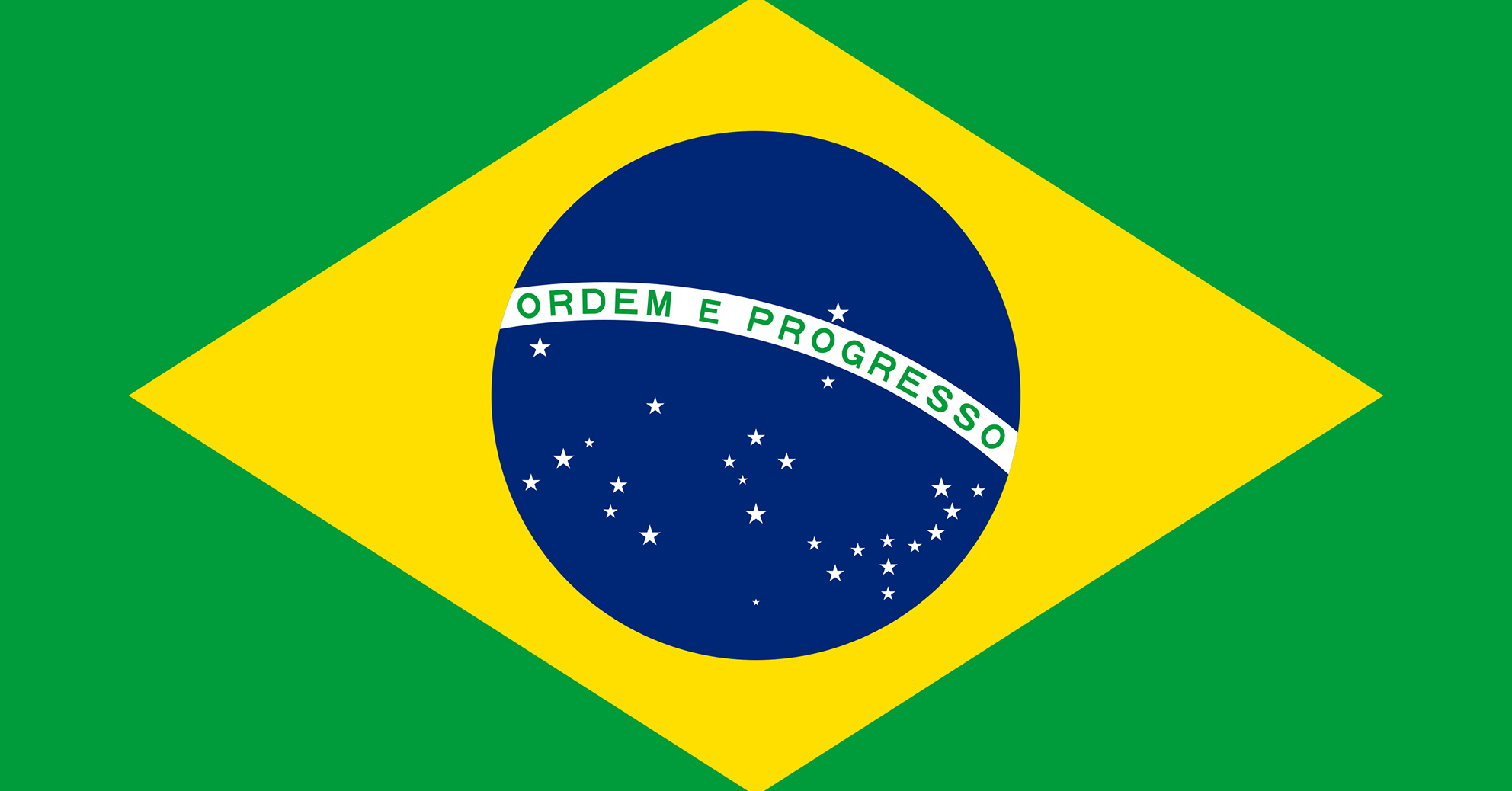}}{\textrm{C}}\xspace}
\newcommand{\CANADA}{\scalerel*{\bbx{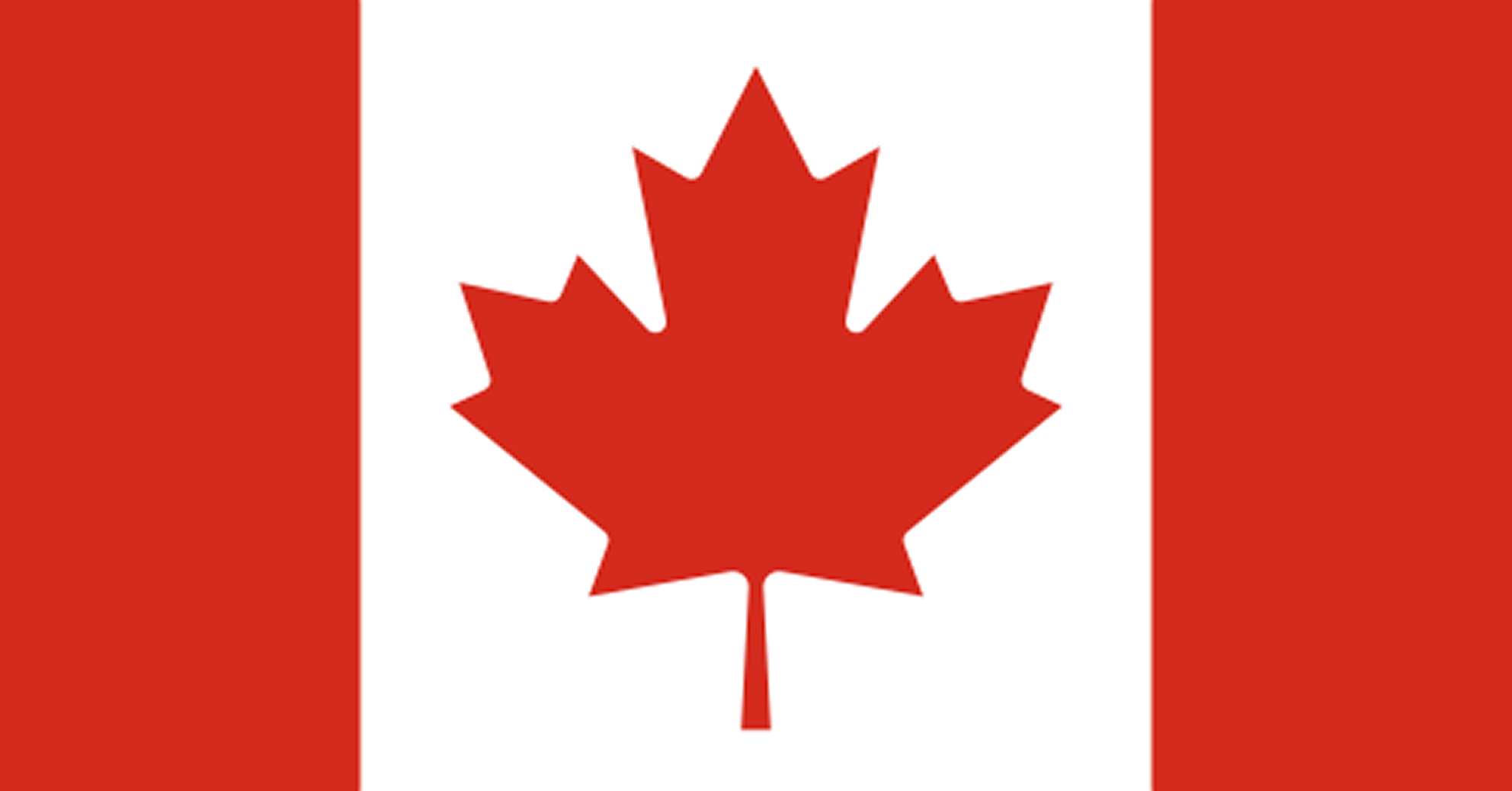}}{\textrm{C}}\xspace}
\newcommand{\CHINA}{\scalerel*{\bbx{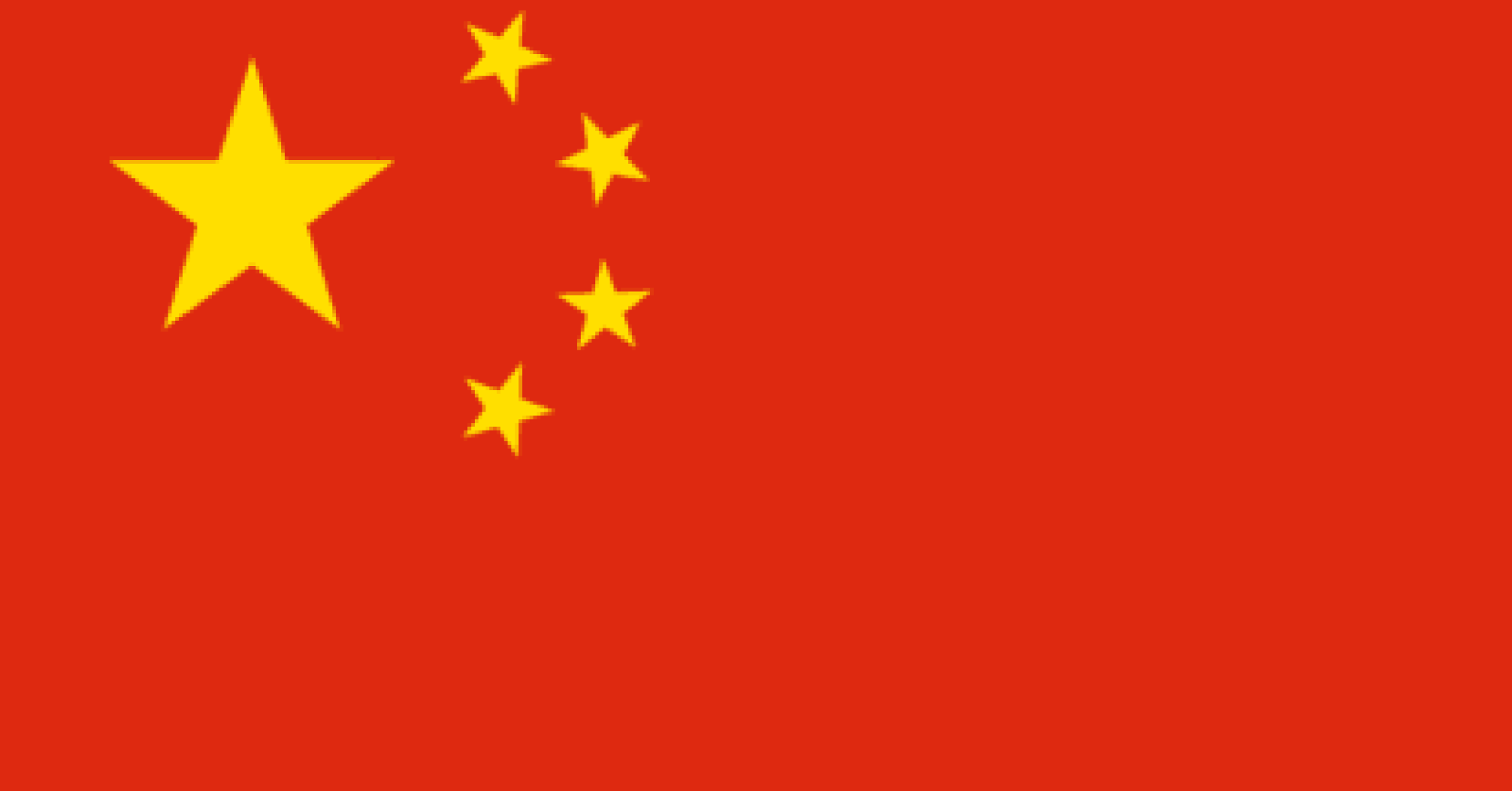}}{\textrm{C}}\xspace}
\newcommand{\COLOMBIA}{\scalerel*{\bbx{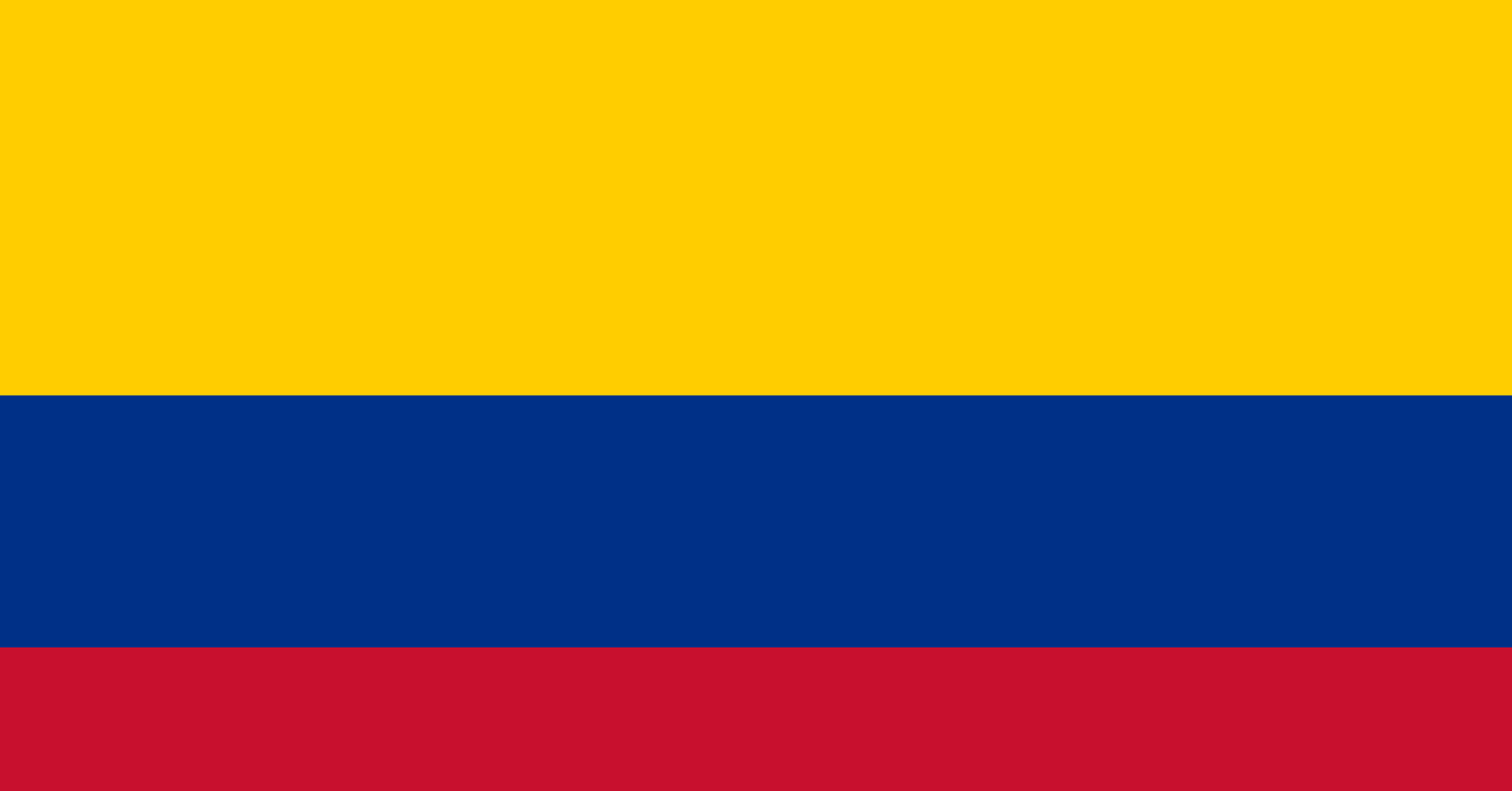}}{\textrm{C}}\xspace}
\newcommand{\EGYPT}{\scalerel*{\bbx{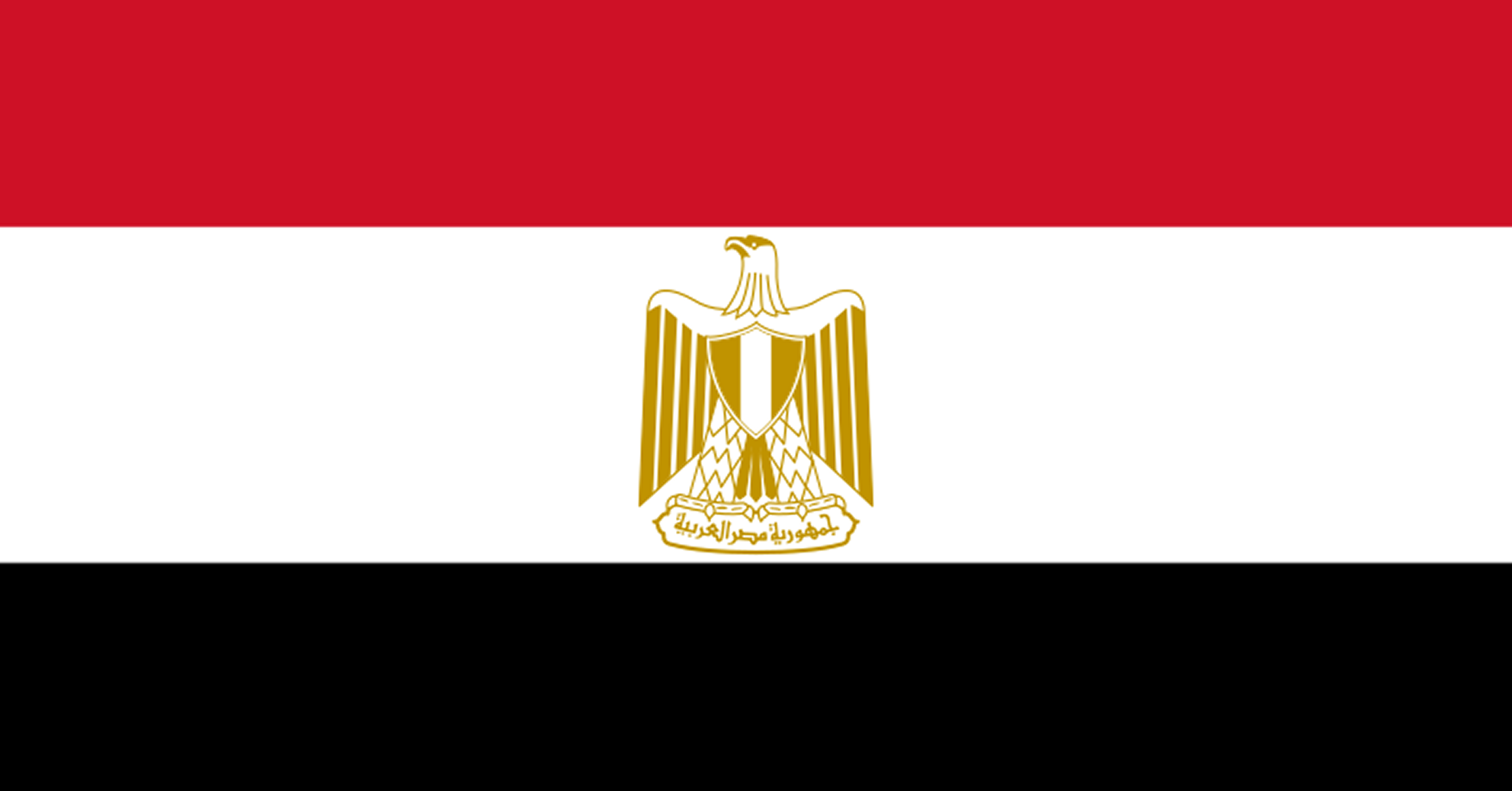}}{\textrm{C}}\xspace}
\newcommand{\EU}{\scalerel*{\bbx{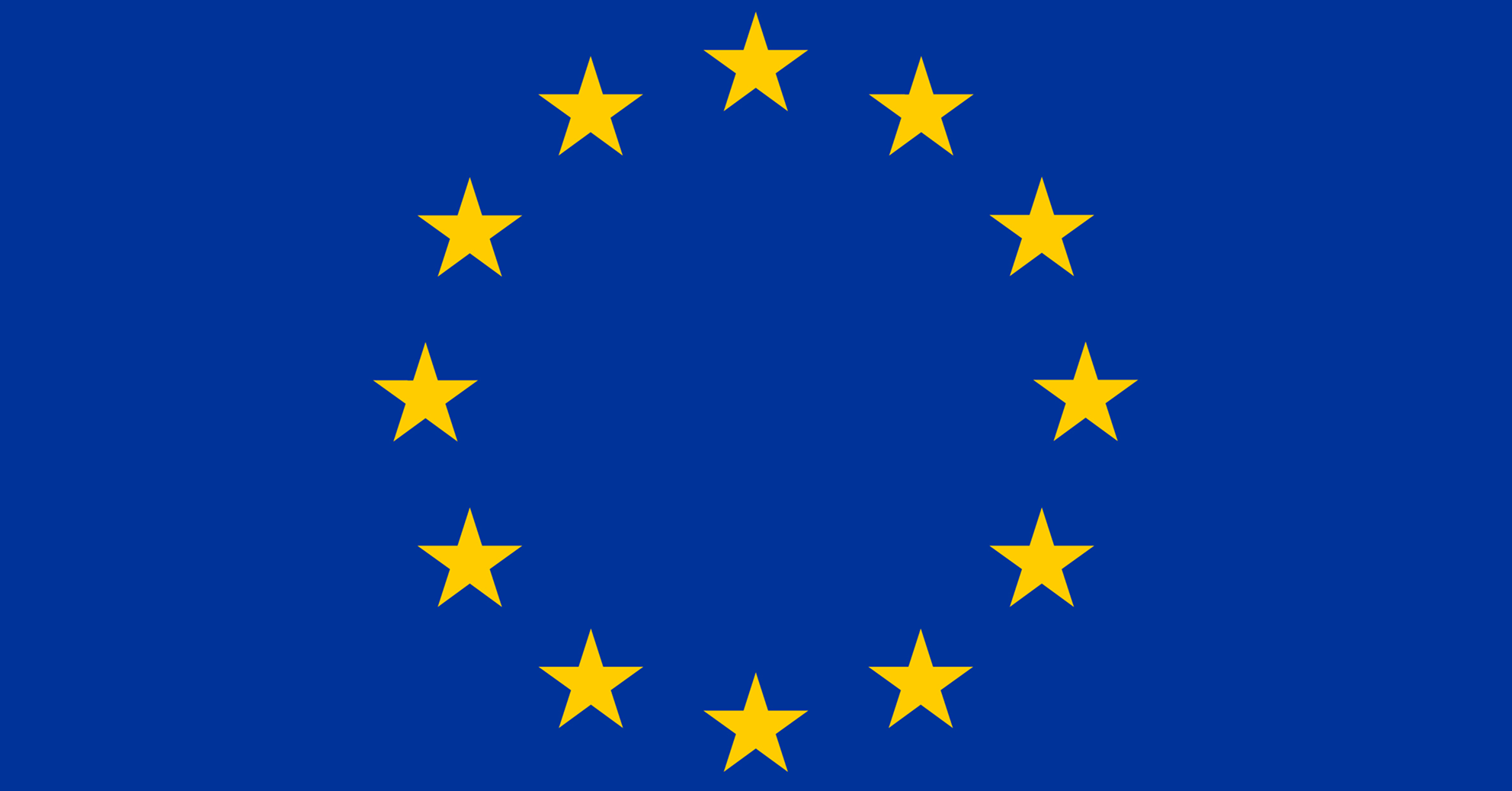}}{\textrm{C}}\xspace}
\newcommand{\INDIA}{\scalerel*{\bbx{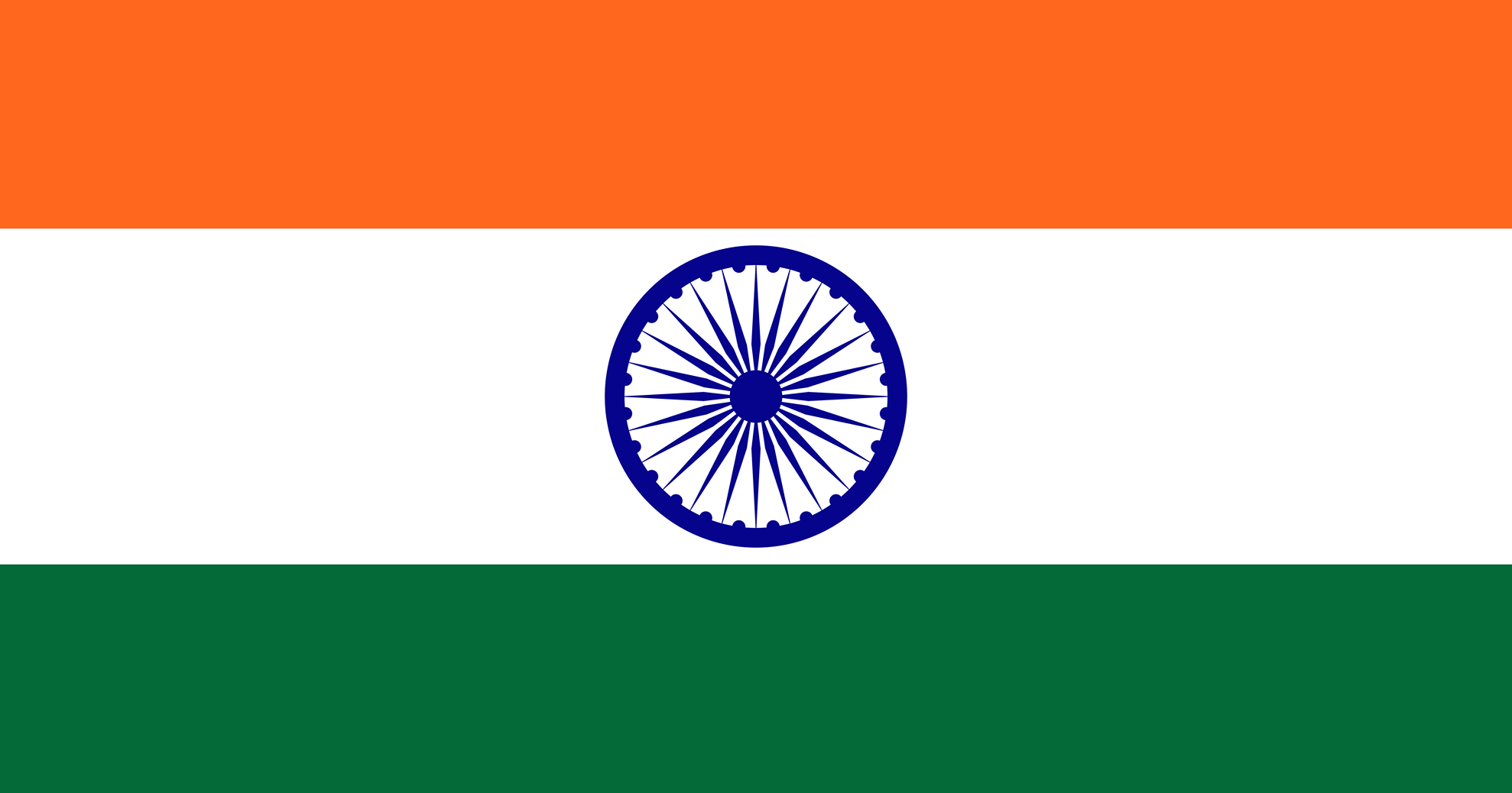}}{\textrm{C}}\xspace}
\newcommand{\MALAYSIA}{\scalerel*{\bbx{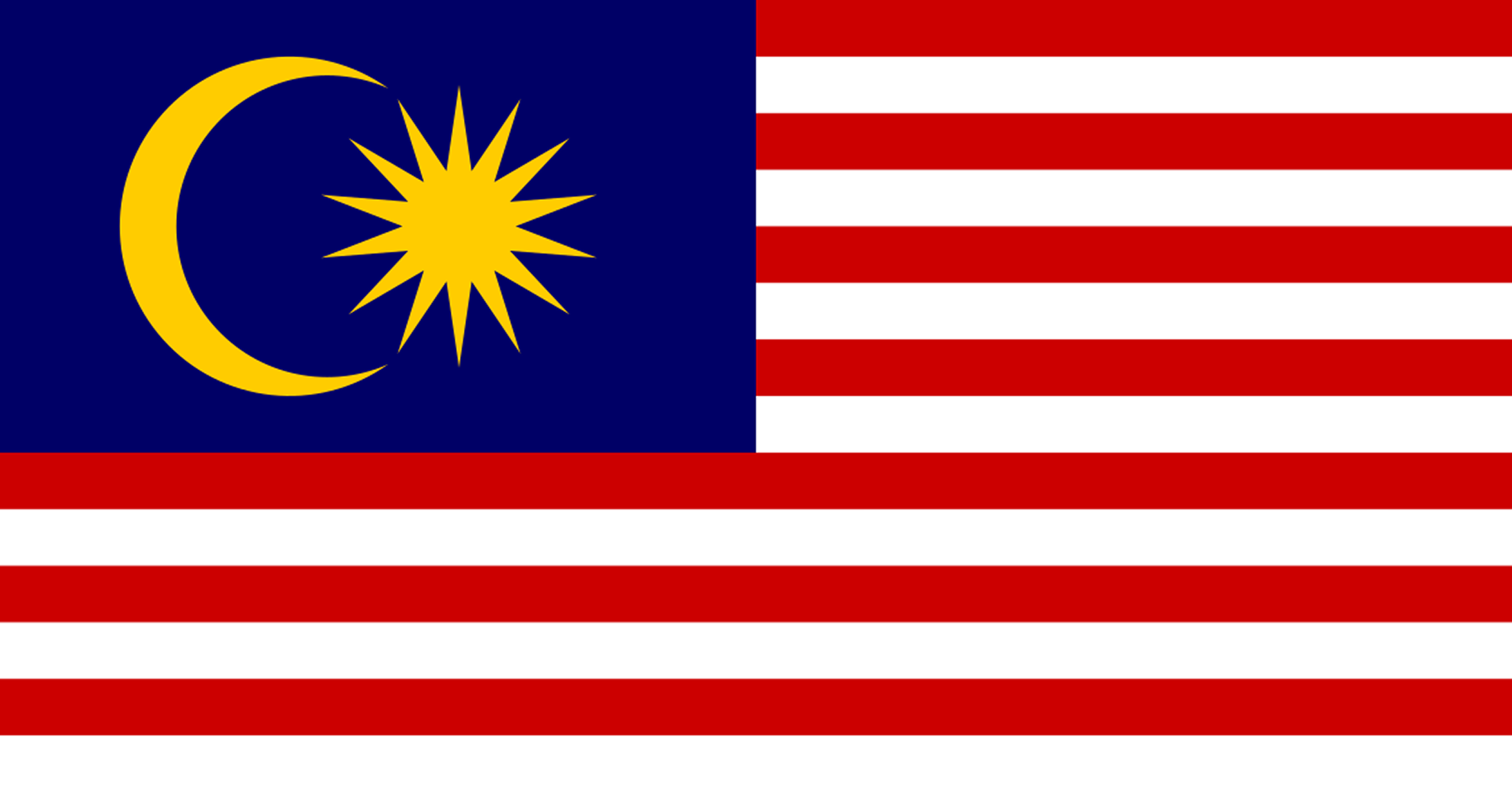}}{\textrm{C}}\xspace}
\newcommand{\MEXICO}{\scalerel*{\bbx{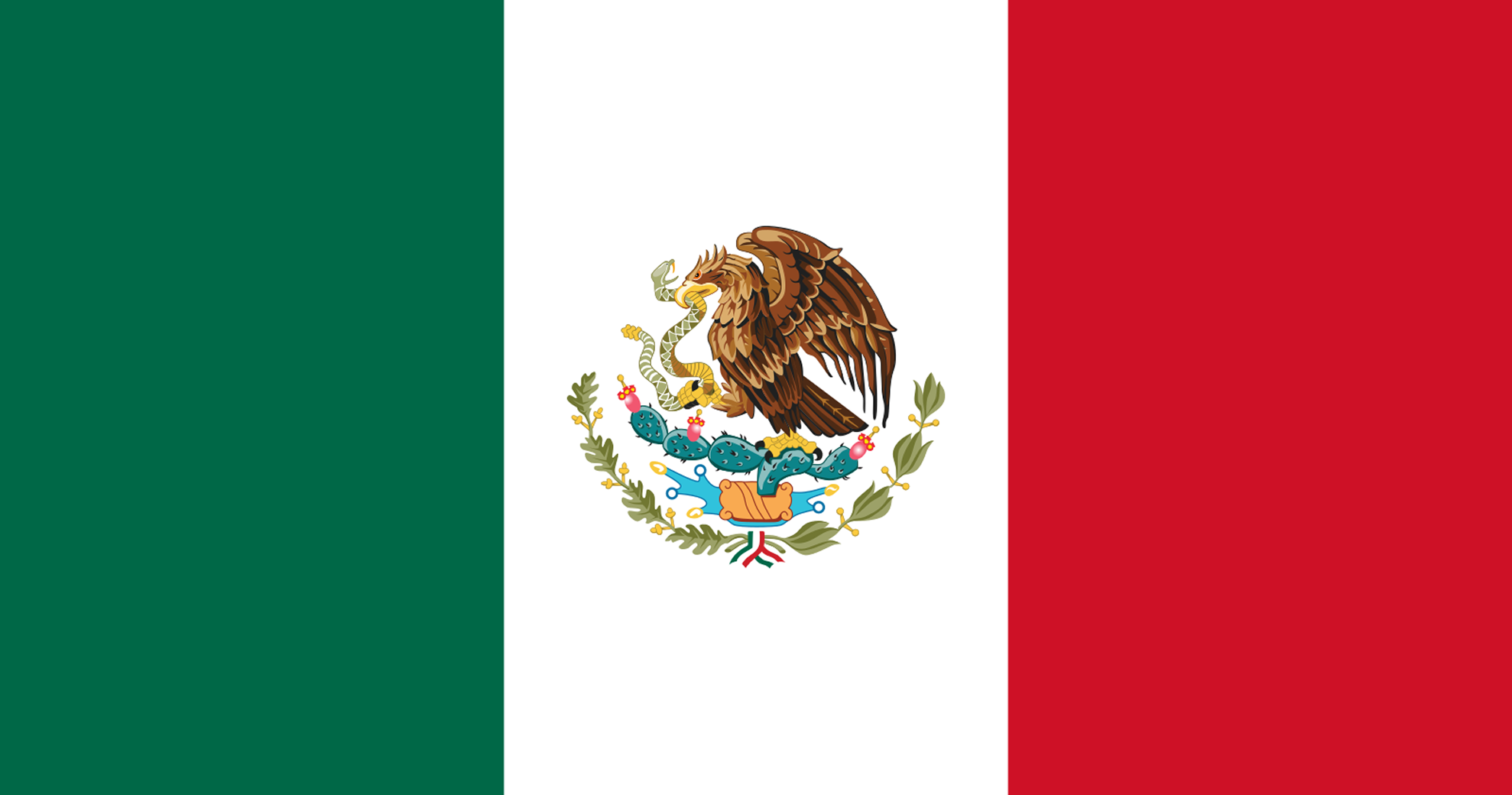}}{\textrm{C}}\xspace}
\newcommand{\PERU}{\scalerel*{\bbx{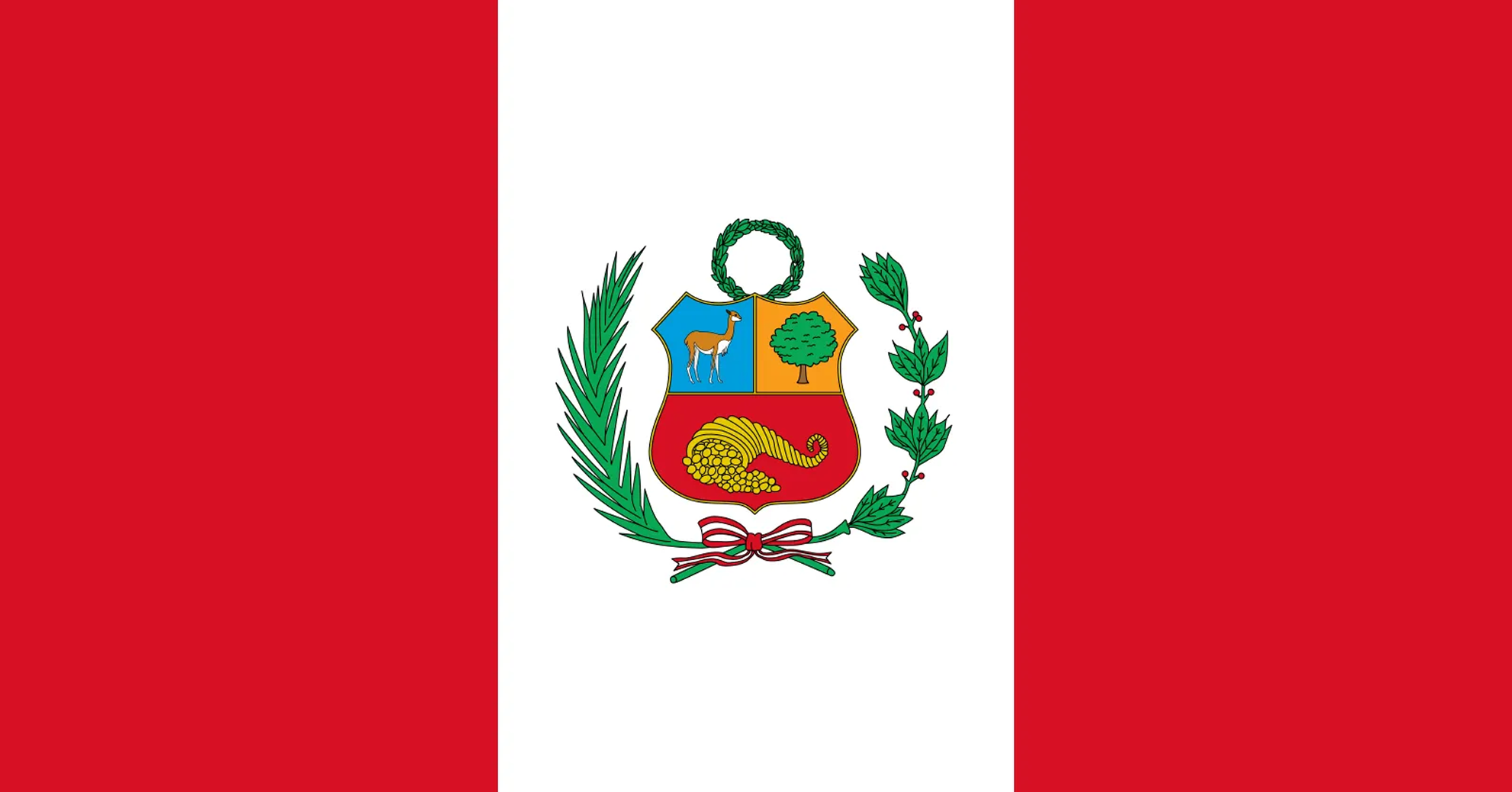}}{\textrm{C}}\xspace}
\newcommand{\PHILIPPINES}{\scalerel*{\bbx{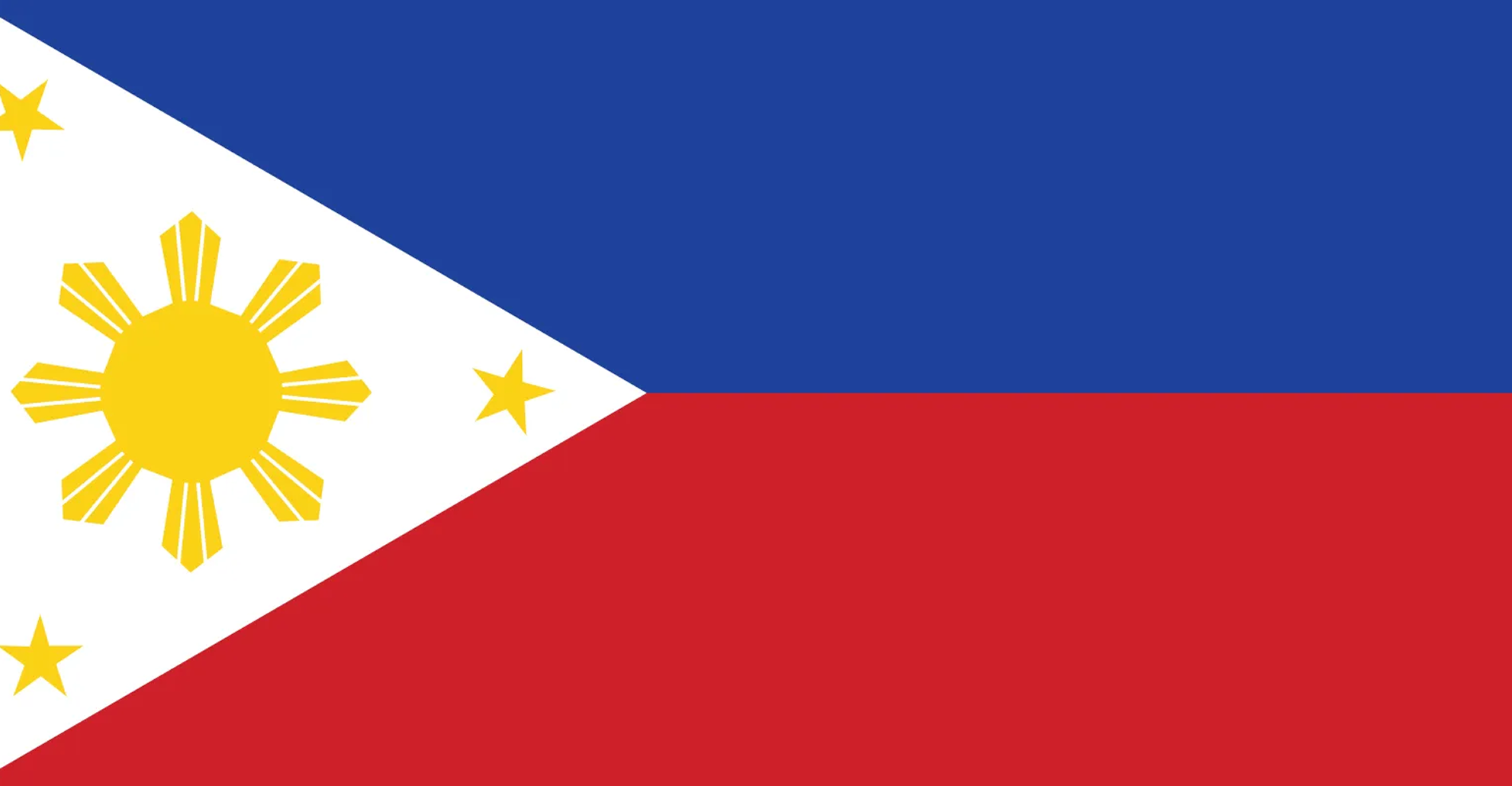}}{\textrm{C}}\xspace}
\newcommand{\POLAND}{\scalerel*{\bbx{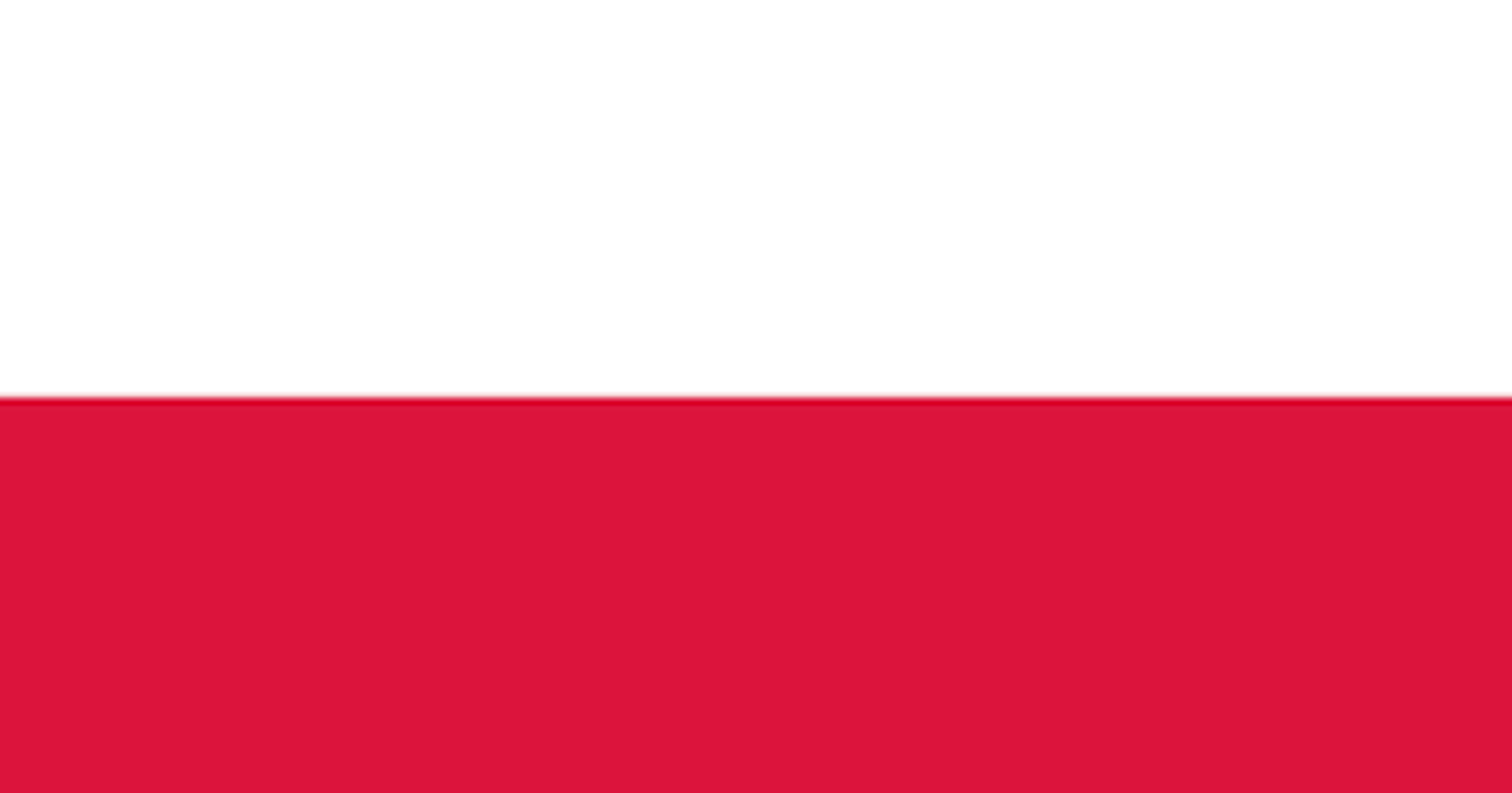}}{\textrm{C}}\xspace}
\newcommand{\SINGAPORE}{\scalerel*{\bbx{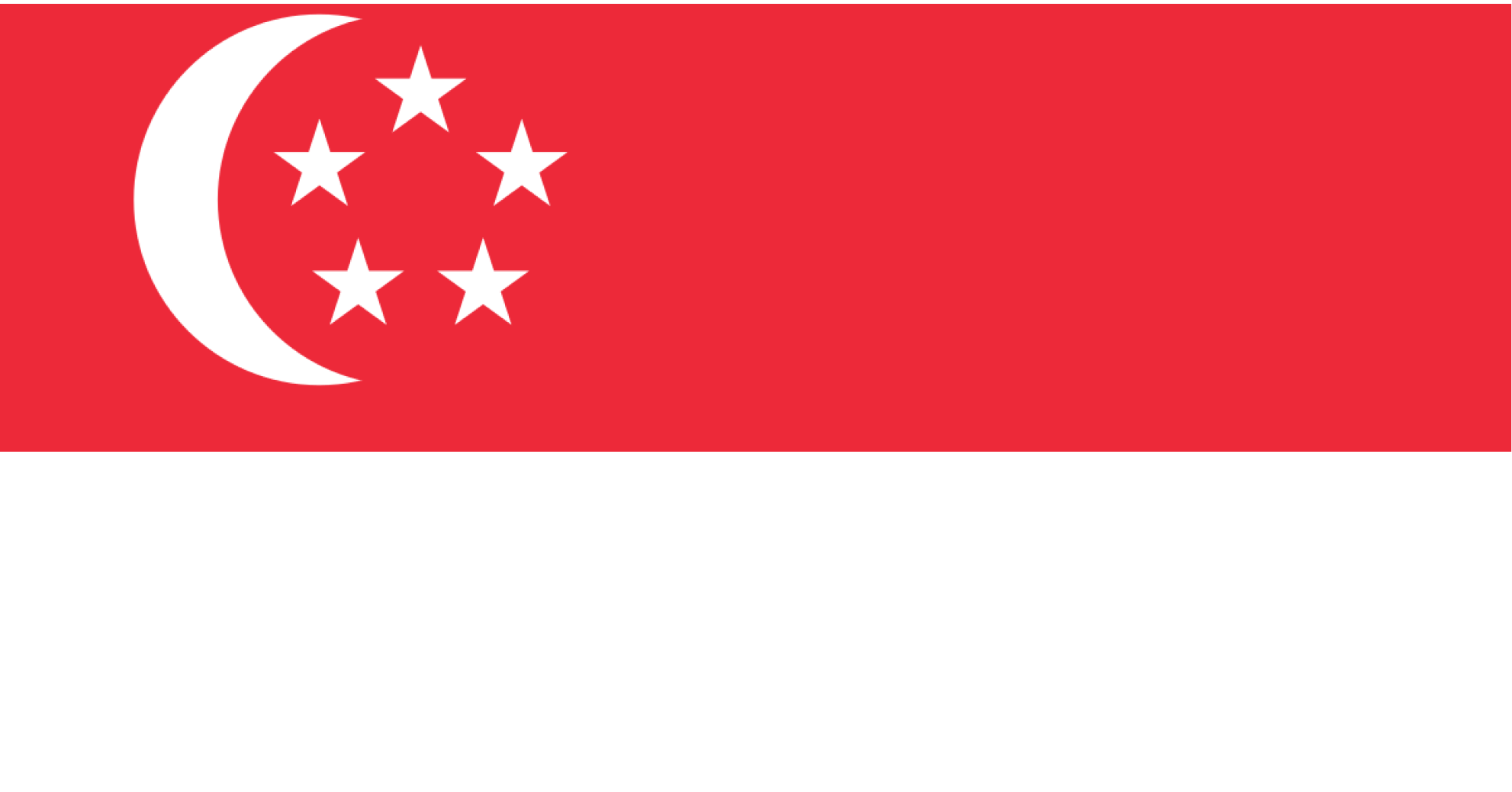}}{\textrm{C}}\xspace}
\newcommand{\SOUTHKOREA}{\scalerel*{\bbx{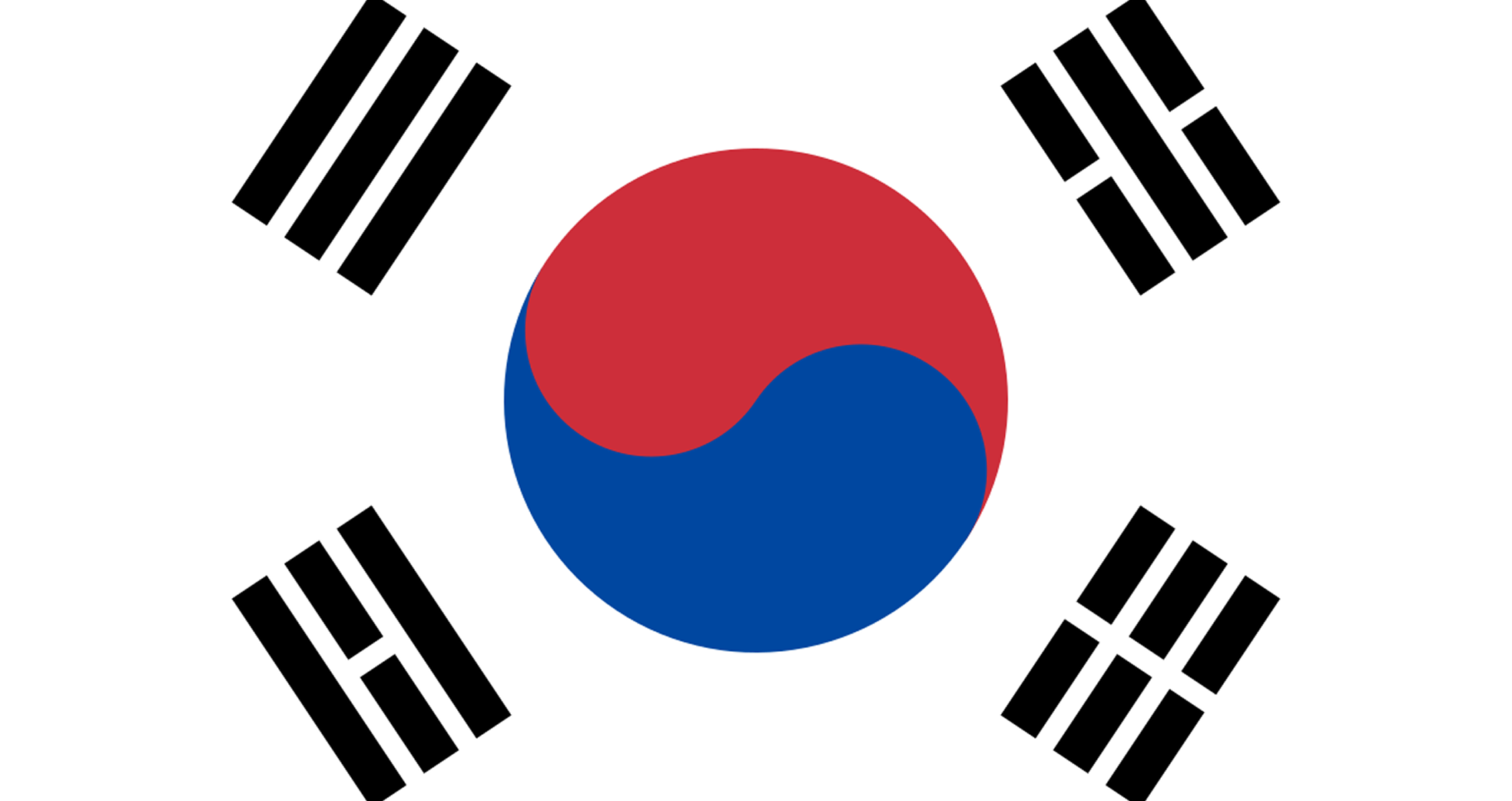}}{\textrm{C}}\xspace}
\newcommand{\SWITZERLAND}{\scalerel*{\bbx{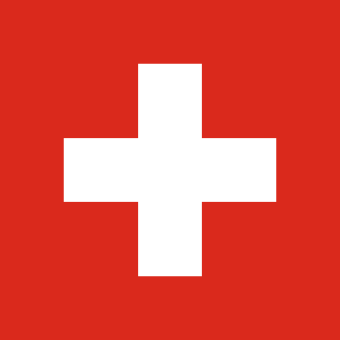}}{\textrm{C}}\xspace}
\newcommand{\THAILAND}{\scalerel*{\bbx{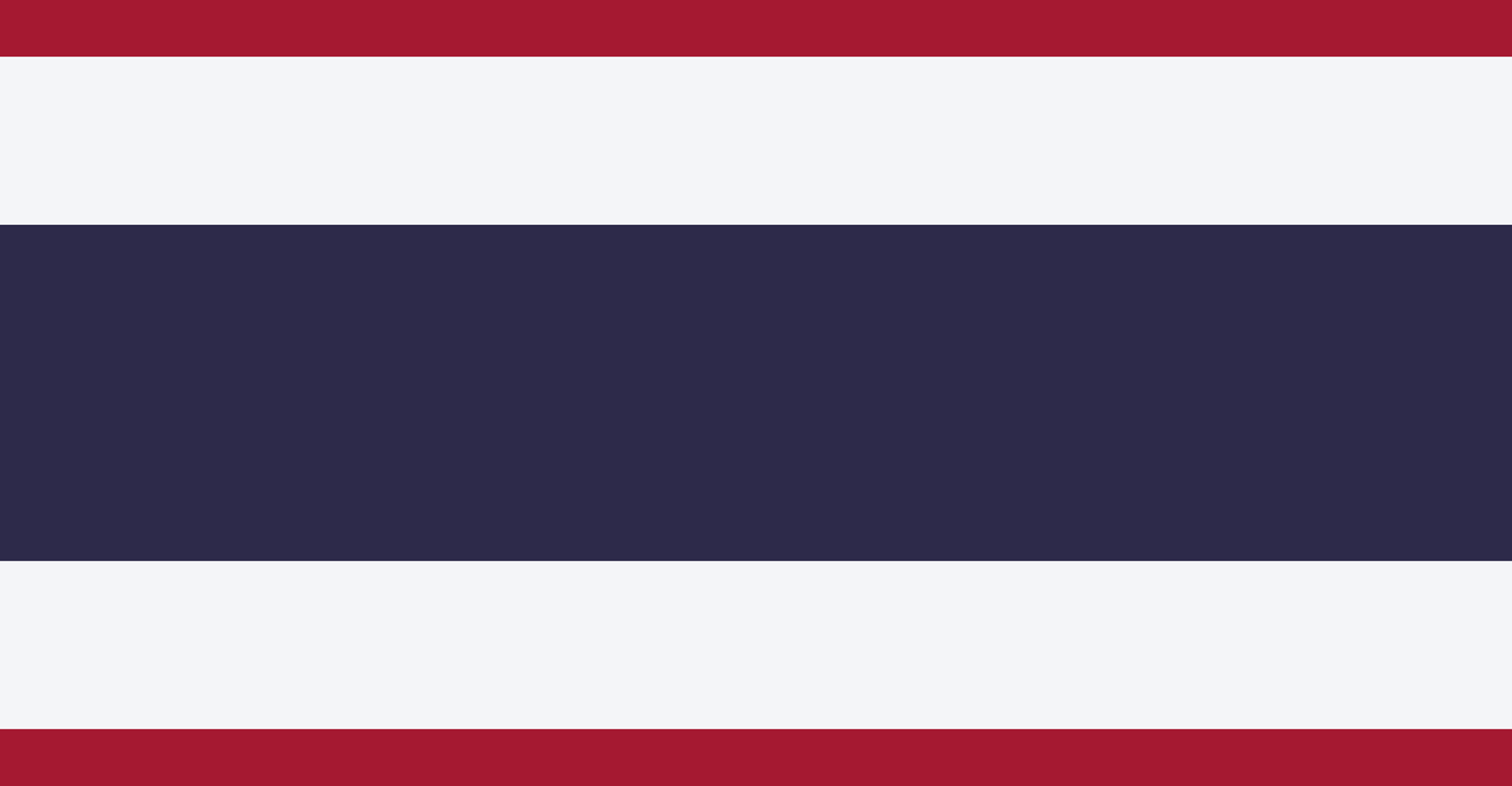}}{\textrm{C}}\xspace}
\newcommand{\UK}{\scalerel*{\bbx{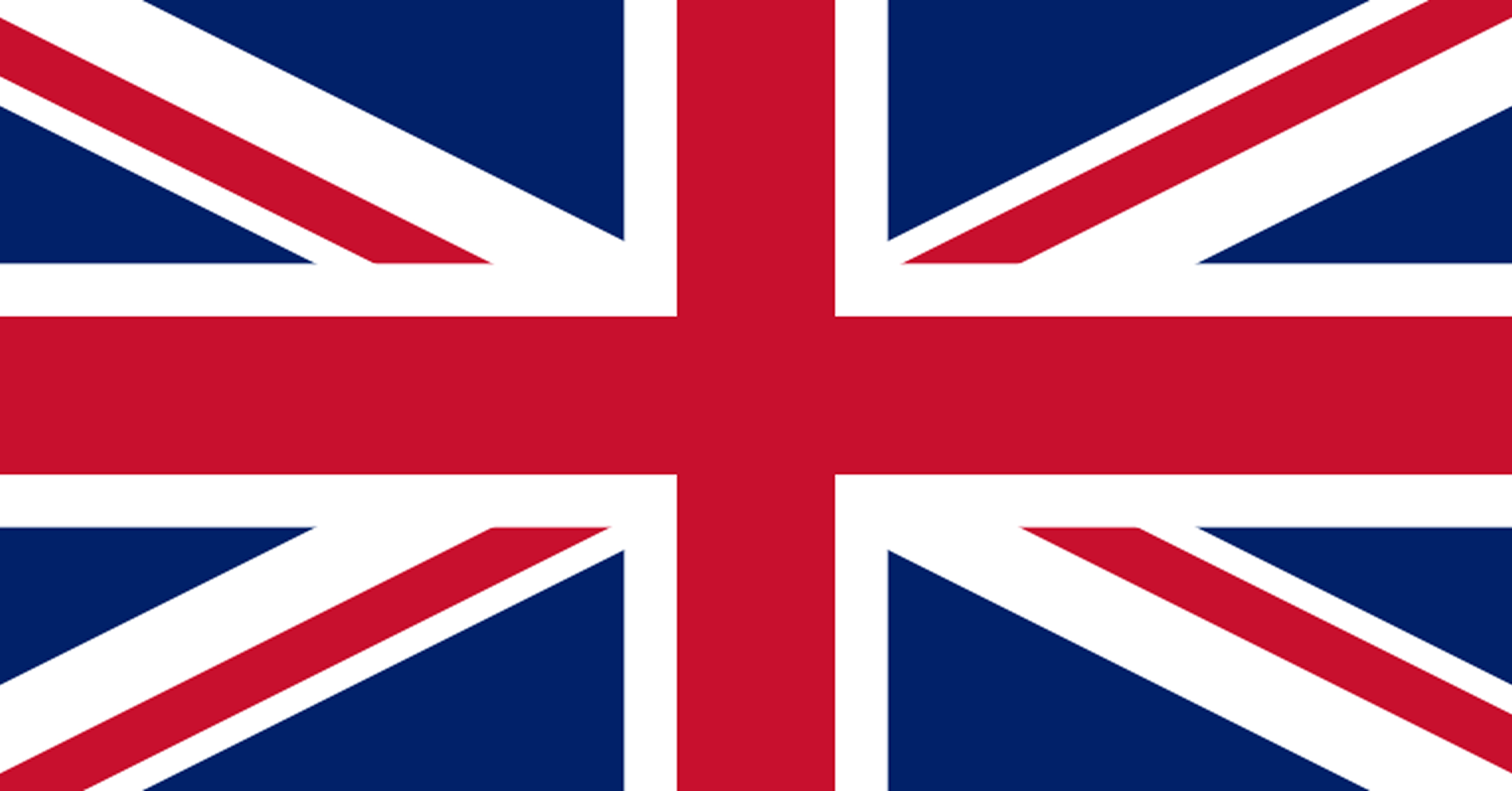}}{\textrm{C}}\xspace}
\newcommand{\TURKEY}{\scalerel*{\bbx{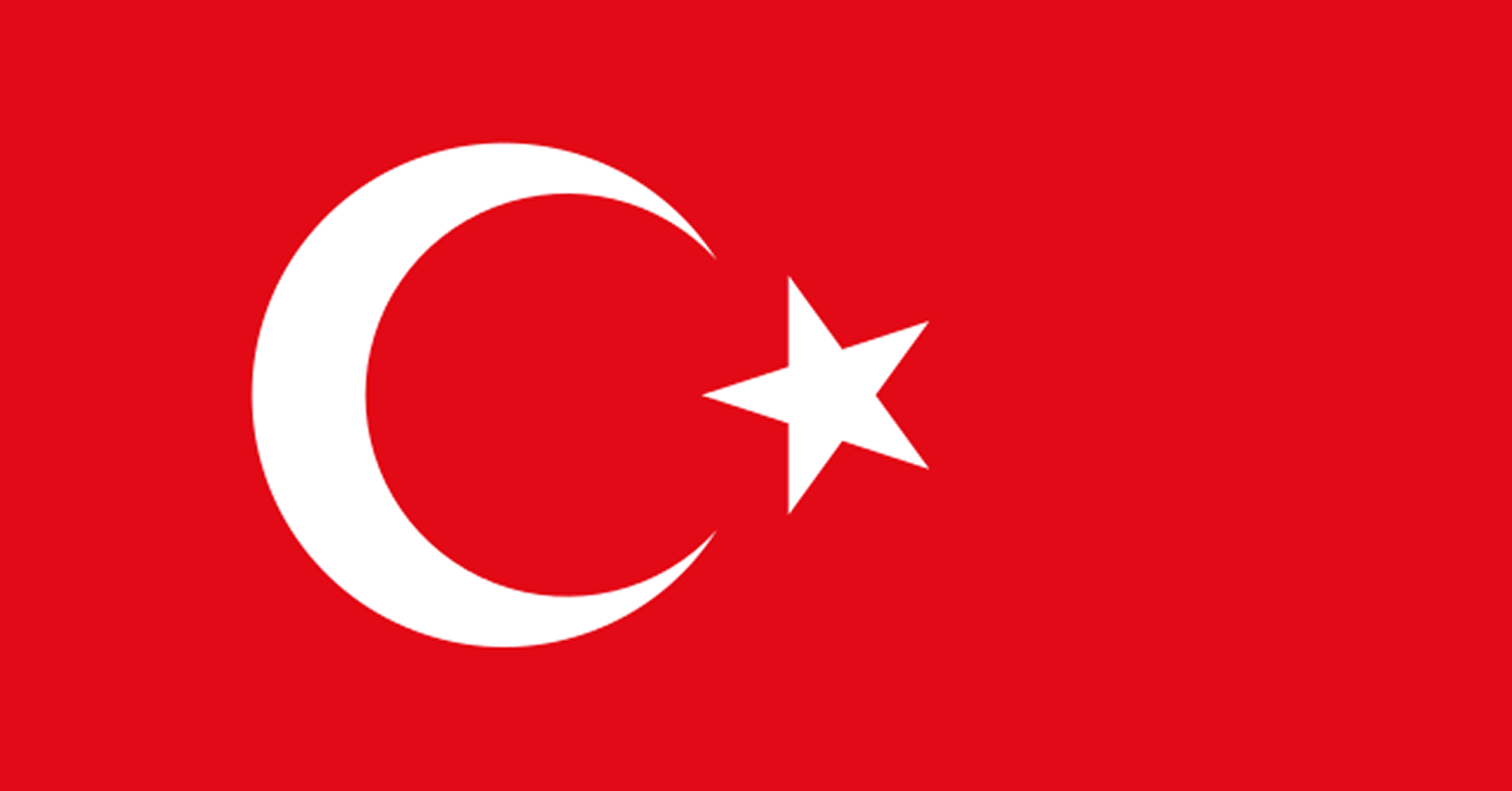}}{\textrm{C}}\xspace}
\newcommand{\TAIWAN}{\scalerel*{\bbx{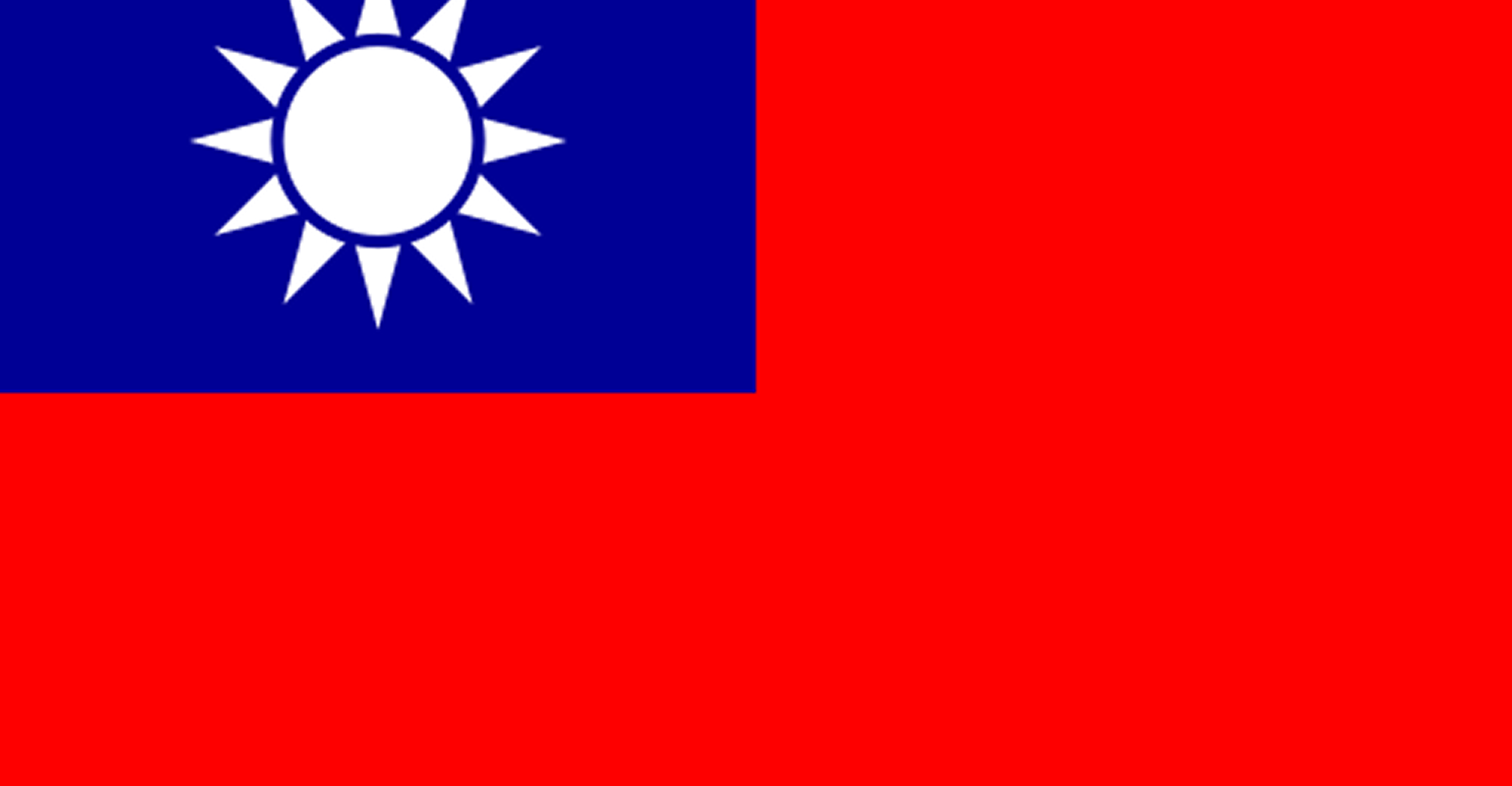}}{\textrm{C}}\xspace}
\newcommand{\ISRAEL}{\scalerel*{\bbx{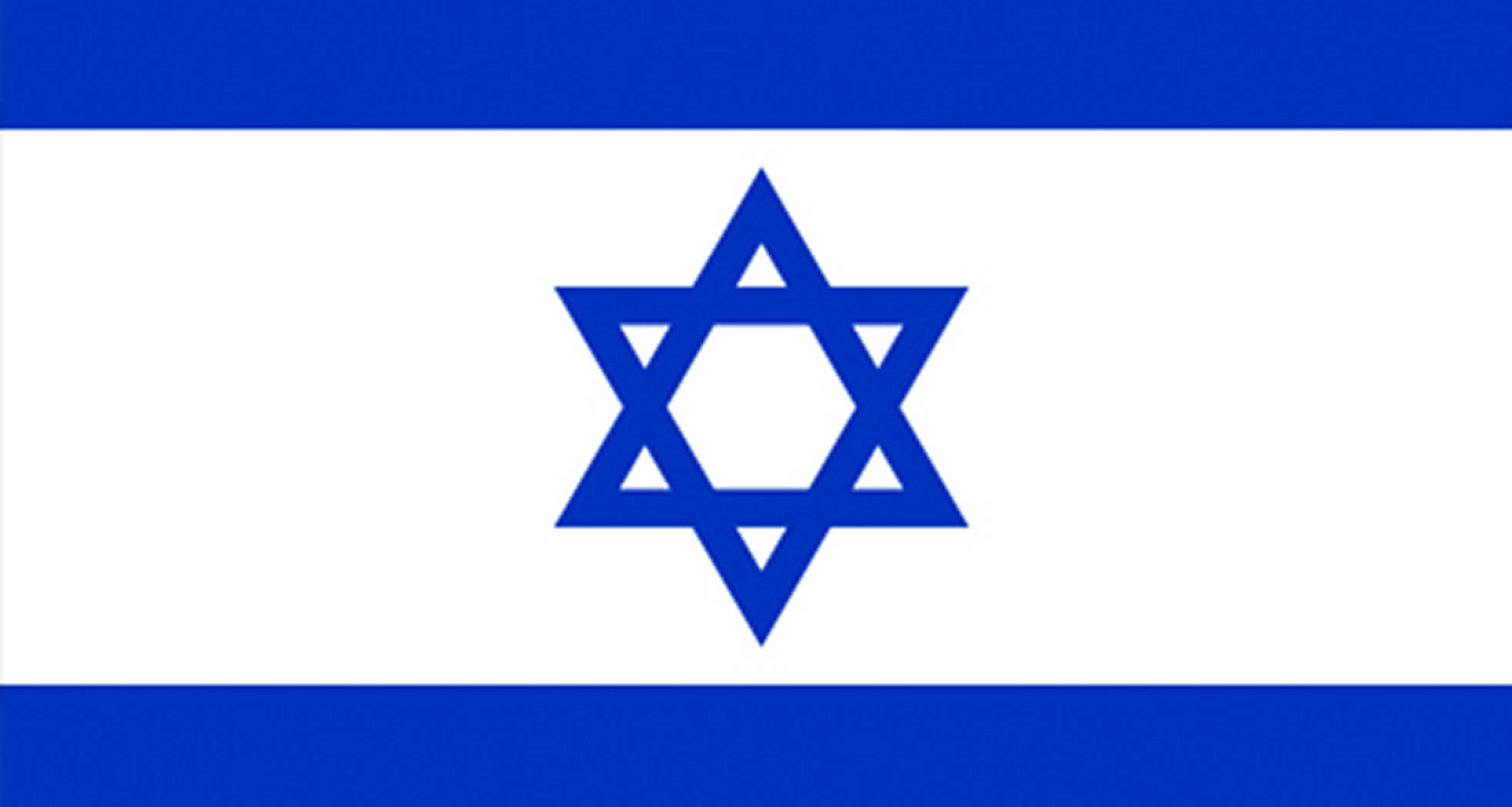}}{\textrm{C}}\xspace}
\newcommand{\JAPAN}{\scalerel*{\bbx{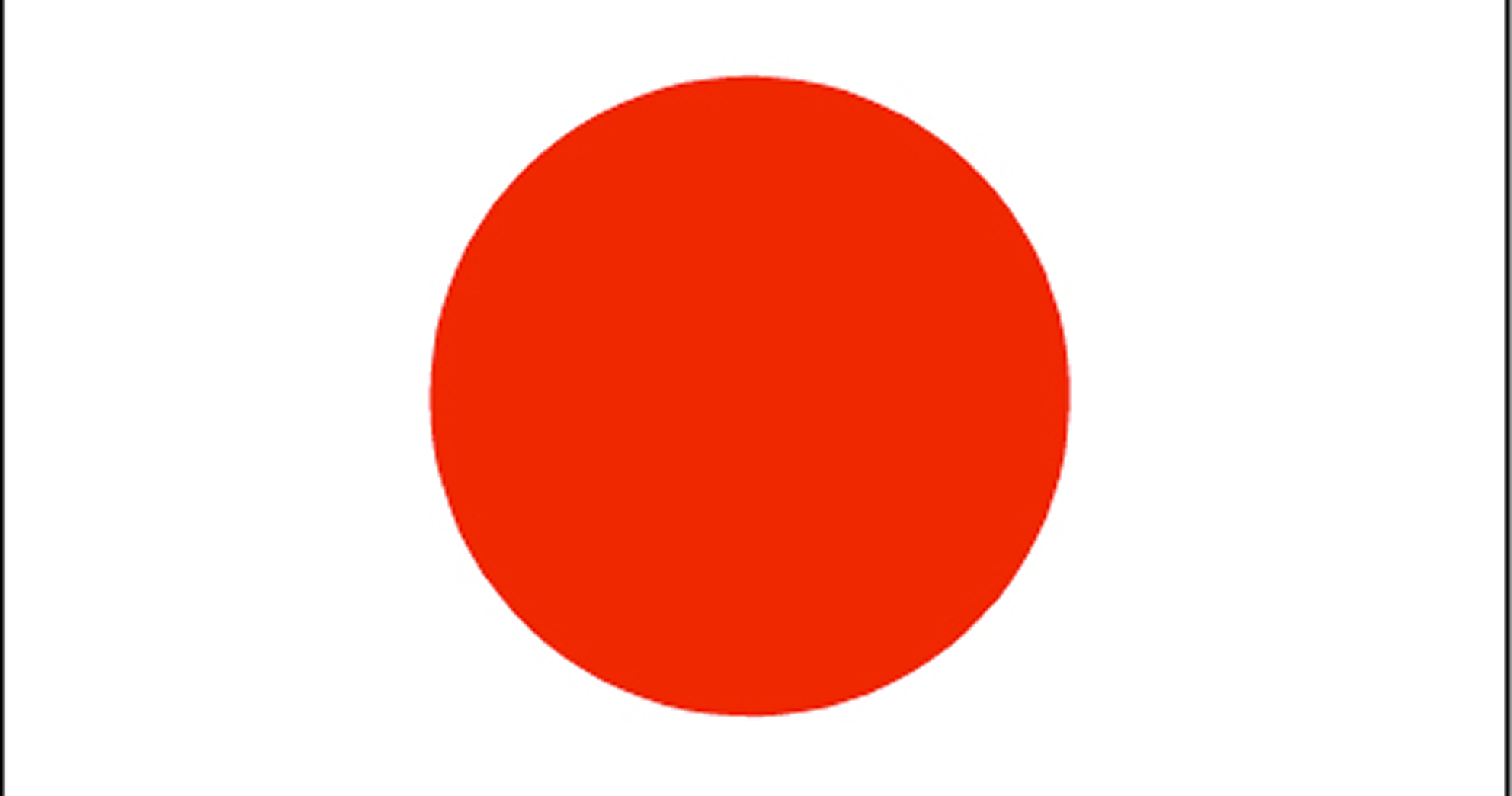}}{\textrm{C}}\xspace}
\newcommand{\RUSSIA}{\scalerel*{\bbx{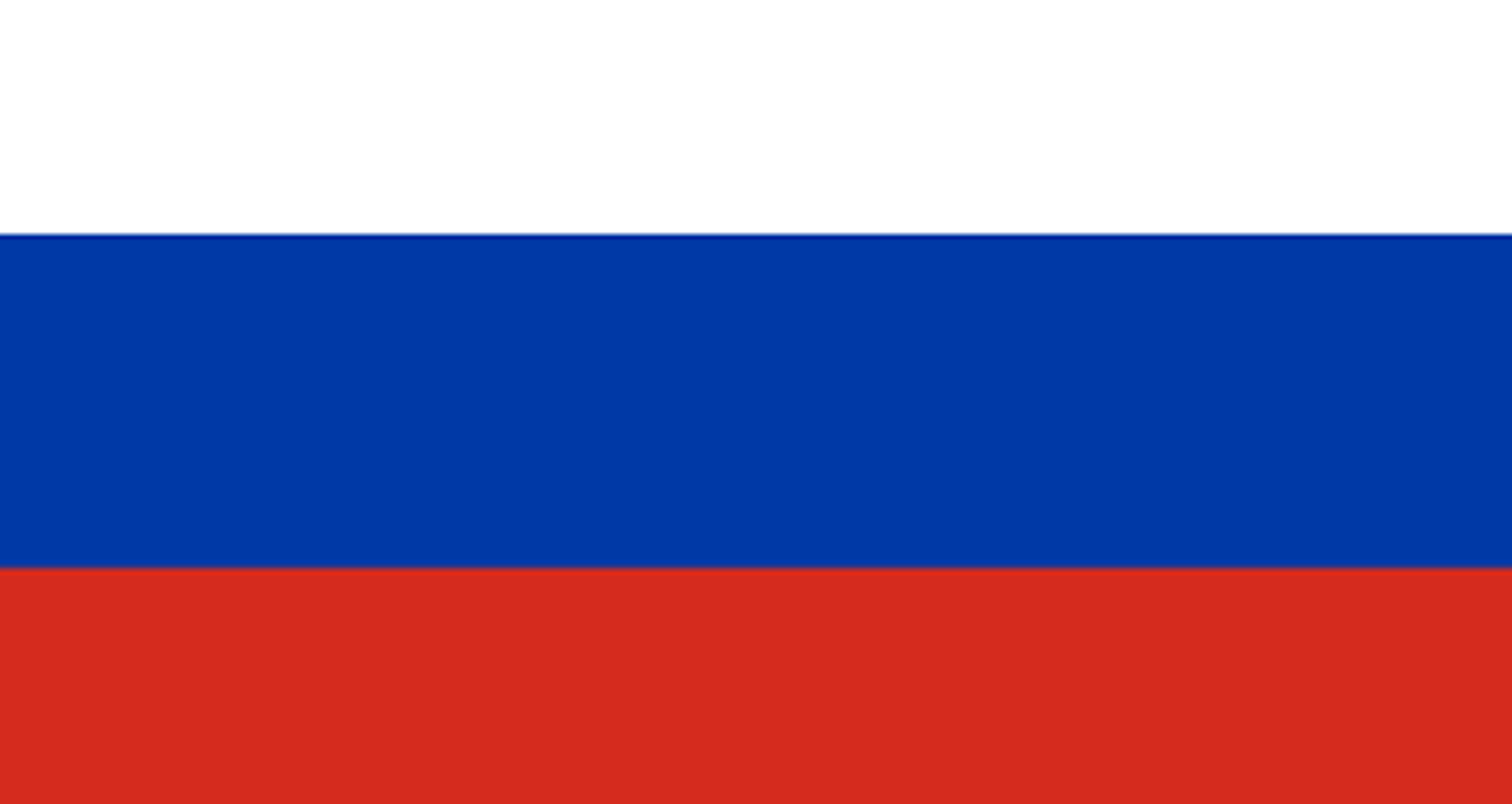}}{\textrm{C}}\xspace}
\newcommand{\huggingface}{\scalerel*{\simplebbx{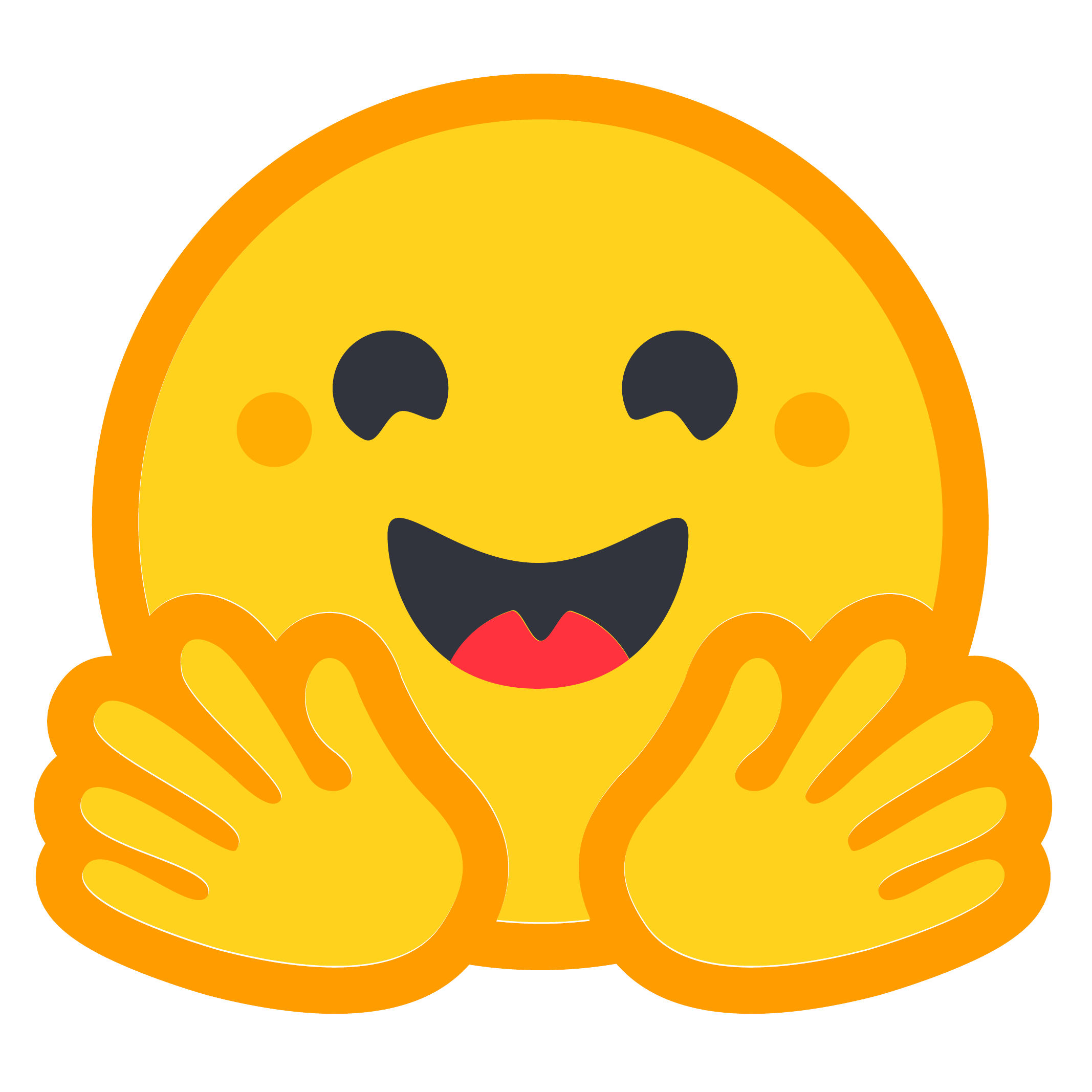}}{\textrm{C}}\xspace}
\newcommand{\github}{\scalerel*{\simplebbx{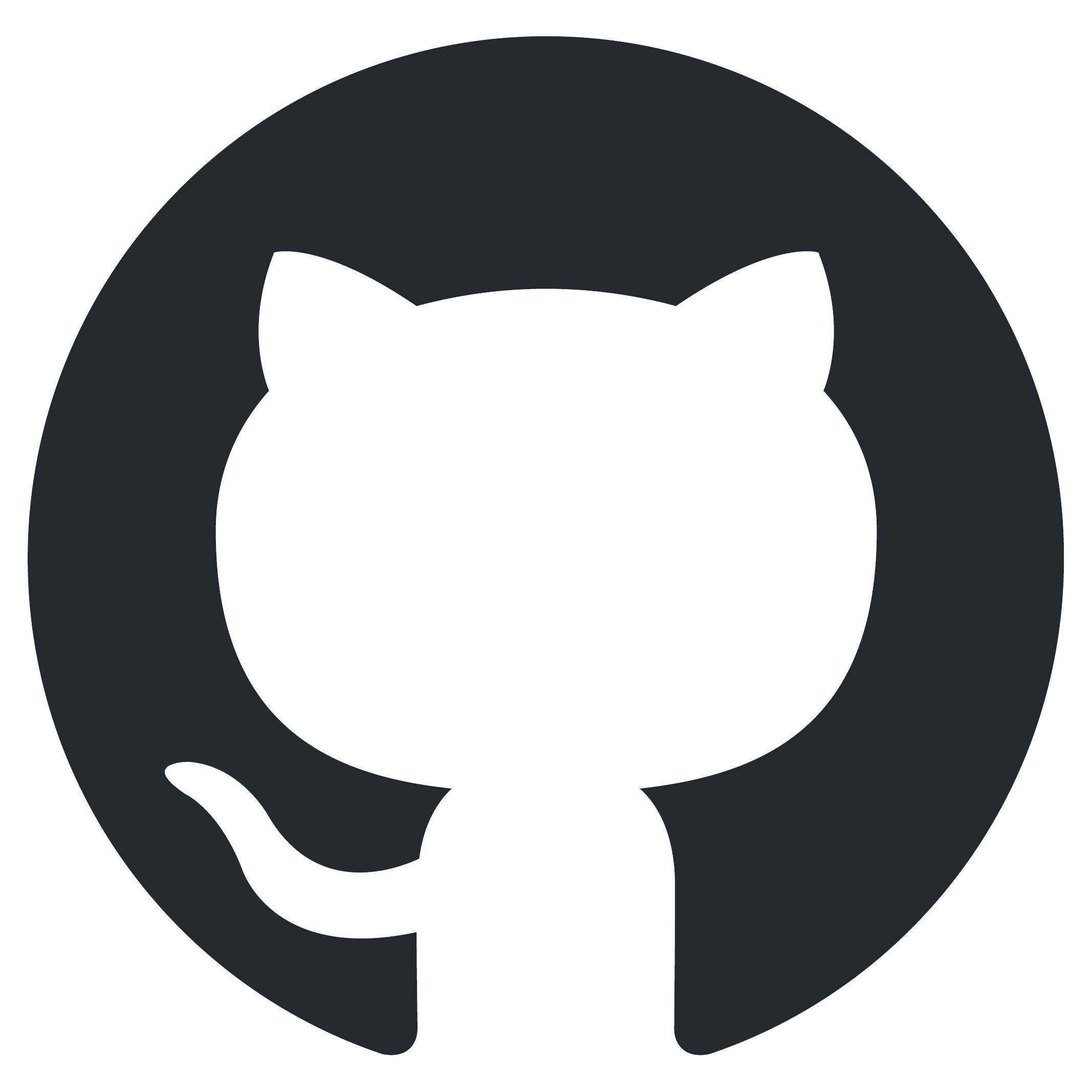}}{\textrm{C}}\xspace}
\newcommand{\logoweb}{\scalerel*{\simplebbx{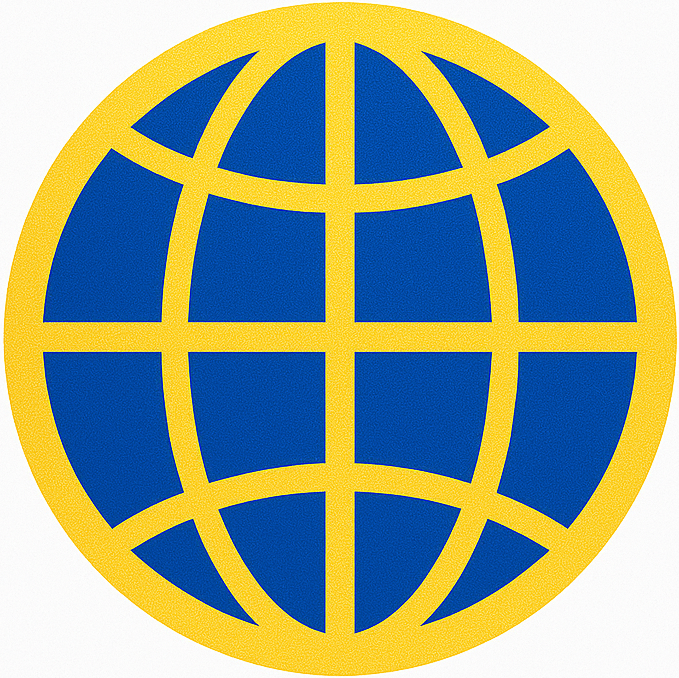}}{\textrm{C}}\xspace}
\newcommand{\Buzz}{\scalebox{1.2}{\scalerel*{\simplebbx{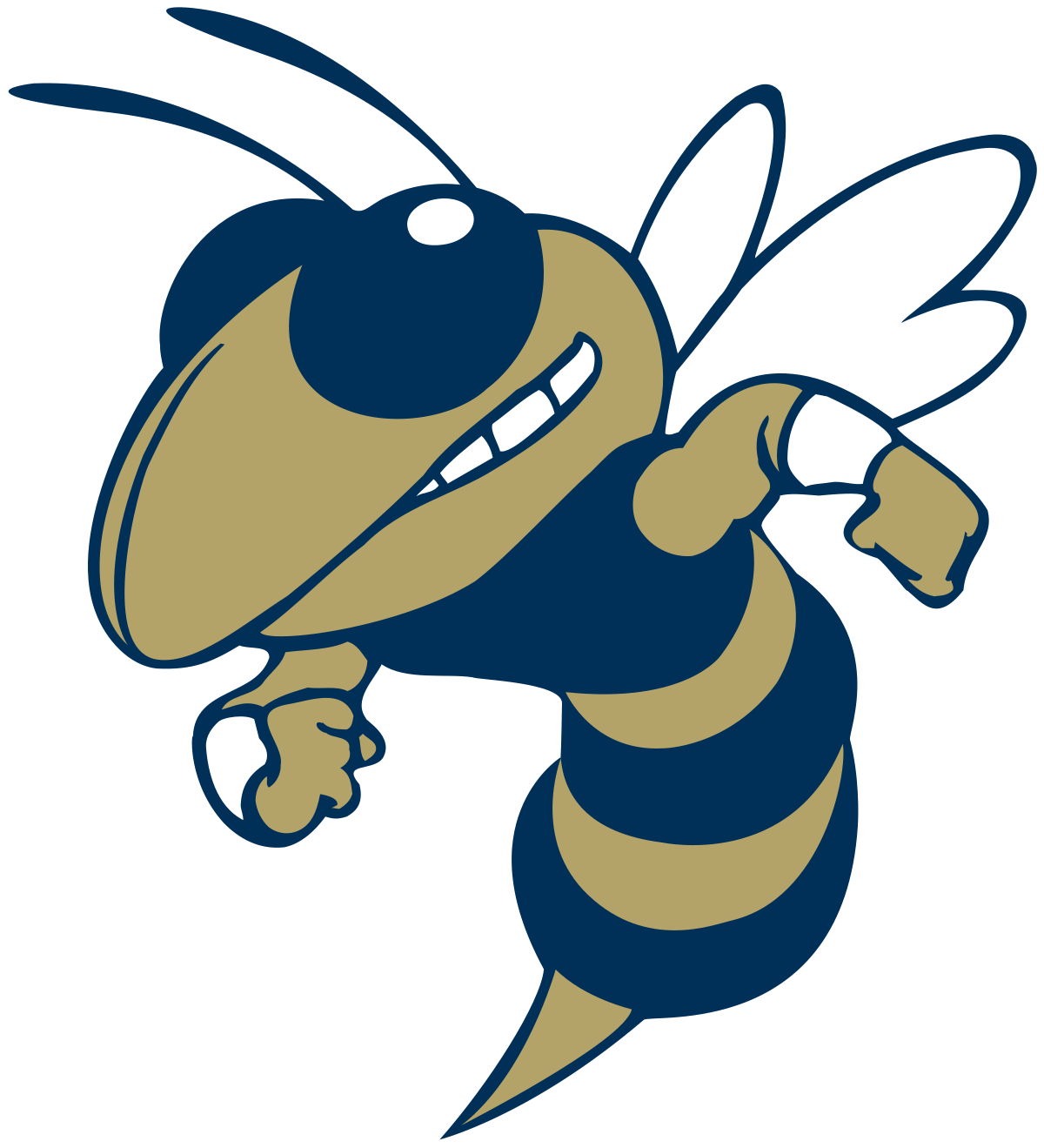}}{\textrm{C}}}\xspace}
\newcommand{\stanford}{\scalerel*{\simplebbx{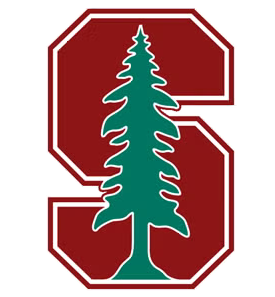}}{\textrm{C}}\xspace}
\newcommand{\duke}{\scalerel*{\simplebbx{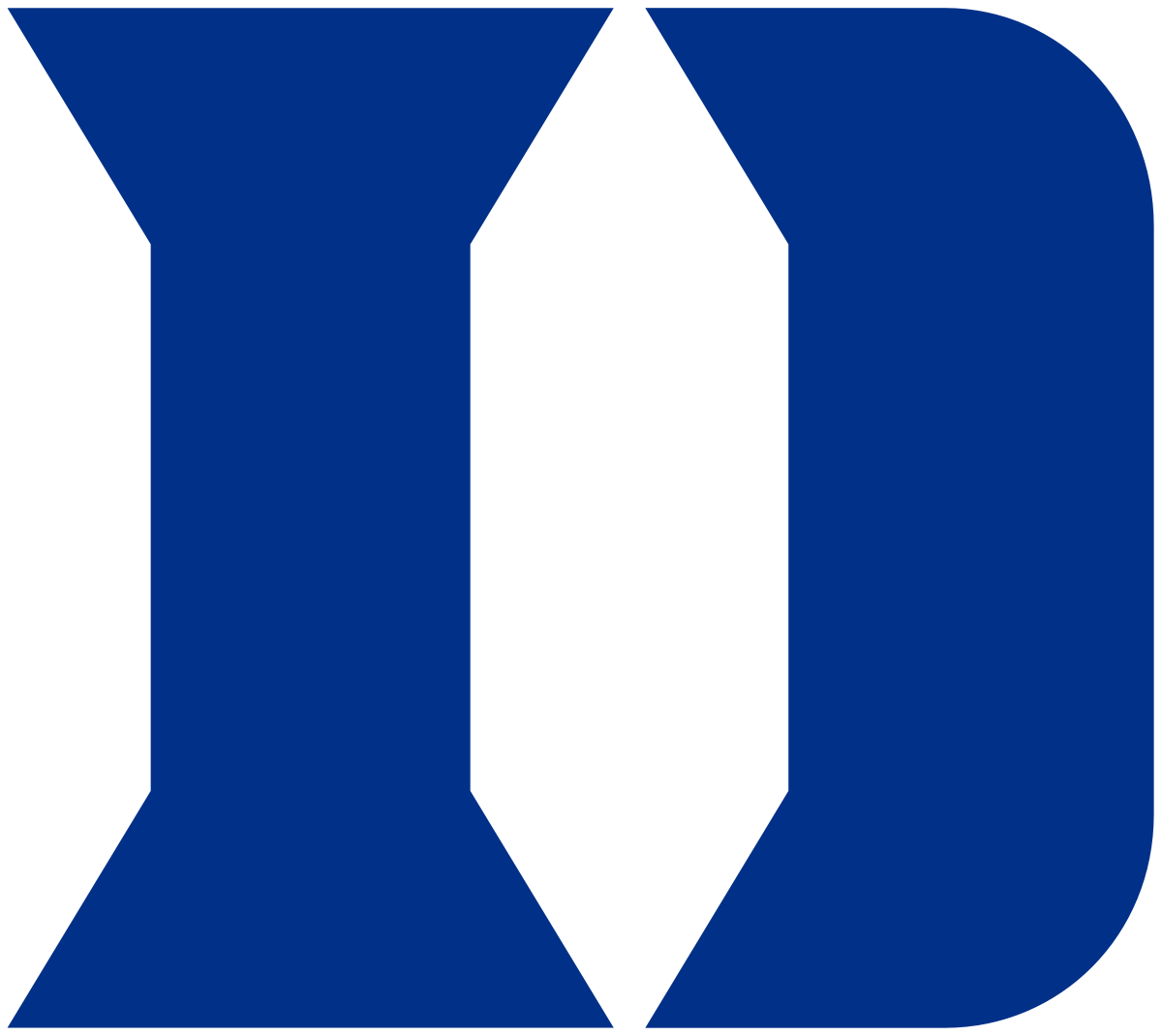}}{\textrm{C}}\xspace}
\newcommand{\msft}{\scalerel*{\simplebbx{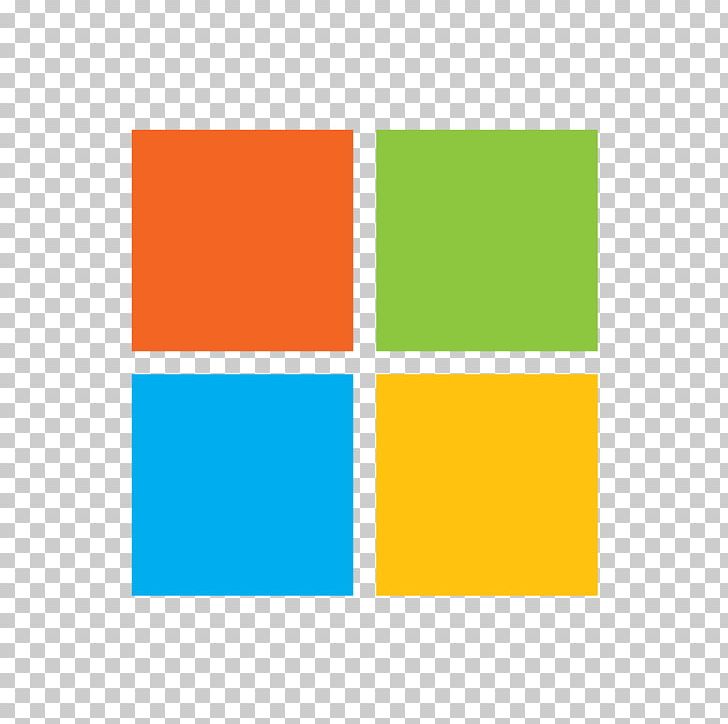}}{\textrm{C}}\xspace}

\title{Words That Unite The World: A Unified Framework for Deciphering Central Bank Communications}

\author{%
Agam Shah\textsuperscript{\Letter}, Siddhant Sukhani\textsuperscript{\Letter \stanford}, Huzaifa Pardawala\textsuperscript{\Letter}, Saketh Budideti\textsuperscript{$\dagger$}, Riya Bhadani\textsuperscript{$\dagger$}, \\
{\bf
Rudra Gopal\textsuperscript{$\dagger$}, Siddhartha Somani\textsuperscript{$\dagger$}, Rutwik Routu\textsuperscript{$\dagger$ \duke}, Michael Galarnyk\textsuperscript{$\dagger$}, Soungmin Lee\textsuperscript{$\dagger$}, } \\\\ 
{\bf
Arnav Hiray, Akshar Ravichandran, Eric Kim, Pranav Aluru, Joshua Zhang, } \\
{\bf
Sebastian Jaskowski\textsuperscript{\msft}, Veer Guda, Meghaj Tarte, Liqin Ye, 
Spencer Gosden, } \\
{\bf
Rachel Yuh, Sloka Chava, Sahasra Chava, Dylan Kelly, } \\
{\bf
Aiden Chiang, Harsit Mittal, Sudheer Chava} \\\\
    \Letter\;Equal First Authors: \{shahagam4, siddhantsukhani5, huzaifahp7\}@gmail.com\\
    \Letter\;and $\dagger$ Indicates core contributors.\\\\
    {\large All Authors: \Buzz Georgia Institute of Technology}\\ 
    Additional Affiliations: \stanford Stanford University; \duke Duke University; \msft Microsoft Corp.\\\\
    \huggingface \href{https://huggingface.co/collections/gtfintechlab/wcb-678965e38178c63158b45fdf}{Data and Models}\;\;\github \href{https://github.com/gtfintechlab/WorldCentralBanks}{Code}\;\;\logoweb \href{https://gcb-web-bb21b.web.app/}{Website}
}

\def\@trackname{Conference on Neural Information Processing Systems (NeurIPS 2025).}

\begin{document}

\maketitle

\begin{abstract}
Central banks around the world play a crucial role in maintaining economic stability. Deciphering policy implications in their communications is essential, especially as misinterpretations can disproportionately impact vulnerable populations. To address this, we introduce the World Central Banks (WCB) dataset, the most comprehensive monetary policy corpus to date, comprising over 380k sentences from 25 central banks across diverse geographic regions, spanning 28 years of historical data. After uniformly sampling 1k sentences per bank (25k total) across all available years, we annotate and review each sentence using dual annotators, disagreement resolutions, and secondary expert reviews. We define three tasks: \SD, \TC, and \CE, with each sentence annotated for all three. We benchmark seven Pretrained Language Models (PLMs) and nine Large Language Models (LLMs) (Zero-Shot, Few-Shot, and with annotation guide) on these tasks, running 15,075 benchmarking experiments. We find that a model trained on aggregated data across banks significantly surpasses a model trained on an individual bank's data, confirming the principle \textit{"the whole is greater than the sum of its parts."} Additionally, rigorous human evaluations, error analyses, and predictive tasks validate our framework's economic utility. Our artifacts are accessible through the HuggingFace, and GitHub under the \texttt{CC-BY-NC-SA 4.0} license.
\end{abstract}
\addtocontents{toc}{\protect\setcounter{tocdepth}{0}}

\section{Introduction}
\label{sec:intro}
Over the past three decades, shocks such as the Dot-Com crash (2000–02), the Global Financial Crisis (2008), the European Sovereign Debt Crisis (2010), negative oil prices (2020), the COVID-19 pandemic (2020), and most recently, the Trump administration’s tariffs (2025) have left their mark on the global economy. During these episodes, central banks responded by first stabilizing volatile markets by expanding the money supply \citep{friedman2010learning,buiter2008central} and later by curbing inflation by tightening monetary conditions \citep{hooley2023quasi,occhino2020quantitative}. The effectiveness of these monetary policies hinges on central banks’ ability to pursue a balance between two key goals: controlling inflation \citep{gerald2007central} and maximizing employment \citep{epstein2007central}. High inflation disproportionately harms lower-income populations, whose wages typically lag in catching up with rapidly rising prices. Conversely, inadequate employment severely limits economic opportunities for these same vulnerable groups, exacerbating social inequalities \citep{jayashankar2023inflation}. 
\begin{figure}[htbp]
    \includegraphics[width=\linewidth]{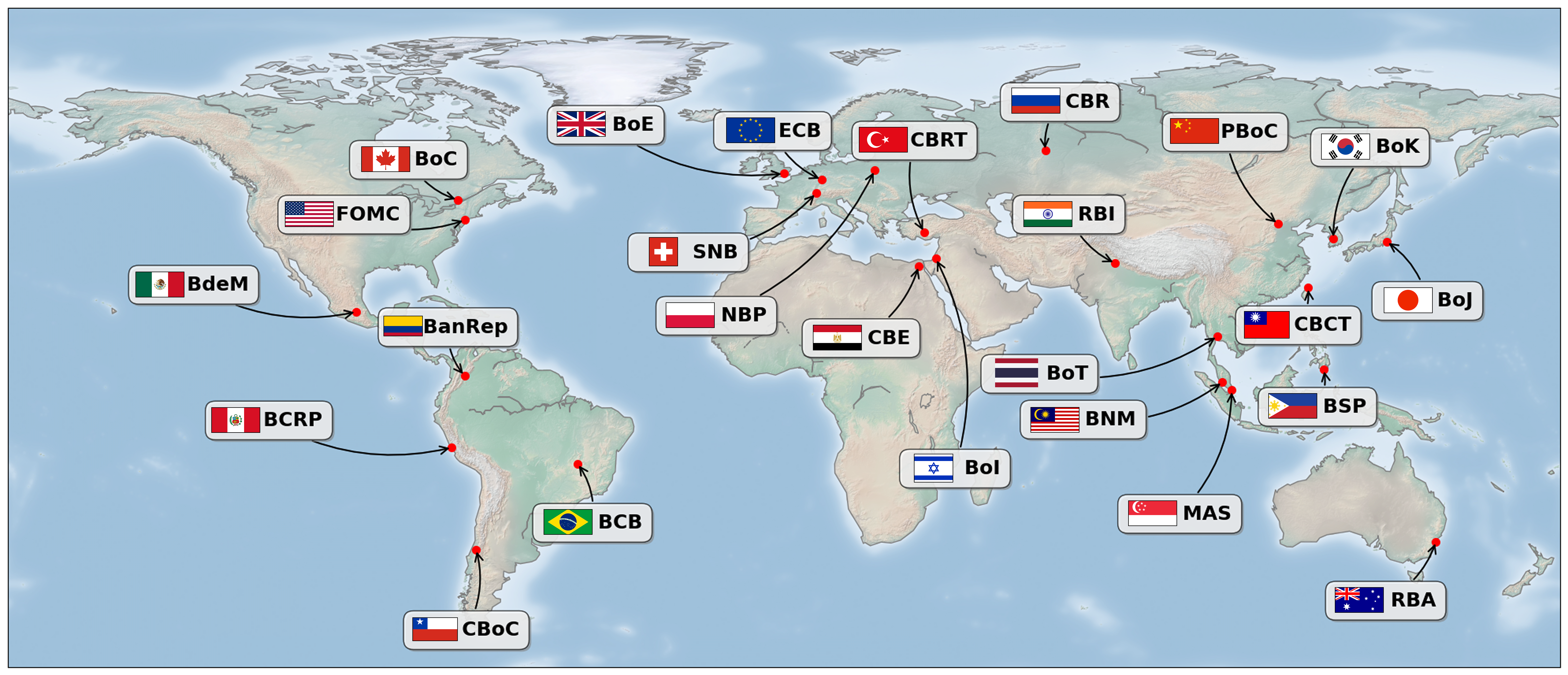}
    
 {\tiny
    \hspace*{0.125em}
    \begin{tabularx}{\dimexpr\linewidth}{ll|ll|ll}
        \multicolumn{2}{c}{\cellcolor{LightBlue1}\textbf{Dataset}} &
        \multicolumn{2}{c}{\cellcolor{Khaki1}\textbf{Models}} &
        \multicolumn{2}{c}{\cellcolor{DarkSeaGreen2}\textbf{Annotations}} \\
        \hhline{------}
        Central Banks & 25 & Pretrained Language Models & 7 & Annotators & 104 \\
        
        Years & 1996--2024 & Large Language Models & 9 & Annotation Guides & 26 \\
        
        Scraped Sentences & 380,200 & Best Stance Model$^*$ & \texttt{RoBERTa-Large} (0.740) & Annotation Steps & 6 \\
        
        Annotated Sentences & 25,000 & Best Temporal Model$^*$ & \texttt{RoBERTa-Base} (0.868) & \cellcolor{celadon} \textbf{Tasks} & 
        \\
        
        \cline{5-5}

        Total Words & 10,289,163 & Best Uncertainty Model$^*$ & \texttt{RoBERTa-Large} (0.846) & Stance Detection & Hawkish, Dovish,\\
       
        Corpus Size & 2,661,400 & Benchmarking Experiments & 15,075 &  & Neutral, Irrelevant\\
        
        Sentences/Year$^*$ & 13,110.34 & Few Shot & \gcheck & Temporal Classification & (Not) Forward Looking \\
        
        Words/Sentence$^*$ & 27.06 & Few Shot + Ann. Guide & \gcheck & Uncertainty Estimation & (Un)certain \\
        \hhline{------}
    \end{tabularx}
}

    \caption{\textbf{A summary of the World Central Bank (WCB)} dataset and experiments. We systematically collect, clean and research the communications from 1996–2024 of 25 central banks at a sentence level, leading to 380,200 sentences (avg. 27.06 words/sentence) in our corpus. We present an annotated dataset consisting of 25,000 sentences across three tasks (\texttt{Stance Detection}, \texttt{Temporal Classification}, and \texttt{Uncertainty Estimation}) using comprehensive individual annotation guides and detailed instructions for annotation. We benchmark seven PLMs and eight LLMs on these tasks, under a bank-specific (1,000 bank-specific annotated sentences) and global setup (25,000 annotated sentences). The performance of the General (All-Banks) Setup model for each task is showcased in the figure. In these tables, $^*$ represents that it is an average.}
    \label{fig:main-figure}
\end{figure}

However, despite the global implications of monetary policy, existing research and datasets used to train artificial intelligence (AI) models within the economic and financial space exhibit a pronounced geographical bias and substandard data quality \citep{tatarinov2025languagemodelingfuturefinance}. As highlighted by \citet{longpre2024bridging}, most text datasets employed in financial and economic machine learning applications originate predominantly from web-crawled sources originating from a narrow geographical region, resulting in significant Western-centric biases. This limited representation within financial data restricts the generalizability and applicability of research insights globally \citep{manvi2024largelanguagemodelsgeographically}. To address this critical gap, we introduce a new dataset WCB that significantly expands the scope of prior resources by collecting extensive monetary policy communications from a geographically diverse set of 25 central banks. Figure \ref{fig:main-figure} shows a summary of our dataset and experiments. The dataset spans 28 years and includes more than 380k sentences, averaging over 27 words each.

To ensure the utility and versatility of our dataset, we annotated 1k uniformly sampled sentences per bank (25k total) on three core tasks: \texttt{Stance Detection, Temporal Classification, Uncertainty Estimation.}. These labels jointly capture economic and monetary policy outlook (\SD), when actions are signaled (\TC), and how confidently decisions are framed (\CE), providing the essential dimensions for interpretation of central-bank communications across the world. Through these annotations and our experiments, our analysis reveals that the aggregated insights substantially exceed what could be inferred by considering each data subset independently. This \textit{“whole is greater than the sum of its parts”} effect highlights the benefit of cross-bank transfer learning and confirms that training on a unified corpus of monetary policy data captures shared linguistic patterns that individual bank models miss. Furthermore, we demonstrate the broad applicability of our best-performing fine-tuned PLMs beyond finance and economics, highlighting their potential for evaluating uncertainty and temporal orientation in other critical contexts, such as Congressional hearings on Conservation, Climate, Forestry, and Natural Resources (see Appendix~\ref{app:esg_transfer_learning}).

Our paper makes several key contributions to the existing literature. First, we introduce a comprehensive dataset of central bank communications alongside a systematic automated collection and cleaning method outlined in Section~\ref{sec:data_methodology}. This is coupled with extensive annotation guides specific to each bank (Appendix~\ref{app:annotation_guides}) and thorough annotation instructions (Appendix~\ref{app:instructions}). Second, we present thorough annotation frameworks and release 78 fine-tuned PLMs capable of extracting nuanced insights from textual data, demonstrating their applicability across various contexts. We additionally perform extensive experiments and analysis under two evaluation setups: General (All-Banks) and Bank-Specific as defined in Section~\ref{sec:results}. These experiments include the generation of meeting minutes using LLMs (Appendix~\ref{app:gen_mm}), economic analysis with inflation data (Appendix~\ref{app:ext_econ_analysis}), error analysis (Appendix~\ref{app:china_error}) and performance gain analysis for \texttt{Stance Detection} model using the General Setup (Appendix~\ref{app:whole_perf_gain}). Finally, our work highlights and addresses significant biases inherent in existing datasets and methodologies that focus on the specific central bank communications \citep{kumar2024wordsmarketsquantifyingimpact, TADLE2022106021,ROZKRUT2007176, KAZINNIK2021308, vega2020assessing, oshima2018monetary}, as a result promoting greater transparency and responsible AI development. The subsequent sections detail our data collection and annotation processes, benchmarking frameworks, empirical results, and discussions of broader implications.

\section{Task Definition}
\label{sec:task_definition}

We define three sentence-level tasks designed to capture key dimensions of central bank communication: \texttt{Stance Detection, Temporal Classification, }and \texttt{Uncertainty Estimation}. The three tasks are defined as follows:

\begin{enumerate}
    \item \textbf{Stance Detection} Each sentence is labeled as \texttt{Hawkish}, \texttt{Dovish}, \texttt{Neutral} (defined within Appendix \ref{app:glossary}), or \texttt{Irrelevant} to capture the monetary stance of the central bank. The first three categories are adopted from \citet{shah-etal-2023-trillion}. However, we introduce the \texttt{Irrelevant} class to capture salutations, logistical details, and operational jargon that do not convey any relevant information pertaining to monetary policy. This addition improves the quality of the collected corpus by preventing contamination from non-informative content without relying on a strict word dictionary. This is beneficial for our analysis as observed within Appendix~\ref{app:Word_Filter}, wherein we illustrate the increased information retained by avoiding a word filter. \SD serves as the central task in our work and is used across several experiments and analyses. 

    \item \textbf{Temporal Classification} Each sentence is classified as \texttt{Forward Looking} or \texttt{Not Forward Looking}. \texttt{Forward Looking} sentences refer to anticipatory and predictive sentences, often indicating future policy direction or economic expectations. \texttt{Not Forward Looking} sentences focus on past or contemporaneous observations. This task captures the temporal framing of the sentences and aids in understanding the temporal orientation of the sentence.

    \item \textbf{Uncertainty Estimation} Each sentence is labeled as \texttt{Certain} or \texttt{Uncertain}, based on the presence of speculative language. It is designed to estimate the level of confidence conveyed in policy statements, which may reflect the clarity or ambiguity of the central bank’s economic outlook.
\end{enumerate}

\section{Dataset Construction}
\label{sec:data_methodology}
\begin{figure}[h!]
    \centering
    \includegraphics[width=\linewidth]{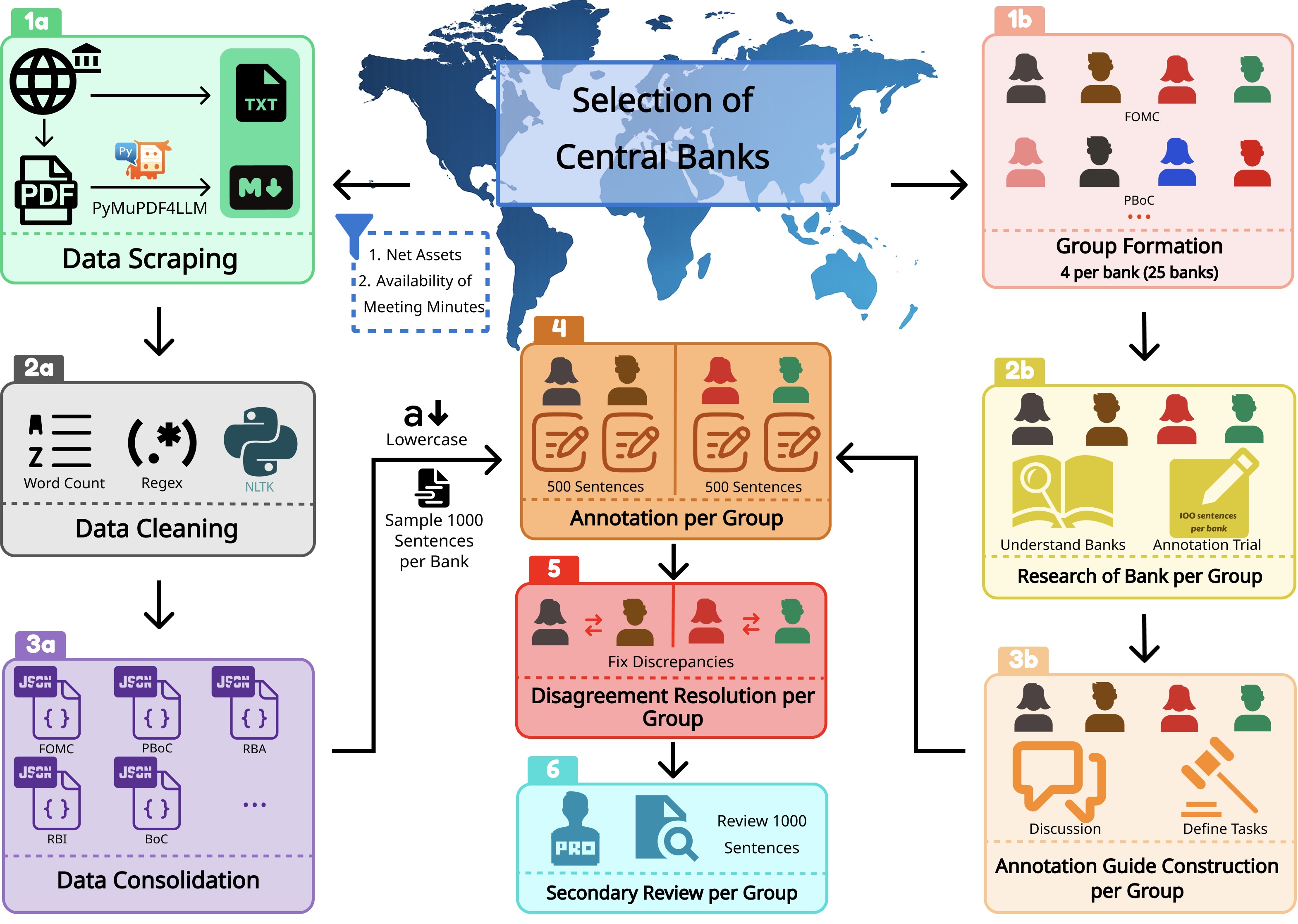}
    \caption{{The dataset generation process for each central bank across the corpus of 25 central banks consists of three stages. (1a) We collected data from the central bank communications and converted them to Markdown files if they were PDFs. (2a) The data was cleaned using regex patterns and tokenized as described in Appendix~\ref{app:datasetconstructioncleaning}. (3a) The data for each central bank was consolidated into \texttt{JSON} files as described in \ref{app:Metadata}. (1b) Annotators were divided into 25 groups of four. (2b) They researched their assigned central bank and labeled a sample of 100 sentences. (3b) Individually drafted annotation guides by each group member were consolidated into a collaborative document for each bank. (4) Independently, the group members annotated the corresponding sentences using their annotation guide. (5) In pairs, they compared their annotations and resolved disagreements (Appendix \ref{app:annotation_guides}). (6) An expert annotator performed a final review of all the annotations by each group.}}
    \label{fig:annotation_process}
\end{figure}

We develop our dataset using a thorough and systematic approach, adhering to strict methodological steps for both data collection and annotation, as shown in Figure~\ref{fig:annotation_process} and detailed below.
\paragraph{Central Bank Selection}
\label{BankChoose}
We selected our initial set of banks using the central bank list provided by \citet{SWF_CentralBank_2025}. Filtering based on the banks' assets in descending order, we prioritize central banks that had either monetary policy meeting minutes or equivalent documents in both format and content available in English to ensure consistent annotations across central banks. A comprehensive reasoning on the selection of central banks is provided in Table~\ref{tab:excluded_banks} Appendix~\ref{app:Banks_excl}. This selection process yielded a final corpus covering 26 central banks' data, out of which one bank's data (Czech National Bank) was chosen at random as a hold-out dataset for ablation studies. The data from the remaining 25 banks was used to create the actual dataset. 

\paragraph{Data Collection, Cleaning, and Consolidation}
The raw documents were collected from the official central banks' websites, targeting meeting minutes or their closest equivalents, as outlined in Table~\ref{tab:BanksChosen} Appendix~\ref{app:datasetconstructioncleaning}. Documents were extracted in either Markdown or plain text form and processed to retain only policy-relevant content. This involved removing contact details and other non-textual artifacts. The cleaned sentences from these documents were then consolidated into \texttt{JSON} files, with full metadata for each document stored as outlined in Appendix~\ref{app:Metadata}. The full collection, cleaning, and consolidation pipeline is described extensively in Appendix~\ref{app:datasetconstructioncleaning}.
\paragraph{Annotation Methodology}
\label{par:annotation} 
Whilst the data was being preprocessed, the entire team of annotators was split into subgroups (four annotators per bank) and assigned various responsibilities required to complete the annotation procedure. Instructions given to the annotators and the full sequence of steps leading to the final annotations are outlined in Appendix~\ref{app:instructions}, including participation terms, initial research, annotation guide development, annotation steps, and disagreement resolution. A key contribution of the dataset also lies in the creation a thorough annotation guide frameworks, to the best of our knowledge. For each of the 26 central banks, annotators developed bank-specific labeling criteria after conducting extensive research on monetary policy mandates, economic history, and communication style. Initial annotation guides were drafted independently and then consolidated into a detailed document. Annotations were carried out in a double-blind manner, with pairwise comparison for disagreement resolution. This was followed by the review of all annotations by an expert. This entire process is visualized in Figure~\ref{fig:annotation_process}. Further details on the annotation guides, annotator names and contact information, as well as inter-annotator agreement statistics for all three tasks are provided in Appendix~\ref{app:annotation_guides}. These details are reported separately for each central bank.

\section{Benchmarking}
\label{sec:benchmarking}
To evaluate the performance of Language Models (LMs) on the tasks (as defined in Section~\ref{sec:task_definition}), we benchmark a suite of seven PLMs and nine LLMs on our annotated dataset, with full specifications provided in Table~\ref{tab:model_abb} Appendix~\ref{app:Model_abb}. For each experiment, we use a 700-150-150 train-test-validation split. For PLMs, we run a grid search across the following parameters: \texttt{seeds = [5768, 78516, 944601]\footnote{These specific seeds were chosen based on the work of \citet{shah-etal-2022-flue} and \citet{shah-etal-2023-trillion}. }, batch sizes = [32, 16], learning rates = [1e-5, 1e-6]}. Full training specifications, compute environment, and LLM API usage are detailed in Appendix~\ref{app:technical_details_benchmarking}, and the different prompts used for LLM inference are listed in Appendix~\ref{app:promptsbenchmarking}.

\subsection{Models}
\label{sec:models}

 The list of models used is given below. Model-specific details are provided in Table~\ref{tab:model_abb}, Appendix~\ref{app:Model_abb}.

    \textbf{PLMs} \texttt{Bert-Large} \citep{devlin2019bert}, \texttt{BERT-Base} \citep{devlin2019bert}, \texttt{RoBERTa-Large} \citep{liu2019roberta}, \texttt{RoBERTa-Base} \citep{liu2019roberta}, \texttt{ModernBert-large} \citep{warner2024smarterbetterfasterlonger}, \texttt{ModernBert-Base }\citep{warner2024smarterbetterfasterlonger}, \texttt{FinBert-Pretrain} \citep{Finbert}.
    
    \textbf{Closed Source LLMs} \texttt{Gemini 2.0 flash} \citep{gemini2.0flash2024}, \texttt{GPT-4o (gpt-4o-2024-08-06)} \citep{hurst2024gpt}, \texttt{GPT 4.1 (gpt-4.1-2025-04-14)} and \texttt{GPT 4.1 mini (gpt-4.1-mini-2025-04-14)} \citep{gpt4.12025}.
    
    \textbf{Open Source (open weights) LLMs} \texttt{Llama-3-70B-Chat} \citep{llama3modelcard}, \texttt{DeepSeek-V3 (DeepSeek-V3-0324)} \citep{deepseekai2025deepseekv3technicalreport}, \texttt{Qwen2.5-72B-Instruct-Turbo} \citep{bai2023qwentechnicalreport}, \texttt{FinMA} \citep{xie2023pixiulargelanguagemodel} and \texttt{Llama-4-Scout-17B-16E-Instruct} \citep{meta2024llama4scout}.
    
    \textbf{Domain Specific Models} FinBert-Pretrain \citep{Finbert} and FinMA \citep{xie2023pixiulargelanguagemodel}.
\subsection{Results}
\label{sec:results}
We present benchmarking results for all three tasks: \SD, \TC, and \CE across multiple PLMs and LLMs. LLMs were evaluated under one setup whereas PLMs were evaluated under two distinct setups defined as follows:

\textbf{General (All-Banks) Setup} In this setup, models are evaluated on a single combined dataset containing all annotated sentences from all 25 central banks. The benchmarking results for \SD are shown in Table~\ref{tab:plm_f1_stance_all}, \TC in Table~\ref{tab:plm_f1_time_all} Appendix~\ref{app:time_all_result}, and \CE in Table~\ref{tab:plm_f1_certain_all} Appendix~\ref{app:certain_all_result}. 

To assess how the stance detection label model trained in the general seup behaves across varying amounts of training data, we ran experiments with sample sizes ranging from 40 to 700. As shown in Figure \ref{fig:size_ablation} below, the F1-Score begins to plateau at $0.74$, indicating that changing the sample size does not impact performance after 600 samples. 

\textbf{Bank-Specific Setup} In this setup, each central bank's data is treated as an independent dataset. Models are evaluated separately on each bank's data, and the average weighted F1-Score across all 25 banks is reported. Results for \texttt{Stance Detection} are shown in Table~\ref{tab:plm_f1_stance_specific}, \texttt{Temporal Classification} in Table~\ref{tab:plm_f1_time_specific} Appendix~\ref{app:time_specific_result}, and \texttt{Uncertainty Estimation} in Table~\ref{tab:plm_f1_certain_specific} Appendix~\ref{app:certain_specific_result}. 
\begin{table}[ht!]
\tiny
\centering
\caption{F1-Scores across the banks for the \SD label of \hawk, \dov, 
\neut, \irr with the standard deviation in brackets. These results are based on the All-Banks General Dataset Setup, as mentioned in Section \ref{sec:results}, on each bank. The best performing PLM and LLM  in each row is highlighted in blue and green respectively. The best performing model for a given bank is bold. The model names are abbreviated in accordance with Table \ref{tab:model_abb}, Appendix \ref{app:Model_abb}. The LLMs are evaluated using a zero-shot prompt.}
\label{tab:plm_f1_stance_all}
    \begin{tabularx}{\textwidth}{
    l 
    *{4}{X}                   
    : *{3}{X}                   
    | *{4}{X}                 
    : *{5}{X}                 
    }
    \toprule
    & \multicolumn{4}{c}{\textbf{Base}} 
    & \multicolumn{3}{c}{\textbf{Large}}
    & \multicolumn{4}{c}{\textbf{Closed-Source}} 
    & \multicolumn{5}{c}{\textbf{Open-Source}} \\
    \cmidrule(lr){2-5}\cmidrule(lr){6-8}\cmidrule(lr){9-12}\cmidrule(lr){13-17}
    \textbf{Bank} 
    & \MB MBB
    & \google BB
    & \finbert FB 
    & \meta RBB 
    & \MB MBL
    & \google BL
    & \meta RBL
    & \google Gem
    & \openai 4o
    & \openai 4.1m
    & \openai 4.1
    & \deepseek DS
    & \qwen Qwen
    & \FinMA FM
    & \meta L3
    & \meta L4S \\
    \midrule
\BRAZIL~BCB & \cellcolor{blue!20}\textbf{.678 (.039)} & .635 (.057) & .609 (.008) & .655 (.018) & .673 (.016) & .636 (.014) & .634 (.050) & .528 (.027) & .498 (.017) & .462 (.020) & .504 (.017) & \cellcolor{green!20}.613 (.021) & .525 (.028) & .350 (.034) & .503 (.021) & .589 (.049) \\

\PERU~BCRP     
& .788 (.021) 
& .781 (.008) 
& .764 (.009) 
& .779 (.028) 
& .798 (.013) 
& .801 (.024) 
& \cellcolor{blue!20}\textbf{.821 (.015)}
& \cellcolor{green!20}.675 (.004) & .628 (.008) & .634 (.035) & .665 (.004) & .666 (.010) & .503 (.062) & .301 (.031) & .641 (.014) & .620 (.043) \\

\MALAYSIA~BNM  
& .626 (.029) 
& .629 (.017) 
& \cellcolor{blue!20}\textbf{.653 (.023)}
& .601 (.012) 
& .630 (.041) 
& .644 (.013) 
& .640 (.009)
& .409 (.006) & .443 (.027) & .430 (.007) & .409 (.025) & .475 (.013) & .333 (.026) & .160 (.033) & \cellcolor{green!20}.567 (.025)& .435 (.005) \\

\PHILIPPINES~BSP  
& .741 (.015) 
& .697 (.020) 
& .707 (.028) 
& .749 (.010) 
& .724 (.049) 
& .698 (.025) 
& \cellcolor{blue!20}\textbf{.784 (.017)}
& .424 (.042) & .420 (.069) & .451 (.039) & .514 (.076) & .534 (.029) & .380 (.015) & .245 (.027) & \cellcolor{green!20}.584 (.042)& .500 (.035) \\

\COLOMBIA~BanRep  
& .685 (.031) 
& .638 (.022) 
& .679 (.017) 
& .673 (.013) 
& \cellcolor{blue!20}\textbf{.702 (.001) }
& .650 (.017) 
& .701 (.013)
& .515 (.021) & .455 (.031) & .520 (.036) & .553 (.015) & .570 (.037) & .450 (.023) & .230 (.033) & \cellcolor{green!20}.573 (.038)& .423 (.033) \\
\CANADA~BoC   
& .722 (.033) 
& .740 (.006) 
& .750 (.010) 
& .745 (.032) 
& .721 (.004) 
& .755 (.005) 
& \cellcolor{blue!20}\textbf{.785 (.010)}
& .629 (.069) & .647 (.052) & .641 (.052) & .657 (.028) & .657 (.026) & .524 (.038) & .264 (.029) & \cellcolor{green!20}.669 (.043) & .644 (.024) \\

\UK~BoE      
& .686 (.031) 
& .706 (.056) 
& .734 (.052) 
& .769 (.019) 
& .735 (.039) 
& .755 (.009) 
& \cellcolor{blue!20}\textbf{.785 (.034)}
& .543 (.026) & .524 (.031) & .537 (.048) & .602 (.031) & .543 (.070) & .396 (.021) & .129 (.026) & \cellcolor{green!20}.661 (.044) & .518 (.045) \\

\ISRAEL~BoI  
& .652 (.003) 
& .642 (.023) 
& .616 (.019) 
& .614 (.008) 
& .658 (.023) 
& .628 (.033) 
& \cellcolor{blue!20}\textbf{.689 (.017)} 
& .474 (.011) & .460 (.005) & .433 (.032) & .526 (.013) & .482 (.001) & .329 (.025) & .085 (.011) & \cellcolor{green!20}.594 (.023) & .430 (.008) \\

\JAPAN~BoJ   
& .691 (.020) 
& .662 (.044) 
& .629 (.047) 
& .660 (.020) 
& \cellcolor{blue!20}\textbf{.708 (.042)}
& .683 (.033) 
& .702 (.025)
& .524 (.010) & .545 (.021) & .465 (.028) & .565 (.008) & .498 (.009) & .406 (.040) & .157 (.033) & \cellcolor{green!20}.574 (.027)& .507 (.026) \\

\SOUTHKOREA~BoK  
& .723 (.056) 
& .664 (.009) 
& .679 (.011) 
& .706 (.018) 
& .740 (.009) 
& .700 (.028) 
& \cellcolor{blue!20}\textbf{.755 (.019)}
& .646 (.040) & .648 (.030) & .594 (.066) & \cellcolor{green!20}.678 (.016)& .629 (.016) & .466 (.076) & .181 (.031) & .632 (.032) & .592 (.047) \\

\MEXICO~BdeM  
& .696 (.020) 
& .694 (.048) 
& .642 (.032) 
& .724 (.023) 
& .716 (.008) 
& .684 (.026) 
& \cellcolor{blue!20}\textbf{.735 (.030)} 
& .596 (.009) & .602 (.024) & .509 (.023) & .626 (.016) & \cellcolor{green!20}.669 (.034)& .447 (.047) & .118 (.018) & .642 (.012) & .552 (.013) \\

\THAILAND~BoT 
& .717 (.046) 
& .696 (.064) 
& .723 (.021) 
& .735 (.057) 
& .733 (.071) 
& .728 (.006) 
& \cellcolor{blue!20}\textbf{.741 (.017)}
& .547 (.004) & .549 (.032) & .551 (.039) & .573 (.012) & .581 (.029) & .484 (.038) & .258 (.026) & \cellcolor{green!20}.596 (.009)& .577 (.030) \\

\TAIWAN~CBCT 
& .641 (.049) 
& .637 (.031) 
& .635 (.029) 
& .667 (.015) 
& .616 (.035) 
& .678 (.032) 
& \cellcolor{blue!20}\textbf{.688 (.024)}
& .451 (.044) & .474 (.044) & .474 (.031) & .485 (.026) & .522 (.022) & .388 (.049) & .180 (.037) & \cellcolor{green!20}.556 (.015)& .475 (.051) \\

\EGYPT~CBE   
& .773 (.027) 
& .783 (.026) 
& .790 (.004) 
& .810 (.016) 
& .822 (.015) 
& .788 (.029) 
& \cellcolor{blue!20}\textbf{.836 (.014)}
& .629 (.037) & .672 (.036) & .581 (.045) & .648 (.036) & .636 (.007) & .352 (.056) & .142 (.014) & \cellcolor{green!20}.702 (.021)& .594 (.024) \\

\RUSSIA~CBR  
& .763 (.031) 
& .754 (.017) 
& .750 (.020) 
& .798 (.032) 
& .811 (.023) 
& .779 (.022) 
& \cellcolor{blue!20}\textbf{.835 (.011)}
& .759 (.015) & .749 (.027) & .693 (.026) & .701 (.049) & \cellcolor{green!20}.794 (.028)& .573 (.035) & .146 (.022) & .772 (.029) & .665 (.013) \\

\TURKEY~CBRT 
& .717 (.015) 
& .743 (.016) 
& .724 (.012) 
& .746 (.011) 
& .746 (.014) 
& .724 (.032) 
& \cellcolor{blue!20}\textbf{.762 (.027)}
& .495 (.006) & .421 (.014) & .424 (.018) & .475 (.015) & .539 (.030) & .277 (.006) & .133 (.020) & \cellcolor{green!20}.653 (.036)& .416 (.032) \\

\CHILE~CBoC  
& .760 (.048) 
& .747 (.006) 
& .743 (.054) 
& .779 (.049) 
& .792 (.043) 
& .793 (.058) 
& \cellcolor{blue!20}\textbf{.799 (.037)}
& .668 (.032) & .604 (.027) & .605 (.033) & .678 (.057) & .676 (.038) & .559 (.072) & .223 (.040) & \cellcolor{green!20}.685 (.019)& .539 (.071) \\

\EU~ECB      
& .707 (.040) 
& .699 (.048) 
& .668 (.030) 
& .724 (.050) 
& .724 (.014) 
& .713 (.022) 
& \cellcolor{blue!20}\textbf{.755 (.024)}
& .638 (.023) & .599 (.019) & .610 (.038) & \cellcolor{green!20}.660 (.021) & .637 (.005) & .548 (.017) & .206 (.052) & .613 (.020) & .595 (.016) \\

\USA~FOMC    
& .671 (.029) 
& .674 (.035) 
& .675 (.046) 
& .747 (.020) 
& .732 (.037) 
& .685 (.044) 
& \cellcolor{blue!20}\textbf{.749 (.047)}
& .572 (.021) & .584 (.025) & .564 (.018) & .649 (.023) & \cellcolor{green!20}.653 (.023) & .512 (.023) & .170 (.025) & .599 (.012) & .498 (.015) \\

\SINGAPORE~MAS  
& .656 (.049) 
& .681 (.043) 
& .666 (.005) 
& .666 (.066) 
& .698 (.042) 
& .680 (.049) 
& \cellcolor{blue!20}\textbf{.703 (.033)}
& .553 (.046) & .581 (.026) & .588 (.034) & .569 (.015) & \cellcolor{green!20}.689 (.041)& .540 (.026) & .347 (.024) & .638 (.035) & .646 (.023) \\

\POLAND~NBP  
& .685 (.016) 
& .690 (.013) 
& .705 (.026) 
& \cellcolor{blue!20}\textbf{.731 (.013) }
& .725 (.009) 
& .697 (.010) 
& .695 (.020)
& .637 (.015) & .631 (.043) & .614 (.063) & \cellcolor{green!20}.665 (.031) & .660 (.002) & .508 (.028) & .118 (.017) & .618 (.035) & .597 (.015) \\
\CHINA~PBoC  
& .791 (.033) 
& .763 (.008) 
& .742 (.046) 
& .787 (.026) 
& \cellcolor{blue!20}\textbf{.813 (.014)}
& .793 (.007) 
& .786 (.017)
& .492 (.046) & .559 (.037) & .531 (.026) & .535 (.033) & .592 (.037) & .379 (.017) & .128 (.018) & \cellcolor{green!20}.613 (.033)& .446 (.024) \\
\AUS~RBA     
& .685 (.023) 
& .681 (.019) 
& .682 (.027) 
& .672 (.015) 
& .707 (.029) 
& .695 (.024) 
& \cellcolor{blue!20}\textbf{.741 (.028)}
& .531 (.049) & .478 (.079) & .483 (.058) & .553 (.074) & .537 (.055) & .358 (.049) & .133 (.020) & \cellcolor{green!20}.614 (.034)& .495 (.057) \\

\INDIA~RBI   
& .604 (.037) 
& .649 (.035) 
& .653 (.041) 
& .628 (.034) 
& .633 (.032) 
& .655 (.039) 
& \cellcolor{blue!20}\textbf{.668 (.044)}
 & .489 (.025) & .519 (.041) & .509 (.026) & .495 (.027) & .542 (.016) & .431 (.008) & .231 (.058) & \cellcolor{green!20}.581 (.030)& .557 (.032) \\

\SWITZERLAND~SNB & .692 (.017) & .685 (.028) & .653 (.016) & .685 (.009) & \cellcolor{blue!20}\textbf{.725 (.018)} & .698 (.030) & .713 (.020) & .635 (.003) & .601 (.015) & .640 (.037) & \cellcolor{green!20}.652 (.016) & .643 (.024) & .554 (.029) & .252 (.008) & .612 (.014) & .607 (.051) \\
\midrule
~Average
& .702 (.030) 
& .695 (.028)
& .691 (.025)
& .714 (.025) 
& .723 (.026) 
& .710 (.025) 
& \cellcolor{blue!20}\textbf{.740 (.024)}
& .562 (.085) & .556 (.084) & .542 (.075) & .586 (.078) & .601 (.075) & .449 (.083) & .196 (.071) & \cellcolor{green!20}.620 (.053)& .541 (.074) \\
\bottomrule
\end{tabularx}
\end{table}

\begin{figure}[h!]
    \centering
    \includegraphics[width=\linewidth]{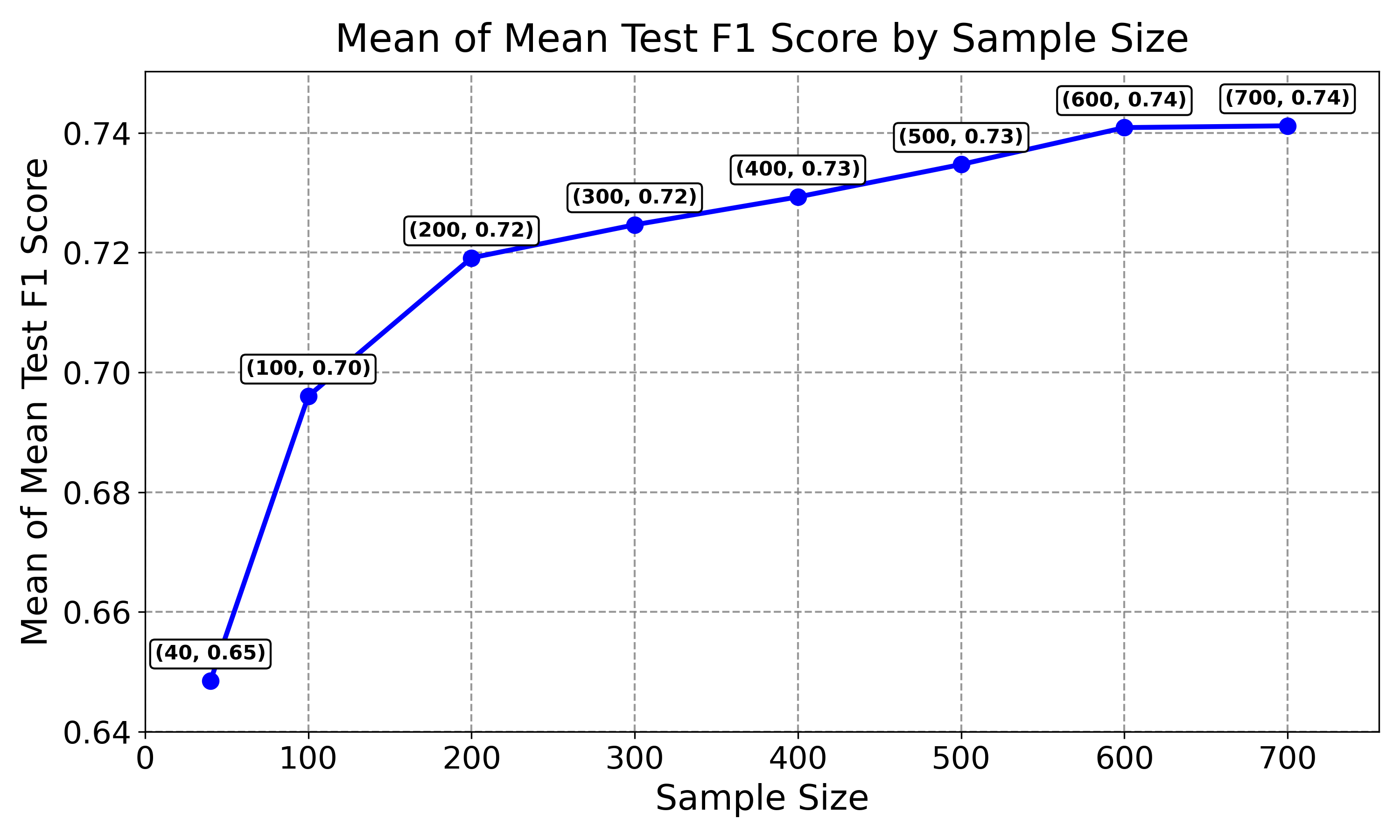}
    \caption{Impact of varying training sample sizes on average test F1-Score of the best performing model using the entire annotated dataset. Performance rapidly improves until 600 samples, after which performance starts to plateau. We first average the results over the 3 random seeds for each bank, and then take the mean of those averages across all banks.}
    \label{fig:size_ablation}
\end{figure}

\subsection{Model Performance Analysis}
\label{sec:modelanalysismain}

\paragraph{PLM Performance Analysis} For PLMs in the General Setup for \texttt{Stance Detection}, we observe superior performance from \texttt{RoBERTa-Large}, achieving an average weighted F1-Score of 0.740. Moreover, larger models consistently outperform their base model counterparts.

Interestingly, \texttt{ModernBERT-Large} does not emerge as the best-performing model for most banks. It only achieves the highest performance for four banks. This suggests that, for specific domains, it may not generalize well, and \texttt{RoBERTa} models (Base and Large) offer greater performance for this particular task. Another significant observation is the relatively poor performance of \texttt{FinBERT-Pretrain}, despite being fine-tuned on financial data. Its lower than average weighted F1-Score indicates that our \SD task involves complexities beyond the recognition of financial jargon, highlighting substantial differences from traditional sentiment analysis tasks and emphasizing the nuanced linguistic complexities inherent to monetary policy communications.

\paragraph{LLM Performance Analysis (Zero-Shot prompting)}
Using Zero-Shot prompting, \texttt{Llama-3-70B} outperforms all other LLMs, achieving an average weighted F1-Score of 0.620 on the \texttt{Stance Detection} task. In contrast, \texttt{Llama-4-Scout} performs worse across all banks, showing limited domain adaptation in this context. \texttt{GPT-4.1} and \texttt{DeepSeek-V3} perform the best for a combined total of eight banks, while \texttt{Gemini-2.0-Flash} achieves top performance on one bank. The domain-specific model \texttt{FinMA} has the lowest average performance across all LLMs. We attribute this to the absence of monetary policy-specific data in FinMA's training corpus. For \SD, Table~\ref{tab:plm_f1_stance_all} clearly shows that PLMs consistently outperform LLMs using Zero-Shot prompting on average.

\section{Additional Experiments}
\label{sec:additionalExperiments}
\paragraph {Performance Gain Analysis: The whole is greater than the sum of its parts}
To understand why the best-performing General Setup model consistently outperforms the best-performing Bank-Specific Setup models for the \SD task, we conduct qualitative and statistical analyses. For the qualitative analysis, we focus on test-set sentences that were misclassified by the best Bank-Specific model but correctly classified by the General model (\texttt{RoBERTa-Large}). For each bank, we sample such sentences, compute their sentence embeddings using \texttt{voyage-finance-2} model \cite{voyage-finance-2}, and find the most semantically similar sentence in the training corpus. Representative examples  within Table~\ref{tab:error_examples} Appendix~\ref{app:qual_gain} show that the General Setup model benefits from semantically related sentences across banks, allowing it to learn from a much larger and diverse sample. 

For the statistical analysis, we compute TF-IDF vectors for each document and calculate cross-bank cosine similarity at a document level. Regressing the change in weighted F1-Score (General Setup – Specific Setup) for \texttt{RoBERTa-Large} on normalized average similarity reveals a statistically significant relationship ($p = 0.016$), as shown in Appendix~\ref{app:statistical_gain}. This confirms that semantic similarity across banks correlates with larger gains from the General Setup model. Together, these results (discussed in Appendix~\ref{app:whole_perf_gain}) confirm that the General Setup model leverages cross-bank semantic patterns to improve performance. This implies that the more a bank’s language resembles that of others, the more it benefits from a model trained on the aggregated dataset. This highlights the value of shared linguistic structure in monetary policy communication.

\paragraph{Error Analysis}
We conduct error analysis on the annotated sentences from People's Bank of China, focusing on the the Zero-Shot output from the best performing LLM (\texttt{Llama-3-70B-Chat}) on the three tasks. We select this central bank for two reasons: (1) it represents one of the world’s largest banks in terms of assets \citep{SWF_CentralBank_2025} and (2) we find that while the overall policy tone is generally dovish, many statements appear linguistically neutral, making \SD particularly challenging. We examine the subset of sentences that were misclassified by \texttt{Llama-3-70B-Chat}, aiming to characterize common errors in judgment. The split of selected sentences, types of errors, analysis procedure, and detailed results are provided in Appendix~\ref{app:china_error}.

\paragraph{Benchmarking with Few-Shot and prompting with Annotation Guide}
In Section~\ref{sec:modelanalysismain}, LLMs are evaluated using basic Zero-Shot prompting. To analyze the impact of sophisticated prompting strategies on LLM performance across all three tasks, we further evaluate the best-performing LLM on average (\texttt{Llama-3-70B-Chat}) using two advanced prompting strategies: Few-Shot prompting and prompting with Annotation Guide. The detailed procedures and prompt templates for Few-Shot prompting are provided in \ref{app:few_shot_prompting_structure} and for prompting with Annotation Guide are provided in Appendix~\ref{app:annotation_shot_prompting_structure}. As reported in Table~\ref{tab:Ann_fewshot} Appendix~\ref{app:results}, both prompting strategies result in marginal improvements on the \SD task but lead to performance declines on \TC and \CE. A detailed breakdown of results is provided in Appendix~\ref{app:results}.

\paragraph{Generating Meeting Minutes Documents Using an LLM}
To extend our analysis beyond the annotated dataset, we generate meeting minutes for a target document whose release date is beyond the knowledge cutoff of May $31^{\text{st}}$, 2024. We use \texttt{GPT-4.1}, selected for its one million token context window and strong performance relative to other models with sufficiently large context windows on the \SD task. Each generation includes example inputs, an example output and actual context inputs as defined in Appendix~\ref{app:gen_mm_method}; due to the frequency and timing of meetings which may differ across various central banks, some documents may overlap across the example and actual input sets. We evaluate the generated minutes by computing their hawkishness scores, as defined in Section~\ref{sec:economic_analysis}, and compare them to the hawkishness measures of the corresponding target documents labeled by our best-performing fine-tuned \SD model. This quantifies alignment with real-world policy tone and demonstrates a novel use case for our fine-tuned models: as evaluators of generated policy content. The full generation procedure is discussed in \ref{app:gen_mm_method} and associated results are presented in Appendix~\ref{app:gen_mm_hawkishness}.
\paragraph{Human Evaluation}
To benchmark against human performance, we conduct an evaluation on the European Central Bank's (ECB) communications, selected due to its direct influence on 20 member nations. We compare the best-performing LLM for each task:  \texttt{Llama-3-70B-Chat} for \texttt{Temporal Classification} and \texttt{Uncertainty Estimation} and \texttt{GPT 4.1} for \texttt{Stance Detection} against a human annotator without access to the annotation guide. Using 50 samples per task, the LLM consistently outperforms the human across all three tasks, despite being evaluated with Zero-Shot prompting. The methodology for this experiment as well as results are detailed in Appendix~\ref{app:HumanEval}.

\section{Economic Analysis}
\label{sec:economic_analysis}
To understand the impact that the stance of a central bank's communications can have on an economy, we inference our best performing model for \SD (\texttt{RoBERTa-Large}) on our corpus of over 380k sentences and generate the following measure of hawkishness \citet{shah-etal-2023-trillion} at a document level and compare it to the inflation within the specific central bank's economy.
\begin{align}
    \text{Hawkishness Measure} &= \frac{\text{\# Hawkish Sentences}-\text{\# Dovish Sentences}}{\text{\# Total Sentences} - \text{\# Irrelevant Sentences}}
\end{align}
This was computed for each document associated with every central bank. As seen in Appendix~\ref{app:ext_econ_analysis}, we then plot the change of hawkishness measure over time alongside the percentage change in Consumer Price Index (CPI) for each economy to determine if our measure aligns with individual economies' inflation. Building on the findings of \citet{shah-etal-2023-trillion, kumar2024wordsmarketsquantifyingimpact}, our analysis shows that the hawkishness measures derived from monetary policy documents closely track inflation trends within each economy, particularly during periods of financial distress such as the 2001 and 2008 crises and the COVID-19 pandemic as seen in Appendix~\ref{app:ext_econ_analysis}.

\section{Related Works}
\label{sec:related_works}
\begin{table}[h]
\centering
\scriptsize
\caption{{Comparison of related datasets. The \textit{Corpus} is the aggregated data size and the number of DP (\textit{Data Points}) is the annotated dataset size. CB (\textit{Central Banks}) is the number of central banks included in the data set. \textit{Data} is the content which was analyzed in the paper. S (\textit{Stance}), T (\textit{Temporal}), U (\textit{Uncertainty})
 specify if the corresponding label is present. AG (\textit{Annotation Guide}) details if an annotation guide was used and is provided.}}
\label{tab:dataset_comp_table}
\begin{tabular}{
    p{0.195\textwidth}  
    p{0.09\textwidth}  
    p{0.045\textwidth}  
    p{0.035\textwidth}  
    p{0.01\textwidth}  
    p{0.3\textwidth}  
    p{0.005\textwidth}  
    p{0.005\textwidth}  
    p{0.005\textwidth}  
    p{0.01\textwidth}  
}
\toprule
\textbf{Paper} & \textbf{Year Range} & \textbf{Corpus} & \textbf{DP} &
\textbf{CB} & \textbf{Data} & \textbf{S} & \textbf{T} &
\textbf{U} & \textbf{AG} \\
\midrule
\citet{DVN/23957-2013} & 1995–2012 & 2,544 & 2,544 & 6 &
Policy Transcripts & $\times$  & $\times$  & $\times$  & $\times$  \\
\citet{ARMELIUS2020102116} & 2002–2017 & 12,024 & -- & 23 &
Speeches, Transcripts & \checkmark & $\times$  & $\times$  & $\times$  \\
\citet{mathur2022monopoly} & 2009–2022 & 24,180 & 24,180 & 6 &
MP Conferences & \checkmark & $\times$  & $\times$  & $\times$  \\
\citet{Kirti2022COVIDPolicies} & 2020 & 5,462 & 5,462 & 74 &
Gov.\ Websites, MP Reports, News & $\times$  & $\times$  & $\times$  & $\times$  \\
\citet{shah-etal-2023-trillion} & 1996–2022 & 216,926 & 7,440 & 1 &
FOMC Minutes, Speeches, Conferences & \checkmark & $\times$  & $\times$  & \checkmark \\
\citet{Bolhuis2024New} & 2000–2022 & 3,545 & 3,545 & 20 &
MP Announcements, Finance Data & $\times$  & $\times$  & $\times$  & $\times$  \\
\citet{barigozzi2024large} & 2000–2024 & 800 & 800 & 1 &
National Accounts & $\times$  & $\times$  & $\times$  & $\times$  \\
\citet{zhang2025camefcausalaugmentedmultimodalityeventdriven} & 2008–2024 & 25,690 & -- & 0 &
MacroEcon Announcements, Finance data & $\times$  & $\times$  & $\times$  & $\times$  \\
\midrule
\textbf{WCB (Ours)} & \textbf{1996–2024} & \textbf{2,661,400} & \textbf{100,000} & \textbf{25} &
\textbf{Monetary Policy Minutes }& \textbf{\gcheck} & \textbf{\gcheck} & \textbf{\gcheck} & \textbf{\gcheck} \\
\bottomrule
\end{tabular}
\end{table}

Over the last two decades, Natural Language Processing (NLP) has significantly impacted fields like finance and economics \citep{tatarinov2025languagemodelingfuturefinance, ehrmann2007communication}. In finance, developments such as specialized financial sentiment lexicons \citep{LoughranMcDonald2010} have enabled detailed analyses of textual datasets, facilitating research on monetary policy, \texttt{Uncertainty Estimation} \citep{10.1257/jel.20181020, baker2016measuring} and \texttt{Temporal Classification} \citep{hansen2019long}, and monetary policy \texttt{Stance Detection} \citep{shah-etal-2023-trillion}. 

Central bank communications, including press conferences, speeches, minutes, and economic news, have been analyzed using NLP \citep{10.1257/aer.20220129, barbaglia2022forecastingeconomicnews, AHRENS2024105921, ApelGrimaldi+2014+53+76, ehrmann2007communication, shah-etal-2023-trillion,pardawala2025subjectiveqameasuringsubjectivityearnings}. These studies help identify monetary policy shocks \citep{aruoba2024identifying}, assess sentiment \citep{hilscher2024information}, and explore the implications of transparency on market expectations \citep{10.1257/aer.20220129}. Recent works leverage transformer-based models like \texttt{BERT} and \texttt{RoBERTa} \citep{shah-etal-2023-trillion, gambacorta2024cb} alongside methods like LDA \citep{jegadeesh2017deciphering}, word embeddings \citep{matsui-etal-2021-using}, and graph clustering \citep{zhou2017graph}. 

Current analyses often focus on individual central banks such as RBI, FOMC, and ECB, providing localized insights \citep{kumar2024wordsmarketsquantifyingimpact, TADLE2022106021, tobback2017between, KAZINNIK2021308, vega2020assessing, ROZKRUT2007176, oshima2018monetary, iglesias2017emerging, aguilar2022communication}. This approach, however, limits the understanding of broader global trends \citep{ehrmann2020starting, WONG2011432, ARMELIUS2020102116}. Our study fills this gap by analyzing texts from 25 central banks across multiple economic crises, offering comprehensive global insights. Table \ref{tab:dataset_comp_table} shows that our dataset represents the largest monetary policy dataset created to date. Additional related work and detailed dataset comparisons are available in Table \ref{tab:dataset_comp_table_extra} Appendix~\ref{app:extended_related_works}.

\section{Discussion}
\label{sec:discussion}
To the best of our knowledge, this is the first dataset in the finance and economics domain covering 26 central banks across six continents with 25k annotated sentences (more than 380k sentences total), structured metadata (Appendix~\ref{app:Metadata}) and three core tasks. We provide detailed, bank-specific annotation guides (Appendix~\ref{app:annotation_guides}), annotator instructions (Appendix~\ref{app:instructions}), and documentation of the annotation process (Section~\ref{par:annotation}) to facilitate reproducibility. Beyond being a benchmark, our work enables rich downstream analysis: improved performance via generalization across banks (Appendix~\ref{app:whole_perf_gain}), meeting minutes' generation using LLMs (Appendix~\ref{app:gen_mm}), economic analysis (Appendix~\ref{app:ext_econ_analysis}), human evaluation (Appendix~\ref{app:HumanEval}), and error analysis (Appendix~\ref{app:china_error}). We also show that fine-tuned PLMs outperform LLMs on all three tasks, even with Few-Shot or Annotation-Guide prompting (Appendix~\ref{app:few_shot_ann_guide}).

\paragraph{Global Coverage Gaps}
Despite our efforts to ensure wide geographical representation, the broader data ecosystem remains skewed. As highlighted by \citet{longpre2024bridging}, over 93\% of textual data (tokens) still originates from North American and European institutions, while data from Africa and South America accounts for less than 0.2\% of content. This persistent imbalance limits the robustness and fairness of language technologies. Our dataset takes a step toward bridging this gap, but more work is needed to balance representation.

\paragraph{Broader Impacts}
To demonstrate broader applicability of our work \textbf{beyond finance}, we evaluate the \texttt{Temporal Classification} and \texttt{Uncertainty Estimation} tasks on U.S. Congressional Committee hearings. Shown in Appendix~\ref{app:esg_transfer_learning}, our best-performing PLM achieves F1-Scores of 0.879 and 0.683, respectively, indicating strong generalization. Additionally, to assess transferability \textbf{beyond the chosen 25 central banks' data}, we evaluate the \texttt{Stance Detection} task on cleaned data from the 26th bank, Czech National Bank, and observe a mean weighted F1-Score of 0.800, as detailed in Appendix~\ref{app:PLM_Czech}. This confirms that our models retain high performance even on out-of-sample bank's data.

\paragraph{Open Access and Expansion}
We currently cover 26 central banks and initiate this project along with its publicly available Hugging Face repository and GitHub repository as an open research effort. Our goal is to expand beyond the current 26 central banks through community collaboration and lower entry barriers, especially for independent researchers. To support this, we also provide cleaned data for two additional central banks (Appendix \ref{app:Banks_excl}) along with a reproducible pipeline that enables others to build upon and extend our work.
\section*{Acknowledgments and Disclosure of Funding}
\label{app:Acknowledgments}
We would like to thank all the annotators who annotated the sentences and contributed to initial drafts of the annotation guides. We also appreciate the Georgia Tech Library for giving us access to the Bloomberg Terminal, Partnership for an Advanced Computing Environment (PACE) at Georgia Tech, and especially Label Studio. 

\section*{Disclaimer}
\label{app:disclaimer}
We provide the following disclaimers regarding our work. 

\textbf{Geopolitical Disclaimer} All references to country names, territories, flags, boundaries, or other geopolitical identifiers in this work are drawn solely from publicly available sources, principally datasets and publications produced by international organizations such as the International Monetary Fund (IMF), the World Bank, and the United Nations. Their inclusion is solely for descriptive or analytical purposes.

Neither the authors nor their affiliated institution(s):
\begin{itemize}
    \item Endorse or oppose any political stance, territorial claim, or status implied by these identifiers;
    \item Express any opinion concerning the legal status of any country, territory, city, or area, or the delimitation of its frontiers or boundaries;
    \item Imply recognition, approval, or disapproval of the policies, positions, or governments of the entities mentioned.
\end{itemize}

\textbf{Financial Disclaimer} The analyses and interpretations presented in this paper are provided only for academic discussion. They do not constitute official economic policy advice, legal guidance, or a recommendation to implement specific actions. Readers should exercise their own judgment before making decisions based on the information herein.

\textbf{Data Disclaimer} The datasets and their subsets utilized in this study were specifically collected and annotated for academic research and educational purposes. Over 15,000 experiments were conducted on these datasets, leading to insights that we believe are valuable to the deep learning community. Readers who wish to explore extended or specialized applications of these datasets are encouraged to contact the corresponding authors for further information. 

\bibliographystyle{plainnat}
{
\clearpage
\small
\bibliography{main}
}

\clearpage
\section*{NeurIPS Paper Checklist}
\begin{enumerate}

\item {\bf Claims}
    \item[] Question: Do the main claims made in the abstract and introduction accurately reflect the paper's contributions and scope?
    \item[] Answer: \answerYes{} 
    \item[] Justification: Abstract (on page 1) and introduction (Section 1) accurately reflect the paper's contributions and scope. 

\item {\bf Limitations}
    \item[] Question: Does the paper discuss the limitations of the work performed by the authors?
    \item[] Answer: \answerYes{} 
    \item[] Justification: In Section \ref{sec:discussion} and Section \ref{app:ethicalconsideration}. 

\item {\bf Theory assumptions and proofs}
    \item[] Question: For each theoretical result, does the paper provide the full set of assumptions and a complete (and correct) proof?
    \item[] Answer: \answerNA{} 
    \item[] Justification: The paper is submitted to the Dataset and Benchmark Track, and we have conducted extensive experiments to back any claim we make. 

    \item {\bf Experimental result reproducibility}
    \item[] Question: Does the paper fully disclose all the information needed to reproduce the main experimental results of the paper to the extent that it affects the main claims and/or conclusions of the paper (regardless of whether the code and data are provided or not)?
    \item[] Answer: \answerYes{} 
    \item[] Justification: We believe we do one of the most extensive experiments and reporting on experiments and the annotation process. The details on the models experimented with are provided in Section \ref{sec:models}, Appendix \ref{app:Model_abb}, and Appendix \ref{app:technical_details_benchmarking}. Prompts are provided in Appendix \ref{app:promptsbenchmarking}. Instructions given to annotators are provided in Appendix \ref{app:instructions}, while the interface used is provided in \ref{app:interface}. The complete list is too long to be included here, so we would like to direct you to the Table of Contents on Page 14. 

\item {\bf Open access to data and code}
    \item[] Question: Does the paper provide open access to the data and code, with sufficient instructions to faithfully reproduce the main experimental results, as described in supplemental material?
    \item[] Answer: \answerYes{} 
    \item[] Justification: Page 1 provides URLs to both our Code, Model, and Dataset pages. The Model and Dataset pages also provide code on how to use them. We go above and beyond to create a website for the non-technical audience to use our work. 

\item {\bf Experimental setting/details}
    \item[] Question: Does the paper specify all the training and test details (e.g., data splits, hyperparameters, how they were chosen, type of optimizer, etc.) necessary to understand the results?
    \item[] Answer: \answerYes{} 
    \item[] Justification: The details on dataset splits and other settings are provided in Section \ref{sec:models}, Appendix \ref{app:Model_abb}, and Appendix \ref{app:technical_details_benchmarking}. Prompts are provided in Appendix \ref{app:promptsbenchmarking}. Hyperparamter details are provided in Appendix \ref{app:hyperparams}. 

\item {\bf Experiment statistical significance}
    \item[] Question: Does the paper report error bars suitably and correctly defined or other appropriate information about the statistical significance of the experiments?
    \item[] Answer: \answerYes{} 
    \item[] Justification: Every experiment is conducted over multiple seeds, and we report the standard deviation of metrics to provide understanding of the statistical significance of the experiments. Additionally, whenever we run regression or correlation, we provide the p-value. 

\item {\bf Experiments compute resources}
    \item[] Question: For each experiment, does the paper provide sufficient information on the computer resources (type of compute workers, memory, time of execution) needed to reproduce the experiments?
    \item[] Answer: \answerYes{}{} 
    \item[] Justification: We provide extensive details on packages used, hardware used, and outside services used in Appendix \ref{app:technical_details_benchmarking}. 
    
\item {\bf Code of ethics}
    \item[] Question: Does the research conducted in the paper conform, in every respect, with the NeurIPS Code of Ethics \url{https://neurips.cc/public/EthicsGuidelines}?
    \item[] Answer: \answerYes{} 
    \item[] Justification: We provide an author statement in the Appendix \ref{app:authorstatement}, Disclaimer in Appendix \ref{app:disclaimer}, Ethical Consideration in \ref{app:ethicalconsideration}, and Scraping Policy Consideration in \ref{app:scraping_clause}. 

\item {\bf Broader impacts}
    \item[] Question: Does the paper discuss both potential positive societal impacts and negative societal impacts of the work performed?
    \item[] Answer: \answerYes{} 
    \item[] Justification: In Section \ref{sec:discussion}. 

\item {\bf Safeguards}
    \item[] Question: Does the paper describe safeguards that have been put in place for responsible release of data or models that have a high risk for misuse (e.g., pretrained language models, image generators, or scraped datasets)?
    \item[] Answer: \answerYes{} 
    \item[] Justification: We provide the dataset clause in Appendix \ref{app:datasetclause}. For models, we fine-tune existing models and cite them. 

\item {\bf Licenses for existing assets}
    \item[] Question: Are the creators or original owners of assets (e.g., code, data, models), used in the paper, properly credited and are the license and terms of use explicitly mentioned and properly respected?
    \item[] Answer: \answerYes{} 
    \item[] Justification: We cite all the models used in Section \ref{sec:models}. We collect the original new data with Scraping Policy Consideration in Appendix \ref{app:scraping_clause}. 

\item {\bf New assets}
    \item[] Question: Are new assets introduced in the paper well documented and is the documentation provided alongside the assets?
    \item[] Answer: \answerYes{} 
    \item[] Justification: Appendix \ref{app:hosting} provide details on this extensively. 

\item {\bf Crowdsourcing and research with human subjects}
    \item[] Question: For crowdsourcing experiments and research with human subjects, does the paper include the full text of instructions given to participants and screenshots, if applicable, as well as details about compensation (if any)? 
    \item[] Answer: \answerYes{} 
    \item[] Justification: Instructions given to annotators are provided in Appendix \ref{app:instructions}, while the interface used is provided in \ref{app:interface}. 

\item {\bf Institutional review board (IRB) approvals or equivalent for research with human subjects}
    \item[] Question: Does the paper describe potential risks incurred by study participants, whether such risks were disclosed to the subjects, and whether Institutional Review Board (IRB) approvals (or an equivalent approval/review based on the requirements of your country or institution) were obtained?
    \item[] Answer: \answerNA{} 
    \item[] Justification: We do not perform any study that requires it. 

\item {\bf Declaration of LLM usage}
    \item[] Question: Does the paper describe the usage of LLMs if it is an important, original, or non-standard component of the core methods in this research? Note that if the LLM is used only for writing, editing, or formatting purposes and does not impact the core methodology, scientific rigorousness, or originality of the research, declaration is not required.
    \item[] Answer: \answerNA{} 
    \item[] Justification: We only use LLM for performance benchmarking. 

\end{enumerate}

\clearpage
\appendix
\addtocontents{toc}{\protect\setcounter{tocdepth}{-1}}
\addcontentsline{toc}{section}{Appendix}
\addtocontents{toc}{\protect\setcounter{tocdepth}{3
}}
\phantomsection
\hypertarget{toc}{}
\tableofcontents

\clearpage
\section{Release and Responsibility Statements}

\subsection{Author Statement}
\label{app:authorstatement}
We confirm that the WCB dataset is licensed under the Creative Commons Attribution-NonCommercial-ShareAlike 4.0 International (\texttt{CC-BY-NC-SA 4.0}) license, which allows others to share, copy, distribute, and transmit the work, as well as to adapt the work, provided that appropriate credit is given for non-commercial use. Additionally, the derived work needs to be made available under a similar license. 
 
\subsection{Hosting, Licensing and Maintenance}
\label{app:hosting}

The \textbf{World Central Banks (WCB)} dataset is distributed under the \textit{Creative Commons Attribution-NonCommercial-ShareAlike 4.0 International} (CC-BY-NC-SA 4.0) license. This license allows users to share, adapt, and build upon the dataset for non-commercial purposes, provided proper attribution is given and derivative works are shared alike.

To ensure broad accessibility and reproducibility, we host the dataset, codebase, models, and visualizations across three major platforms:

\begin{center}
  \fbox{%
    \parbox{\textwidth}{%
      \begin{center}
        \textbf{The WCB dataset and resources are available at:}
        \vspace{0.6em}

        \textbf{GitHub}:\\
        \url{https://github.com/gtfintechlab/WorldCentralBanks}

        \vspace{0.8em}

        \textbf{Hugging Face Collection} (Models + Datasets):\\
        \url{https://huggingface.co/collections/gtfintechlab/wcb-678965e38178c63158b45fdf}

        \vspace{0.8em}

        \textbf{Kaggle Dataset:}\\
        \url{https://www.kaggle.com/datasets/gtfintechlab/wcb-dataset-annotated/data}

        \vspace{0.8em}

        \textbf{Website}:\\
        \url{https://gcb-web-bb21b.web.app/}
      \end{center}
    }
  }
\end{center}

\vspace{0.5em}
The Hugging Face collection includes:
\begin{itemize}
  \item \textbf{Four datasets}: 
  \begin{itemize}
    \item A unified corpus of 380,200 scraped and cleaned sentences 
    \item An annotated dataset of 25,000 sentences across three tasks 
    \item A Czech National Bank subset (CNB)
    \item Bank-specific subsets for 25 central banks
  \end{itemize}
  \item \textbf{78 trained models}:
  \begin{itemize}
    \item 75 bank-specific models (trained on 3 tasks for each of the 25 central banks)
    \item 3 models trained on the general dataset (trained on 3 tasks using the aggregated dataset of all central banks)
  \end{itemize}
\end{itemize}

The links to the 1k annotated sentences for all 25 central banks, their corresponding Bank-Specific Setup models, the aggregated 25k annotated sentences dataset, and the three General Setup models are provided in Table~\ref{tab:banks_data_models}.

The GitHub repository offers comprehensive access to:
\begin{itemize}
  \item All raw, cleaned, and sanitized data
  \item Structured metadata
  \item Complete benchmarking and training pipelines
  \item Instructions to reproduce results from benchmarking as well as additional experiments
\end{itemize}


Each folder contains data for each of the 25 banks. Furthermore, there is a global \textbf{Master Metadata JSON} file which is structured as seen in Figure \ref{fig:metadata} Appendix \ref{app:Metadata}. It is a large dictionary where the doc\_id is mapped to the unique documents within a specific bank for a specific year. More information on the metadata is reported in \ref{app:Metadata}.

The WCB website complements these resources with dynamic dashboards showcasing model performance and interactive visualizations.

\begin{table}[h!]
\centering
\caption{{The datasets for each central banks and best performing models for each task (\texttt{Stance Detection, Temporal Classification}, and \texttt{Uncertainty Estimation}) are available hyperlinked in the table below. The specific hyperparameter used for each model are summarized in Tables \ref{tab:best_hypers_certain},
\ref{tab:best_hypers_stance}, \ref{tab:best_hypers_time} in Appendix \ref{app:hyperparams}.}}
\label{tab:banks_data_models}
\begin{tabular}{l l c c c}
\toprule
& & \multicolumn{3}{c}{\textbf{Model}} \\
\cmidrule(lr){3-5}
\textbf{Bank} & \textbf{Dataset} & \textbf{Stance} & \textbf{Temporal} & \textbf{Uncertainty} \\
\midrule
General Setup
    & \href{https://huggingface.co/datasets/gtfintechlab/all_annotated_sentences_25000}{Dataset}
    & \href{https://huggingface.co/gtfintechlab/model_WCB_stance_label}{Model}
    & \href{https://huggingface.co/gtfintechlab/model_WCB_time_label}{Model}
    & \href{https://huggingface.co/gtfintechlab/model_WCB_certain_label}{Model} \\

\USA Federal Open Market Committee
    & \href{https://huggingface.co/datasets/gtfintechlab/fomc_communication}{Dataset}
    & \href{https://huggingface.co/gtfintechlab/model_federal_reserve_system_stance_label}{Model}
    & \href{https://huggingface.co/gtfintechlab/model_federal_reserve_system_time_label}{Model}
    & \href{https://huggingface.co/gtfintechlab/model_federal_reserve_system_certain_label}{Model} \\

\CHINA People's Bank of China
    & \href{https://huggingface.co/datasets/gtfintechlab/peoples_bank_of_china}{Dataset}
    & \href{https://huggingface.co/gtfintechlab/model_peoples_bank_of_china_stance_label}{Model}
    & \href{https://huggingface.co/gtfintechlab/model_peoples_bank_of_china_time_label}{Model}
    & \href{https://huggingface.co/gtfintechlab/model_peoples_bank_of_china_certain_label}{Model} \\

\JAPAN Bank of Japan
    & \href{https://huggingface.co/datasets/gtfintechlab/bank_of_japan}{Dataset}
    & \href{https://huggingface.co/gtfintechlab/model_bank_of_japan_stance_label}{Model}
    & \href{https://huggingface.co/gtfintechlab/model_bank_of_japan_time_label}{Model}
    & \href{https://huggingface.co/gtfintechlab/model_bank_of_japan_certain_label}{Model} \\

\UK Bank of England
    & \href{https://huggingface.co/datasets/gtfintechlab/bank_of_england}{Dataset}
    & \href{https://huggingface.co/gtfintechlab/model_bank_of_england_stance_label}{Model}
    & \href{https://huggingface.co/gtfintechlab/model_bank_of_england_time_label}{Model}
    & \href{https://huggingface.co/gtfintechlab/model_bank_of_england_certain_label}{Model} \\

\SWITZERLAND Swiss National Bank
    & \href{https://huggingface.co/datasets/gtfintechlab/swiss_national_bank}{Dataset}
    & \href{https://huggingface.co/gtfintechlab/model_swiss_national_bank_stance_label}{Model}
    & \href{https://huggingface.co/gtfintechlab/model_swiss_national_bank_time_label}{Model}
    & \href{https://huggingface.co/gtfintechlab/model_swiss_national_bank_certain_label}{Model} \\

\BRAZIL Central Bank of Brazil
    & \href{https://huggingface.co/datasets/gtfintechlab/central_bank_of_brazil}{Dataset}
    & \href{https://huggingface.co/gtfintechlab/model_central_bank_of_brazil_stance_label}{Model}
    & \href{https://huggingface.co/gtfintechlab/model_central_bank_of_brazil_time_label}{Model}
    & \href{https://huggingface.co/gtfintechlab/model_central_bank_of_brazil_certain_label}{Model} \\

\INDIA Reserve Bank of India
    & \href{https://huggingface.co/datasets/gtfintechlab/reserve_bank_of_india}{Dataset}
    & \href{https://huggingface.co/gtfintechlab/model_reserve_bank_of_india_stance_label}{Model}
    & \href{https://huggingface.co/gtfintechlab/model_reserve_bank_of_india_time_label}{Model}
    & \href{https://huggingface.co/gtfintechlab/model_reserve_bank_of_india_certain_label}{Model} \\

\EU European Central Bank
    & \href{https://huggingface.co/datasets/gtfintechlab/european_central_bank}{Dataset}
    & \href{https://huggingface.co/gtfintechlab/model_european_central_bank_stance_label}{Model}
    & \href{https://huggingface.co/gtfintechlab/model_european_central_bank_time_label}{Model}
    & \href{https://huggingface.co/gtfintechlab/model_european_central_bank_certain_label}{Model} \\

\RUSSIA Central Bank of the Russian Federation
    & \href{https://huggingface.co/datasets/gtfintechlab/central_bank_of_the_russian_federation}{Dataset}
    & \href{https://huggingface.co/gtfintechlab/model_central_bank_of_the_russian_federation_stance_label}{Model}
    & \href{https://huggingface.co/gtfintechlab/model_central_bank_of_the_russian_federation_time_label}{Model}
    & \href{https://huggingface.co/gtfintechlab/model_central_bank_of_the_russian_federation_certain_label}{Model} \\

\TAIWAN Central Bank of China
    & \href{https://huggingface.co/datasets/gtfintechlab/central_bank_of_china_taiwan}{Dataset}
    & \href{https://huggingface.co/gtfintechlab/model_central_bank_of_china_taiwan_stance_label}{Model}
    & \href{https://huggingface.co/gtfintechlab/model_central_bank_of_china_taiwan_time_label}{Model}
    & \href{https://huggingface.co/gtfintechlab/model_central_bank_of_china_taiwan_certain_label}{Model} \\

\SINGAPORE Monetary Authority of Singapore
    & \href{https://huggingface.co/datasets/gtfintechlab/monetary_authority_of_singapore}{Dataset}
    & \href{https://huggingface.co/gtfintechlab/model_monetary_authority_of_singapore_stance_label}{Model}
    & \href{https://huggingface.co/gtfintechlab/model_monetary_authority_of_singapore_time_label}{Model}
    & \href{https://huggingface.co/gtfintechlab/model_monetary_authority_of_singapore_certain_label}{Model} \\

\SOUTHKOREA Bank of Korea
    & \href{https://huggingface.co/datasets/gtfintechlab/bank_of_korea}{Dataset}
    & \href{https://huggingface.co/gtfintechlab/model_bank_of_korea_stance_label}{Model}
    & \href{https://huggingface.co/gtfintechlab/model_bank_of_korea_time_label}{Model}
    & \href{https://huggingface.co/gtfintechlab/model_bank_of_korea_certain_label}{Model} \\

\AUS Reserve Bank of Australia
    & \href{https://huggingface.co/datasets/gtfintechlab/reserve_bank_of_australia}{Dataset}
    & \href{https://huggingface.co/gtfintechlab/model_reserve_bank_of_australia_stance_label}{Model}
    & \href{https://huggingface.co/gtfintechlab/model_reserve_bank_of_australia_time_label}{Model}
    & \href{https://huggingface.co/gtfintechlab/model_reserve_bank_of_australia_certain_label}{Model} \\

\ISRAEL Bank of Israel
    & \href{https://huggingface.co/datasets/gtfintechlab/bank_of_israel}{Dataset}
    & \href{https://huggingface.co/gtfintechlab/model_bank_of_israel_stance_label}{Model}
    & \href{https://huggingface.co/gtfintechlab/model_bank_of_israel_time_label}{Model}
    & \href{https://huggingface.co/gtfintechlab/model_bank_of_israel_certain_label}{Model} \\

\CANADA Bank of Canada
    & \href{https://huggingface.co/datasets/gtfintechlab/bank_of_canada}{Dataset}
    & \href{https://huggingface.co/gtfintechlab/model_bank_of_canada_stance_label}{Model}
    & \href{https://huggingface.co/gtfintechlab/model_bank_of_canada_time_label}{Model}
    & \href{https://huggingface.co/gtfintechlab/model_bank_of_canada_certain_label}{Model} \\

\MEXICO Bank of Mexico
    & \href{https://huggingface.co/datasets/gtfintechlab/bank_of_mexico}{Dataset}
    & \href{https://huggingface.co/gtfintechlab/model_bank_of_mexico_stance_label}{Model}
    & \href{https://huggingface.co/gtfintechlab/model_bank_of_mexico_time_label}{Model}
    & \href{https://huggingface.co/gtfintechlab/model_bank_of_mexico_certain_label}{Model} \\

\POLAND National Bank of Poland
    & \href{https://huggingface.co/datasets/gtfintechlab/national_bank_of_poland}{Dataset}
    & \href{https://huggingface.co/gtfintechlab/model_national_bank_of_poland_stance_label}{Model}
    & \href{https://huggingface.co/gtfintechlab/model_national_bank_of_poland_time_label}{Model}
    & \href{https://huggingface.co/gtfintechlab/model_national_bank_of_poland_certain_label}{Model} \\

\TURKEY Central Bank of Turkey
    & \href{https://huggingface.co/datasets/gtfintechlab/central_bank_republic_of_turkey}{Dataset}
    & \href{https://huggingface.co/gtfintechlab/model_central_bank_republic_of_turkey_stance_label}{Model}
    & \href{https://huggingface.co/gtfintechlab/model_central_bank_republic_of_turkey_time_label}{Model}
    & \href{https://huggingface.co/gtfintechlab/model_central_bank_republic_of_turkey_certain_label}{Model} \\

\THAILAND Bank of Thailand
    & \href{https://huggingface.co/datasets/gtfintechlab/bank_of_thailand}{Dataset}
    & \href{https://huggingface.co/gtfintechlab/model_bank_of_thailand_stance_label}{Model}
    & \href{https://huggingface.co/gtfintechlab/model_bank_of_thailand_time_label}{Model}
    & \href{https://huggingface.co/gtfintechlab/model_bank_of_thailand_certain_label}{Model} \\

\EGYPT Central Bank of Egypt
    & \href{https://huggingface.co/datasets/gtfintechlab/central_bank_of_egypt}{Dataset}
    & \href{https://huggingface.co/gtfintechlab/model_central_bank_of_egypt_stance_label}{Model}
    & \href{https://huggingface.co/gtfintechlab/model_central_bank_of_egypt_time_label}{Model}
    & \href{https://huggingface.co/gtfintechlab/model_central_bank_of_egypt_certain_label}{Model} \\

\MALAYSIA Bank Negara Malaysia
    & \href{https://huggingface.co/datasets/gtfintechlab/bank_negara_malaysia}{Dataset}
    & \href{https://huggingface.co/gtfintechlab/model_bank_negara_malaysia_stance_label}{Model}
    & \href{https://huggingface.co/gtfintechlab/model_bank_negara_malaysia_time_label}{Model}
    & \href{https://huggingface.co/gtfintechlab/model_bank_negara_malaysia_certain_label}{Model} \\

\PHILIPPINES Central Bank of the Philippines
    & \href{https://huggingface.co/datasets/gtfintechlab/central_bank_of_the_philippines}{Dataset}
    & \href{https://huggingface.co/gtfintechlab/model_central_bank_of_the_philippines_stance_label}{Model}
    & \href{https://huggingface.co/gtfintechlab/model_central_bank_of_the_philippines_time_label}{Model}
    & \href{https://huggingface.co/gtfintechlab/model_central_bank_of_the_philippines_certain_label}{Model} \\

\CHILE Central Bank of Chile
    & \href{https://huggingface.co/datasets/gtfintechlab/central_bank_of_chile}{Dataset}
    & \href{https://huggingface.co/gtfintechlab/model_central_bank_of_chile_stance_label}{Model}
    & \href{https://huggingface.co/gtfintechlab/model_central_bank_of_chile_time_label}{Model}
    & \href{https://huggingface.co/gtfintechlab/model_central_bank_of_chile_certain_label}{Model} \\

\PERU Central Reserve Bank of Peru
    & \href{https://huggingface.co/datasets/gtfintechlab/central_reserve_bank_of_peru}{Dataset}
    & \href{https://huggingface.co/gtfintechlab/model_central_reserve_bank_of_peru_stance_label}{Model}
    & \href{https://huggingface.co/gtfintechlab/model_central_reserve_bank_of_peru_time_label}{Model}
    & \href{https://huggingface.co/gtfintechlab/model_central_reserve_bank_of_peru_certain_label}{Model} \\

\COLOMBIA Bank of the Republic
    & \href{https://huggingface.co/datasets/gtfintechlab/bank_of_the_republic_colombia}{Dataset}
    & \href{https://huggingface.co/gtfintechlab/model_bank_of_the_republic_colombia_stance_label}{Model}
    & \href{https://huggingface.co/gtfintechlab/model_bank_of_the_republic_colombia_time_label}{Model}
    & \href{https://huggingface.co/gtfintechlab/model_bank_of_the_republic_colombia_certain_label}{Model} \\
\bottomrule
\end{tabular}
\end{table} 
\subsection{Ethical Considerations}
\label{app:ethicalconsideration}

In conducting our research, we adhere to strict ethical and legal policies and do not identify any risks associated with the research carried out. While our training process was designed to minimize the introduction of any bias, we acknowledge that residual bias may persist due to underlying biases in our data sources. Furthermore, we acknowledge that our research is subject to certain limitations and potential biases inherent in our annotation and study process. 

\begin{itemize}
    \item \textbf{Educational Bias} Most of the annotators came from STEM fields, and this could have impacted the annotation process.

    \item \textbf{Geographic Bias} NLP datasets often exhibit geographical biases, due to disproportionately sourced data from specific regions, cultural contexts, and languages \cite{tatarinov2025languagemodelingfuturefinance}. In our research, we sourced our data from each individual central bank's official website to minimize disproportion as much as possible. 
    
    \item \textbf{Data Ethics} All the names, data, and views are sourced exclusively from the official website of the central banks and the World Bank. All data were obtained ethically and legally. Some central banks restrict access based on geographical factors, so to align with ethical research practices, we did not collect data from those central banks. As a result, our study is limited to 25 central banks whose data can be ethically accessed. We also included an additional bank, the Czech National Bank, whose data was accessed ethically and utilized exclusively for ablation studies.
    
    \item \textbf{Annotation Ethics} The annotation of the dataset was completed ethically. All annotation work was completed on a voluntarily basis, and annotators were made aware of the tasks and purpose of the project in advance. None of the annotators were paid to do the annotations.
    
    \item \textbf{Publicly Available Data} All the data utilized for this research will be made publicly available, along with the appropriate licenses under which it may be shared.
    
    \item \textbf{Language Model Ethics} All the language models employed for this research are publicly available and we complied with their license categories for our intended purposes. To curb the carbon footprint associated with large‑scale pre‑training, we only fine‑tuned existing PLMs and ran prompt‑based evaluations on LLMs.  We cited all the models used.
    
    \item \textbf{Hyperparameter Reporting} All hyperparameter used in training the models are specified within Appendix \ref{app:hyperparams},
    thus making our model setup transparent and providing detailed information on our training process.
    
    \item \textbf{Model Ethics} Responsible AI practices guided our use of both PLMs and LLMs, including transparency, fairness, and reproducibility.
\end{itemize}

The research team is committed to promoting transparency, fairness, and accessibility by openly acknowledging any limitations in the findings, promoting ethical integrity, and encouraging responsible research practices.

\subsection{Dataset Clause and Terms of Use}
\label{app:datasetclause}

\textbf{Purpose} \quad The Dataset is provided for the purpose of research and educational use in the field of NLP, Machine Learning, and related areas; and can be used to further the development, evaluation, or benchmarking of AI.

\textbf{Usage Restriction} \quad Users of the Dataset should adhere to the terms of use for specific models or tasks. This includes respecting any limitations or use case prohibitions set forth by the original model’s and guide's creators.

\textbf{Content Warning} \quad The Dataset comprises of raw monetary policy filings from 25 central banks spanning from 1996 to 2024. These documents may reflect sensitive economic positions, institutional biases, or historically outdated policy perspectives. Therefore, users should exercise caution by applying appropriate filtering, contextual analysis, and moderation protocols when using this dataset for model training or decision-making tasks to ensure adherence to ethical standards.

\textbf{No Endorsement of Content} \quad The conversations and data within this Dataset do not reflect the views or opinions of the Dataset creators or any affiliated institutions. The dataset is provided as a neutral resource for financial and economic research and should not be construed as endorsing any specific viewpoints.

\textbf{Limitations of Liability} \quad The authors of this Dataset will not be liable for any claims, damages, or other liabilities arising from the use of the dataset, including but not limited to the misuse, interpretation, or reliance on any data contained within.

\subsection{Scraping Policy Consideration}
\label{app:scraping_clause}

The data gathered for this research were collected using legal and ethical practices. We have ensured that all of our datasets were gathered in full compliance with each institution's ethical requirements. Specifically, for each bank we reviewed their official Terms of Use or Disclaimer page on their websites and also inspected their site's \texttt{robots.txt} (if available) file to confirm that automated crawling of public press releases and meeting-minutes pages was permitted. Care was taken to respect access rules, avoid disallowed areas, and maintain server load limits through crawl delays. Therefore, our dataset respects each bank's data-use policy. As the central banks’ websites and policies are subject to change, we do not assume responsibility for any liabilities arising from their use.
\clearpage
\section{Extended Data Construction Process}
\label{app:Extended_data_construction}

\subsection{Acronyms for Bank Names}
We utilize the following acronyms within this paper in order to present our results in a clear and concise manner. The acronyms are based on the official abbreviation or standard usage. 
\label{app:acronyms}
\begin{table}[h!]
\caption{{List of central banks and their corresponding acronyms.}}
\label{tab:BankAcronyms}
\centering
\begin{tabular}{ll}
\toprule
\textbf{Bank Name} & \textbf{Acronym} \\
\midrule
\USA Federal Open Market Committee & FOMC \\
\CHINA People's Bank of China & PBoC \\
\JAPAN Bank of Japan & BoJ \\
\UK Bank of England & BoE \\
\SWITZERLAND Swiss National Bank & SNB \\
\BRAZIL Central Bank of Brazil & BCB \\
\INDIA Reserve Bank of India & RBI \\
\EU European Central Bank & ECB \\
\RUSSIA Central Bank of the Russian Federation & CBR \\
\TAIWAN Central Bank of China (Taiwan) & CBCT \\
\SINGAPORE Monetary Authority of Singapore & MAS \\
\SOUTHKOREA Bank of Korea & BoK \\
\AUS Reserve Bank of Australia & RBA \\
\ISRAEL Bank of Israel & BoI \\
\CANADA Bank of Canada & BoC \\
\MEXICO Bank of Mexico & BdeM \\
\POLAND Narodowy Bank Polski & NBP \\
\TURKEY Central Bank of Turkey & CBRT \\
\THAILAND Bank of Thailand & BoT \\
\EGYPT Central Bank of Egypt & CBE \\
\MALAYSIA Bank Negara Malaysia & BNM \\
\PHILIPPINES Central Bank of the Philippines & BSP \\
\CHILE Central Bank of Chile & CBoC \\
\PERU Central Reserve Bank of Peru & BCRP \\
\COLOMBIA Bank of the Republic & BanRep \\
\bottomrule
\end{tabular}
\end{table}

\subsection{Bank Selection}

\label{app:Banks_excl}
We had a particular set of criteria for the selection of the central banks within our corpus. We describe the reason for not selecting certain banks in Table \ref{tab:excluded_banks}. Banks/Economies that fall under the jurisdiction of the European Central Bank were excluded from this dataset as their monetary policy is governed by the ECB and they use a standardized currency (the Euro). We observed that certain documents from the Bank of Chile in 2022 were in Spanish with no official translation, and 2018 documents from the Central Bank of the Republic of Turkey were unavailable.

\paragraph{Community Involvement}
After extracting and cleaning the data, we select 28 central banks, including the 26 within our analysis and two additional banks: Sveriges Riksbank and Bank Indonesia. The Czech National Bank was used as a hold-out dataset, with only 500 sentences annotated for evaluation. Out of the corpus of cleaned data for 28 banks, we randomly sample the latter two banks (Sveriges Riksbank and Bank Indonesia) that we open to the community of AI, NLP, and economic researchers to analyze and annotate. This step aims to increase engagement of our work within the community.
\begin{table}[h!]
\small
\centering
\caption{{List of excluded central banks and reasons for their exclusion. Rows marked with * fall under the jurisdiction of the European Central Bank, as stated by \cite{ECBOurMoney}.}}
\label{tab:excluded_banks}
\begin{tabularx}{\textwidth}{
    p{0.37\textwidth}
    p{0.6\textwidth}
}
\toprule
\textbf{Bank} & \textbf{Reason for Exclusion} \\
\midrule
Bank of Finland                        & Bank falls under jurisdiction of European Central Bank* \\ 
Bank of France                         & Bank falls under jurisdiction of European Central Bank* \\ 
Bank of Greece                         & Bank falls under jurisdiction of European Central Bank* \\ 
Bank of Italy                          & Bank falls under jurisdiction of European Central Bank* \\ 
Bank of Portugal                       & Bank falls under jurisdiction of European Central Bank* \\ 
Bank of Spain                          & Bank falls under jurisdiction of European Central Bank* \\ 
Central Bank of Ireland                & Bank falls under jurisdiction of European Central Bank* \\ 
Central Bank of Luxembourg             & Bank falls under jurisdiction of European Central Bank* \\ 
Deutsche Bundesbank                    & Bank falls under jurisdiction of European Central Bank* \\ 
National Bank of Belgium               & Bank falls under jurisdiction of European Central Bank* \\ 
National Bank of the Republic of Austria & Bank falls under jurisdiction of European Central Bank* \\ 
Netherlands Bank                       & Bank falls under jurisdiction of European Central Bank* \\

Saudi Central Bank                     & Documents inaccessible at scale \\ 
Central Bank of Iran                   & Documents inaccessible at scale \\ 
Central Bank of Iraq                   & Documents inaccessible at scale \\

Norges Bank                            & Lack of consistent meeting minute level equivalent document \\ 
Central Bank of Lebanon               & Lack of consistent meeting minute level equivalent document \\ 
National Bank of Denmark              & Lack of consistent meeting minute level equivalent document \\ 
State Bank of Vietnam                  & Lack of consistent meeting minute level equivalent document \\ 
Qatar Central Bank                     & Lack of consistent meeting minute level equivalent document \\

Bank of Algeria                        & No meeting minute level document \\ 
Hong Kong Monetary Authority           & No meeting minute level document \\ 
Central Bank of Nigeria                & No meeting minute level document \\ 
Central Bank of UAE                    & No meeting minute level document \\

Central Bank of Argentina              & Insufficient number of documents \\ 
\bottomrule
\end{tabularx}

\end{table}
\clearpage

\subsection{Dataset Construction and Cleaning}
\label{app:datasetconstructioncleaning}

\paragraph{Extraction}
\label{par:extraction}
Based on our selection of the central banks as highlighted in Table \ref{tab:BanksChosen}, we collected data from the central banks' websites for the necessary documents. We targeted meeting minute documents or their closest equivalent in terms of content and formatting. The documents for each central bank, alongside the link we used to find these documents, can be found in Table \ref{tab:BanksChosen}. For the data collection, we used the Python library \texttt{selenium} to navigate through the pages and download the data. It is important to note that some central banks' data was in PDF format, whilst some were just text files. This led to a split in the data wherein certain banks' data was downloaded as a PDF and converted to Markdown by using the \texttt{PyMuPDF4LLM} Python library. If the data was in text format, it was kept as is, and any Microsoft Word documents (Docx) were converted to text.

\paragraph{Cleaning and Normalizing}
\label{par:clean_token}
Once all the documents were downloaded and sorted based on year, we began cleaning the documents by removing everything apart from the actual sentences of the document. This involved removing any charts and other visuals, indiscernible characters, phone numbers, email addresses, and other such unnecessary information that may be present in the document. 

After the data was cleaned, the sentences for each central bank were consolidated into separate JSON files, with the years as keys and the list of sentences as the values. Each sentence was then normalized by stripping the whitespace and removing any special characters. A sampling script was then used to randomly sample 1000 sentences from each central bank, with equal weight being given to each year. The final dataset consisted of 1000 normalized sentences from each central bank. This process enabled us to have a clean and rich dataset that was ready to annotate. 

\begin{table}[h!]
\caption{{The central banks that were chosen alongside their type of document and format. Each central bank name is hyperlinked to the source of that central bank's data.}}
\label{tab:BanksChosen}
\begin{tabular}{l l l}
\toprule 
\textbf{Bank} & \textbf{Type of Document} & \textbf{Format}\\
\midrule
\USA \href{https://www.federalreserve.gov/monetarypolicy/fomcminutes.htm}{Federal Reserve System} & Monetary Policy Minutes & Text \\
\CHINA \href{http://www.pbc.gov.cn/en/3688229/3688311/3688329/index.html}{People's Bank of China} & Monetary Policy Minutes & Text \\
\JAPAN \href{https://www.boj.or.jp/en/mopo/mpmsche_minu/index.htm}{Bank of Japan} & Monetary Policy Minutes & PDF \& Text \\
\UK \href{https://www.bankofengland.co.uk/monetary-policy-summary-and-minutes}{Bank of England} & Monetary Policy Minutes & PDF \\
\SWITZERLAND \href{https://www.snb.ch/en/the-snb/mandates-goals/monetary-policy}{Swiss National Bank} & Monetary policy assessments & PDF \\
\BRAZIL \href{https://www.bcb.gov.br/en/publications/copomminutes}{Central Bank of Brazil} & Monetary Policy Minutes & PDF\\
\INDIA \href{https://www.rbi.org.in/commonman/english/scripts/Minutes.aspx}{Reserve Bank of India} & Monetary Policy Minutes & PDF\\
\EU \href{https://www.ecb.europa.eu/press/accounts/html/index.en.html}{European Central Bank} & Monetary policy accounts & Text \\
\RUSSIA \href{https://www.cbr.ru/eng/dkp/mp_dec/}{Central Bank of the Russian Federation} & Key rate decisions & Text \\
\TAIWAN \href{https://www.cbc.gov.tw/en/lp-1025-2.html}{Central Bank of China (Taiwan)} & Monetary Policy Minutes & PDF \& Docx \\
\SINGAPORE \href{https://www.mas.gov.sg/monetary-policy/past-monetary-policy-statements-and-press-releases}{Monetary Authority of Singapore} & Monetary policy statements & Text \\
\SOUTHKOREA \href{https://www.bok.or.kr/eng/singl/newsDataEng/list.do}{Bank of Korea} & Monetary Policy Minutes & PDF \\
\AUS \href{https://www.rba.gov.au/monetary-policy/rba-board-minutes.html}{Reserve Bank of Australia} & Monetary Policy Minutes & Text \\
\ISRAEL \href{https://www.boi.org.il/en/communication-and-publications/interest-rate-decisions/Pages/default.aspx}{Bank of Israel} & Monetary Policy Minutes & PDF \\
\CANADA \href{https://www.bankofcanada.ca/press/press-releases/}{Bank of Canada} & Monetary Policy Press releases & Text \\
\MEXICO \href{https://www.banxico.org.mx/publications-and-press/minutes-of-the-board-of-governors-meetings/}{Bank of Mexico} & Monetary Policy Minutes & PDF \\
\POLAND \href{https://nbp.pl/en/monetary-policy/mpc-documents/}{National Bank of Poland} & Monetary Policy Minutes & PDF \\
\TURKEY \href{https://www.tcmb.gov.tr/wps/wcm/connect/EN/TCMB+EN/Main+Menu/Policy+Text+and+Monetary+Policy+Meetings/Meeting+Summary/}{Central Bank of Turkey} & Monetary Policy Meeting summary & Text \\
\THAILAND \href{https://www.bot.or.th/en/our-roles/monetary-policy/monetary-policy-committee}{Bank of Thailand} & Monetary Policy Minutes & PDF \\
\EGYPT \href{https://www.cbe.org.eg/en/monetary-policy/mpc-meeting-minutes/Pages/default.aspx}{Central Bank of Egypt} & Monetary Policy Press releases & PDF \\
\MALAYSIA \href{https://www.bnm.gov.my/monetary-stability/mpc-meeting-highlights/}{Bank Negara Malaysia} & Monetary policy statements & Text \\
\PHILIPPINES \href{https://www.bsp.gov.ph/SitePages/PriceStability/Minutes.aspx}{Central Bank of the Philippines} & Monetary Board highlights & PDF \\
\CHILE \href{https://www.bcentral.cl/areas/politica-monetaria/minutas/}{Central Bank of Chile} & Monetary Policy Minutes & PDF \\
\PERU \href{https://www.bcrp.gob.pe/en/monetary-policy/information-mpc.html}{Central Reserve Bank of Peru} & Monetary Policy Statements & PDF \\
\COLOMBIA \href{https://www.banrep.gov.co/en/minutes}{Bank of the Republic (Colombia)} & Monetary Policy Minutes & PDF \& Text \\
\bottomrule
\end{tabular}
\end{table}
\clearpage
\subsection{Dataset Statistics}
\begin{table*}[h]
\scriptsize
\caption{Statistics for various central banks over different year ranges. The ‘*’ indicates a corpus-wide average taken over the whole dataset.}
\label{tab:central_banks_stats_data}
\begin{tabular}{
    p{0.085\linewidth} p{0.12\linewidth} p{0.03\linewidth}
    >{\centering\arraybackslash}p{0.08\linewidth}
    >{\centering\arraybackslash}p{0.08\linewidth}
    >{\centering\arraybackslash}p{0.08\linewidth}
    >{\centering\arraybackslash}p{0.08\linewidth}
    >{\centering\arraybackslash}p{0.08\linewidth}
    >{\centering\arraybackslash}p{0.08\linewidth}}
\toprule
\multicolumn{3}{l}{} &
\multicolumn{3}{c}{\textbf{Total}} &
\multicolumn{3}{c}{\textbf{Average}} \\
\cmidrule(lr){4-6} \cmidrule(lr){7-9}
\textbf{Bank} & \textbf{Year Range} & \textbf{Years} &
\textbf{Sentences} & \textbf{Characters} & \textbf{Words} &
\textbf{Chars/Sent} & \textbf{Words/Sent} & \textbf{Sents/Year} \\
\midrule
\USA FOMC                  & 1996--2024 & 29    & 47,740 & 8,784,484  & 1,331,461 & 184.01 & 27.89 & 1,646.21 \\
\CHINA PBoC                & 2002--2024 & 21    & 1,189  & 264,511    & 39,082    & 222.47 & 32.87 & 56.62 \\
\JAPAN BoJ                 & 1998--2024 & 27    & 70,941 & 13,854,901 & 2,174,537 & 195.30 & 30.65 & 2,627.44 \\
\UK BoE                    & 1997--2024 & 28    & 68,281 & 10,136,875 & 1,642,593 & 148.46 & 24.06 & 2,438.61 \\
\SWITZERLAND SNB           & 2001--2024 & 24    & 3,377  & 369,886    & 59,585    & 109.53 & 17.64 & 140.71 \\
\BRAZIL BCB                & 2000--2024 & 24    & 11,623 & 2,086,611  & 316,659   & 179.52 & 27.24 & 484.29 \\
\INDIA RBI                 & 2016--2024 & 9     & 11,102 & 1,613,434  & 258,117   & 145.33 & 23.25 & 1,233.56 \\
\EU ECB                    & 2015--2024 & 10    & 21,044 & 3,734,051  & 582,948   & 177.44 & 27.70 & 2,104.40 \\
\RUSSIA CBR                & 2014--2024 & 10    & 3,430  & 433,937    & 68,111    & 126.51 & 19.86 & 343.00 \\
\TAIWAN CBCT                & 2017--2024 & 8     & 5,317  & 955,055    & 148,257   & 179.62 & 27.88 & 664.62 \\
\SINGAPORE MAS             & 2001--2024 & 24    & 2,183  & 300,453    & 47,987    & 137.63 & 21.98 & 90.96 \\
\SOUTHKOREA BoK            & 2010--2024 & 15    & 11,120 & 2,592,561  & 403,395   & 233.14 & 36.28 & 741.33 \\
\AUS RBA                   & 2006--2024 & 19    & 25,544 & 3,912,327  & 619,701   & 153.16 & 24.26 & 1,344.42 \\
\ISRAEL BoI                & 2006--2024 & 18    & 13,199 & 2,280,543  & 372,816   & 172.78 & 28.25 & 733.28 \\
\CANADA BoC                & 2011--2024 & 14    & 1,499  & 197,542    & 31,575    & 131.78 & 21.06 & 107.07 \\
\MEXICO BdeM               & 2018--2024 & 7     & 14,290 & 2,450,950  & 377,409   & 171.52 & 26.41 & 2,041.43 \\
\POLAND NBP                & 2007--2024 & 18    & 11,388 & 2,101,567  & 334,635   & 184.54 & 29.38 & 632.67 \\
\TURKEY CBRT               & 2015--2024 & 10    & 9,801  & 1,457,271  & 221,108   & 148.69 & 22.56 & 980.10 \\
\THAILAND BoT              & 2011--2024 & 14    & 7,828  & 1,290,443  & 191,807   & 164.85 & 24.50 & 559.14 \\
\EGYPT CBE                 & 2005--2024 & 20    & 2,369  & 399,815    & 62,496    & 168.77 & 26.38 & 118.45 \\
\MALAYSIA BNM              & 2004--2024 & 21    & 2,344  & 311,502    & 46,766    & 132.89 & 19.95 & 111.62 \\
\PHILIPPINES BSP           & 2002--2024 & 23    & 15,697 & 2,634,059  & 411,611   & 167.81 & 26.22 & 682.48 \\
\CHILE CBoC                & 2018--2024 & 7     & 3,907  & 657,712    & 104,124   & 168.34 & 26.65 & 558.14 \\
\PERU BCRP                 & 2001--2024 & 24    & 4,199  & 695,774    & 111,294   & 165.70 & 26.50 & 174.96 \\
\COLOMBIA BanRep           & 2008--2024 & 17    & 10,788 & 2,054,048  & 331,089   & 190.40 & 30.69 & 634.59 \\
\midrule
WCB                        & 1996--2024 & 17.64$^*$    & 380,200 & 65,570,312 & 10,289,163 & 172.46$^*$ & 27.06$^*$ & 13,110.34$^*$ \\
\bottomrule
\end{tabular}
\end{table*}

\clearpage
\subsection{Metadata Reporting}
\label{app:Metadata}

We format our metadata for each central bank and the corpus of 380,200 sentences as a whole in a standardized format. Our metadata is reported in both \texttt{CSV} and \texttt{JSON} format on our GitHub and HuggingFace for each central bank's dataset. The metadata reported is for all sentences for a given bank rather than just the annotated sentences. The formatting of the metadata in \texttt{JSON} is shown below in Figure \ref{fig:metadata}. The metadata in \texttt{CSV} format is similarly formatted, with each row containing the following columns: \texttt{sentence, bank, year, original\_document\_name, release\_date, start\_date, end\_date, minutes\_link, cleaned\_document\_name, doc\_id, sentence\_length}.

\begin{figure}[h!]
    \centering
    \footnotesize
    \begin{tcolorbox}
    [colback=gray!5,colframe=black!75,title=Metadata \texttt{JSON} formatting]
        \small
        \begin{verbatim}
    { 
      "central_bank_name": { 
        "year": { 
          "document_key": {
            "release_date": "DD-MM-YYYY",
            "start_date": "DD-MM-YYYY",
            "end_date": "DD-MM-YYYY",
            "minutes_link": "URL to the source document",
            "cleaned_document_name": "Filename of cleaned document",
            "original_document_name": "Filename of original document",
            "sentences": [
              "First sentence from the document.",
              "Second sentence from the document.",
              "..."
            ]
          }
        }
      }
    }
        \end{verbatim}
    \end{tcolorbox}
    \caption{{Template of \texttt{JSON}-style formatting for the metadata of each central bank in the corpus.}}
    \label{fig:metadata}
\end{figure}

\subsection*{Central Bank Document Limitations and Observations}
\label{app:metadata_limit}
\begin{itemize}
    \item \textbf{People’s Bank of China:} Many documents display vague dates like ``days ago'' or ``several days ago.'' In such cases, we defaulted to using the release date as the meeting date.
    
    \item \textbf{Central Reserve Bank of Peru:} There is a typographical error in the document dated July 6, 2002. Another document intended for March 7, 2002 is incorrectly labeled as March 7, 2001 on the central bank's website.

    \item \textbf{Bank of Russia}: Two documents' names are written in Russian but the actual sentences are in english. Their filenames have been manually renamed to maintain consistency. The two files were \href{https://www.cbr.ru/press/PR/?file=29042016_133001eng_dkp2016-04-29T13_07_53.htm}{document 1} and 
\href{https://www.cbr.ru/press/PR/?file=16062017_133001eng_keyrate2017-06-16T13_21_58.htm}{document 2}. 
    
    \item \textbf{Central Bank of Chile:} Some document filenames include the start date, but the actual meeting end date may be one day later. In rare cases, the date in the filename corresponds to the end date instead. These were handled accordingly. Additionally, a document from May 2022 is protected by Microsoft, which prevented access.
    
    \item \textbf{Bank of Israel:} No meeting documents were found for 2017, only a general report. In many cases, the link date and actual discussion date do not align clearly. For example:
    
    \begin{quote}
        ``Report to the public of the Bank of Israel’s discussions prior to setting the interest rate for December 2006. The discussions took place on November 26, 2006.''
    \end{quote}
    
    This makes automated date extraction unreliable due to lack of consistent formatting.
    
    \item \textbf{Central Bank of Egypt:} A document from March 28, 2004 was found to be irrelevant as a whole and thus excluded from processing. The document only mentioned the change of date of the meeting and nothing more.
    
    \item \textbf{Central Bank of Brazil:} One document (\texttt{MIN200044-COPOM20000215-44th COPOM Minutes}) from February 15, 2000 has a release date earlier than the end date, which is an anomaly. A document from the 49\textsuperscript{th} meeting in 2000 is too noisy and unreadable using standard PDF libraries and was therefore excluded.
\end{itemize}

\subsection{Instructions Given}
\label{app:instructions}

\paragraph{Terms of Participation}
The annotators in this project were volunteers or students working toward research credits at the Georgia Institute of Technology. Individuals who consistently demonstrated high-quality contributions were awarded co-authorship on this paper. Participants may distribute their own work but must not reference the lab or affiliates without prior consent. All external distributions must comply with applicable intellectual property rights and data sharing agreements.

The following subsections describe the instructions provided to participants based on the tasks assigned.

\subsubsection{Initial Research}
\label{app:research_assignment}
The participants were tasked with writing a summary of their assigned central bank to help shape the initial annotation guideline. Annotators were expected to read several minutes or equivalent documents from Monetary Policy Meetings of their assigned central bank in order to gain insight into the intended structure and content of the annotation guideline. The annotators were instructed to cite all sources used and to not refer to any third-party data sources for their research.

\textbf{Researching Central Bank}

\begin{enumerate}
    \item \textbf{Summary of the Monetary Policy:} Summarize the Monetary Policy of the central bank. This should be about 2-3 paragraphs long briefly mentioning the central bank's money supply, interest rates, and a description of the central bank's key historical events that have shaped the policy, as well as any other indicators. Determine whether the central bank claims a specific mandate. For example, the FOMC has a dual mandate: maximum employment and price stability. 
    \item \textbf{Numerical:} List the key numbers mentioned in the monetary policy such as Inflation rate, Interest rate, and so on. Be as specific as you can, providing dates with each number.
    \item \textbf{Trends:} Try to identify any trends within the monetary policy. This will require you to research your respective central bank's policy over the last 5 to 10 years.
    \item \textbf{Impact of Events:} Use a few lines to explain the impact of various global events such as the COVID pandemic and the 2008 crisis on your specific central bank's monetary policy. Additionally, observe any impacts that domestic events have had on the monetary policy. (For example, demonetization for India).
\end{enumerate}

\textbf{Guidelines Provided to Annotators}
\begin{itemize}
    \item It is of utmost importance that you \textbf{do not not communicate} with your peers regarding this task related to your central bank.
    \item The annotation guide should be written in clear and concise language. It must be in a PDF format.
\end{itemize}

\subsubsection{Sample Annotation Task}

Each annotator was expected to complete the 100 assigned sentences and become familiar with the Label Studio interface (Figure \ref{fig:heartex-interface} Appendix \ref{app:interface}). Annotators followed the instructions for annotations. These instructions were also made available on Label Studio. The annotators were told that each sentence must be annotated along the following three dimensions:

\begin{itemize}
    \item \textbf{Stance Detection:} \texttt{Hawkish, Dovish, Neutral}, or \texttt{Irrelevant}.
    \item \textbf{Temporal Classification:} \texttt{Forward Looking} or \texttt{Not Forward Looking}.
    \item \textbf{Uncertainty Estimation:} \texttt{Certain} or \texttt{Uncertain}.
\end{itemize}

\textbf{Examples for Each Dimension}
\begin{enumerate}
    \item \textbf{Annotations:} Annotate the given data according to the three dimensions. Each sentence will have these three dimensions associated with it.
    
\begin{itemize}
    \item \textbf{Stance Detection:} \texttt{Hawkish, Dovish, Neutral, or Irrelevant}
    \begin{itemize}
        \item  "The Committee then turned to a discussion of the economic and financial outlook, the ranges for the growth of money and debt in 1996, and the implementation of monetary policy over the intermeeting period ahead." \texttt{Neutral}
        \item "To support the Committee's decision to raise the target range for the federal funds rate, the Board of Governors voted unanimously to raise the interest rates on required and excess reserve balances to 2.": \texttt{Hawkish}
        \item "Inflation has moderated since earlier in the year, and longer-term inflation expectations have remained stable.": \texttt{Dovish}
        \item " It was agreed that the next meeting of the Committee would be held on Tuesday,  August 21, 2001.": \texttt{Irrelevant}
    \end{itemize}

    \item \textbf{Temporal Classification:} \texttt{Forward Looking or Not Forward Looking}
        \begin{itemize}
        \item  "Although the pace of economic recovery is likely to be moderate for a time, the Committee anticipates a gradual return to higher levels of resource utilization in a context of price stability.": \texttt{Forward Looking}
        \item "Commodity prices had been mixed recently after trending down earlier.": \texttt{Not Forward Looking}
        \end{itemize}
    \item \textbf{Uncertainty Estimation:} Certain or Uncertain.
        \begin{itemize}
            \item  "To support the Committee's decision to raise the target range for the federal funds rate, the Board of Governors of the Federal Reserve System voted unanimously to raise the interest rate paid on reserve balances to 0.": \texttt{Certain}
            \item "Some participants were concerned that inflation could rise as the recovery continued, and some business contacts had reported that producers expected to see an increase in pricing power over time.": \texttt{Uncertain}
        \end{itemize}
        
\end{itemize}
\end{enumerate}

\textbf{Guidelines Provided to Annotators}
\begin{itemize}
    \item It is of utmost importance that you \textbf{do not communicate} with your peers regarding the annotations related to your assigned countries.
    \item The annotations must be your own work. Progress will be monitored through Label Studio.
\end{itemize}

\subsubsection{Annotation Guide Task}
\label{app:annotation_guide_assignment}
The goal of this task was to create an annotation guide that teams would utilize for the final round of annotations. Specifically, the annotators were tasked with generating a guide describing the various labels that the sentences were to be annotated with. The annotators were instructed to specify and justify what the labels mean for their central bank and were encouraged to consider their initial annotation process and the first annotation guide while building this. 

\textbf{Guidelines Provided for Annotators}
\begin{itemize}
    \item The submission must be your own work, but you are welcome to work in teams and use LLMs. 
    \item Provide both the initial and final annotation guide.
\end{itemize}
\subsubsection{Annotation Task}
\label{app:annotation_assignment}
Through the Label Studio interface, the annotators were tasked with annotating each sentence assigned to them on the following three tasks: 
\begin{enumerate} 
    \item \texttt{Stance Detection:} \texttt{Hawkish, Dovish, Neutral, Irrelevant} 
    \item \texttt{Temporal Classification:} \texttt{Forward Looking, Not Forward Looking}
    \item \texttt{Uncertainty Estimation:} \texttt{Certain, Uncertain}
\end{enumerate}

\subsubsection{Annotation Disagreement Resolution}
\label{app:disagreement_assignment}
The goal of this task was to resolve disagreements in the final annotations using Label Studio. The annotators were teamed up in pairs. 

The following instructional steps were provided to the annotators for resolving disagreements in the annotations:
\begin{enumerate}

\item Collaborate with the other annotator on your project. Strictly adhere to the latest annotation guide for your assigned central bank and refer to it while resolving disagreements.

\item Open the data manager in your project and review annotations where there was a disagreement (where annotator agreement falls below 100\%). These can be identified in the “Agreement” column in the data manager.
\item Ensure that the entire sentence is selected for all annotations. If there is a disagreement, the annotator with the correct annotation should keep it. For example, if annotator A and annotator B disagree and annotator A’s annotation is correct, then annotator B should accept A’s annotation, and annotator A should fix and accept B’s annotations as corrected.

\item Confirm that all annotations have been accepted and that each annotator’s name is marked with a tick in the “Annotators” column in the data manager, indicating completion.
\end{enumerate}
\subsection{Annotation Interface}
\label{app:interface}
\begin{figure}[h!]
\centering
    \includegraphics[width=\linewidth]{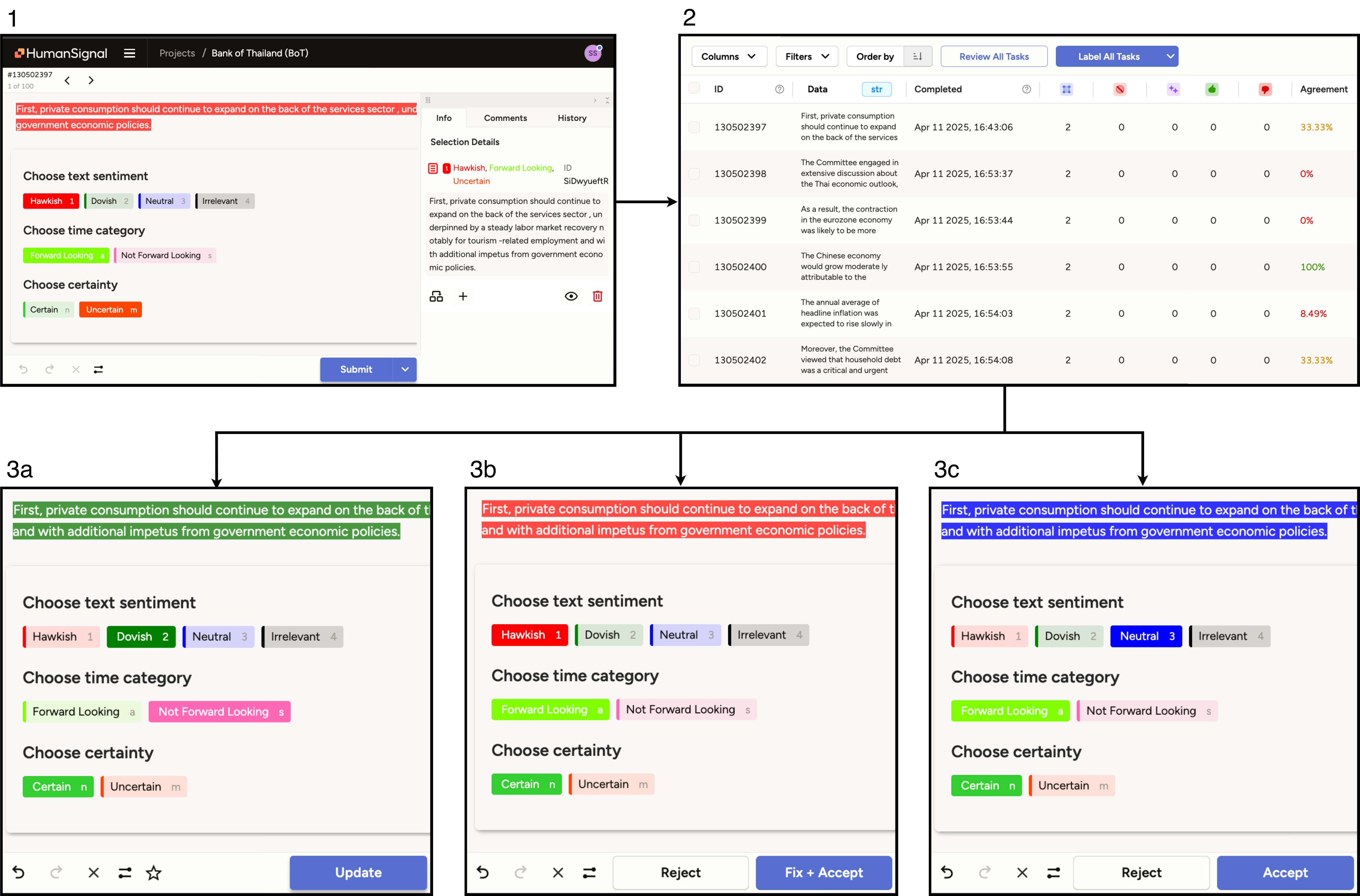}
    \caption{Tile 1 displays the interface to annotate sentences. Tile 2 displays the ``Reviewer'' view, allowing annotators to view disagreements after their initial annotations are complete. Tiles 3a, 3b, and 3c display the process of updating the annotation after pairs have reached a consensus.}
    \label{fig:heartex-interface}
\end{figure}

Tile one in Figure~\ref{fig:heartex-interface} shows the Label Studio interface presented to annotators during the initial labeling phase. Annotators were required to select one label per category without knowledge of their partner's annotations. Tile 2 of Figure~\ref{fig:heartex-interface} shows the interface used during the review and disagreement resolution phrase, where annotators could view disagreement statistics and their partners labels. During this phrase, they were allowed to discuss, edit their partner's labels, or accept them directly, which are displayed in tiles 3a, 3b, and 3c of Figure~\ref{fig:heartex-interface}. All annotations and review were completed on the same platform \cite{labelstudio2024} (Label Studio) to ensure consistency.

\subsection{Why Unique Annotation Guides?}
\label{app:why_annot}

\subsubsection{Manual Cross Validation}
\label{app:manual_check_annotation_guide}
To assess whether each central bank requires its own annotation guide, we conducted an experiment in which we first drew a random sample of 500 sentences from the Central Bank of Egypt (CBE) corpus. Two annotators from the CBE subgroup then applied the Bank of Canada (BoC) guide to these sentences, producing an initial set of labels. Independently, a separate team of annotators using the correct CBE guide labeled the same 500 sentences to create a gold-standard reference. Finally, we compared the BoC-guided annotations against this CBE ground truth to quantify the loss in labeling accuracy when using the wrong guide, thereby demonstrating the necessity of bank-specific annotation instructions.

As a result of this experiment, we observe that about a third of the sentences for \texttt{Stance Detection} and about $10\%$ for {\texttt{Temporal Classification}} were mislabeled. Since the \CE task is relatively similar for each bank, we observe that the number of mislabeled sentences is only $37/500$. This demonstrates that the annotation guides have significant differences that cannot be overlooked. It is apparent that misunderstanding specific details such as inflation targets and policy mandates can significantly impact the annotation label, highlighting the need for unique annotation guides. A sample of mislabeled data is in Table \ref{tab:mislabel_example}.

\begin{table}[h!]
\centering
\caption{Sample of mislabeled sentences between groups one and two. }
\label{tab:mislabel_example}
\begin{tabular}{p{0.325\textwidth}p{0.2\textwidth}p{0.2\textwidth}p{0.15\textwidth}}
\toprule
\textbf{Sentence} & \textbf{Task} & \textbf{Incorrect Label} & \textbf{Correct Label} \\
\midrule
annual headline urban inflation declined in october and november 2021, to record 6.3 percent and 5.6 percent from 6.6 percent in september 2021, respectively. & \SD & Neutral & Dovish \\
meanwhile, core inflation declined to 7.9 percent in november 2018 from 8.9 percent in october 2018, recording the lowest rate since february 2016. & \SD & Dovish & Hawkish \\
in addition to oil price developments, risks from the external economy continue to include the pace of tightening financial conditions. & \TC & Not forward looking & Forward looking \\
moreover, the mpc is mindful of the underlying inflationary pressures which continue to tilt the balance of risks to the upside. & \CE & Uncertain & Certain \\
\bottomrule
\end{tabular}

\end{table}

\subsubsection{4-Gram Semantic Overlap}
\label{app:4gram-sem-overlap}

\begin{figure}[h!]
\centering
    \includegraphics[width=\linewidth]{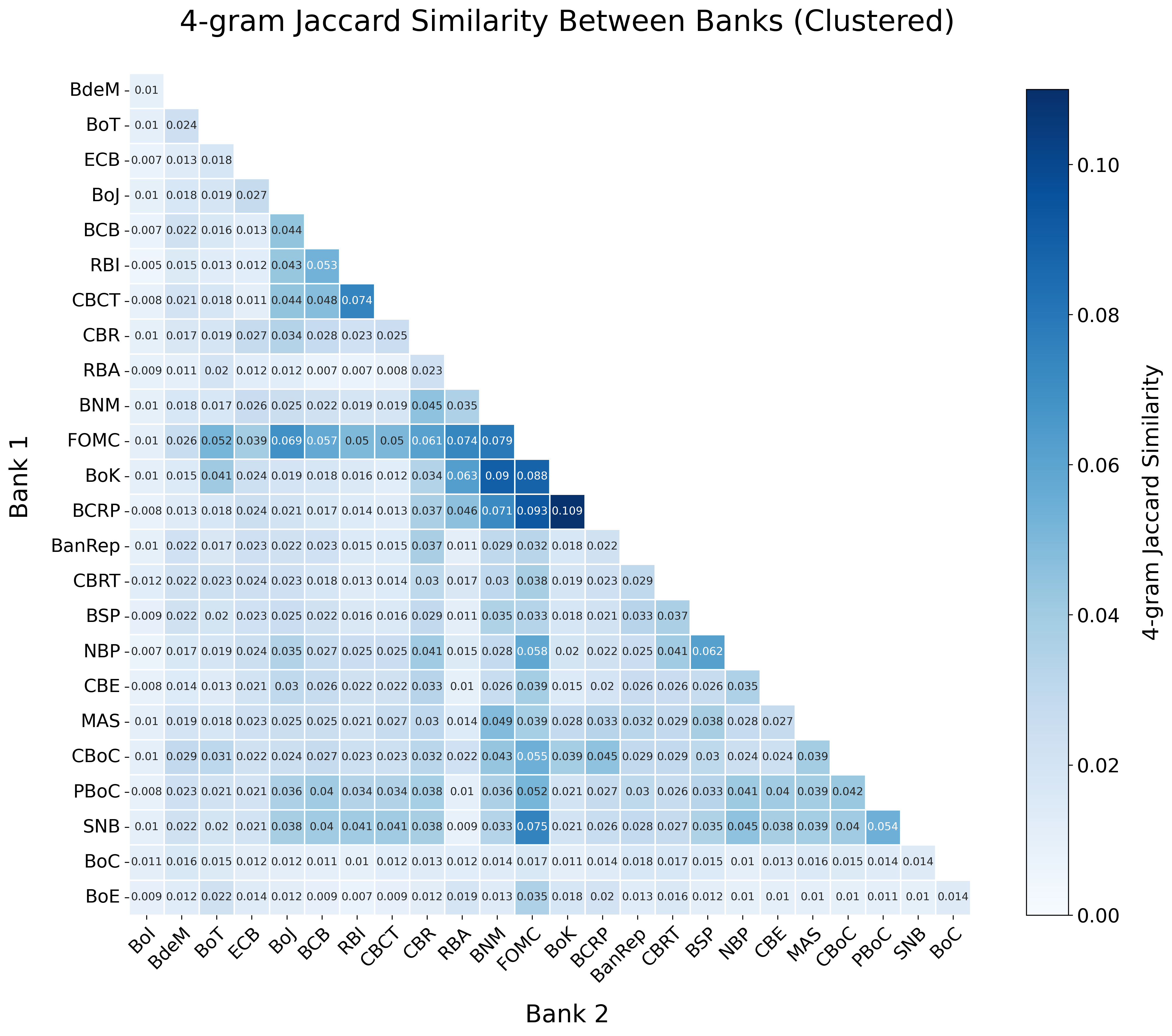}
    \caption{Pairwise 4-gram Jaccard similarity scores between annotation guides show extremely low overlap, ranging from $0.0055$ to $0.1088$. These results support our decision for unique annotation guides.}
    \label{fig:NGram-annotation}
\end{figure}
To further evaluate the degree of similarity between the annotation guides, we compute the pairwise 4-gram Jaccard similarity scores, inspired by the work of \citep{shah2024finerordfinancialnamedentity, choubey-etal-2023-lexical}. After converting the annotation guides (Appendix \ref{app:annotation_guides}) to text and removing the latex jargon, we run the N-gram Jaccard semantic between pairs of all central banks' annotation guides. As Figure \ref{fig:NGram-annotation} illustrates, there is extremely low similarity, ranging from 0.0055 to 0.1088, between the central banks' annotation guides. Despite having the same formatting guidelines, each annotation guide has unique concepts and textual information, reinforcing the need for unique annotation guides.
\subsection{Why Irrelevant Label for Stance? }
\label{app:Word_Filter}

Unlike other research done in the past, we utilize an alternative approach to clean the sentence level data that serves as the input for our annotation work. As mentioned in Appendix \ref{app:datasetconstructioncleaning}, we filter the data on token length (remove obvious bad sentences), and then use the sentence corpus as a whole rather than breaking it down using word filtering as suggested by the work of \cite{gorodnichenko2023voice}. Instead, we use a novel approach by adding an additional label in our annotations for \texttt{Stance Detection} called \texttt{Irrelevant}. Since we are analyzing the monetary policy communications from central banks across the world in which a specific dictionary such as \cite{gorodnichenko2023voice} has not previously been created or defined, it is not appropriate for us to generalize the dictionary to central bank communications from around the world. 
Additionally, we measure the central bank communications across additional features namely forward looking nature and certainty for which word filtering is not applicable. 

Moreover, we observe that using the dictionary from \cite{yuri} rather than our classification of \irr sentences as a label causes a significant number of sentences to be left out, as seen in Table \ref{tab:fed_chair_filter}. Using the dictionary would limit the scope of our work and cause false negatives. We include a sample of sentences in Table \ref{tab:yuri_failure} that remain in our dataset and contain valuable central bank communication.

\begin{table}[ht]
\centering
\renewcommand{\arraystretch}{1.2}
\caption{Comparison of number of relevant and irrelevant sentences using Word Filtering method versus the \irr label for sentences pertaining to each FOMC Chair. This analysis was specifically conducted for the FOMC meeting minute data.}
\label{tab:fed_chair_filter}
\begin{tabular}{lccc}
\toprule
\multicolumn{4}{c}{\textbf{FED Chair — Alan Greenspan}} \\
\midrule
 & \textbf{Relevant} & \textbf{Irrelevant} & \textbf{Sum} \\
\midrule
Word Filter         & 118 & 122 & 230 \\
Ground Truth  & 228 &  12 & 230 \\
\midrule
\multicolumn{4}{c}{\textbf{FED Chair — Ben Bernanke}} \\
\midrule
Word Filter       & 131 & 169 & 300 \\
Ground Truth  & 289 &  11 & 300 \\
\midrule
\multicolumn{4}{c}{\textbf{FED Chair — Janet Yellen}} \\
\midrule
Word Filter          & 103 &  88 & 191 \\
Ground Truth  & 183 &   8 & 191 \\
\midrule
\multicolumn{4}{c}{\textbf{FED Chair — Jerome Powell}} \\
\midrule
Word Filter         & 117 & 136 & 252 \\
Ground Truth  & 245 &   8 & 252 \\
\midrule
\multicolumn{4}{c}{\textbf{Total}} \\
\midrule
Word Filter          & 469 & 531 & 1000 \\
Ground Truth  & 961 &  39 & 1000 \\
\bottomrule
\end{tabular}
\end{table}

\begin{table}[ht]
\centering
\small
\caption{Sample of sentences that were removed according to Yuriy's dictionary but were included in our sampling corpus.}
\label{tab:yuri_failure}
\begin{tabular}{p{0.25\textwidth} ccc}
\toprule
\textbf{Sentence} & \textbf{Stance Detection} & \textbf{Temporal Classification} & \textbf{Uncertainty Estimation} \\
\midrule
In June, preliminary data indicated declining nominal goods exports and imports. 
  & \neut & \notforward & \certain\\
The core component of the CPI also accelerated in January and on a year-over-year basis, 
but by lesser amounts than did the total index.
  & \hawk & \notforward & \certain\\
Exports of industrial supplies and consumer goods also rose strongly, 
while exports of services expanded modestly.
  & \dov & \notforward & \certain\\
Real GDP was anticipated to increase at a rate noticeably below its potential in 2008.
  & \dov & \forward & \uncertain\\\\
\bottomrule
\end{tabular}

\end{table}

\clearpage
\section{Models and Benchmarking}
\label{app:models_and_benchmarking}
\subsection{Prompts for Benchmarking}
\label{app:promptsbenchmarking}
We present all prompts used to benchmark our experiments, including zero-shot prompts, few-shot prompts, and prompts with annotation guides.
\subsubsection{Zero-Shot Prompt: Stance Detection }
\label{app:prompt_stance}

\begin{figure}[h!]
    \centering
    \footnotesize
    \begin{tcolorbox}
    [colback=gray!5,colframe=black!75,title= \SD Annotation Prompt]
        \small
\begin{verbatim}
"""You are given a sentence related to the {bank_name}'s monetary policy meeting. 
Your task is to classify its monetary policy stance and briefly justify 
your choice.

Input:
- Sentence: {sentence}

Instructions:
1. Assign one of the following labels under the `label` key:
"hawkish", "neutral", "dovish", or "irrelevant".

2. Provide a concise explanation for your classification using the 
   `justification` key. Limit the justification to one sentence.

3. Your output must follow this structure:
{
"label": "hawkish | dovish | neutral | irrelevant",
"justification": "One-sentence explanation for the assigned label"
}"""
\end{verbatim}
    \end{tcolorbox}
    \caption{Zero-shot prompt used for \texttt{Stance Detection} by an LLM across all central banks. For banks with names ending in \textit{s}, such as the Central Bank of the Philippines, an apostrophe was added without the \textit{s} (e.g., \texttt{{bank\_name}'}) to ensure grammatical correctness.}
    \label{fig:stance_prompt}
\end{figure}
\clearpage
\subsubsection{Zero-Shot Prompt: Temporal Classification}
\label{app:prompt_time}
\begin{figure}[h!]
    \centering
    \footnotesize
    \begin{tcolorbox}
    [colback=gray!5,colframe=black!75,title=\TC Annotation Prompt]
        \small
\begin{verbatim}
"""You are given a sentence related to the {bank_name}'s monetary policy meeting. 
Your task is to classify whether it is forward looking or not forward looking 
and briefly justify your choice.

Input:
- Sentence: {sentence}

Instructions:
1. Assign one of the following labels under the `label` key:
"forward looking", "not forward looking".

2. Provide a concise explanation for your classification using the 
   `justification` key. Limit the justification to one sentence.

3. Your output must follow this structure:
{
"label": "forward looking | not forward looking",
"justification": "One-sentence explanation for the assigned label"
}"""
\end{verbatim}
    \end{tcolorbox}
       \caption{Zero-shot prompt used for \texttt{Temporal Classification} by an LLM across all central banks. For banks with names ending in \textit{s}, such as the Central Bank of the Philippines, an apostrophe was added without the \textit{s} (e.g., \texttt{{bank\_name}'}) to ensure grammatical correctness.}
    \label{fig:time_prompt}
\end{figure}

\clearpage
\subsubsection{Zero-Shot Prompt: Uncertainty Estimation}
\label{app:prompt_certain}
\begin{figure}[h!]
    \centering
    \footnotesize
    \begin{tcolorbox}
    [colback=gray!5,colframe=black!75,title=\CE Annotation Prompt]
        \small
\begin{verbatim}
"""You are given a sentence related to the {bank_name}'s monetary policy meeting. 
Your task is to classify whether it is certain or uncertain 
and briefly justify your choice.

Input:
- Sentence: {sentence}

Instructions:
1. Assign one of the following labels under the `label` key:
"certain", "uncertain".

2. Provide a concise explanation for your classification using the
`justification` key. Limit the justification to one sentence.

3. Your output must follow this structure:
{
"label": "certain | uncertain",
"justification": "One-sentence explanation for the assigned label"
}"""
\end{verbatim}
    \end{tcolorbox}
    \caption{Zero-shot prompt used for \texttt{Uncertainty Estimation} by an LLM across all central banks. For banks with names ending in \textit{s}, such as the Central Bank of the Philippines, an apostrophe was added without the \textit{s} (e.g., \texttt{{bank\_name}'}) to ensure grammatical correctness.}  \label{fig:certain_prompt}
\end{figure}
\clearpage

\subsubsection{Prompting Structure for Few-Shot prompting}
\label{app:few_shot_prompting_structure}
\begin{figure}[h!]
    \centering
    \footnotesize
    \begin{tcolorbox}[
        colback=gray!5,
        colframe=black!75,
        title=Prompt Structure for Few-Shot prompting. This exact structure was used across all tasks.
    ]
\textbf{[SYSTEM INPUT]}
\begin{verbatim}
"""You are given one sentence related to {bank_name}'s monetary policy meeting. 
Your task is to classify whether it is {lbl_map[feature]} and briefly 
justify your choice.

Instructions:
1. Assign one of the following labels under the `label` key: {lbl_map[feature]}
2. Provide a concise explanation for your classification using the 
`justification` key. Limit the justification to one sentence.
3. Your output must follow this structure:
{
  "label": "{instruct_map[feature]}",
  "justification": "One-sentence explanation for the assigned label"
}"""
\end{verbatim}
\vspace{0.5em}

{\color{orange}
Example:\\
bank\_name: Bank Negara Malaysia.\\
label: stance}

\vspace{1em}
\textbf{[USER INPUT]} \\
{
Input:\\
    \#\#\# Few‑shot Examples\\ \texttt{\{examples\}}\\ \\
    \#\#\# Sentence\\ \texttt{\{sentence\}} \\}

\vspace{0.5em}
{\color{orange}
Example:\\\\
\#\#\#Few-shot Examples\\\\
Sentence: the ceiling and floor rates of the corridor of the opr are correspondingly reduced to 2.75 percent and 2.25 percent, respectively\\
Label: neutral\\

Sentence: since march 2020, bank negara malaysia has provided additional liquidity of approximately rm42 billion into the domestic financial markets, via various tools including outright purchase of government securities, reverse repos and the reduction in statutory reserve requirement\\
Label: dovish\\

Sentence: the meeting also approved the schedule of mpc meetings for 2014\\
Label: irrelevant\\

Sentence: downside risks to this outlook, however, continue to persist\\
Label: hawkish\\\\
\#\#\#Sentence\\
the mpc recognises that there are downside risks in the global economic and financial environment and is closely monitoring and assessing their implications on domestic price stability and growth}
    \end{tcolorbox}
    \caption{Prompting Structure for Few-Shot prompting using few-shot examples from the training set for that particular seed.}
    \label{fig:few_shot_prompt}
\end{figure}

\clearpage
\subsubsection{Prompting Structure for prompting with Annotation Guides}
\label{app:annotation_shot_prompting_structure}
\begin{figure}[h!]
    \centering
    \footnotesize
    \begin{tcolorbox}[
        colback=gray!5,
        colframe=black!75,
        title=Prompt Structure for prompting with Annotation Guides. This exact structure was used across all tasks.
    ]
\textbf{[SYSTEM INPUT]} \\
{
You are given one sentence related to \texttt{\{bank\_name\}}'’s monetary policy
meeting. You are also given an annotation guide for the task: \texttt{\{task\}}.
Strictly follow the guide; do not invent new criteria or labels. Your task
is to classify whether it is \texttt{\{lbl\_map[feature]\}} and briefly justify your choice.\\
\\
Instructions:\\
1. Assign one of the following labels under the `label` key: \texttt{\{lbl\_map[feature]\}}
\\
2. Provide a concise explanation for your classification using the `justification` key. Limit the justification to one sentence.
\\
3. Your output must follow this structure:\\
\{\{\\
"label": "\texttt{\{instruct\_map[feature]\}}",\\
"justification": "One-sentence explanation for the assigned label"\\
\}\}
}
\vspace{0.5em}

{\color{orange}
Example:\\
bank\_name: Reserve Bank of India.\\
label: stance}

\vspace{1em}
\textbf{[USER INPUT]} \\
{
Input:\\
    \#\#\# Annotation Guide: \texttt{\{guide\_text\}}\\ \\
    \#\#\# Sentence\\ \texttt{\{sentence\}}} \\

\vspace{0.5em}
{\color{orange}
Example:\\
\begin{verbatim}

###Annotation Guide

\mptext{rbi}{fourteen} Inflation, 
Repo Rate, Reverse Repo Rate, Cash Reserve Ratio, Statutory Liquidity Ratio, 
GDP Growth Forecast, Monetary Policy Measures, 
Export Performance, Manufacturing Activity, Consumer Demand, 
Employment Levels, Commodity Prices, 
Credit Growth, Exchange Rate:

\begin{itemize}
\item \emph{Inflation}: A sentence pertaining to the general increase 
in prices for goods and services; the decrease in the purchasing power 
of a currency...

###Sentence
the outlook for aggregate demand is progressively improving but the slack is 
large: output is still below pre-covid level and the recovery is uneven and 
critically dependent upon policy support.
\end{verbatim}
}

    \end{tcolorbox}
    \caption{Prompting Structure for prompting with Annotation Guides using the annotation guide for that particular central bank and task.}
    \label{fig:annotation_shot_prompt}
\end{figure}
\subsection{Model Details}
Details about each model such as model type, size, abbreviations used in results' tables, quantization settings for LLMs, context windows (in tokens) for LLMs, and cutoff dates for LLMs are provided in Table~\ref{tab:model_abb}.
\label{app:Model_abb}
\begin{table}[h!]
\centering
\scriptsize
\caption{{Comparison of fine-tuned models used to classify sentences across notable characteristics. \textit{Type} is the type of model, whether base or large, or closed or open source. \textit{Size} is number of parameters in the model. \textit{Abbreviation} is the abbreviated name of the model used in our paper. \textit{Note} specifies if a model incorporates reasoning model or a mixture of exports (MOE) architectures. \textit{*} means that the date for the knowledge cutoff was unable to be found. Since FinMA is fine-tuned on Llama-7B, we use the \textit{cutoff date} for Llama-7B, despite FinMA undergoing further training. The horizontal separation is to separate the PLMs (top) and the LLMs (bottom) for further analysis on the specific LLMs used.}}
\label{tab:model_abb}

\begin{tabular}{p{0.27\textwidth}p{0.1\textwidth}p{0.05\textwidth}p{0.08\textwidth}p{0.07\textwidth}p{0.1\textwidth}p{0.02\textwidth}p{0.05\textwidth}}
\toprule
Model & Type & Size & Abbreviation & Quantization & Context Window(Tokens) & Note & Cutoff \\
\midrule
\MB \texttt{ModernBERT-base} & Base & 150m & MBB & \\
\google \texttt{bert-base-uncased} & Base & 110m & BB & \\
\finbert \texttt{finbert-pretrain} & Base & 110m & FB & \\
\meta \texttt{roberta-base} & Base & 125m & RBB & \\
\MB \texttt{ModernBERT-large} & Large & 396m & MBL & \\
\google \texttt{bert-large} & Large & 340m & BL & \\
\meta \texttt{roberta-large} & Large & 355m & RBL & \\
\midrule
\FinMA \texttt{finma-7b-full} & Closed Source & - & FM & - & - & - & 3/\text{*}/2023 \\
\google \texttt{gemini-2.0-flash} & Closed Source & - & Gem & - & 1M & - & 8/\text{*}/2024 \\
\openai \texttt{gpt-4o-2024-08-06} & Closed Source & - & 4o & - & 128k & - & 9/30/2023 \\
\openai \texttt{gpt-4.1-2025-04-14} & Closed Source & - & 4.1 & - & 1M & - & 5/31/2024 \\
\openai \texttt{gpt-4.1-mini-2025-04-14} & Closed Source & - & 4.1M & - & 1M & - & 5/31/2024  \\
\deepseek \texttt{DeepSeek-V3(DeepSeek-V3-0324)} & Open Source & 671b & DS & FP8 & 131k & MoE & 7/1/2024\\
\qwen \texttt{Qwen2.5-72B-Instruct-Turbo} & Open Source & 72.7b & Qwen & FP8 & 32.768k & - & - \\
\meta \texttt{Llama-3-70b-chat-hf} & Open Source & 70b & L3 & FP16 & 8k & - & 12/\text{*}/2023\\
\meta \texttt{Llama-4-Scout-17B-16E-Instruct} & Open Source & 405b & L4S & FP16 & 1M & - & 8/\text{*}/2024\\

\bottomrule
\end{tabular}

\end{table}

\subsection{Model Specifications, Compute, and Providers}
\label{app:technical_details_benchmarking}

We use a 700–150–150 train–validation–test split for all PLM experiments. All models are trained using cross-entropy loss as the criteria and optimized with AdamW \citep{loshchilov2019decoupledweightdecayregularization}. Training runs for a maximum of 50 epochs with early stopping based on validation performance. The best-performing weights and results are saved locally when running experiments for each hyperparameter combination.

\paragraph{Pretrained Language Models (PLMs).}
All PLM experiments are conducted using PyTorch \citep{paszke2019pytorch} on NVIDIA H200 GPUs. Models are initialized from their publicly available checkpoints via the Hugging Face Transformers library \citep{wolf-etal-2020-transformers}. We perform a grid search over seeds, batch sizes, and learning rates:
\begin{itemize}
    \item \texttt{seeds = [5768, 78516, 944601]}
    \item \texttt{batch sizes = [16, 32]}
    \item \texttt{learning rates = [1e-5, 1e-6]}
\end{itemize}

\paragraph{Large Language Models (LLMs).}
For LLMs, we use \texttt{LiteLLM} \citep{litellm2024} as a unified API router. Inference is conducted with the following hyperparameters unless otherwise noted:
\begin{itemize}
    \item \texttt{max tokens = 128}.
    \item \texttt{temperature = 0.0}
    \item Other parameters such as \texttt{top-p}, \texttt{top-k}, and \texttt{repetition penalty} are kept at model-specific defaults.
\end{itemize}
Note: Temperature is fixed to 0.0 across all models. For models that are proprietary, closed-source, and closed-weights, actual behavior may vary slightly due to backend constraints.

\paragraph{LLM Providers.}
\begin{itemize}
    \item \textbf{Together.ai API:} Used for open-source models including the Llama family, \texttt{DeepSeek}, and \texttt{Qwen2.5}.
    \item \textbf{OpenAI API:} Used for \texttt{GPT-4o}, \texttt{GPT-4.1}, and \texttt{GPT-4.1-Mini}.
    \item \textbf{Gemini SDK:} Used for \texttt{Gemini 2.0 Flash}; accessed via the official Google Gemini Python client, not routed through LiteLLM.
\end{itemize}

\subsection{Results}
\label{app:results}
We present the results of the benchmarking experiments for each task (\SD, \TC, and \CE) for both General and Bank-Specific Setups. The results for the \SD under the General Setup have already been displayed in Section \ref{sec:results}.
\subsubsection{General (All-Banks) Setup Models}
\label{app:all_model}
\textbf{Temporal Classification Label}

\paragraph{General Setup – Temporal Classification Analysis.}
In the General (All-Banks) Setup evaluation, Table \ref{tab:plm_f1_time_all} shows F1-Score ($\pm$ std) for the \texttt{Temporal Classification} across 25 central banks, comparing four base PLMs, three large PLMs, four closed‑source LLMs, and five open‑source LLMs. On average, \texttt{RoBERTa-Base} ($\overline{F1} = 0.868$) is the best-performing PLM  while \texttt{Gemini-2.0-Flash} ($\overline{F1} = 0.864$) is the best-performing LLM. Although \texttt{RoBERTa-Base} slightly outperforms \texttt{Gemini} across most metrics, the PLM-LLM gap remains small ($<$0.005 F1 on average), suggesting that both families can tackle this binary task equally well when fine‑tuned on pooled data.

\label{app:time_all_result}
\begin{table}[h!]
\centering
\tiny

\caption{F1-Scores across the banks for the \TC task with the standard deviation in the brackets. These results are based on the testing of a singular model on each bank's data. The best-performing PLM and LLM in each row is highlighted in blue and green, respectively. The best-performing overall model is bolded. The model names are abbreviated as described in Table \ref{tab:model_abb}, Appendix \ref{app:Model_abb}.}

\label{tab:plm_f1_time_all}
\begin{tabularx}{\textwidth}{
l 
*{4}{X}                   
: *{3}{X}                   
| *{4}{X}                 
: *{5}{X}                 
}
\toprule
& \multicolumn{4}{c}{\textbf{Base}} 
& \multicolumn{3}{c}{\textbf{Large}}
& \multicolumn{4}{c}{\textbf{Closed-Source}} 
& \multicolumn{5}{c}{\textbf{Open-Source}} \\
\cmidrule(lr){2-5}\cmidrule(lr){6-8}\cmidrule(lr){9-12}\cmidrule(lr){13-17}
\textbf{Bank} 
& \MB MBB
& \google BB
& \finbert FB 
& \meta RBB 
& \MB MBL
& \google BL
& \meta RBL
& \google Gem
& \openai 4o
& \openai 4.1m
& \openai 4.1
& \deepseek DS
& \qwen Qwen
& \FinMA FM
& \meta L3
& \meta L4S \\
\midrule
\BRAZIL~BCB & .857 (.019) & .859 (.017) & .844 (.016) & .870 (.024) & \cellcolor{blue!20}\textbf{.884 (.039)} & .870 (.032) & .851 (.022) &  
  .835 (.031) & .840 (.025) & .809 (.027) & .822 (.021) & \cellcolor{green!20}{.845 (.014)} & .830 (.027) & .182 (.010) & .821 (.005) & .826 (.014) \\  

  \PERU~BCRP & .858 (.034) & .878 (.028) & .869 (.037) & .878 (.026) & .867 (.033) & .878 (.024) & \cellcolor{blue!20}\textbf{.884 (.010)} & .847 (.022) & .861 (.016) & .855 (.032) & \cellcolor{green!20}.869 (.034) & .851 (.037) & .838 (.020) & .113 (.014) & .854 (.040) & .848 (.036) \\  

  \MALAYSIA~BNM & .885 (.005) & .876 (.010) & .889 (.011) & \cellcolor{blue!20}\textbf{.903 (.019)} & .886 (.016) & .870 (.007) & .892 (.017) & .856 (.018) & .877 (.016) & .868 (.017) & \cellcolor{green!20}.889 (.021) & .880 (.020) & .872 (.028) & .553 (.013) & .888 (.025) & .879 (.025) \\  

  \PHILIPPINES~BSP & \cellcolor{blue!20}\textbf{.921 (.030)} & .907 (.022) & .907 (.023) & .919 (.020) & .906 (.009) & .900 (.011) & .901 (.015) & .886 (.012) & \cellcolor{green!20}.916 (.013) & .894 (.020) & .912 (.019) & .911 (.019) & .916 (.015) & .166 (.016) & .905 (.011) & .901 (.011) \\  

  \COLOMBIA~BanRep & .885 (.025) & .884 (.016) & .866 (.031) & .883 (.027) & \cellcolor{blue!20}\textbf{.909 (.025)} & .870 (.017) & .885 (.016) & .845 (.054) & \cellcolor{green!20}.890 (.045) & .884 (.020) & .885 (.042) & .875 (.026) & .881 (.025) & .148 (.024) & .877 (.029) & .867 (.036) \\  

  \CANADA~BoC & .876 (.031) & .864 (.031) & .868 (.036) & .890 (.039) & .875 (.034) & \cellcolor{blue!20}\textbf{.891 (.027)} & .882 (.040) & .852 (.051) & \cellcolor{green!20}.882 (.035) & .830 (.030) & .871 (.018) & .861 (.037) & .876 (.036) & .353 (.023) & .858 (.036) & .848 (.028) \\  

  \UK~BoE & .782 (.018) & \cellcolor{blue!20}.787 (.024) & .768 (.018) & .780 (.019) & .782 (.006) & .765 (.022) & .778 (.012) & .715 (.018) & .761 (.008) & .771 (.022) & \cellcolor{green!20}\textbf{.793 (.014)} & .763 (.019) & .777 (.003) & .142 (.030) & .765 (.017) & .766 (.006) \\  

  \ISRAEL~BoI & .854 (.022) & .874 (.041) & \cellcolor{blue!20}.878 (.007) & .863 (.015) & .873 (.028) & .866 (.030) & .865 (.036) & .829 (.026) & .875 (.028) & \cellcolor{green!20}\textbf{.896 (.025)} & .879 (.014) & .884 (.036) & .858 (.021) & .260 (.024) & .851 (.018) & .875 (.027) \\  

  \JAPAN~BoJ & .788 (.054) & .779 (.025) & \cellcolor{blue!20}\textbf{.792 (.025)} & .790 (.025) & .792 (.026) & .783 (.024) & .786 (.029) & .753 (.008) & .786 (.022) & .775 (.037) & .778 (.017) & .778 (.017) & \cellcolor{green!20}.789 (.037) & .322 (.017) & .763 (.031) & .764 (.020) \\  

  \SOUTHKOREA~BoK & .882 (.007) & .875 (.014) & .885 (.017) & \cellcolor{blue!20}.896 (.005) & .888 (.022) & .882 (.009) & .891 (.008) & .872 (.005) & .900 (.009) & .885 (.030) & .900 (.009) & \cellcolor{green!20}\textbf{.908 (.012)} & .904 (.008) & .369 (.020) & .891 (.008) & .853 (.014) \\  

  \MEXICO~BdeM & .864 (.023) & .859 (.009) & .868 (.024) & .852 (.032) & .871 (.022) & .853 (.025) & \cellcolor{blue!20}\textbf{.875 (.011)} & .809 (.016) & \cellcolor{green!20}.875 (.018) & .862 (.033) & .858 (.027) & .840 (.038) & .862 (.028) & .307 (.019) & .862 (.028) & .830 (.026) \\  

  \THAILAND~BoT & \cellcolor{blue!20}\textbf{.770 (.056)} & .762 (.043) & .743 (.054) & .753 (.048) & .763 (.046) & .759 (.030) & .750 (.053) & .738 (.025) & .748 (.029) & .722 (.025) & \cellcolor{green!20}.752 (.025) & .732 (.035) & .751 (.018) & .427 (.007) & .727 (.033) & .735 (.031) \\  

  \TAIWAN~CBCT & \cellcolor{blue!20}\textbf{.833 (.029)} & .829 (.039) & .831 (.029) & .824 (.039) & .806 (.030) & .823 (.039) & .828 (.029) & .831 (.028) & \cellcolor{green!20}.831 (.019) & .810 (.027) & .820 (.019) & .822 (.037) & .818 (.035) & .306 (.025) & .825 (.021) & .809 (.015) \\  

  \EGYPT~CBE & \cellcolor{blue!20}.926 (.021) & .913 (.020) & .925 (.019) & .914 (.017) & .923 (.025) & .904 (.024) & .923 (.028) & .913 (.011) & .935 (.006) & .943 (.011) & .933 (.009) & \cellcolor{green!20}\textbf{.943 (.014)} & .928 (.007) & .182 (.025) & .935 (.008) & .936 (.006) \\  

  \RUSSIA~CBR & .887 (.020) & .887 (.018) & .880 (.032) & .876 (.018) & .887 (.020) & \cellcolor{blue!20}\textbf{.904 (.023)} & .885 (.030) & .857 (.040) & .851 (.008) & .838 (.029) & .849 (.019) & .840 (.016) & .857 (.031) & .235 (.027) & .851 (.027) & \cellcolor{green!20}.861 (.024) \\  

  \TURKEY~CBRT & .926 (.007) & .946 (.008) & .927 (.024) & .947 (.005) & .923 (.009) & .946 (.006) & \cellcolor{blue!20}\textbf{.953 (.008)} & .919 (.006) & \cellcolor{green!20}.936 (.008) & .901 (.015) & .922 (.027) & .919 (.010) & .931 (.006) & .132 (.024) & .935 (.009) & .915 (.015) \\  

  \CHILE~CBoC & .811 (.014) & .812 (.011) & .808 (.014) & \cellcolor{blue!20}.820 (.027) & .818 (.015) & .789 (.025) & .802 (.028) & .798 (.009) & \cellcolor{green!20}\textbf{.835 (.007)} & .793 (.021) & .832 (.008) & .813 (.020) & .809 (.027) & .214 (.017) & .797 (.007) & .794 (.011) \\  

  \EU~ECB & .795 (.010) & .783 (.018) & .774 (.010) & \cellcolor{blue!20}\textbf{.811 (.011)} & .790 (.024) & .791 (.008) & .774 (.009) & .741 (.045) & .781 (.022) & .763 (.023) & .784 (.014) & .771 (.013) & .779 (.023) & .267 (.051) & \cellcolor{green!20}.788 (.019) & .780 (.011) \\  

  \USA~FOMC & .881 (.017) & \cellcolor{blue!20}.904 (.005) & .888 (.027) & .899 (.019) & .887 (.024) & .890 (.007) & .890 (.023) & .844 (.022) & .907 (.019) & .865 (.022) & .904 (.017) & .890 (.017) & \cellcolor{green!20}\textbf{.909 (.025)} & .184 (.010) & .882 (.019) & .883 (.008) \\  

  \SINGAPORE~MAS & .936 (.003) & \cellcolor{blue!20}.942 (.023) & .938 (.017) & .933 (.018) & .936 (.013) & .932 (.009) & .928 (.011) & .903 (.015) & .927 (.020) & .926 (.014) & .931 (.017) & .933 (.025) & \cellcolor{green!20}\textbf{.951 (.017)} & .371 (.050) & .920 (.009) & .900 (.011) \\  

  \POLAND~NBP & .887 (.011) & \cellcolor{blue!20}.901 (.015) & .879 (.028) & .894 (.007) & .899 (.010) & .887 (.013) & .887 (.006) & .868 (.009) & .880 (.016) & .861 (.011) & \cellcolor{green!20}\textbf{.913 (.005)} & .877 (.006) & .900 (.014) & .395 (.019) & .896 (.019) & .876 (.011) \\  

  \CHINA~PBoC & .914 (.011) & .913 (.008) & .907 (.006) & .919 (.012) & .904 (.019) & .913 (.019) & \cellcolor{blue!20}\textbf{.923 (.015)} & .863 (.017) & .914 (.008) & .900 (.013) & .908 (.021) & .903 (.026) & \cellcolor{green!20}.917 (.015) & .580 (.036) & .900 (.016) & .879 (.007) \\  

  \AUS~RBA & .913 (.028) & \cellcolor{blue!20}.935 (.022) & .909 (.004) & .920 (.027) & .924 (.013) & .918 (.035) & .930 (.033) & .891 (.021) & \cellcolor{green!20}\textbf{.935 (.009)} & .917 (.017) & .929 (.008) & .921 (.009) & .920 (.015) & .183 (.032) & .924 (.008) & .912 (.018) \\  

  \INDIA~RBI & .812 (.005) & .828 (.013) & .798 (.018) & .817 (.011) & .826 (.019) & .813 (.016) & \cellcolor{blue!20}\textbf{.835 (.021)} & .784 (.020) & .802 (.006) & .795 (.011) & .807 (.019) & .807 (.022) & .800 (.026) & .389 (.010) & \cellcolor{green!20}.807 (.010) & .790 (.012) \\  

  \SWITZERLAND~SNB & .837 (.020) & .843 (.013) & \cellcolor{blue!20}.850 (.014) & .848 (.010) & .836 (.013) & .827 (.010) & .838 (.011) & .800 (.018) & .853 (.019) & .854 (.009) & .850 (.023) & .859 (.013) & .853 (.006) & .425 (.047) & \cellcolor{green!20}\textbf{.860 (.011)} & .822 (.026) \\

  \midrule  

  Average & .863 (.049) & .866 (.052) & .860 (.053) & \cellcolor{blue!20}\textbf{.868 (.051)} & .866 (.050) & .861 (.053) & .865 (.054) & .834 (.054) & \cellcolor{green!20}.864 (.054) & .849 (.056) & .863 (.052) & .857 (.055) & .861 (.055) & .288 (.126) & 0.855 (0.055) & .846 (.052) \\  
  \bottomrule
\end{tabularx}
\end{table}

\clearpage
\textbf{Uncertainty Estimation Task}

\paragraph{General Setup – Uncertainty Estimation Analysis.}
Table \ref{tab:plm_f1_certain_all} reports F1-Score ($\pm$std) on the \CE for four base PLMs, three large PLMs, four closed‑source LLMs, and five open‑source LLMs, tested on pooled all‑banks data. \texttt{RoBERTa-Large} ($\overline{F1} = 0.846$) again ranks highest among PLMs, with \texttt{Llama-3-70B-Chat} ($\overline{F1} = 0.816$) performing best among LLMs. The overall F1-Scores for this task are higher than those for \SD, but the relative ordering across models resembles the \SD task: PLMs outperform LLMs on average. \CE proves to be more difficult than \TC due to lower average scores across all models.

\label{app:certain_all_result}
\begin{table}[h!]
    \centering
    \tiny
    \caption{F1-Scores across the banks for the \CE task with the standard deviation in the brackets. These results are based on the testing of a singular model on each bank's data. The best-performing PLM and LLM in each row is highlighted in blue and green, respectively. The best-performing overall model is bolded. The model names are abbreviated as described in Table \ref{tab:model_abb}, Appendix \ref{app:Model_abb}.}
    \label{tab:plm_f1_certain_all}
    \begin{tabularx}{\textwidth}{
    l 
    *{4}{X}                   
    : *{3}{X}                   
    | *{4}{X}                 
    : *{5}{X}                 
    }
    \toprule
    & \multicolumn{4}{c}{\textbf{Base}} 
    & \multicolumn{3}{c}{\textbf{Large}}
    & \multicolumn{4}{c}{\textbf{Closed-Source}} 
    & \multicolumn{5}{c}{\textbf{Open-Source}} \\
    \cmidrule(lr){2-5}\cmidrule(lr){6-8}\cmidrule(lr){9-12}\cmidrule(lr){13-17}
    \textbf{Bank} 
    & \MB MBB
    & \google BB
    & \finbert FB 
    & \meta RBB 
    & \MB MBL
    & \google BL
    & \meta RBL
    & \google Gem
    & \openai 4o
    & \openai 4.1m
    & \openai 4.1
    & \deepseek DS
    & \qwen Qwen
    & \FinMA FM
    & \meta L3
    & \meta L4S \\
    \midrule
\BRAZIL~BCB & .863 (.002) & .869 (.006) & .879 (.005) & .890 (.006) & .888 (.023) & \cellcolor{blue!20}\textbf{.894 (.031)} & .873 (.005) & .741 (.025) & .607 (.035) & .779 (.020) & .749 (.006) & .692 (.004) & .670 (.014) & .696 (.029) & \cellcolor{green!20}.819 (.010) & .766 (.019) \\
\PERU~BCRP & .912 (.032) & .902 (.030) & .916 (.034) & .913 (.037) & \cellcolor{blue!20}\textbf{.918 (.031)} & .916 (.032) & .903 (.032) & .852 (.011) & .816 (.030) & .857 (.025) & .847 (.024) & .838 (.034) & .807 (.027) & .884 (.026) & \cellcolor{green!20}.898 (.016) & .888 (.025) \\
\MALAYSIA~BNM & .813 (.036) & \cellcolor{blue!20}\textbf{.813 (.029)} & .805 (.028) & .812 (.026) & .808 (.029) & .799 (.046) & .812 (.020) & .693 (.006) & .676 (.002) & .769 (.009) & .793 (.010) & .760 (.015) & .616 (.045) & .660 (.021) & .784 (.010) & \cellcolor{green!20}.808 (.013) \\
\PHILIPPINES~BSP & .889 (.019) & .893 (.010) & .893 (.032) & .879 (.020) & .881 (.017) & \cellcolor{blue!20}\textbf{.894 (.010)} & .876 (.010) & .822 (.033) & .690 (.033) & .789 (.024) & .779 (.028) & .788 (.024) & .753 (.030) & .785 (.024) & \cellcolor{green!20}.877 (.024) & .828 (.007) \\
\COLOMBIA~BanRep & .868 (.013) & .862 (.010) & .850 (.023) & .872 (.018) & \cellcolor{blue!20}\textbf{.882 (.007)} & .864 (.006) & .880 (.011) & .822 (.019) & .737 (.040) & .843 (.005) & .821 (.031) & .798 (.022) & .764 (.031) & .718 (.019) & \cellcolor{green!20}.855 (.030) & .782 (.044) \\
\CANADA~BoC & .823 (.048) & .837 (.015) & .842 (.033) & .857 (.027) & .821 (.039) & .855 (.028) & \cellcolor{blue!20}\textbf{.858 (.014)} & .746 (.010) & .725 (.022) & \cellcolor{green!20}.825 (.031) & .799 (.035) & .737 (.014) & .702 (.010) & .506 (.041) & .751 (.011) & .699 (.035) \\
\UK~BoE & .692 (.019) & .710 (.036) & .696 (.033) & .676 (.031) & .700 (.028) & .713 (.024) & \cellcolor{blue!20}.721 (.025) & .669 (.035) & .663 (.019) & .681 (.043) & .687 (.043) & .662 (.025) & .698 (.028) & .595 (.036) & \cellcolor{green!20}\textbf{.730 (.022)} & .702 (.013) \\
\ISRAEL~BoI & .712 (.010) & .729 (.015) & .732 (.016) & \cellcolor{blue!20}\textbf{.733 (.005)} & .705 (.013) & .705 (.021) & .732 (.014) & .665 (.017) & .643 (.015) & .709 (.026) & .685 (.010) & .666 (.015) & .651 (.001) & .660 (.026) & \cellcolor{green!20}.724 (.005) & .673 (.015) \\
\JAPAN~BoJ & .823 (.022) & .835 (.006) & \cellcolor{blue!20}\textbf{.840 (.005)} & .825 (.010) & .828 (.004) & .836 (.005) & .821 (.018) & .621 (.025) & .553 (.028) & .625 (.042) & .652 (.029) & .512 (.045) & .604 (.048) & .670 (.015) & \cellcolor{green!20}.782 (.009) & .718 (.021) \\
\SOUTHKOREA~BoK & .856 (.031) & .860 (.031) & .861 (.021) & .866 (.028) & .865 (.014) & .846 (.018) & \cellcolor{blue!20}\textbf{.879 (.033)} & .658 (.026) & .561 (.054) & .650 (.030) & .721 (.031) & .582 (.042) & .637 (.027) & .686 (.020) & \cellcolor{green!20}.801 (.008) & .760 (.019) \\
\MEXICO~BdeM & .736 (.036) & .713 (.044) & .722 (.049) & \cellcolor{blue!20}.738 (.039) & .713 (.053) & .723 (.040) & .719 (.056) & .666 (.055) & .541 (.027) & .685 (.023) & .703 (.035) & .454 (.007) & .575 (.040) & .629 (.067) & \cellcolor{green!20}\textbf{.750 (.012)} & .639 (.051) \\
\THAILAND~BoT & .657 (.025) & .662 (.011) & .662 (.025) & .665 (.024) & \cellcolor{blue!20}\textbf{.683 (.009)} & .661 (.012) & .681 (.005) & .590 (.021) & .541 (.045) & .571 (.028) & .600 (.023) & .571 (.043) & .564 (.036) & .623 (.024) & \cellcolor{green!20}.662 (.011) & .630 (.031) \\
\TAIWAN~CBCT & .800 (.001) & .821 (.031) & \cellcolor{blue!20}\textbf{.823 (.035)} & .802 (.022) & .791 (.031) & .816 (.012) & .810 (.019) & .780 (.031) & .719 (.033) & .785 (.036) & \cellcolor{green!20}.800 (.026) & .792 (.026) & .757 (.025) & .652 (.020) & .784 (.039) & .756 (.011) \\
\EGYPT~CBE & \cellcolor{blue!20}\textbf{.920 (.002)} & .900 (.015) & .919 (.023) & .906 (.012) & .908 (.012) & .911 (.021) & .917 (.031) & .829 (.031) & .815 (.014) & .862 (.020) & .852 (.024) & .839 (.015) & .838 (.026) & .806 (.017) & \cellcolor{green!20}.904 (.027) & .874 (.021) \\
\RUSSIA~CBR & .910 (.032) & .913 (.034) & .895 (.027) & .908 (.023) & .899 (.017) & \cellcolor{blue!20}\textbf{.916 (.025)} & .902 (.022) & .701 (.043) & .662 (.029) & .778 (.030) & .767 (.055) & .807 (.054) & .713 (.037) & .817 (.014) & \cellcolor{green!20}.830 (.041) & .797 (.024) \\
\TURKEY~CBRT & .927 (.018) & .935 (.028) & .936 (.009) & .935 (.008) & .932 (.016) & .928 (.009) & \cellcolor{blue!20}\textbf{.951 (.010)} & .851 (.021) & .780 (.011) & .884 (.023) & .884 (.014) & .843 (.007) & .843 (.010) & .792 (.009) & \cellcolor{green!20}.906 (.019) & .870 (.014) \\
\CHILE~CBoC & .862 (.022) & .859 (.036) & .857 (.025) & \cellcolor{blue!20}\textbf{.866 (.021)} & .864 (.038) & .854 (.012) & .855 (.017) & .754 (.041) & .577 (.048) & .764 (.045) & .741 (.038) & .623 (.068) & .682 (.060) & .713 (.019) & \cellcolor{green!20}.816 (.028) & .705 (.042) \\
\EU~ECB & \cellcolor{blue!20}\textbf{.885 (.027)} & .863 (.031) & .878 (.023) & .876 (.029) & .858 (.025) & .861 (.058) & .881 (.036) & .735 (.019) & .645 (.006) & .722 (.027) & .762 (.023) & .669 (.039) & .735 (.010) & .701 (.036) & .841 (.007) & \cellcolor{green!20}.787 (.013) \\
\USA~FOMC & .808 (.020) & \cellcolor{blue!20}.838 (.012) & .791 (.025) & .809 (.020) & .818 (.010) & .792 (.031) & .835 (.019)  & .818 (.019) & .763 (.045) & .806 (.016) & .779 (.034) & .718 (.048) & .818 (.032) & .671 (.033) & \cellcolor{green!20}\textbf{.839 (.022)} & .825 (.011) \\
\SINGAPORE~MAS & .827 (.029) & .843 (.013) & \cellcolor{blue!20}.847 (.016) & .834 (.023) & .826 (.042) & .837 (.019) & .840 (.027) & .798 (.043) & .803 (.027) & .822 (.036) & .828 (.035) & .809 (.024) & .800 (.029) & .602 (.039) & \cellcolor{green!20}\textbf{.858 (.035)} & .857 (.031) \\
\POLAND~NBP & .840 (.016) & .850 (.016) & .838 (.015) & .856 (.006) & .846 (.015) & \cellcolor{blue!20}\textbf{.866 (.005)} & .851 (.011) & .760 (.013) & .610 (.021) & .725 (.007) & .745 (.023) & .623 (.026) & .689 (.033) & .607 (.023) & \cellcolor{green!20}.813 (.016) & .774 (.019) \\
\CHINA~PBoC & .944 (.008) & .948 (.011) & .945 (.012) & .940 (.004) & .943 (.007) & .945 (.006) & \cellcolor{blue!20}.950 (.007) & .641 (.046) & .749 (.017) & .851 (.013) & .801 (.023) & .869 (.022) & .844 (.023) & .927 (.021) & \cellcolor{green!20}\textbf{.955 (.010)} & .942 (.012) \\
\AUS~RBA & .867 (.057) & .882 (.054) & .834 (.050) & \cellcolor{blue!20}\textbf{.894 (.043)} & .842 (.068) & .865 (.054) & .890 (.021) & .847 (.015) & .782 (.021) & .851 (.031) & .852 (.032) & .778 (.019) & .824 (.029) & .671 (.022) & \cellcolor{green!20}.854 (.039) & .847 (.019) \\
\INDIA~RBI & \cellcolor{blue!20}\textbf{.862 (.028)} & .835 (.027) & .850 (.023) & .834 (.015) & .851 (.025) & .833 (.018) & .851 (.023)  & .662 (.018) & .584 (.017) & .700 (.024) & .686 (.021) & .616 (.020) & .599 (.042) & .750 (.030) & \cellcolor{green!20}.755 (.023) & .702 (.029) \\
\SWITZERLAND~SNB & .853 (.014) & .868 (.015) & .863 (.011) & \cellcolor{blue!20}\textbf{.869 (.015)} & .855 (.019) & .852 (.014) & .864 (.017) & .703 (.017) & .680 (.001) & .779 (.007) & .759 (.010) & .721 (.015) & .651 (.010) & .733 (.008) & \cellcolor{green!20}.805 (.006) & .779 (.025) \\
\midrule
Average & .838 (.073) & .842 (.071) & .839 (.072) & .842 (.073) & .837 (.072) & .839 (.073) & \cellcolor{blue!20}\textbf{.846 (.070)}  & .737 (.077) & .677 (.088) & .764 (.079) & .764 (.067) & .711 (.109) & .713 (.087) & .702 (.093) & \cellcolor{green!20}.816 (.066) & .776 (.078) \\
    \bottomrule
\end{tabularx}
\end{table}

\clearpage
\subsubsection{Specific Bank Setup Models}
\label{app:specific_model}
\textbf{Stance Label}

\paragraph{Bank-Specific Setup – Stance Detection.}
Table \ref{tab:plm_f1_stance_specific} shows bank‑by‑bank F1-Score ($\pm$ std) for \hawk/\dov/\neut/\irr, comparing four base PLMs, three large PLMs, four closed‑source LLMs, and five open‑source LLMs, each fine‑tuned on that bank’s data. \texttt{RoBERTa-Large} ($\overline{F1} = 0.694$) remains the best-performing PLM, and \texttt{Llama-3-70B-Chat} ($\overline{F1} = 0.620$) leads the LLMs, consistent with the results from the General Setup. The relative ranking of models is largely preserved, with PLMs outperforming LLMs overall. While the performance gap between PLMs and LLMs narrows slightly (the mean difference is $0.074$) in this setting, it remains clear that PLMs benefit more from fine-tuning on the aggregated dataset. It is important to note that LLMs are evaluated using the same zero-shot prompts in both the General and Bank-Specific Setups.

\label{app:stance_specific_result}
\begin{table}[h!]
\tiny
\centering
\caption{{F1-Scores of models classifying sentences from each central bank with the monetary policy stance label of \hawk/\dov/\neut/\irr with standard deviation in the parenthesis. The best-performing PLM and LLM in each row is highlighted in blue and green, respectively. The best-performing overall model is bolded. The model names are abbreviated as described in Table \ref{tab:model_abb}, Appendix \ref{app:Model_abb}.}}
\label{tab:plm_f1_stance_specific}
    \begin{tabularx}{\textwidth}{
    l 
    *{4}{X}                   
    : *{3}{X}                   
    | *{4}{X}                 
    : *{5}{X}                 
    }
    \toprule
    & \multicolumn{4}{c}{\textbf{Base}} 
    & \multicolumn{3}{c}{\textbf{Large}}
    & \multicolumn{4}{c}{\textbf{Closed-Source}} 
    & \multicolumn{5}{c}{\textbf{Open-Source}} \\
    \cmidrule(lr){2-5}\cmidrule(lr){6-8}\cmidrule(lr){9-12}\cmidrule(lr){13-17}
    \textbf{Bank} 
    & \MB MBB
    & \google BB
    & \finbert FB 
    & \meta RBB 
    & \MB MBL
    & \google BL
    & \meta RBL
    & \google Gem
    & \openai 4o
    & \openai 4.1m
    & \openai 4.1
    & \deepseek DS
    & \qwen Qwen
    & \FinMA FM
    & \meta L3
    & \meta L4S \\
    \midrule
\BRAZIL~BCB & .525 (.023) & .456 (.023) & .518 (.067) & .572 (.042) & .507 (.055) & .527 (.064) & \cellcolor{blue!20}\textbf{.623 (.012)} & .528 (.027) & .498 (.017) & .462 (.020) & .504 (.017) & \cellcolor{green!20}.613 (.021) & .525 (.028) & .350 (.034) & .503 (.021) & .589 (.049) \\
\PERU~BCRP & .721 (.035) & .769 (.029) & .768 (.033) & .784 (.032) & .729 (.089) & .741 (.030) & \cellcolor{blue!20}\textbf{.809 (.032)} & \cellcolor{green!20}.675 (.004) & .628 (.008) & .634 (.035) & .665 (.004) & .666 (.010) & .503 (.062) & .301 (.031) & .641 (.014) & .620 (.043) \\
\MALAYSIA~BNM & .590 (.046) & .632 (.076) & .600 (.062) & .630 (.023) & \cellcolor{blue!20}\textbf{.667 (.026)} & .607 (.078) & .642 (.058) & .409 (.006) & .443 (.027) & .430 (.007) & .409 (.025) & .475 (.013) & .333 (.026) & .160 (.033) & \cellcolor{green!20}.567 (.025) & .435 (.005) \\
\PHILIPPINES~BSP & .521 (.021) & .608 (.076) & .667 (.073) & .677 (.043) & .540 (.028) & .560 (.073) & \cellcolor{blue!20}\textbf{.698 (.048)} & .424 (.042) & .420 (.069) & .451 (.039) & .514 (.076) & .534 (.029) & .380 (.015) & .245 (.027) & \cellcolor{green!20}.584 (.042) & .500 (.035) \\
\COLOMBIA~BanRep & .472 (.009) & .585 (.031) & .644 (.025) & .637 (.088) & .515 (.058) & .558 (.042) & \cellcolor{blue!20}\textbf{.691 (.037)} & .515 (.021) & .455 (.031) & .520 (.036) & .553 (.015) & .570 (.037) & .450 (.023) & .230 (.033) & \cellcolor{green!20}.573 (.038) & .423 (.033) \\
\CANADA~BoC & .582 (.073) & .633 (.091) & .653 (.021) & .693 (.034) & .635 (.050) & .708 (.022) & \cellcolor{blue!20}\textbf{.728 (.009)} & .629 (.069) & .647 (.052) & .641 (.052) & .657 (.028) & .657 (.026) & .524 (.038) & .264 (.029) & \cellcolor{green!20}.669 (.043) & .644 (.024) \\
\UK~BoE & .560 (.033) & .595 (.054) & .646 (.049) & .677 (.049) & .659 (.041) & .653 (.018) & \cellcolor{blue!20}\textbf{.702 (.041)} & .543 (.026) & .524 (.031) & .537 (.048) & .602 (.031) & .543 (.070) & .396 (.021) & .129 (.026) & \cellcolor{green!20}.661 (.044) & .518 (.045) \\
\ISRAEL~BoI & .460 (.082) & .543 (.050) & .556 (.062) & .587 (.019) & .539 (.071) & .570 (.044) & \cellcolor{blue!20}\textbf{.635 (.048) } & .474 (.011) & .460 (.005) & .433 (.032) & .526 (.013) & .482 (.001) & .329 (.025) & .085 (.011) & \cellcolor{green!20}.594 (.023) & .430 (.008) \\
\JAPAN~BoJ & .495 (.009) & .510 (.069) & .602 (.040) & \cellcolor{blue!20}\textbf{.608 (.017) } & .544 (.036) & .515 (.034) & .606 (.032) & .524 (.010) & .545 (.021) & .465 (.028) & .565 (.008) & .498 (.009) & .406 (.040) & .157 (.033) & \cellcolor{green!20}.574 (.027) & .507 (.026) \\
\SOUTHKOREA~BoK & .556 (.048) & .572 (.077) & .548 (.056) & .645 (.012) & .581 (.026) & .610 (.030) & \cellcolor{blue!20}\textbf{.704 (.052)} & .646 (.040) & .648 (.030) & .594 (.066) & \cellcolor{green!20}.678 (.016) & .629 (.016) & .466 (.076) & .181 (.031) & .632 (.032) & .592 (.047) \\
\MEXICO~BdeM & .517 (.026) & .437 (.082) & .596 (.009) & .618 (.051) & .559 (.031) & .536 (.022) & \cellcolor{blue!20}\textbf{.684 (.059) } & .596 (.009) & .602 (.024) & .509 (.023) & .626 (.016) & \cellcolor{green!20}.669 (.034) & .447 (.047) & .118 (.018) & .642 (.012) & .552 (.013) \\
\THAILAND~BoT & .597 (.010) & .604 (.033) & .666 (.025) & .654 (.118) & .636 (.045) & .599 (.014) & \cellcolor{blue!20}\textbf{.680 (.065)  } & .547 (.004) & .549 (.032) & .551 (.039) & .573 (.012) & .581 (.029) & .484 (.038) & .258 (.026) & \cellcolor{green!20}.596 (.009) & .577 (.030) \\
\TAIWAN~CBCT & .488 (.067) & .564 (.014) & .613 (.027) & .631 (.037) & .564 (.054) & .557 (.038) & \cellcolor{blue!20}\textbf{.644 (.032)  } & .451 (.044) & .474 (.044) & .474 (.031) & .485 (.026) & .522 (.022) & .388 (.049) & .180 (.037) & \cellcolor{green!20}.556 (.015) & .475 (.051) \\
\EGYPT~CBE & .708 (.052) & .676 (.029) & .729 (.076) & .749 (.014) & .733 (.035) & .736 (.062) & \cellcolor{blue!20}\textbf{.783 (.028)  } & .629 (.037) & .672 (.036) & .581 (.045) & .648 (.036) & .636 (.007) & .352 (.056) & .142 (.014) & \cellcolor{green!20}.702 (.021) & .594 (.024) \\
\RUSSIA~CBR & .675 (.085) & .652 (.042) & .666 (.059) & .750 (.035) & .671 (.085) & .716 (.027) & \cellcolor{blue!20}\textbf{.796 (.042)  } & .759 (.015) & .749 (.027) & .693 (.026) & .701 (.049) & \cellcolor{green!20}.794 (.028) & .573 (.035) & .146 (.022) & .772 (.029) & .665 (.013) \\
\TURKEY~CBRT & .505 (.010) & .600 (.018) & .665 (.067) & .673 (.015) & .596 (.034) & .650 (.044) & \cellcolor{blue!20}\textbf{.716 (.048)  } & .495 (.006) & .421 (.014) & .424 (.018) & .475 (.015) & .539 (.030) & .277 (.006) & .133 (.020) & \cellcolor{green!20}.653 (.036) & .416 (.032) \\
\CHILE~CBoC & .485 (.106) & .574 (.041) & .606 (.069) & .653 (.019) & .569 (.041) & .603 (.082) & \cellcolor{blue!20}\textbf{.721 (.085)  } & .668 (.032) & .604 (.027) & .605 (.033) & .678 (.057) & .676 (.038) & .559 (.072) & .223 (.040) & \cellcolor{green!20}.685 (.019) & .539 (.071) \\
\EU~ECB & .479 (.028) & .491 (.036) & .530 (.027) & .601 (.013) & .533 (.119) & .518 (.040) & \cellcolor{blue!20}.654 (.006)  & .638 (.023) & .599 (.019) & .610 (.038) & \cellcolor{green!20}\textbf{.660 (.021)} & .637 (.005) & .548 (.017) & .206 (.052) & .613 (.020) & .595 (.016) \\
\USA~FOMC & .455 (.008) & .505 (.022) & .559 (.046) & .589 (.030) & .558 (.067) & .563 (.012) & \cellcolor{blue!20}\textbf{.691 (.011)  } & .572 (.021) & .584 (.025) & .564 (.018) & .649 (.023) & \cellcolor{green!20}.653 (.023) & .512 (.023) & .170 (.025) & .599 (.012) & .498 (.015) \\
\SINGAPORE~MAS & .492 (.009) & .598 (.065) & .632 (.072) & \cellcolor{blue!20}\textbf{.704 (.010)} & .601 (.035) & .589 (.003) & .690 (.034) & .553 (.046) & .581 (.026) & .588 (.034) & .569 (.015) & \cellcolor{green!20}.689 (.041) & .540 (.026) & .347 (.024) & .638 (.035) & .646 (.023) \\
\POLAND~NBP & .586 (.015) & .540 (.017) & .594 (.014) & .666 (.026) & .555 (.060) & .617 (.027) & \cellcolor{blue!20}\textbf{.674 (.045)  } & .637 (.015) & .631 (.043) & .614 (.063) & \cellcolor{green!20}.665 (.031) & .660 (.002) & .508 (.028) & .118 (.017) & .618 (.035) & .597 (.015) \\
\CHINA~PBoC & .727 (.025) & .742 (.030) & .736 (.023) & .762 (.025) & .744 (.019) & .730 (.014) & \cellcolor{blue!20}\textbf{.764 (.030)} & .492 (.046) & .559 (.037) & .531 (.026) & .535 (.033) & .592 (.037) & .379 (.017) & .128 (.018) & \cellcolor{green!20}.613 (.033) & .446 (.024) \\
\AUS~RBA & .531 (.072) & .516 (.041) & .586 (.059) & \cellcolor{blue!20}\textbf{.685 (.023)} & .549 (.024) & .553 (.022) & .642 (.038) & .531 (.049) & .478 (.079) & .483 (.058) & .553 (.074) & .537 (.055) & .358 (.049) & .133 (.020) & \cellcolor{green!20}.614 (.034) & .495 (.057) \\
\INDIA~RBI & .519 (.050) & .593 (.015) & .584 (.043) & .640 (.056) & .608 (.067) & .590 (.038) & \cellcolor{blue!20}\textbf{.673 (.043)} & .489 (.025) & .519 (.041) & .509 (.026) & .495 (.027) & .542 (.016) & .431 (.008) & .231 (.058) & \cellcolor{green!20}.581 (.030) & .557 (.032) \\
\SWITZERLAND~SNB & .573 (.026) & .612 (.046) & .680 (.023) & .653 (.022) & .530 (.056) & .674 (.013) & \cellcolor{blue!20}\textbf{.704 (.025)} & .635 (.003) & .601 (.015) & .640 (.037) & \cellcolor{green!20}.652 (.016) & .643 (.024) & .554 (.029) & .252 (.008) & .612 (.014) & .607 (.051) \\
\midrule
~Average & .553 (.081) & .584 (.077) & .636 (.064) & .662 (.056) & .597 (.070) & .611 (.071) & \cellcolor{blue!20}\textbf{.694 (.052)} & .562 (.085) & .556 (.084) & .542 (.075) & .586 (.078) & .601 (.075) & .449 (.083) & .196 (.071) & \cellcolor{green!20}.620 (.053) & .541 (.074) \\
\bottomrule
\end{tabularx}
\end{table}

\clearpage
\textbf{Temporal Classification Label}

\paragraph{Bank-Specific Setup – Temporal Classification Analysis.}
Table \ref{tab:plm_f1_time_specific} reports per‑bank F1-Score ($\pm$std) for the \TC  across four base PLMs, three large PLMs, four closed‑source LLMs, and five open‑source LLMs—each fine‑tuned solely on that bank’s data. \texttt{BERT-Large} ($\overline{F1} = 0.861$) achieves the best PLM score for \TC, while \texttt{GPT-4o} ($\overline{F1} = 0.864$) leads among LLMs. The average performance between PLMs and LLMs is very similar ($<$0.005 F1-Score on the best performing models). 

\label{app:time_specific_result}
\begin{table}[h!]
\centering
\tiny
\caption{{F1-Scores of models classifying sentences from each central bank as \forward/\notforward, with standard deviation in the parenthesis. The best-performing PLM and LLM in each row is highlighted in blue and green, respectively. The best-performing overall model is bolded. The model names are abbreviated as described in Table \ref{tab:model_abb}, Appendix \ref{app:Model_abb}.}}
\label{tab:plm_f1_time_specific}
    \begin{tabularx}{\textwidth}{
    l 
    *{4}{X}                   
    : *{3}{X}                   
    | *{4}{X}                 
    : *{5}{X}                 
    }
    \toprule
    & \multicolumn{4}{c}{\textbf{Base}} 
    & \multicolumn{3}{c}{\textbf{Large}}
    & \multicolumn{4}{c}{\textbf{Closed-Source}} 
    & \multicolumn{5}{c}{\textbf{Open-Source}} \\
    \cmidrule(lr){2-5}\cmidrule(lr){6-8}\cmidrule(lr){9-12}\cmidrule(lr){13-17}
    \textbf{Bank} 
    & \MB MBB
    & \google BB
    & \finbert FB 
    & \meta RBB 
    & \MB MBL
    & \google BL
    & \meta RBL
    & \google Gem
    & \openai 4o
    & \openai 4.1m
    & \openai 4.1
    & \deepseek DS
    & \qwen Qwen
    & \FinMA FM
    & \meta L3
    & \meta L4S \\
    \midrule
\BRAZIL~BCB    & .818 (.027) & .821 (.060) & .863 (.015) & .843 (.050) & .840 (.021) & .866 (.027) & \cellcolor{blue!20}\textbf{.869 (.023)} & .835 (.031) & .840 (.025) & .809 (.027) & .822 (.021) & \cellcolor{green!20}.845 (.014) & .830 (.027) & .182 (.010) & .821 (.005) & .826 (.014) \\
\PERU~BCRP     & .866 (.044) & .867 (.047) & .868 (.038) & .847 (.028) & .872 (.024) & .867 (.036) & \cellcolor{blue!20}\textbf{.882 (.034)} & .847 (.022) & .861 (.016) & .855 (.032) & \cellcolor{green!20}.869 (.034) & .851 (.037) & .838 (.020) & .113 (.014) & .854 (.040) & .848 (.036) \\
\MALAYSIA~BNM  & .850 (.025) & .867 (.029) & .877 (.006)  & .877 (.003) & \cellcolor{blue!20}\textbf{.899 (.010)} & .878 (.034) & .868 (.023) & .856 (.018) & .877 (.016) & .868 (.017) & \cellcolor{green!20}.889 (.021) & .880 (.020) & .872 (.028) & .553 (.013) & .888 (.025) & .879 (.025) \\
\PHILIPPINES~BSP  & .906 (.017) & .899 (.016) & .894 (.031) & .904 (.030) & .913 (.011) & .908 (.012) & \cellcolor{blue!20}.916 (.016) & .886 (.012) & \cellcolor{green!20}\textbf{.916 (.013)} & .894 (.020) & .912 (.019) & .911 (.019) & .916 (.015) & .166 (.016) & .905 (.011) & .901 (.011) \\
\COLOMBIA~BanRep  & .857 (.045) & .849 (.013) & .863 (.013) & .847 (.015) & .865 (.038) & \cellcolor{blue!20}\textbf{.896 (.036)} & .868 (.018) & .845 (.054) & \cellcolor{green!20}.890 (.045) & .884 (.020) & .885 (.042) & .875 (.026) & .881 (.025) & .148 (.024) & .877 (.029) & .867 (.036) \\
\CANADA~BoC   & .856 (.056) & .852 (.050) & .864 (.059) & .864 (.078) & .869 (.042) & \cellcolor{blue!20}.879 (.050) & .878 (.041) & .852 (.051) & \cellcolor{green!20}\textbf{.882 (.035)} & .830 (.030) & .871 (.018) & .861 (.037) & .876 (.036) & .353 (.023) & .858 (.036) & .848 (.028) \\
\UK~BoE      & .736 (.028) & .732 (.031) & .777 (.020) & .760 (.014) & .757 (.003)  & \cellcolor{blue!20}.781 (.009) & .761 (.029) & .715 (.018) & .761 (.008) & .771 (.022) & \cellcolor{green!20}\textbf{.793 (.014)} & .763 (.019) & .777 (.003) & .142 (.030) & .765 (.017) & .766 (.006) \\
\ISRAEL~BoI  & .822 (.021) & .803 (.046) & .847 (.003) & .850 (.063) & \cellcolor{blue!20}.873 (.036) & .855 (.030) & .864 (.038) & .829 (.026) & .875 (.028) & \cellcolor{green!20}\textbf{.896 (.025)} & .879 (.014) & .884 (.036) & .858 (.021) & .260 (.024) & .851 (.018) & .875 (.027) \\
\JAPAN~BoJ   & .776 (.033) & .765 (.013) & .785 (.015) & .752 (.002) & .771 (.006)  & \cellcolor{blue!20}\textbf{.792 (.022)} & .767 (.025) & .753 (.008) & .786 (.022) & .775 (.037) & .778 (.017) & .778 (.017) & \cellcolor{green!20}.789 (.037) & .322 (.017) & .763 (.031) & .764 (.020) \\
\SOUTHKOREA~BoK  & .861 (.012) & .854 (.024) & .879 (.011) & .852 (.021) & .859 (.007)  & .874 (.022) & \cellcolor{blue!20}.881 (.010) & .872 (.005) & .900 (.009) & .885 (.030) & .900 (.009) & \cellcolor{green!20}\textbf{.908 (.012)} & .904 (.008) & .369 (.020) & .891 (.008) & .853 (.014) \\
\MEXICO~BdeM  & .801 (.032) & .794 (.026) & .829 (.044) & .822 (.058) & .821 (.016)  & \cellcolor{blue!20}.864 (.024) & .837 (.035) & .809 (.016) & \cellcolor{green!20}\textbf{.875 (.018)} & .862 (.033) & .858 (.027) & .840 (.038) & .862 (.028) & .307 (.019) & .862 (.028) & .830 (.026) \\
\THAILAND~BoT & .712 (.062) & .700 (.034) & .734 (.085) & .719 (.040) & .719 (.061)  & .734 (.086) & \cellcolor{blue!20}.742 (.067) & .738 (.025) & .748 (.029) & .722 (.025) & \cellcolor{green!20}\textbf{.752 (.025)} & .732 (.035) & .751 (.018) & .427 (.007) & .727 (.033) & .735 (.031) \\
\TAIWAN~CBCT & .783 (.041) & .786 (.016) & .807 (.047) & \cellcolor{blue!20}.829 (.032) & .824 (.041)  & .816 (.046) & .817 (.038) & .831 (.028) & \cellcolor{green!20}\textbf{.831 (.019)} & .810 (.027) & .820 (.019) & .822 (.037) & .818 (.035) & .306 (.025) & .825 (.021) & .809 (.015) \\
\EGYPT~CBE   & .927 (.014) & \cellcolor{blue!20}.930 (.020) & .913 (.022) & .912 (.036) & .919 (.009)  & .928 (.027) & .925 (.006) & .913 (.011) & .935 (.006) & .943 (.011) & .933 (.009) & \cellcolor{green!20}\textbf{.943 (.014)} & .928 (.007) & .182 (.025) & .935 (.008) & .936 (.006) \\
\RUSSIA~CBR  & .874 (.026) & .876 (.048) & \cellcolor{blue!20}\textbf{.901 (.023)} & .891 (.042) & .888 (.031)  & .888 (.037) & .887 (.042) & .857 (.040) & .851 (.008) & .838 (.029) & .849 (.019) & .840 (.016) & .857 (.031) & .235 (.027) & .851 (.027) & \cellcolor{green!20}.861 (.024) \\
\TURKEY~CBRT & .905 (.021) & .904 (.039) & .910 (.012) & .906 (.022) & .898 (.003)  & \cellcolor{blue!20}.925 (.015) & .907 (.020) & .919 (.006) & \cellcolor{green!20}\textbf{.936 (.008)} & .901 (.015) & .922 (.027) & .919 (.010) & .931 (.006) & .132 (.024) & .935 (.009) & .915 (.015) \\
\CHILE~CBoC  & .773 (.009) & .777 (.045) & .790 (.037) & .758 (.023) & .760 (.007)  & \cellcolor{blue!20}.797 (.013) & .787 (.026) & .798 (.009) & \cellcolor{green!20}\textbf{.835 (.007)} & .793 (.021) & .832 (.008) & .813 (.020) & .809 (.027) & .214 (.017) & .797 (.007) & .794 (.011) \\
\EU~ECB      & .770 (.073) & .706 (.044) & .789 (.040) & .752 (.038) & .774 (.040)  & \cellcolor{blue!20}\textbf{.808 (.047)} & .803 (.047) & .741 (.045) & .781 (.022) & .763 (.023) & .784 (.014) & .771 (.013) & .779 (.023) & .267 (.051) & \cellcolor{green!20}.788 (.019) & .780 (.011) \\
\USA~FOMC    & .850 (.035) & .863 (.040) & .886 (.027) & .866 (.020) & .875 (.040)  & .866 (.022) & \cellcolor{blue!20}.895 (.033) & .844 (.022) & .907 (.019) & .865 (.022) & .904 (.017) & .890 (.017) & \cellcolor{green!20}\textbf{.909 (.025)} & .184 (.010) & .882 (.019) & .883 (.008) \\
\SINGAPORE~MAS  & .920 (.013) & .929 (.020) & .944 (.011) & .944 (.022) & \cellcolor{blue!20}\textbf{.960 (.015) } & .942 (.023) & .938 (.017) & .903 (.015) & .927 (.020) & .926 (.014) & .931 (.017) & .933 (.025) & \cellcolor{green!20}.951 (.017) & .371 (.050) & .920 (.009) & .900 (.011) \\
\POLAND~NBP  & .861 (.004) & .873 (.011) & .886 (.032) & .863 (.018) & .871 (.020)  & .875 (.034) & \cellcolor{blue!20}.908 (.028) & .868 (.009) & .880 (.016) & .861 (.011) & \cellcolor{green!20}\textbf{.913 (.005)} & .877 (.006) & .900 (.014) & .395 (.019) & .896 (.019) & .876 (.011) \\
\CHINA~PBoC  & \cellcolor{blue!20}\textbf{.919 (.017) }& .913 (.024) & .907 (.023) & .911 (.011) & .916 (.019)  & .912 (.008) & .910 (.018) & .863 (.017) & .914 (.008) & .900 (.013) & .908 (.021) & .903 (.026) & \cellcolor{green!20}.917 (.015) & .580 (.036) & .900 (.016) & .879 (.007) \\
\AUS~RBA     & .913 (.025) & .917 (.040) & .909 (.021) & \cellcolor{blue!20}.922 (.026) & .893 (.029)  & .916 (.037) & \cellcolor{blue!20}.922 (.038)  & .891 (.021) & \cellcolor{green!20}\textbf{.935 (.009)} & .917 (.017) & .929 (.008) & .921 (.009) & .920 (.015) & .183 (.032) & .924 (.008) & .912 (.018) \\
\INDIA~RBI   & .781 (.016) & .780 (.042) & .807 (.042) & .792 (.022) & .822 (.028)  & \cellcolor{blue!20}\textbf{.833 (.014) }& .803 (.034) & .784 (.020) & .802 (.006) & .795 (.011) & .807 (.019) & .807 (.022) & .800 (.026) & .389 (.010) & \cellcolor{green!20}.807 (.010) & .790 (.012) \\
\SWITZERLAND~SNB  & .811 (.018) & .794 (.039) & .817 (.035) & .773 (.038) & .823 (.030)  & .819 (.021) & \cellcolor{blue!20}.846 (.016)  & .800 (.018) & .853 (.019) & .854 (.009) & .850 (.023) & .859 (.013) & .853 (.006) & .425 (.047) & \cellcolor{green!20}\textbf{.860 (.011)} & .822 (.026) \\
\midrule
Average & .838 (.061) & .834 (.068) & .854 (.053) & .842 (.062) & .851 (.059) & \cellcolor{blue!20}.861 (.052) & .859 (.056) & .834 (.054) & \cellcolor{green!20}\textbf{.864 (.054)} & .849 (.056) & .863 (.052) & .857 (.055) & .861 (.055) & .288 (.126) & 0.855 (0.055) & .846 (.052) \\
\bottomrule
\end{tabularx}
\end{table}

\clearpage
\textbf{Uncertainty Estimation}

\paragraph{Bank-Specific Setup – Uncertainty Estimation Analysis.}
Table \ref{tab:plm_f1_certain_specific} reports per‑bank F1-Score ($\pm$std) on the \CE task for four base PLMs, three large PLMs, four closed‑source LLMs, and five open‑source LLMs—each fine‑tuned only on that bank’s data. The \CE task exhibits the highest variance across models. \texttt{Llama-3-70B-Chat} ($\overline{F1} = 0.816$) is the best-performing LLM, while PLM performance is more mixed. \texttt{RoBERTa-Large} ($\overline{F1} = 0.820$) and \texttt{FinBERT-Pretrain} ($\overline{F1} = 0.820$) achieve the best single-bank result. \texttt{RoBERTa-Large} has a higher standard deviation (0.100) compared to \texttt{FinBERT-Pretrain} (0.093) but PLMs on average still outperform LLMs (the mean difference is $0.004$). These findings indicate that \CE as a task is especially sensitive to changes in central bank phrasing, which hampers the models’ ability to generalize on this task.

\label{app:certain_specific_result}
\begin{table}[h!]
\centering
\tiny
\caption{{F1-Scores of models classifying sentences from each central bank as \uncertain/\certain, with standard deviation in the parenthesis. The best-performing PLM and LLM in each row is highlighted in blue and green, respectively. The best-performing overall model is bolded. The model names are abbreviated as described in Table \ref{tab:model_abb}, Appendix \ref{app:Model_abb}.}}
\label{tab:plm_f1_certain_specific}
    \begin{tabularx}{\textwidth}{
    l 
    *{4}{X}                   
    : *{3}{X}                   
    | *{4}{X}                 
    : *{5}{X}                 
    }
    \toprule
    & \multicolumn{4}{c}{\textbf{Base}} 
    & \multicolumn{3}{c}{\textbf{Large}}
    & \multicolumn{4}{c}{\textbf{Closed-Source}} 
    & \multicolumn{5}{c}{\textbf{Open-Source}} \\
    \cmidrule(lr){2-5}\cmidrule(lr){6-8}\cmidrule(lr){9-12}\cmidrule(lr){13-17}
    \textbf{Bank} 
    & \MB MBB
    & \google BB
    & \finbert FB 
    & \meta RBB 
    & \MB MBL
    & \google BL
    & \meta RBL
    & \google Gem
    & \openai 4o
    & \openai 4.1m
    & \openai 4.1
    & \deepseek DS
    & \qwen Qwen
    & \FinMA FM
    & \meta L3
    & \meta L4S \\
    \midrule
\BRAZIL~BCB    & .804 (.052) & .823 (.053) & \cellcolor{blue!20}\textbf{.876 (.034) }& .845 (.058) & .860 (.029) & .798 (.110) & .861 (.007) & .741 (.025) & .607 (.035) & .779 (.020) & .749 (.006) & .692 (.004) & .670 (.014) & .696 (.029) & \cellcolor{green!20}{.819 (.010)} & .766 (.019) \\
\PERU~BCRP     & .910 (.032) & .908 (.036) & .912 (.033) & .915 (.035) & \cellcolor{blue!20}\textbf{.923 (.027)} & .920 (.016) & .916 (.038) & .852 (.011) & .816 (.030) & .857 (.025) & .847 (.024) & .838 (.034) & .807 (.027) & .884 (.026) & \cellcolor{green!20}.898 (.016) & .888 (.025) \\
\MALAYSIA~BNM  & .738 (.021) & .762 (.027) & .790 (.039) & .761 (.036) & .808 (.007) & .807 (.041) & \cellcolor{blue!20}\textbf{.809 (.024)} & .693 (.006) & .676 (.002) & .769 (.009) & .793 (.010) & .760 (.015) & .616 (.045) & .660 (.021) & .784 (.010) & \cellcolor{green!20}.808 (.013) \\
\PHILIPPINES~BSP  & \cellcolor{blue!20}\textbf{.886 (.024) }& .858 (.026) & .882 (.014) & .873 (.026) & .879 (.044) & .867 (.054) & .869 (.026)  & .822 (.033) & .690 (.033) & .789 (.024) & .779 (.028) & .788 (.024) & .753 (.030) & .785 (.024) & \cellcolor{green!20}{.877 (.024)} & .828 (.007) \\
\COLOMBIA~BanRep  & .792 (.041) & .845 (.013) & .848 (.021) & .832 (.019) & .843 (.024) & .850 (.012) & \cellcolor{blue!20}\textbf{.868 (.020)} & .822 (.019) & .737 (.040) & .843 (.005) & .821 (.031) & .798 (.022) & .764 (.031) & .718 (.019) & \cellcolor{green!20}.855 (.030) & .782 (.044) \\
\CANADA~BoC   & .794 (.040) & .839 (.033) & .838 (.020) & \cellcolor{blue!20}\textbf{.859 (.029) }& .849 (.028) & .819 (.074) & .843 (.042) & .746 (.010) & .725 (.022) & \cellcolor{green!20}{.825 (.031)} & .799 (.035) & .737 (.014) & .702 (.010) & .506 (.041) & .751 (.011) & .699 (.035) \\
\UK~BoE      & .641 (.012) & .681 (.026) & \cellcolor{blue!20}{.699 (.025) }& .656 (.083) & .670 (.051) & .677 (.040) & .694 (.067) & .669 (.035) & .663 (.019) & .681 (.043) & .687 (.043) & .662 (.025) & .698 (.028) & .595 (.036) & \cellcolor{green!20}\textbf{.730 (.022)} & .702 (.013) \\
\ISRAEL~BoI  & \cellcolor{blue!20}{.644 (.047) }& .633 (.040) & .624 (.034) & .622 (.037) & .624 (.034) & .624 (.034) & .624 (.034) & .665 (.017) & .643 (.015) & .709 (.026) & .685 (.010) & .666 (.015) & .651 (.001) & .660 (.026) & \cellcolor{green!20}\textbf{.724 (.005)} & .673 (.015) \\
\JAPAN~BoJ   & .767 (.065) & .779 (.036) & .751 (.054) & .747 (.046) & .775 (.047) & \cellcolor{blue!20}\textbf{.786 (.031) }& .783 (.046) & .621 (.025) & .553 (.028) & .625 (.042) & .652 (.029) & .512 (.045) & .604 (.048) & .670 (.015) & \cellcolor{green!20}{.782 (.009)} & .718 (.021) \\
\SOUTHKOREA~BoK  & .801 (.049) & .837 (.034) & \cellcolor{blue!20}\textbf{.853 (.018) }& .779 (.108) & .828 (.029) & .771 (.105) & .844 (.048) & .658 (.026) & .561 (.054) & .650 (.030) & .721 (.031) & .582 (.042) & .637 (.027) & .686 (.020) & \cellcolor{green!20}{.801 (.008)} & .760 (.019) \\
\MEXICO~BdeM  & .651 (.017) & .620 (.053) & .673 (.094) & .682 (.035) & .656 (.062) & \cellcolor{blue!20}{.702 (.054) }& .625 (.048) & .666 (.055) & .541 (.027) & .685 (.023) & .703 (.035) & .454 (.007) & .575 (.040) & .629 (.067) & \cellcolor{green!20}\textbf{.750 (.012)} & .639 (.051) \\
\THAILAND~BoT & .520 (.013) & .563 (.038) & \cellcolor{blue!20}{.589 (.023) }& .567 (.051) & .552 (.052) & .539 (.051) & .530 (.087) & .590 (.021) & .541 (.045) & .571 (.028) & .600 (.023) & .571 (.043) & .564 (.036) & .623 (.024) & \cellcolor{green!20}\textbf{.662 (.011)} & .630 (.031) \\
\TAIWAN~CBCT & .680 (.039) & .721 (.027) & .792 (.021) & .794 (.008) & \cellcolor{blue!20}\textbf{.819 (.007)} & .792 (.010) & .805 (.030) & .780 (.031) & .719 (.033) & .785 (.036) & \cellcolor{green!20}.800 (.026) & .792 (.026) & .757 (.025) & .652 (.020) & .784 (.039) & .756 (.011) \\
\EGYPT~CBE   & .842 (.015) & .830 (.038) & .854 (.053) & .858 (.024) & .858 (.011) & .822 (.016) & \cellcolor{blue!20}.863 (.046) & .829 (.031) & .815 (.014) & .862 (.020) & .852 (.024) & .839 (.015) & .838 (.026) & .806 (.017) & \cellcolor{green!20}\textbf{.904 (.027)} & .874 (.021) \\
\RUSSIA~CBR  & .880 (.013) & .896 (.022) & \cellcolor{blue!20}\textbf{.919 (.022) }& .884 (.028) & .909 (.012) & .908 (.022) & .916 (.028) & .701 (.043) & .662 (.029) & .778 (.030) & .767 (.055) & .807 (.054) & .713 (.037) & .817 (.014) & \cellcolor{green!20}{.830 (.041)} & .797 (.024) \\
\TURKEY~CBRT & .891 (.011) & .898 (.030) & \cellcolor{blue!20}\textbf{.945 (.014) }& .916 (.018) & .914 (.030) & .938 (.016) & .937 (.015) & .851 (.021) & .780 (.011) & .884 (.023) & .884 (.014) & .843 (.007) & .843 (.010) & .792 (.009) & \cellcolor{green!20}{.906 (.019)} & .870 (.014) \\
\CHILE~CBoC  & .806 (.034) & .775 (.041) & .793 (.037) & .794 (.031) & .823 (.054) & .820 (.011) & \cellcolor{blue!20}\textbf{.834 (.066) } & .754 (.041) & .577 (.048) & .764 (.045) & .741 (.038) & .623 (.068) & .682 (.060) & .713 (.019) & \cellcolor{green!20}{.816 (.028)} & .705 (.042) \\
\EU~ECB      & .838 (.027) & .878 (.052) & .888 (.041) & .877 (.036) & \cellcolor{blue!20}\textbf{.888 (.025)} & .873 (.015) & .876 (.011) & .735 (.019) & .645 (.006) & .722 (.027) & .762 (.023) & .669 (.039) & .735 (.010) & .701 (.036) & .841 (.007) & \cellcolor{green!20}.787 (.013) \\
\USA~FOMC    & .773 (.025) & .777 (.008) & .818 (.025) & .803 (.011) & .826 (.023) & .824 (.008) & \cellcolor{blue!20}{.837 (.021) }  & .818 (.019) & .763 (.045) & .806 (.016) & .779 (.034) & .718 (.048) & .818 (.032) & .671 (.033) & \cellcolor{green!20}\textbf{.839 (.022)} & .825 (.011) \\
\SINGAPORE~MAS  & .803 (.050) & .825 (.057) & \cellcolor{blue!20}{.852 (.019) }& .816 (.017) & .833 (.031) & .846 (.045) & .843 (.020) & .798 (.043) & .803 (.027) & .822 (.036) & .828 (.035) & .809 (.024) & .800 (.029) & .602 (.039) & \cellcolor{green!20}\textbf{.858 (.035)} & .857 (.031) \\
\POLAND~NBP  & .803 (.020) & .803 (.029) & \cellcolor{blue!20}\textbf{.855 (.009) }& .849 (.026) & .814 (.006) & .840 (.017) & .844 (.021) & .760 (.013) & .610 (.021) & .725 (.007) & .745 (.023) & .623 (.026) & .689 (.033) & .607 (.023) & \cellcolor{green!20}{.813 (.016)} & .774 (.019) \\
\CHINA~PBoC  & .929 (.005) & \cellcolor{blue!20}.943 (.012) & .941 (.014) & .928 (.006) & .933 (.006) & .937 (.012) & .936 (.004) & .641 (.046) & .749 (.017) & .851 (.013) & .801 (.023) & .869 (.022) & .844 (.023) & .927 (.021) & \cellcolor{green!20}\textbf{.955 (.010)} & .942 (.012) \\
\AUS~RBA     & .861 (.041) & .830 (.054) & .888 (.060) & .878 (.043) & \cellcolor{blue!20}\textbf{.897 (.048)} & .887 (.024) & .885 (.065) & .847 (.015) & .782 (.021) & .851 (.031) & .852 (.032) & .778 (.019) & .824 (.029) & .671 (.022) & \cellcolor{green!20}.854 (.039) & .847 (.019) \\
\INDIA~RBI   & .754 (.028) & .780 (.075) & .812 (.040) & .813 (.080) & .829 (.007) & .813 (.099) & \cellcolor{blue!20}\textbf{.836 (.068)}  & .662 (.018) & .584 (.017) & .700 (.024) & .686 (.021) & .616 (.020) & .599 (.042) & .750 (.030) & \cellcolor{green!20}.755 (.023) & .702 (.029) \\
\SWITZERLAND~SNB  & .815 (.021) & .813 (.012) & .798 (.041) & \cellcolor{blue!20}\textbf{.860 (.019) }& .806 (.056) & .824 (.035) & .822 (.010) & .703 (.017) & .680 (.001) & .779 (.007) & .759 (.010) & .721 (.015) & .651 (.010) & .733 (.008) & \cellcolor{green!20}{.805 (.006)} & .779 (.025) \\
\midrule
Average & .785 (.097) & .797 (.093) & \cellcolor{blue!20}\textbf{.820 (.093)} & .808 (.093) & .819 (.092) & .811 (.019) & \cellcolor{blue!20}.820 (.100)  & .737 (.077) & .677 (.088) & .764 (.079) & .764 (.067) & .711 (.109) & .713 (.087) & .702 (.093) & \cellcolor{green!20}.816 (.066) & .776 (.078) \\
\bottomrule
\end{tabularx}
\end{table}

\clearpage
\subsection{Few shot \& Annotation Guide Few Shots Prompting}
\label{app:few_shot_ann_guide}
\begin{table}[h!]
\centering
\small
\caption{F1-Scores across all three tasks (\SD, \TC, and \CE) are reported using the best-performing LLM on average, \texttt{Llama-3-70B-Chat}. This is evaluated under Few-Shot prompting and prompting with Annotation Guides. Standard deviations are shown in parentheses.}
\label{tab:Ann_fewshot}
\begin{tabularx}{\textwidth}{
    l                
    *{3}{X}          
    :*{3}{X}         
}
\toprule
& \multicolumn{3}{c}{\textbf{Few‑Shot Prompting}} 
& \multicolumn{3}{c}{\textbf{Annotation‑Guide Few‑Shot Prompting}} \\
\cmidrule(lr){2-4}\cmidrule(lr){5-7}
\textbf{Bank} & Stance & Temporal & Certain & Stance & Temporal & Certain \\
\midrule
\BRAZIL~BCB            & .627 (.068) & .757 (.047) & .780 (.104) & .524 (.105) & .836 (.035) & .767 (.071) \\
\PERU~BCRP             & .641 (.080) & .809 (.061) & .878 (.307) & .627 (.094) & .869 (.052) & .896 (.262) \\
\MALAYSIA~BNM          & .577 (.129) & .865 (.037) & .745 (.104) & .588 (.080) & .875 (.039) & .762 (.065) \\
\PHILIPPINES~BSP       & .586 (.106) & .902 (.029) & .885 (.162) & .562 (.145) & .920 (.028) & .857 (.159) \\
\COLOMBIA~BanRep       & .628 (.056) & .858 (.048) & .829 (.164) & .563 (.084) & .869 (.046) & .856 (.119) \\
\CANADA~BoC            & .661 (.195) & .820 (.048) & .624 (.092) & .718 (.075) & .869 (.031) & .769 (.055) \\
\UK~BoE                & .682 (.063) & .757 (.074) & .728 (.065) & .620 (.120) & .762 (.077) & .719 (.072) \\
\ISRAEL~BoI            & .585 (.159) & .882 (.033) & .729 (.143) & .427 (.135) & .844 (.032) & .695 (.205) \\
\JAPAN~BoJ             & .583 (.104) & .718 (.023) & .784 (.167) & .616 (.078) & .781 (.022) & .808 (.101) \\
\SOUTHKOREA~BoK        & .655 (.058) & .884 (.017) & .860 (.100) & .586 (.135) & .867 (.021) & .843 (.085) \\
\MEXICO~BdeM           & .528 (.104) & .851 (.028) & .657 (.089) & .678 (.160) & .861 (.029) & .722 (.079) \\
\THAILAND~BoT          & .573 (.188) & .723 (.067) & .671 (.083) & .596 (.213) & .711 (.050) & .684 (.070) \\
\TAIWAN~CBCT           & .611 (.147) & .840 (.024) & .665 (.118) & .542 (.087) & .833 (.016) & .745 (.056) \\
\EGYPT~CBE             & .641 (.158) & .934 (.019) & .859 (.171) & .808 (.063) & .940 (.017) & .851 (.127) \\
\RUSSIA~CBR            & .731 (.142) & .846 (.044) & .905 (.132) & .784 (.127) & .826 (.038) & .909 (.123) \\
\TURKEY~CBRT           & .647 (.085) & .847 (.049) & .884 (.148) & .645 (.110) & .934 (.031) & .901 (.132) \\
\CHILE~CBoC            & .663 (.160) & .788 (.041) & .848 (.140) & .698 (.075) & .792 (.048) & .824 (.109) \\
\EU~ECB                & .627 (.096) & .769 (.234) & .839 (.170) & .678 (.057) & .806 (.316) & .874 (.119) \\
\USA~FOMC              & .648 (.052) & .873 (.025) & .799 (.099) & .663 (.044) & .878 (.029) & .792 (.094) \\
\SINGAPORE~MAS         & .663 (.128) & .893 (.019) & .783 (.115) & .583 (.148) & .940 (.010) & .814 (.076) \\
\POLAND~NBP            & .617 (.126) & .900 (.007) & .761 (.044) & .636 (.237) & .891 (.024) & .795 (.027) \\
\CHINA~PBoC            & .639 (.110) & .909 (.268) & .954 (.158) & .589 (.108) & .914 (.030) & .916 (.216) \\
\AUS~RBA               & .572 (.174) & .933 (.033) & .815 (.119) & .636 (.192) & .921 (.032) & .862 (.085) \\
\INDIA~RBI             & .599 (.093) & .800 (.027) & .837 (.128) & .589 (.050) & .788 (.016) & .766 (.124) \\
\SWITZERLAND~SNB       & .648 (.095) & .824 (.023) & .782 (.126) & .613 (.090) & .853 (.021) & .797 (.098) \\
\midrule
Average & .625 (.115) & .839 (.053) & .796 (.130) & .623 (.112) & .855 (.044) & .809 (.109) \\
\bottomrule
\end{tabularx}
\end{table}
\paragraph{Performance Analysis with Few-Shot and Annotation Guide Prompting.}
Table~\ref{tab:Ann_fewshot} shows the effect of Few-Shot and Annotation Guide prompting on \texttt{Llama-3-70B-Chat}. For \CE, performance drops to 0.796 (±0.130) with Few-Shot prompting and 0.809 (±0.109) with Annotation Guide prompting. For \TC, performance shows a similar decline: 0.839 (±0.053) with Few-Shot and 0.855 (±0.044) with Annotation Guide. In contrast, \SD shows a slight improvement, with scores increasing to 0.625 (±0.115) for Few-Shot and 0.623 (±0.112) for Annotation Guide. This translates to relative gains of 0.5\% and 0.3\% respectively. These results reinforce that \SD remains the most challenging of the three tasks, where improved prompting offers minor but consistent gains. However, even with enhanced prompting, LLMs do not surpass PLM performance, suggesting that the performance ceiling for LLMs remains lower than that of fine-tuned PLMs on the \SD task.

\subsection{Whole Model Performance Gain}
\label{app:whole_perf_gain}
In our experiments, we discover that there is a significant improvement in model performance across all three labels when the training corpus of the is expanded from just the specific bank (Bank Specific Setup) to having all banks' data (General Setup). In order to comprehend why this happens, we evaluate the models' increased performance using qualitative and statistical reasoning.
\subsubsection{Qualitative Reasoning}
\label{app:qual_gain}
To understand the improved performance of the model trained on the whole corpus of the annotated data as compared to the specific banks' models, we qualitatively evaluate the \SD labels given by each model and compare them to the ground truth (Table \ref{tab:gain_whole_sentences}). For each bank, we compare its best Bank Specific Setup model against our best General Setup model. This comparison demonstrates the general model benefits from patterns between central bank communications, which suggests that the model is creating meaningful parallels across central banks. These findings highlight the potential for inter-bank learning, where information about one central bank's communications improves the model's ability to interpret another bank's communications.

We then generate embeddings for each sentence within our training corpus and find the sentence whose embeddings are the closest to our selected sentence as well as its label and origin bank as seen within Table \ref{tab:gain_whole_sentences}. We find that there is a significant amount of learning from other banks. When the model is trained on a larger corpus, it is able to learn from the annotations and labels of the other banks. 

\begin{table}[h!]
  \tiny
  \caption{Comparison of the stance labels generated by the whole and specific model (using \texttt{seed = 5768}) for each central bank. The closest bank, sentence and label columns represent the results from generating and comparing embeddings from the training corpus. $^*$ indicates that the general model's label was the ground truth.}
  \label{tab:gain_whole_sentences}
  \begin{tabular}{p{0.08\textwidth}                      p{0.24\textwidth}
                  p{0.05\textwidth} p{0.05\textwidth} |
                  p{0.08\textwidth} p{0.24\textwidth} p{0.05\textwidth}}
    \toprule
    Bank & Sentence & Specific model & General Model$^*$ & Closest Bank & Closest Sentence & Closest Label \\
    \midrule
    \CHILE CBoC      & although establishing such a conclusion required reviewing multiple dimensions, some of them could be mentioned. & Neutral  & Irrelevant & \TAIWAN CBCT   & related discussions are summarized as follows & Irrelevant \\
    \JAPAN BoJ       & meanwhile, business sentiment had generally stayed at a favorable level, although some cautiousness had been observed. & Dovish   & Neutral    & \EGYPT CBE    & moreover, most leading indicators remained in positive territory in 2022 q2. & Hawkish \\
    \TAIWAN CBCT     & related discussions are summarized as follows & Neutral  & Irrelevant & \CHILE CBoC    & although establishing such a conclusion required reviewing multiple dimensions, some of them could be mentioned. & Irrelevant \\
    \MEXICO BdeM     & one member emphasized that its seasonally adjusted and annualized monthly rate was 3.53\% in July 2024. & Hawkish  & Neutral    & \POLAND NBP   & other council members assessed that the recent decline in service price growth had probably been temporary & Neutral \\
    \EU ECB          & US Commodity Futures Trading Commission numbers suggested that the market remained positioned for an appreciation of the euro. & Hawkish  & Dovish     & \USA FOMC     & the nominal deficit on U.S. trade in goods and services continued to widen in April. & Hawkish \\
    \UK BoE          & there were also arguments in favour of a programme of purchases towards the upper end of the range & Irrelevant & Dovish     & \POLAND NBP   & other council members assessed that the recent decline in service price growth had probably been temporary & Neutral \\
    \EGYPT CBE       & moreover, most leading indicators remained in positive territory in 2022 q2. & Neutral  & Hawkish    & \JAPAN BoJ    & meanwhile, business sentiment had generally stayed at a favorable level, although some cautiousness had been observed. & Neutral \\
    \MALAYSIA BNM    & in line with earlier assessments, headline inflation is likely to have peaked in 3q 2022 and is expected to moderate thereafter, albeit remaining elevated & Dovish   & Hawkish    & \EGYPT CBE    & moreover, most leading indicators remained in positive territory in 2022 q2. & Hawkish \\
    \SINGAPORE MAS   & consequently, unit labour costs will increase at a significantly slower pace compared with the preceding two years. & Neutral  & Dovish     & \SOUTHKOREA BoK & increase in the prices of petroleum products and the base effect from the decline in the prices of public services last year. & Hawkish \\
    \COLOMBIA BanRep & in the same period, other commodities exported and imported by Colombia descended. & Neutral  & Dovish     & \SWITZERLAND SNB & what the economy needs to recover, however, is for exports to pick up & Dovish \\
    \THAILAND BoT    & third, the continued economic recovery of trading partners would benefit Thai exports. & Neutral  & Dovish     & \SWITZERLAND SNB & what the economy needs to recover, however, is for exports to pick up & Dovish \\
    \USA FOMC        & the nominal deficit on U.S. trade in goods and services continued to widen in April. & Neutral  & Hawkish    & \SOUTHKOREA BoK & increase in the prices of petroleum products and the base effect from the decline in the prices of public services last year. & Hawkish \\
    \CHINA PBoC      & the PBC will implement monetary policy in a flexible and targeted manner, and strengthen the coordination with fiscal, industrial, and regulatory policies. & Neutral  & Dovish     & \ISRAEL BoI    & in the last twelve months the CPI has increased by 2.5 percent, within the target inflation range. & Neutral \\
    \INDIA RBI       & the latest assessment by the World Trade Organisation (WTO) for Q4 indicates a loss of momentum in global trade due to declining export orders. & Hawkish  & Dovish     & \USA FOMC     & the nominal deficit on U.S. trade in goods and services continued to widen in April. & Hawkish \\
    \SOUTHKOREA BoK  & increase in the prices of petroleum products and the base effect from the decline in the prices of public services last year. & Dovish   & Hawkish    & \ISRAEL BoI    & in the last twelve months the CPI has increased by 2.5 percent, within the target inflation range. & Neutral \\
    \SWITZERLAND SNB & what the economy needs to recover, however, is for exports to pick up & Hawkish  & Dovish     & \THAILAND BoT  & third, the continued economic recovery of trading partners would benefit Thai exports. & Dovish \\
    \TURKEY CBRT     & housing loan rates, which have remained flat in the same period, stood at 41.4\%. & Irrelevant & Neutral    & \RUSSIA CBR   & unsecured consumer lending is supported by higher households' incomes. & Dovish \\
    \PERU BCRP       & these measures have allowed immediate release of funds in domestic currency amounting to a total of 2 billion & Neutral  & Dovish     & \BRAZIL BCB    & the result was creation of 590 thousand jobs in the first half of the year & Hawkish \\
    \PHILIPPINES BSP & the next meeting of the Monetary Board to discuss monetary policy is scheduled on 12 January 2006. & Neutral  & Irrelevant & \CHINA PBoC   & the PBC will implement monetary policy in a flexible and targeted manner, and strengthen coordination with fiscal, industrial, and regulatory policies. & Dovish \\
    \BRAZIL BCB      & the result was creation of 590 thousand jobs in the first half of the year & Neutral  & Hawkish    & \PERU BCRP    & these measures have allowed immediate release of funds in domestic currency amounting to a total of 2 billion & Dovish \\
    \AUS RBA         & however, central banks had also emphasised that policy rates were unlikely to decline over coming months, in contrast to market‑implied expectations. & Dovish   & Hawkish    & \POLAND NBP   & other council members assessed that the recent decline in service price growth had probably been temporary & Neutral \\
    \POLAND NBP      & other council members assessed that the recent decline in service price growth had probably been temporary & Hawkish  & Neutral    & \SOUTHKOREA BoK & increase in the prices of petroleum products and the base effect from the decline in the prices of public services last year. & Hawkish \\
    \RUSSIA CBR      & unsecured consumer lending is supported by higher households' incomes. & Neutral  & Dovish     & \TURKEY CBRT  & housing loan rates, which have remained flat in the same period, stood at 41.4\%. & Neutral \\
    \ISRAEL BoI      & in the last twelve months the CPI has increased by 2.5 percent, within the target inflation range. & Hawkish  & Neutral    & \SOUTHKOREA BoK & increase in the prices of petroleum products and the base effect from the decline in the prices of public services last year. & Hawkish \\
    \CANADA BoC      & GDP growth is forecast to increase in the second half of 2024 and through 2025 & Neutral  & Hawkish    & \SWITZERLAND SNB & what the economy needs to recover, however, is for exports to pick up & Dovish \\
    \bottomrule
  \end{tabular}
\end{table}

\subsubsection{Statistical Reasoning}
\label{app:statistical_gain}
To further understand this improvement in model performance from the specific model to the generalized model trained on all the annotated sentences for all central banks, we hypothesize that there are semantic similarities across central bank meeting level documents that enable the general setup model to generalize effectively. To test our hypothesis, we perform TF-IDF for every central bank at a document level and then using cosine similarity to find the level of similarity between banks' documents. After normalizing these scores and setting them up as the independent variable (variable $x_1$), we compare them using a linear regression with the change in the weighted F1-Score (variable $y$) of the best performing model \texttt{(RoBERTa-large)} for both specific and general cases. This then leads to the following formula:
\begin{align*}
    y = 0.0168x_1 + 0.0511
\end{align*}
This linear regression has a p-value of $0.016$, implying statistical significance under the threshold of $2\%$. These results (with a positive value of the coefficient of variable $x_1$) indicate that semantic overlap between banks' document-level data contributes to a general improvement in model performance, supporting the hypothesis that central bank communications contain consistencies that allow models to learn from other banks' data to improve performance.

\clearpage
\subsection{Hyperparameters}
\label{app:hyperparams}
\subsubsection{WCB Models}
\label{app:hyper_all}
Table \ref{tab:best_hypers_all} displays the best performing hyperparameters for each task given the General Setup. We obtain these hyperparameters through grid search across the following parameters: \texttt{seeds = [5768, 78516, 944601]\footnote{These specific seeds were chosen based on the work of \citet{shah-etal-2022-flue} and \citet{shah-etal-2023-trillion}. }, batch sizes = [32, 16], learning rates = [1e-5, 1e-6]}. 
\begin{table}[h!]
\centering
\caption{Best model hyperparameter for each task for the model trained using the General Setup.}
\label{tab:best_hypers_all}
\begin{tabularx}{\textwidth}{p{0.15\textwidth}p{0.2\textwidth}p{0.2\textwidth}p{0.15\textwidth}p{0.1\textwidth}}
\toprule
\textbf{Task} & \textbf{Model Name} & \textbf{Learning Rate} & \textbf{Batch Size} & \textbf{Best Seed} \\
\midrule

Certain Label & roberta-large & 1e-05 & 16 & 5768 \\
Stance Label & roberta-large & 1e-06 & 32 & 944601 \\
Time Label & roberta-base & 1e-06 & 32 & 944601 \\

\bottomrule
\end{tabularx}
\end{table}

\subsubsection{Specific Bank Models}
\label{app:hyper_specific}
For each bank and task under the specific setup, the hyperparameters that have the best performance for the given top-ranked PLM are shown in the tables below. Table \ref{tab:best_hypers_certain} show the best hyperparameters for the best performing model for \texttt{Uncertainty Estimation} task while Table \ref{tab:best_hypers_stance} and Table \ref{tab:best_hypers_time} show the best hyperparameters for the best performing model for \texttt{Stance Detection} and \texttt{Temporal Classification}, respectively. We obtain these hyperparameters through grid search across the following parameters: \texttt{seeds = [5768, 78516, 944601]\footnote{These specific seeds were chosen based on the work of \citet{shah-etal-2022-flue} and \citet{shah-etal-2023-trillion}. }, batch sizes = [32, 16], learning rates = [1e-5, 1e-6]}. 
\begin{table}[h!]
\centering
\small
\caption{Best performing model and corresponding hyperparameter-learning rate and batch size for \CE.}
\label{tab:best_hypers_certain}
\begin{tabularx}{\textwidth}{p{0.12\textwidth}p{0.18\textwidth}p{0.13\textwidth}p{0.13\textwidth}p{0.13\textwidth}p{0.1\textwidth}}
\toprule
\textbf{Bank} & \textbf{Base Model} & \textbf{Learning Rate} & \textbf{Batch Size} & \textbf{Best Seed} & \textbf{Link}\\
\midrule

\USA FOMC & roberta-large & 1e-06 & 16 & 78516 & \href{https://huggingface.co/gtfintechlab/model_federal_reserve_system_certain_label}{\texttt{HF Link}}\\
\CHINA PBoC & ModernBERT-large & 1e-05 & 16 & 78516 & \href{https://huggingface.co/gtfintechlab/model_peoples_bank_of_china_certain_label}{\texttt{HF Link}} \\
\JAPAN BoJ & roberta-base & 1e-05 & 16 & 944601 & \href{https://huggingface.co/gtfintechlab/model_bank_of_japan_certain_label}{\texttt{HF Link}} \\
\UK BoE & bert-base & 1e-06 & 16 & 944601 & \href{https://huggingface.co/gtfintechlab/model_bank_of_england_certain_label}{\texttt{HF Link}}\\
\SWITZERLAND SNB & bert-large & 1e-05 & 16 & 5768 & \href{https://huggingface.co/gtfintechlab/model_swiss_national_bank_certain_label}{\texttt{HF Link}}\\
\BRAZIL BCB & bert-base & 1e-05 & 32 & 78516 & \href{https://huggingface.co/gtfintechlab/model_central_bank_of_brazil_certain_label}{\texttt{HF Link}} \\
\INDIA RBI & roberta-large & 1e-05 & 16 & 944601 & \href{https://huggingface.co/gtfintechlab/model_reserve_bank_of_india_certain_label}{\texttt{HF Link}}\\
\EU ECB & finbert-pretrain & 1e-06 & 16 & 5768 & \href{https://huggingface.co/gtfintechlab/model_european_central_bank_certain_label}{\texttt{HF Link}}\\
\RUSSIA CBR & bert-base & 1e-05 & 32 & 5768 & \href{https://huggingface.co/gtfintechlab/model_central_bank_of_the_russian_federation_certain_label}{\texttt{HF Link}}\\
\TAIWAN CBCT & finbert-pretrain & 1e-06 & 16 & 78516 & \href{https://huggingface.co/gtfintechlab/model_central_bank_of_china_taiwan_certain_label}{\texttt{HF Link}}\\
\SINGAPORE MAS & bert-base & 1e-06 & 16 & 944601 & \href{https://huggingface.co/gtfintechlab/model_monetary_authority_of_singapore_certain_label}{\texttt{HF Link}}\\
\SOUTHKOREA BoK & bert-base & 1e-06 & 16 & 5768 & \href{https://huggingface.co/gtfintechlab/model_bank_of_korea_certain_label}{\texttt{HF Link}}\\
\AUS RBA & finbert-pretrain & 1e-05 & 16 & 944601 & \href{https://huggingface.co/gtfintechlab/model_reserve_bank_of_australia_certain_label}{\texttt{HF Link}}\\
\ISRAEL BoI & ModernBERT-base & 1e-05 & 32 & 944601 & \href{https://huggingface.co/gtfintechlab/model_bank_of_israel_certain_label}{\texttt{HF Link}} \\
\CANADA BoC & bert-large & 1e-05 & 16 & 944601 & \href{https://huggingface.co/gtfintechlab/model_bank_of_canada_certain_label}{\texttt{HF Link}} \\
\MEXICO BdeM & roberta-base & 1e-06 & 16 & 5768 & \href{https://huggingface.co/gtfintechlab/model_bank_of_mexico_certain_label}{\texttt{HF Link}} \\
\POLAND NBP & bert-base & 1e-05 & 32 & 78516 & \href{https://huggingface.co/gtfintechlab/model_national_bank_of_poland_certain_label}{\texttt{HF Link}}\\
\TURKEY CBRT & bert-base & 1e-05 & 32 & 78516 & \href{https://huggingface.co/gtfintechlab/model_central_bank_republic_of_turkey_certain_label}{\texttt{HF Link}} \\
\THAILAND BoT & bert-base & 1e-05 & 32 & 78516 & \href{https://huggingface.co/gtfintechlab/model_bank_of_thailand_certain_label}{\texttt{HF Link}}\\
\EGYPT CBE & roberta-large & 1e-06 & 16 & 5768 & \href{https://huggingface.co/gtfintechlab/model_central_bank_of_egypt_certain_label}{\texttt{HF Link}}\\
\MALAYSIA BNM & roberta-large & 1e-05 & 32 & 78516 & \href{https://huggingface.co/gtfintechlab/model_bank_negara_malaysia_certain_label}{\texttt{HF Link}}\\
\PHILIPPINES BSP & ModernBERT-base & 1e-05 & 16 & 5768 & \href{https://huggingface.co/gtfintechlab/model_central_bank_of_the_philippines_certain_label}{\texttt{HF Link}}\\
\CHILE CBoC & roberta-large & 1e-05 & 16 & 78516 & \href{https://huggingface.co/gtfintechlab/model_central_bank_of_chile_certain_label}{\texttt{HF Link}}\\
\PERU BCRP & finbert-pretrain & 1e-05 & 32 & 78516 & \href{https://huggingface.co/gtfintechlab/model_central_reserve_bank_of_peru_certain_label}{\texttt{HF Link}}\\
\COLOMBIA BanRep & roberta-large & 1e-06 & 32 & 78516 & \href{https://huggingface.co/gtfintechlab/model_bank_of_the_republic_colombia_certain_label}{\texttt{HF Link}}\\

\bottomrule
\end{tabularx}
\end{table}

\begin{table}[h!]
\centering
\small
\caption{Best performing model and corresponding hyperparameter-learning rate and batch size for \SD.}
\label{tab:best_hypers_stance}
\begin{tabularx}{\textwidth}{p{0.12\textwidth}p{0.18\textwidth}p{0.13\textwidth}p{0.13\textwidth}p{0.13\textwidth}p{0.1\textwidth}}
\toprule
\textbf{Bank} & \textbf{Base Model} & \textbf{Learning Rate} & \textbf{Batch Size} & \textbf{Best Seed} & \textbf{Link}\\
\midrule

\USA FOMC & roberta-large & 1e-06 & 32 & 944601 & \href{https://huggingface.co/gtfintechlab/model_federal_reserve_system_stance_label}{\texttt{HF Link}} \\
\CHINA PBoC & roberta-large & 1e-05 & 16 & 78516 & \href{https://huggingface.co/gtfintechlab/model_peoples_bank_of_china_stance_label}{\texttt{HF Link}} \\
\JAPAN BoJ & roberta-base & 1e-05 & 32 & 78516 & \href{https://huggingface.co/gtfintechlab/model_bank_of_japan_stance_label}{\texttt{HF Link}} \\
\UK BoE & roberta-large & 1e-06 & 16 & 5768 & \href{https://huggingface.co/gtfintechlab/model_bank_of_england_stance_label}{\texttt{HF Link}} \\
\SWITZERLAND SNB & roberta-large & 1e-06 & 16 & 78516 & \href{https://huggingface.co/gtfintechlab/model_swiss_national_bank_stance_label}{\texttt{HF Link}} \\
\BRAZIL BCB & roberta-large & 1e-05 & 16 & 944601 &  \href{https://huggingface.co/gtfintechlab/model_central_bank_of_brazil_stance_label}{\texttt{HF Link}} \\
\INDIA RBI & roberta-large & 1e-06 & 32 & 944601 & \href{https://huggingface.co/gtfintechlab/model_reserve_bank_of_india_stance_label}{\texttt{HF Link}} \\
\EU ECB & roberta-large & 1e-06 & 32 & 5768 & \href{https://huggingface.co/gtfintechlab/model_european_central_bank_stance_label}{\texttt{HF Link}} \\
\RUSSIA CBR & roberta-large & 1e-05 & 16 & 5768 & \href{https://huggingface.co/gtfintechlab/model_central_bank_of_the_russian_federation_stance_label}{\texttt{HF Link}} \\
\TAIWAN CBCT & roberta-large & 1e-06 & 16 & 5768 & \href{https://huggingface.co/gtfintechlab/model_central_bank_of_china_taiwan_stance_label}{\texttt{HF Link}} \\
\SINGAPORE MAS & roberta-base & 1e-06 & 16 & 78516 & \href{https://huggingface.co/gtfintechlab/model_monetary_authority_of_singapore_stance_label}{\texttt{HF Link}} \\
\SOUTHKOREA BoK & roberta-large & 1e-06 & 32 & 5768 & \href{https://huggingface.co/gtfintechlab/model_bank_of_korea_stance_label}{\texttt{HF Link}} \\
\AUS RBA & roberta-base & 1e-05 & 16 & 78516 & \href{https://huggingface.co/gtfintechlab/model_reserve_bank_of_australia_stance_label}{\texttt{HF Link}} \\
\ISRAEL BoI & roberta-large & 1e-06 & 16 & 78516 & \href{https://huggingface.co/gtfintechlab/model_bank_of_israel_stance_label}{\texttt{HF Link}} \\
\CANADA BoC & roberta-large & 1e-06 & 16 & 5768 & \href{https://huggingface.co/gtfintechlab/model_bank_of_canada_stance_label}{\texttt{HF Link}} \\
\MEXICO BdeM & roberta-large & 1e-06 & 16 & 5768 & \href{https://huggingface.co/gtfintechlab/model_bank_of_mexico_stance_label}{\texttt{HF Link}} \\
\POLAND NBP & roberta-large & 1e-05 & 16 & 944601 & \href{https://huggingface.co/gtfintechlab/model_national_bank_of_poland_stance_label}{\texttt{HF Link}} \\
\TURKEY CBRT & roberta-large & 1e-06 & 32 & 944601 & \href{https://huggingface.co/gtfintechlab/model_central_bank_republic_of_turkey_stance_label}{\texttt{HF Link}} \\
\THAILAND BoT & roberta-large & 1e-05 & 32 & 78516 & \href{https://huggingface.co/gtfintechlab/model_bank_of_thailand_stance_label}{\texttt{HF Link}} \\
\EGYPT CBE & roberta-large & 1e-05 & 16 & 78516 & \href{https://huggingface.co/gtfintechlab/model_central_bank_of_egypt_stance_label}{\texttt{HF Link}} \\
\MALAYSIA BNM & ModernBERT-large & 1e-05 & 32 & 5768 & \href{https://huggingface.co/gtfintechlab/model_bank_negara_malaysia_stance_label}{\texttt{HF Link}} \\
\PHILIPPINES BSP & roberta-large & 1e-06 & 16 & 944601 & \href{https://huggingface.co/gtfintechlab/model_central_bank_of_the_philippines_stance_label}{\texttt{HF Link}} \\
\CHILE CBoC & roberta-large & 1e-05 & 16 & 944601 & \href{https://huggingface.co/gtfintechlab/model_central_bank_of_chile_stance_label}{\texttt{HF Link}} \\
\PERU BCRP & roberta-large & 1e-06 & 16 & 5768 & \href{https://huggingface.co/gtfintechlab/model_central_reserve_bank_of_peru_stance_label}{\texttt{HF Link}} \\
\COLOMBIA BanRep & roberta-large & 1e-05 & 16 & 78516 & \href{https://huggingface.co/gtfintechlab/model_bank_of_the_republic_colombia_stance_label}{\texttt{HF Link}} \\

\bottomrule
\end{tabularx}
\end{table}

\begin{table}[h!]
\centering
\small
\caption{Best performing model and corresponding hyperparameter-learning rate and batch size for \TC.}
\label{tab:best_hypers_time}
\begin{tabularx}{\textwidth}{p{0.12\textwidth}p{0.18\textwidth}p{0.13\textwidth}p{0.13\textwidth}p{0.13\textwidth}p{0.1\textwidth}}
\toprule
\textbf{Bank} & \textbf{Base Model} & \textbf{Learning Rate} & \textbf{Batch Size} & \textbf{Best Seed} & \textbf{Link}\\
\midrule

\USA FOMC & roberta-large & 1e-05 & 32 & 5768 & \href{https://huggingface.co/gtfintechlab/model_federal_reserve_system_time_label}{\texttt{HF Link}} \\
\CHINA PBoC & ModernBERT-base & 1e-05 & 16 & 944601 & \href{https://huggingface.co/gtfintechlab/model_peoples_bank_of_china_time_label}{\texttt{HF Link}} \\
\JAPAN BoJ & roberta-base & 1e-06 & 32 & 944601 & \href{https://huggingface.co/gtfintechlab/model_bank_of_japan_time_label}{\texttt{HF Link}} \\
\UK BoE & roberta-base & 1e-05 & 32 & 944601 & \href{https://huggingface.co/gtfintechlab/model_bank_of_england_time_label}{\texttt{HF Link}} \\
\SWITZERLAND SNB & roberta-large & 1e-06 & 16 & 5768 & \href{https://huggingface.co/gtfintechlab/model_swiss_national_bank_time_label}{\texttt{HF Link}} \\
\BRAZIL BCB & roberta-large & 1e-05 & 16 & 944601 & \href{https://huggingface.co/gtfintechlab/model_central_bank_of_brazil_time_label}{\texttt{HF Link}} \\
\INDIA RBI & roberta-base & 1e-06 & 16 & 944601 & \href{https://huggingface.co/gtfintechlab/model_reserve_bank_of_india_time_label}{\texttt{HF Link}} \\
\EU ECB & roberta-base & 1e-06 & 16 & 78516 & \href{https://huggingface.co/gtfintechlab/model_european_central_bank_time_label}{\texttt{HF Link}} \\
\RUSSIA CBR & bert-base & 1e-06 & 16 & 5768 & \href{https://huggingface.co/gtfintechlab/model_central_bank_of_the_russian_federation_time_label}{\texttt{HF Link}} \\
\TAIWAN CBCT & bert-large & 1e-05 & 16 & 944601 & \href{https://huggingface.co/gtfintechlab/model_central_bank_of_china_taiwan_time_label}{\texttt{HF Link}} \\
\SINGAPORE MAS & finbert-pretrain & 1e-05 & 32 & 5768 & \href{https://huggingface.co/gtfintechlab/model_monetary_authority_of_singapore_time_label}{\texttt{HF Link}} \\
\SOUTHKOREA BoK & roberta-large & 1e-06 & 32 & 5768 & \href{https://huggingface.co/gtfintechlab/model_bank_of_korea_time_label}{\texttt{HF Link}}\\
\AUS RBA & roberta-large & 1e-05 & 16 & 78516 & \href{https://huggingface.co/gtfintechlab/model_reserve_bank_of_australia_time_label}{\texttt{HF Link}} \\
\ISRAEL BoI & finbert-pretrain & 1e-05 & 32 & 944601 & \href{https://huggingface.co/gtfintechlab/model_bank_of_israel_time_label}{\texttt{HF Link}} \\
\CANADA BoC & roberta-base & 1e-06 & 32 & 944601 & \href{https://huggingface.co/gtfintechlab/model_bank_of_canada_time_label}{\texttt{HF Link}} \\
\MEXICO BdeM & roberta-base & 1e-06 & 32 & 944601 & \href{https://huggingface.co/gtfintechlab/model_bank_of_mexico_time_label}{\texttt{HF Link}} \\
\POLAND NBP & roberta-large & 1e-05 & 16 & 78516 & \href{https://huggingface.co/gtfintechlab/model_national_bank_of_poland_time_label}{\texttt{HF Link}} \\
\TURKEY CBRT & roberta-base & 1e-05 & 16 & 5768 & \href{https://huggingface.co/gtfintechlab/model_central_bank_republic_of_turkey_time_label}{\texttt{HF Link}} \\
\THAILAND BoT & roberta-large & 1e-06 & 32 & 944601 & \href{https://huggingface.co/gtfintechlab/model_bank_of_thailand_time_label}{\texttt{HF Link}} \\
\EGYPT CBE & ModernBERT-large & 1e-05 & 16 & 78516  & \href{https://huggingface.co/gtfintechlab/model_central_bank_of_egypt_time_label}{\texttt{HF Link}}\\
\MALAYSIA BNM & finbert-pretrain & 1e-06 & 16 & 78516 & \href{https://huggingface.co/gtfintechlab/model_bank_negara_malaysia_time_label}{\texttt{HF Link}} \\
\PHILIPPINES BSP & roberta-large & 1e-06 & 16 & 944601 & \href{https://huggingface.co/gtfintechlab/model_central_bank_of_the_philippines_time_label}{\texttt{HF Link}} \\
\CHILE CBoC & roberta-base & 1e-06 & 32 & 944601 & \href{https://huggingface.co/gtfintechlab/model_central_bank_of_chile_time_label}{\texttt{HF Link}} \\
\PERU BCRP & roberta-large & 1e-05 & 32 & 5768 & \href{https://huggingface.co/gtfintechlab/model_central_reserve_bank_of_peru_time_label}{\texttt{HF Link}} \\
\COLOMBIA BanRep & roberta-base & 1e-06 & 16 & 5768 & \href{https://huggingface.co/gtfintechlab/model_bank_of_the_republic_colombia_time_label}{\texttt{HF Link}} \\

\bottomrule
\end{tabularx}
\end{table}

\clearpage
\section{Extended Experiments and Analysis}
\label{app:Extended Experiments and Analysis}
In our entire study, we run 15,075 experiments. These include benchmarking across both setups for 25 central banks and 16 models. The PLMs were benchmarked using gridsearch across \texttt{seeds = [5768, 78516, 944601], batch sizes = [32, 16], learning rates = [1e-5, 1e-6]}. The LLMs were tested under a single set of hyperparameters but using the same 3 seeds. We also run few shot prompting and annotation guide prompting for the best performing LLM across all banks and the same 3 seeds. In addition to this, we conduct several experiments as described below.
\subsection{LLM Error Analysis using People's Bank of China's Data}
\label{app:china_error}
In order to better understand how well our models classify statements across the three tasks (\texttt{Stance Detection, Temporal Classification}, and \texttt{Uncertainty Estimation}), we conduct an error analysis of \texttt{Llama-3-70b-chat-hf}, the best-performing LLM on the PBoC dataset. We perform error analysis on all the sentences misclassified for \texttt{Temporal Classification} and \texttt{Uncertainty Estimation}. For \texttt{Stance Detection}, we randomly sample 50 statements from the 150 mislabeled sentences with $\texttt{random seed=1}$. A sample of mislabeled sentences and the corresponding reason can be viewed in Table \ref{tab:error_examples}. 

\subsubsection{Stance Detection}
\label{app:error_stance}
For \texttt{Stance Detection}, we identify three common mistakes: 
(1) \textbf{Leaning Dovish} (32/50): The model fails to recognize when a statement is slightly \texttt{Dovish} and incorrectly labels such statements as \texttt{Neutral}.
(2) \textbf{Misidentifies monetary policy} (14/50): The LLM misidentifies statements that advocate for implementation of monetary policy as \texttt{Neutral} or \texttt{Irrelevant} and statements that do not advocate for implementation of monetary policy as \texttt{Hawkish} or \texttt{Dovish}. 
(3) \textbf{Misunderstanding} (4/50): The model misunderstands the statement entirely. 

\subsubsection{Temporal Classification}
\label{app:error_time}
For \texttt{Temporal Classification}, we identify four common errors: 
(1) \textbf{Changing trends} (20/34): The LLM mislabels statements about changing trends in the economy as \texttt{Not Forward Looking} instead of \texttt{Forward Looking}. 
(2) \textbf{Current event} (5/34): The LLM mistakenly labels sentences about current events as \texttt{Not Forward Looking}. 
(3) \textbf{Past tense} (3/34): The LLM misidentifies statements written in past tense as \texttt{Forward Looking}. 
(4) \textbf{Misunderstanding} (11/34): The model misunderstands the statement entirely. 

\subsubsection{Uncertainty Estimation}
\label{app:error_certain}
For \CE, we identify three common errors: 
(1) \textbf{Known economic trends} (7/20): The LLM misidentifies sentences that state the economic trends with certainty as \texttt{Uncertain}. 
(2) \textbf{Uncertain actions} (5/20): The LLM incorrectly labels sentences with no clear monetary policy as \texttt{Certain}.
(3) \textbf{Misunderstanding} (8/20): The model misunderstands the statement entirely. 

\label{app:error_samples}
\begin{table}[h!]
\scriptsize
\caption{Sample of sentences from error analysis. \textit{Correct label} is the actual label for that sentence. \textit{LLM Label} is the label that the Llama-3 model predicted. \textit{Type of Error} is the error that the LLM made.}
\label{tab:error_examples}
\begin{tabular}{p{0.6\textwidth} p{0.1\textwidth} p{0.1\textwidth} p{0.1\textwidth}}
\toprule
\textbf{Sentence} & \textbf{Correct Label} & \textbf{LLM Label} &  \textbf{Type of Error}  \\
\midrule
it was stressed that the sound monetary policy should be continued and the Forward Looking, scientific and effective approach to macro financial management be further strengthened in the coming period. & \dov & \neut & Leaning dovish \\
however, economic stabilization and recovery is not firmly established. & \neut  & \dov & Misidentifies monetary policy \\
\midrule
the principle of making the rmb exchange rate reform a self-initiated, controllable and gradual process should be followed to keep the rmb exchange rate basically stable at an adaptive and equilibrium level. & \forward&\texttt{Not Forward Looking} & Changing trends  \\
the overall performance of the financial sector has remained sound. &\texttt{Forward Looking} &\texttt{Not Forward Looking} & Current event  \\
however, the chinese economy still faced a complex situation, with both favorable and unfavorable factors in sight. &\texttt{Not Forward Looking} &\texttt{Forward Looking} & Past tense \\
\midrule
the solution of the problem needs cooperation among the enterprises, the commercial banks as well as authorities in different sectors. &\texttt{Certain}&\texttt{Uncertain}& Known economic trends\\
while financial support to economic growth must be guaranteed, no negligence should be tolerated in preventing inflation and financial risks. & \texttt{Uncertain} &\texttt{Certain}& Uncertain actions\\
\bottomrule
\end{tabular}
\end{table}

\subsection{Generating Meeting Minutes}
As discussed in Section~\ref{sec:intro}, central bank communications shape expectations about economic stability. To this end, having an early sense of their stance can benefit the broader public. Using our dataset, we explore whether a long‑context LLM can generate a target central bank’s next monetary policy document (minutes or their equivalent). The model receives only (i) the target bank’s own last document issued before that meeting (its most recent pre‑cutoff release) and (ii) documents from the other 24 banks, but only those released prior to the target document's release date. We then test whether the policy stance of the generated document matches that of the actual target document. Alignment is measured with the hawkishness metric defined in Section~\ref{sec:economic_analysis}, and our fine‑tuned \texttt{Stance Detection} model provides the evaluation benchmark.

\label{app:gen_mm}
\subsubsection{Methodology}
\label{app:gen_mm_method}

We generate candidate meeting documents (minutes or their equivalent) with \texttt{GPT‑4.1}, chosen for its one million token context window and top performance among the long‑context models in our benchmark. For each target central bank, we provide documents from the other 24 banks and query the model to produce the target bank’s first post‑cutoff document. We then compute hawkishness scores for the generated text \((\hat z_t)\) and the actual document's \((z_t)\) using the best-performing fine-tuned \texttt{Stance Detection} model. A linear regression of \(\hat z_t\) on \(z_t\) across all banks quantifies how well the generated stance tracks the actual stance in the target document. The full procedure is:

\begin{enumerate}
    \item Let $\mathcal{B} = \{b_1, b_2, \ldots, b_{25}\}$ denote the set of 25 central banks in our dataset.
    
    \item For each target central bank $b_t \in \mathcal{B}$, let $Y_t$ be its first released document after the fixed cutoff date for \texttt{GPT 4.1} (May 31, 2024). This is treated as the \textbf{actual output}.

    \item Let $Y_t'$ be the most recent document from $b_t$ released \textit{before} $Y_t$. This is used as the \textbf{example output}.

    \item For the \textbf{example input set} $X'$, we collect all documents from the remaining 24 banks $\mathcal{B} \setminus \{b_t\}$ that were released before the timestamp of $Y_t'$. This yields a globally available context prior to $Y_t'$.

    \item For the \textbf{actual input set} $X$, we collect all documents from $\mathcal{B} \setminus \{b_t\}$ released before $Y_t$. It is important to note that there may be partial overlap between $X$ and $X'$ depending on each central bank’s document frequency.

    \item The prompt is constructed with $(X', Y_t')$ as an example and $(X)$ as actual input. The LLM is then queried to generate $\hat{Y}_t$, the \textbf{predicted meeting document} for $b_t$. The prompt structure is illustrated in Figure~\ref{fig:prompt_generating_minutes}.

    \item We compute the \textbf{hawkishness measure} for each document $Y_t$ and the generated document $\hat{Y}_t$, denoted $z_t$ and $\hat{z}_t$ respectively. This is done using the hawkishness measure described in Section~\ref{sec:economic_analysis}.
    
    \item Finally, we fit a linear regression model $\hat{z}_t = \alpha + \beta z_t + \epsilon$ across all central banks to test how closely the stance predicted from the generated minutes follows the stance of the corresponding actual document. We use root mean square error (RMSE) as the primary metric and report full regression statistics.
\end{enumerate}

\begin{figure}[h!]
    \centering
    \footnotesize
    \begin{tcolorbox}[
        colback=gray!5,
        colframe=black!75,
        title=Prompt for generating meeting minutes.
    ]
\textbf{[SYSTEM INPUT]} \\
{
You are a precise and formal monetary policy meeting document generator for the central bank called \texttt{\{target\_bank\_nice\_name\}}. \\
Your job is to read meeting document (MM) from 24 other central banks and, based on those, \\
produce meeting documents that reflect what \texttt{\{target\_bank\_nice\_name\}}'s stance would be, in line with the example provided. \\
You should write clearly and professionally, mimicking the tone of central bank communication. \\
Each MM is a list of sentences, separated by a newline (\textbackslash n).
}
\vspace{0.5em}

\textcolor{orange}{Example: You are a precise and formal meeting documents generator for the central bank called Bank of England.}

\vspace{1em}
\textbf{[USER INPUT]} \\
{
Below is one example of an input-output pair, where the input consists of MM from 24 banks, each taken from their most recent meeting \\
prior to the current target date, and the output is the MM generated for \texttt{\{target\_bank\_nice\_name\}} from its own most recent meeting. \\
\\
Example: \\
{[Input from 24 banks]} \\
\texttt{\{example\_input\_str\}} \\
{[Output for \texttt{\{target\_bank\_nice\_name\}]}} \\
\texttt{\{example\_output\_str\}}
}

\vspace{0.5em}
\textcolor{orange}{
Example: \\
{[Input from 24 banks]} \\
Reserve Bank of Australia: The board decided to hold the cash rate at 4.1\% \dots \\
Bank of Canada: The policy rate remains at 5.0\%, inflation pressures persist \dots \\
{[Output for Bank of England]} \\
The MPC decided to maintain the Bank Rate at 5.25\%. Inflation expectations are anchored, but risks remain \dots
}

\vspace{1em}
\textbf{[USER INPUT]} \\
{
Now, here is a new input from 24 banks for \texttt{\{target\_date\}}. Based on the previous example, generate the MM for \texttt{\{target\_bank\_nice\_name\}}. \\
{[Input from 24 banks]} \\
\texttt{\{current\_input\_str\}}
}

\vspace{0.5em}
\textcolor{orange}{
Example: \\
{[Input from 24 banks]} \\
Federal Reserve: The Committee opted to hold, citing labor strength \dots \\
European Central Bank: Inflation continues to moderate, but caution remains \dots
}

\vspace{1em}
\textbf{[EXPECTED OUTPUT]} \\
{
{[Your Output for \texttt{\{target\_bank\_nice\_name\}}]}
}

\vspace{0.5em}
\textcolor{orange}{
Example: \\
{[Your Output for Bank of England]} \\
The MPC voted to keep the Bank Rate unchanged, citing stable growth and easing inflation momentum \dots
}

    \end{tcolorbox}
    \caption{Prompt used to generate meeting minutes using \texttt{GPT 4.1}. Orange examples show how placeholders are populated at inference time. The example sentences shown are illustrative and do not reflect actual central bank communications.}
    \label{fig:prompt_generating_minutes}
\end{figure}

\subsubsection{Hawkishness Measure}
\label{app:gen_mm_hawkishness}

To ensure a fair evaluation, we use our best-performing fine-tuned \texttt{Stance Detection} model (\texttt{RoBERTa-Large}) to label sentences in both the generated and actual documents as \texttt{Hawkish, Dovish, Neutral,} or \texttt{Irrelevant} (Section~\ref{sec:economic_analysis}). We then compute the document-level hawkishness score using the previously defined formula:
\[
z(D) \;=\; \frac{\#\text{Hawkish Sentences} - \#\text{Dovish Sentences}}{\#\text{Total Sentences} - \#\text{Irrelevant Sentences}}\,,
\]
where counts are taken over all sentences in document \(D\). We write \(z_t\) for the true post-cutoff minutes \(Y_t\) and \(\hat z_t\) for the generated minutes \(\widehat Y_t\).

To assess how well the LLM preserves policy stance, we fit the linear regression
\[
\hat z_t \;=\; \alpha + \beta\,z_t + \epsilon
\]
across all 25 banks. We obtain \(\beta = 0.6306\) ($p \leq 10^{-4}$), indicating a highly significant relationship and confirming that higher true hawkishness reliably predicts higher generated hawkishness.

\subsection{Human Evaluation}
\label{app:HumanEval}

\begin{figure}[h!]
    \centering
    \includegraphics[width=\linewidth]{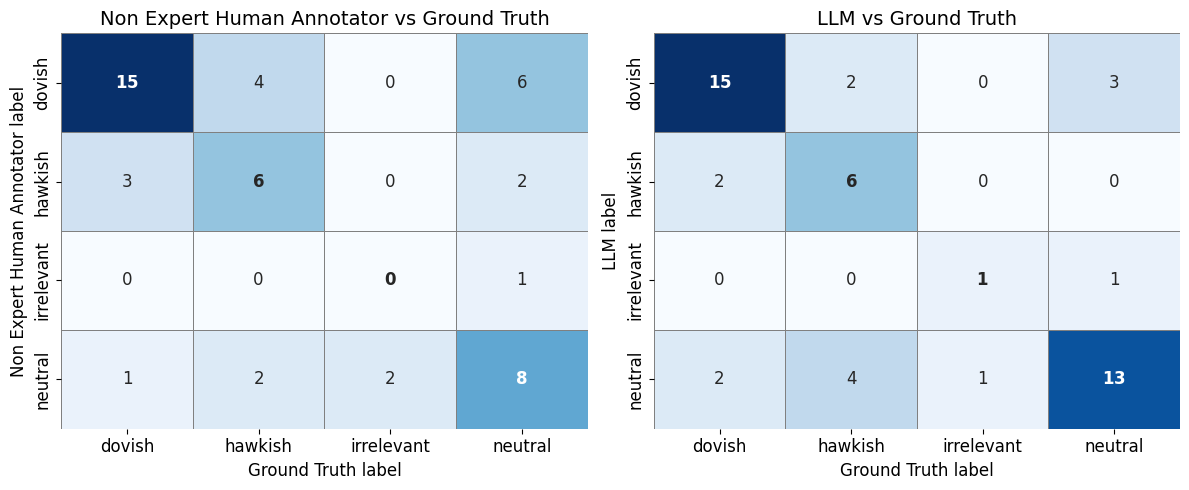}
    \caption{Side-by-side confusion matrices comparing a non-expert human annotator (left) and a large language model (right) against ground-truth stance labels: \texttt{Dovish}, \texttt{Hawkish}, \texttt{Irrelevant}, and \texttt{Neutral}. Each cell indicates the number of sentences for each ground-truth (row) and predicted (column) label pair (e.g. four \texttt{Hawkish} sentences were labeled as \texttt{Neutral} by the LLM).}
    \label{fig:human_eval_stance}
\end{figure}

\begin{figure}[h!]
    \centering
    \includegraphics[width=\linewidth]{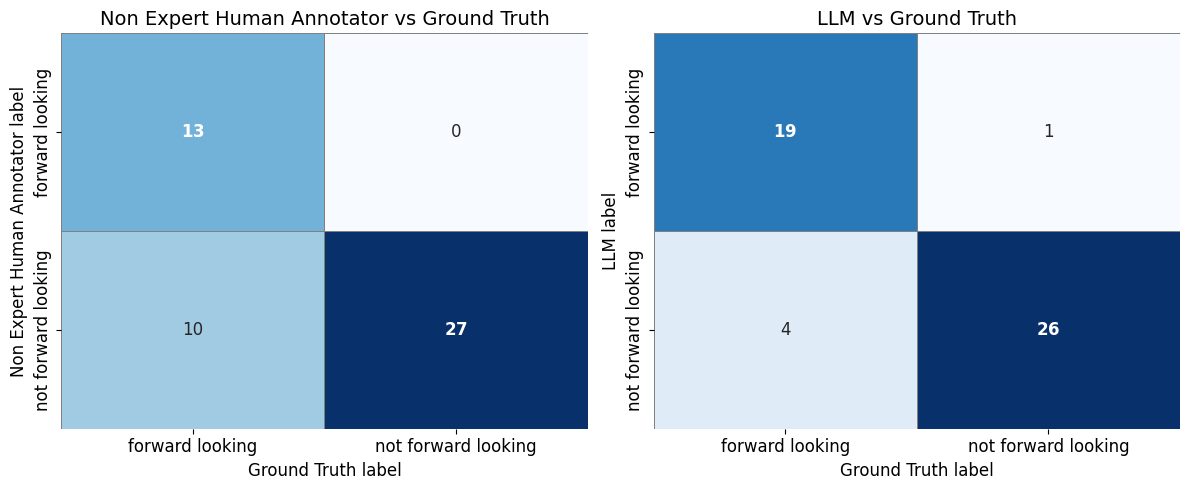}
    \caption{
    Side-by-side confusion matrices comparing a non-expert human annotator (left) and a LLM (right) against ground-truth temporal labels:\texttt{Forward Looking}, \texttt{Not Forward Looking}. Each cell indicates the number of sentences for each ground-truth (row) and predicted (column) label pair (e.g., 10 \texttt{Forward Looking} statements were labeled as \texttt{Not Forward Looking} by the LLM).}
    \label{fig:human_eval_time}
\end{figure}

\begin{figure}[h!]
    \centering
    \includegraphics[width=\linewidth]{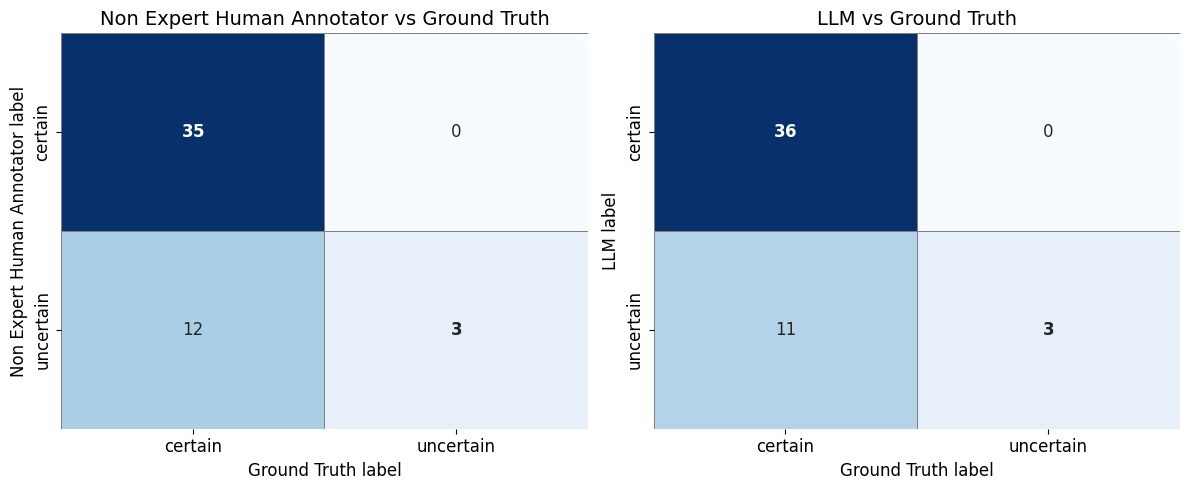}
    \caption{Side-by-side confusion matrices comparing a non-expert human annotator (left) and a LLM (right) against ground-truth certainty labels: \texttt{Uncertain}, \texttt{Certain}. Each cell indicates the number of sentences for each ground-truth (row) and predicted (column) label pair (e.g., 12 \texttt{Certain} statements were labeled as \texttt{Uncertain} looking by the LLM).}
    \label{fig:human_eval_certain}
\end{figure}

We aim to evaluate the ability of our models to understand central communications with significant real-world consequences. We compare the best-performing LLM for each task (\texttt{Stance Detection, Temporal Classification and Uncertainty Estimation}) based on weighted F1-Score on the European Central Bank (ECB) dataset to that of a non-expert annotator with no annotation guide. We specifically chose ECB because its policy moves anchor rates for 20 member states and greatly impact global fixed‑income and foreign exchange markets. The annotator is a PhD-candidate with formal work experience within the financial domain and no prior knowledge of European Union (EU) monetary policy. The best performing LLM for \texttt{Stance Detection} and \texttt{Temporal Classification} is \texttt{Llama-3-70b-chat}  and \texttt{gemini-2.0-flash} for \texttt{Certainty Estimation}. We compare the LLM and human annotations to the ground truth. The results are summarized in Figures \ref{fig:human_eval_stance}, \ref{fig:human_eval_time}, and \ref{fig:human_eval_certain}.


Across all three tasks, the LLM consistently makes fewer mistakes than the non-expert human annotator. For \texttt{Stance Detection}, the non-expert correctly classifies 29 sentences whereas the LLM classifies 35 sentences correctly, with most human errors being confusing \texttt{Hawkish} and \texttt{Neutral} sentences. For \texttt{Uncertainty Estimation}, the human mislabeled 12 \texttt{Uncertain} sentences as \texttt{Certain} while the LLM made 11 such errors. For \texttt{Temporal Classification}, the human mislabeled 10 \texttt{Forward Looking} sentences as \texttt{Not Forward Looking} compared to only four such mistakes by the LLM. These results highlight the inherent difficulty of discerning subtle cues, especially for \texttt{Neutral} or nuanced \texttt{Forward Looking} sentences in the case of the ECB dataset; and our models' emerging strength at maintaining consistency across complex, domain-specific labeling tasks.

\subsection{Transfer Learning}
\subsubsection{Beyond 25 Banks: Czech National Bank}
\label{app:PLM_Czech}
For further test the ability of our best performing models (\texttt{RoBERTa-Large} for \SD and \CE, \texttt{RoBERTa-Base} for \TC) to generalize to different economies, we test our models on sanitized data from the Czech National Bank (CNB). As with all the other central bank data, we scrape, clean, and tokenize the CNB data from 1998 to 2024. We uniformly sample 500 sentences and use the corresponding annotation guide as seen in Appendix \ref{app:czech_annot_guide} to label these sentences. Then, we test our fine-tuned models on the labeled CNB data and obtain the following weighted F1-Scores and their standard deviations:
\begin{itemize}
    \item \textbf{Stance Label:} $0.800	(0.018)$ 
    \item \textbf{Uncertainty Estimation Task:}  $0.702 (0.010)$
    \item \textbf{Time Label:} $0.726 (0.030)$
\end{itemize}
This demonstrates that our models perform well on data from central banks not present within our corpus of 25 banks and highlight the generalizability of our models.

\subsubsection{Beyond Finance: The Future of Conservation Climate, Forestry, and Natural  Resources}
\label{app:esg_transfer_learning}
Since \TC and \CE tasks extend beyond the field of economics, we test the applicability of our best performing models for each on the United States Senate Committee on Agriculture, Nutrition, and Forestry hearing transcripts \citep{CHRG-118shrg57489}. From \citet{CHRG-118shrg57489}, we collect 114 sentences using selenium and process them in the same manner we did for the central bank communications. Using these sentences, we label each sentence as \texttt{Forward Looking} Or \texttt{Not Forward Looking} and \texttt{Certain} or \texttt{Uncertain} using Table \ref{tb:ablation_forward_looking_guide} and \ref{tb:ablation_certainty_guide}. This was done by an expert annotator with the help of an annotation guide. We test our fine-tuned models on the labeled dataset. The results are summarized below. 

\textbf{ESG Annotation Guide}
\label{app:ESG_Ann_guide}
\begin{table}[h!]
\caption{{Annotation guide for classifying ESG-related sentences as \forward or \notforward.}}
\begin{tabular}{p{0.3\textwidth}p{0.3\textwidth}p{0.3\textwidth}}
\toprule
\textbf{Label} & \textbf{Description} & \textbf{Example}\\
\midrule
\textbf{Forward Looking} & When the sentence discusses expectations, projections, or anticipations about future environmental conditions, policy actions, or resource utilization. & “Moving forward, I hope that you will commit to increasing your support  of the valuable locally led initiatives like those of our conservation partners and those in the Colorado Department of Agriculture that you will hear more about from our commissioner of ag.” \\
\midrule
\textbf{Not Forward Looking} & When the sentence reflects on recent or past environmental decisions, trends, or events to describe what has already occurred. & “We do not have the wonderful forest that Colorado  has for a carbon sink, but the soil carbon sinks that we have in our States are just as important.” \\
\bottomrule
\end{tabular}
\label{tb:ablation_forward_looking_guide}
\end{table}

\begin{table}[h!]
\caption{{Annotation guide for classifying sentences from the Agriculture, Nutrition, and Forestry Committees as \certain or \uncertain}}
\begin{tabular}{p{0.3\textwidth}p{0.3\textwidth}p{0.3\textwidth}}
\toprule
\textbf{Label} & \textbf{Description} & \textbf{Example}\\
\midrule
\textbf{Certain} & When an expectation, trend, action, or outcome is definitively stated without ambiguity. Key words include ``is,'' ``are,'' ``decreased,'' and ``will continue.''
 & “Last but not least, Mr. Alexander Funk is the director of  Water Resources and senior counsel for the Theodore Roosevelt  Conservation Partnership.” \\
\midrule
\textbf{Uncertain} & When it expresses concerns, potential risks, or
need to alter monetary policy if a specific event comes to pass. Key words include ``can,'' ``if,'' ``could,'' and ``uncertain.'' & “That said, the conservation title can play a great role in  addressing drought conditions, and there are several  opportunities to help Western farmers and ranchers.” \\
\bottomrule
\end{tabular}
\label{tb:ablation_certainty_guide}
\end{table}
\textbf{Results}
As a result of the testing on the Congress Committee hearings data from the conservation, climate, forestry, and natural resources of United States Senate Committee on Agriculture, Nutrition, and Forestry, we find the following weighted F1-Scores:
\begin{itemize}
    \item \textbf{Time Label:} 0.879 \\
    \item \textbf{Certain Label:} 0.683 \\
\end{itemize}
Furthermore, according to Tables \ref{tab:ESG_labels} and \ref{tab:ESG_confusion_matrix}, the models correctly labeled $100/144 (88\%)$ sentences for \texttt{Temporal Classification} and $87/114 (76\%)$ sentences for \texttt{Uncertainty Estimation}. 
\begin{table}[ht]
\caption{{Confusion matrix comparing predicted and ground truth \TC of ESG-related sentences.}}
\label{tab:ESG_labels}
\centering
\begin{tabular}{lcc}
\toprule
\multicolumn{3}{c}{\textbf{Temporal Classification}} \\
\midrule
 & \textbf{Predicted Not Forward} & \textbf{Predicted Forward} \\
\midrule
True Not Forward  & 35 &  4 \\
True Forward      & 10 & 65 \\
\bottomrule
\end{tabular}

\end{table}

\begin{table}[ht]
\caption{{Confusion matrix comparing predicted and ground truth \CE of sentences from Agriculture, Nutrition, and Forestry Committees.}}
\label{tab:ESG_confusion_matrix}
\centering
\begin{tabular}{lcc}
\toprule
\multicolumn{3}{c}{\textbf{Uncertainty Estimation}} \\
\midrule
 & \textbf{Predicted Certain} & \textbf{Predicted Uncertain} \\
\midrule
True Certain      & 84 &  0 \\
True Uncertain    & 27 &  3 \\
\bottomrule
\end{tabular}
\end{table}
We see that these scores are significant, demonstrating the applicability of our general models fined-tuned for \texttt{Temporal Classification} and \texttt{Uncertainty Estimation} to other domains.
\clearpage
 
\clearpage

\section{Extended Economic Analysis}
\label{app:ext_econ_analysis}
\subsection{Data Collection and Aggregation}
We collect year-over-year (YoY), percent change data for non-seasonally adjusted  Consumer Price Index (CPI) from the Bloomberg Terminal at the Georgia Institute of Technology. We use the non-seasonally adjusted indices because it offers a more complete picture of the macroeconomic trends at that time. All central banks use CPI, with the exception of the European Central Bank and the Bank of England. They use Harmonized CPI which is CPI excluding the rent-equivalent of owner-owned dwellings.  

The table below lists the exact CPI index used for each central bank.
\begin{table*}[h]
\scriptsize
\caption{{Summary of metadata for Consumer Price Index (CPI) data corresponding to the 25 central banks. \textit{Name of Index} refers to the official title of the index. \textit{Index} denotes the Bloomberg ticker. \textit{Source} indicates the government agency responsible for data collection. \textit{Date} and \textit{Time Collected }specify when the data was retrieved from the Bloomberg Terminal.}}
\label{tab:central_banks_stats}
\begin{tabular}{
    p{0.09\linewidth} p{0.25\linewidth} p{0.08\linewidth} 
    p{0.2\linewidth} p{0.1\linewidth} p{0.1\linewidth} 
}
\toprule
\textbf{Bank} & \textbf{Name of Index} & \textbf{Index} & \textbf{Source} & \textbf{Date} & \textbf{Time} \\
\textbf{} & \textbf{} & \textbf{} & \textbf{} & \textbf{Collected} & \textbf{Collected} \\
\midrule
\USA FOMC & US CPI Urban Consumers YoY NSA & CPI YOY & Bureau of Labor Statistics & Apr 22 & 15:07 \\
\CHINA PBoC & China CPI YoY & CNCPIYOY & National Bureau of Statistics of China & Apr 22 & 15:07 \\
\JAPAN BoJ & Japan CPI Nationwide YoY & JNCPIYOY & Ministry of Internal Affairs and Communications & Apr 22 & 15:07 \\
\UK BoE & UK CPI EU Harmonized YoY NSA & UKRPCJYR & UK Office for National Statistics & Apr 22 & 15:07 \\
\SWITZERLAND SNB & Switzerland CPI All Items YoY & SZCPIYOY & Federal Statistics Office of Switzerland & Apr 22 & 15:07 \\
\BRAZIL BCB & Brazil CPI IPCA YoY & BZPIIPCY & The Organization for Economic Co-operation and Development & Apr 22 & 15:07 \\
\INDIA RBI & India CPI Combined YoY & INFUTOTY & Central Statistics Office India & Apr 22 & 15:20 \\
\EU ECB & Eurostat EU HICP All Items YoY NSA & ECCPEUY & Eurostat & Apr 22 & 15:20 \\
\RUSSIA CBR & Russia CPI YoY & RUCPIYOY & Federal Service of State Statistics & Apr 22 & 15:20 \\
\TAIWAN CBCTW & Taiwan CPI YoY NSA & TWCPIYOY & Taiwan Directorate General of Budget Accounting and Statistics & Apr 22 & 15:20 \\
\SINGAPORE MAS & Singapore CPI All Items YoY & SICPIYOY & Singapore Ministry of Trade and Industry & Apr 22 & 15:20 \\
\SOUTHKOREA BoK & South Korea CPI YoY & KOCPIYOY & Bank of Korea & Apr 22 & 15:20 \\
\AUS RBA & Australia CPI All Items YoY & AUCPIYOY & Australian Bureau of Statistics & Apr 22 & 15:20 \\
\ISRAEL BoI & Israel CPI YoY NSA & ISCPIYYN & Israel Central Bureau of Statistics & Apr 22 & 15:20 \\
\CANADA BoC & STCA Canada CPI YoY NSA 2002=100 & CACPIYOY & Statistics Canada & Apr 22 & 15:20 \\
\MEXICO BdeM & Mexico CPI Yearly Percent Change Biweekly & MXBWYOY & National Institute of Statistics and Geography & Apr 22 & 15:20 \\
\POLAND NBP & Poland CPI All Items YoY & POCPIYOY & Polish Statistics Office & Apr 22 & 15:20 \\
\TURKEY CBRT & Turkey CPI YOY \% & TUCPIY & Turkish Statistical Institute & Apr 22 & 15:20 \\
\THAILAND BoT & Thailand CPI All Items YoY & THCPIYOY & Office of the National Economic and Social Development Council & Apr 22 & 15:20 \\
\MALAYSIA BNM & Malaysia CPI YoY 2010=100 & MACPIYOY & Department of Statistics Malaysia & Apr 22 & 15:20 \\
\PHILIPPINES BSP & Philippine CPI All items YoY\% 2018=100& PHC2II& Philippine Statistics Authority & Apr 27& 13:28\\
\EGYPT CBE & Egypt Monthly Urban CPI YoY\% & EGCPYOY & Central Agency for Public Mobilization and Statistics & Apr 22 & 15:20 \\
\CHILE CBoC & Chile CPI YoY Chained & CLINNSYO & Instituto Nacional de Estadistica – Chile & Apr 22 & 15:20 \\
\PERU BCRP & Lima CPI YoY & PRCPYOY & Instituto Nacional De Estadistica E Informatica De Peru & Apr 22 & 15:20 \\
\COLOMBIA BanRep & Colombia CPI YoY & COCPIYOY & Departamento Administrativo Nacional de Estadistica & Apr 22 & 15:20 \\

\bottomrule
\end{tabular}

\end{table*}

\subsection{Hawkishness and Inflation}
\label{app:hawkish-inflation-plots}
We develop several dual‑axis time‑series plot for all banks individually. For all the plots, on the left axis, we present model‑derived hawkishness scores (defined in Section \ref{sec:economic_analysis}) from monetary‑policy communications whereas the right axis represents the specific economy's inflation in terms of Consumer Price Inflation (CPI). 
\begin{figure}
    \centering
    \includegraphics[width=\linewidth]{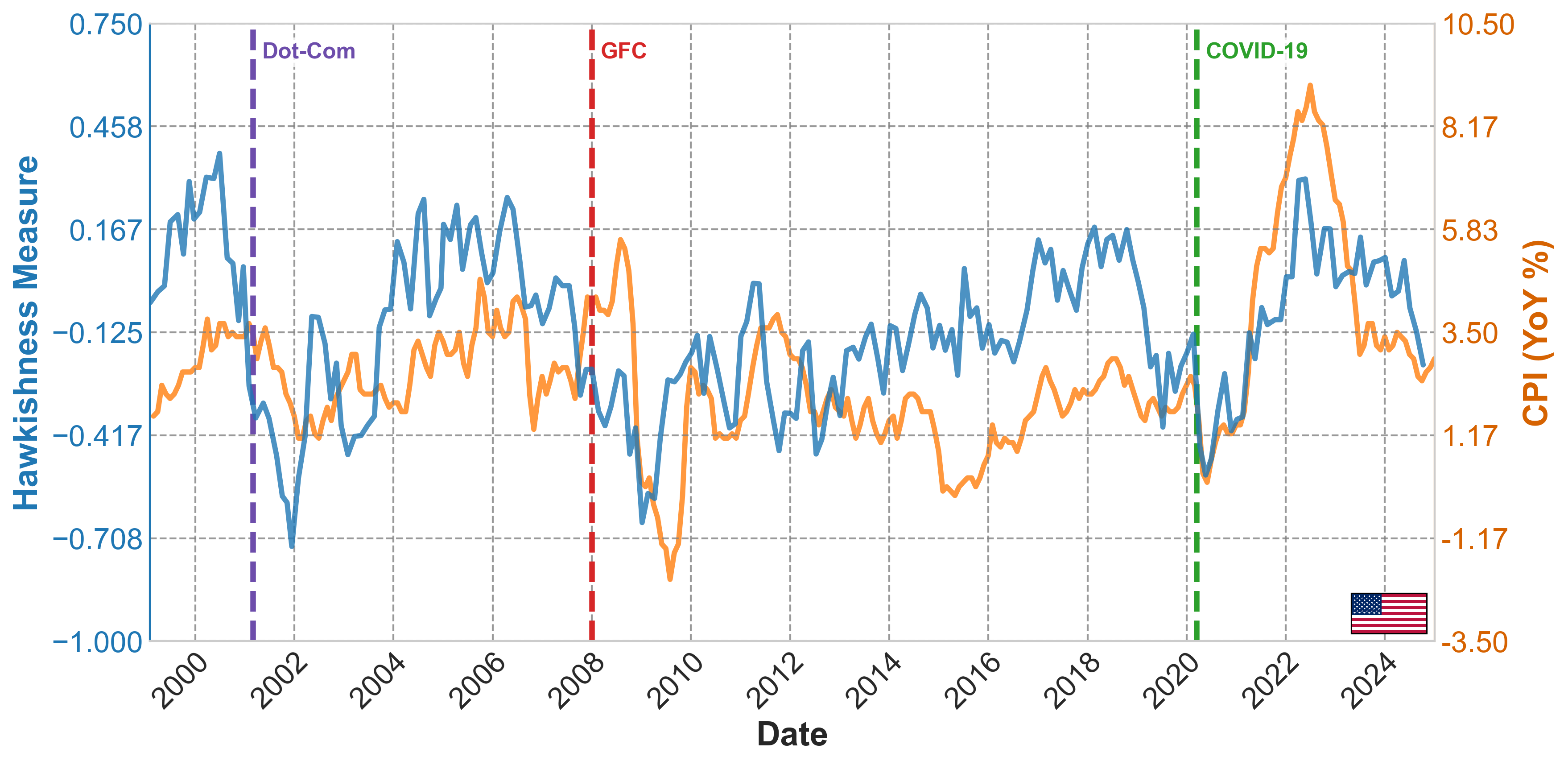}
    \caption{This figure illustrates how the events of the 2001 dot-com bust, 2008 financial crisis, and COVID-19 pandemic affect both the United States's inflation index and our hawkishness measure. These periods were marked by dovish monetary policy as economic growth slowed and unemployment rose, and the FOMC cut rates to encourage spending. Additionally, notice how our hawkish measure aligns with the inflation index post-pandemic. Due to high stimulus spending, supply chain disruptions, and energy price shocks due to the events of Ukraine-Russia, prompted the FOMC to raise rates to bring inflation down to target levels.}
    \label{fig:FOMC-INFLATION}
\end{figure}

\begin{figure}
    \centering
    \includegraphics[width=\linewidth]{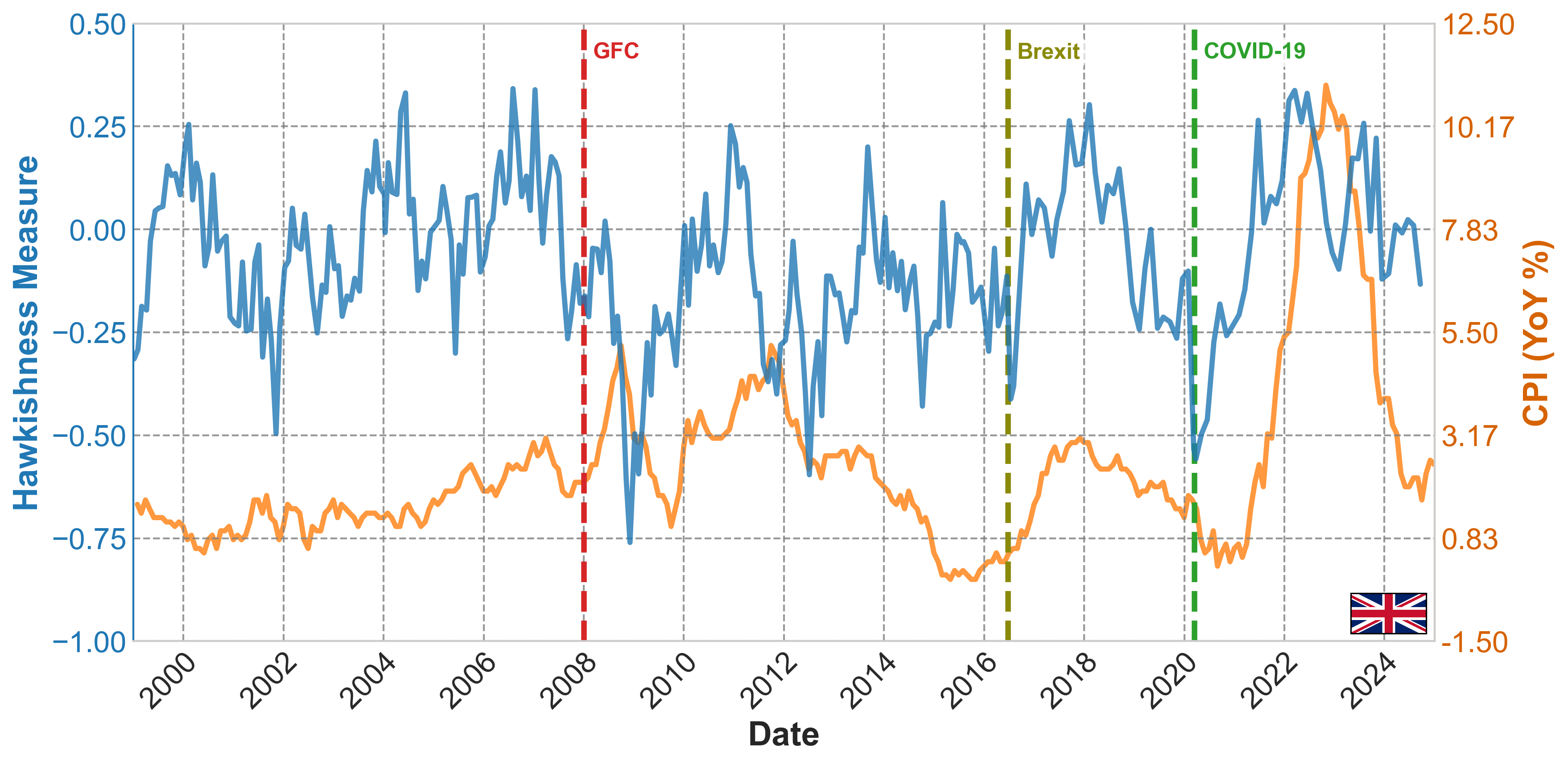}
    \caption{This figure illustrates how the events of the 2008 financial crisis and 2020 COVID-19 pandemic affect both the United Kingdom's inflation index and our hawkishness measure. These events were met with dovish monetary policies by BoE, such as asset buybacks and rate cutting, to stimulate the economy. Both measures also capture the uncertainty caused by Brexit in 2016 and high inflation rate in 2022 caused by supply chain disruptions and high consumer demand, which prompted the BoE to raise rates to stabilize the economy and lower inflation.}
    \label{fig:BoE-INFLATION}
\end{figure}

\begin{figure}
    \centering
    \includegraphics[width=\linewidth]{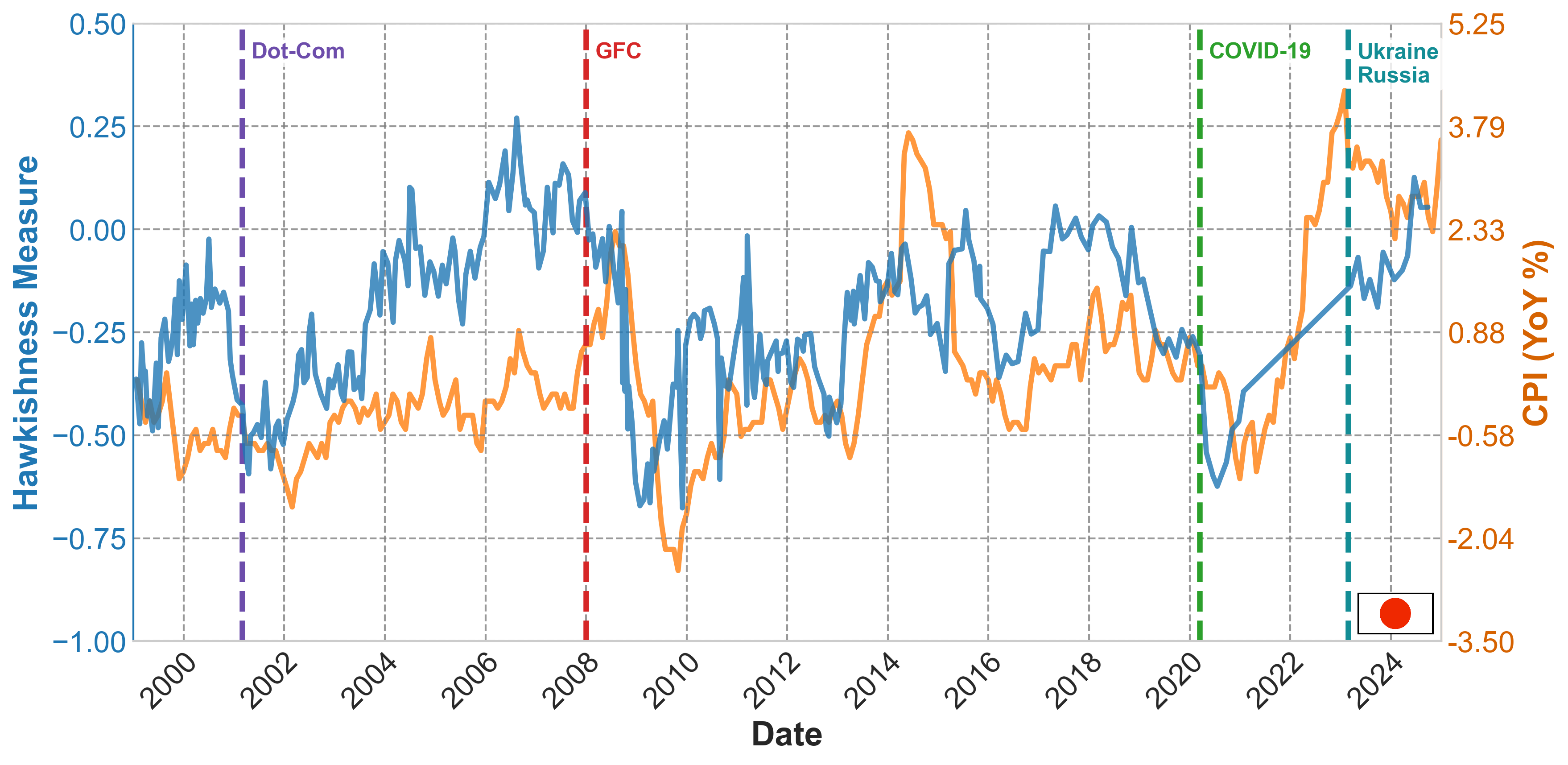}
    \caption{This figure illustrates how the events of the 2001 dot-com bust, 2008 financial crisis, and 2020 COVID-19 pandemic affect both the Japan's inflation index and our hawkishness measure. BoJ was extremely dovish to stimulate the economy during these times. Additionally, both metrics capture the high inflation and subsequent hawkish policy caused by supply chain disruptions due to the pandemic and rising commodity, energy, and food prices following the events of Ukraine-Russia.}
    \label{fig:BoJ-INFLATION}
\end{figure}

\begin{figure}
    \centering
    \includegraphics[width=\linewidth]{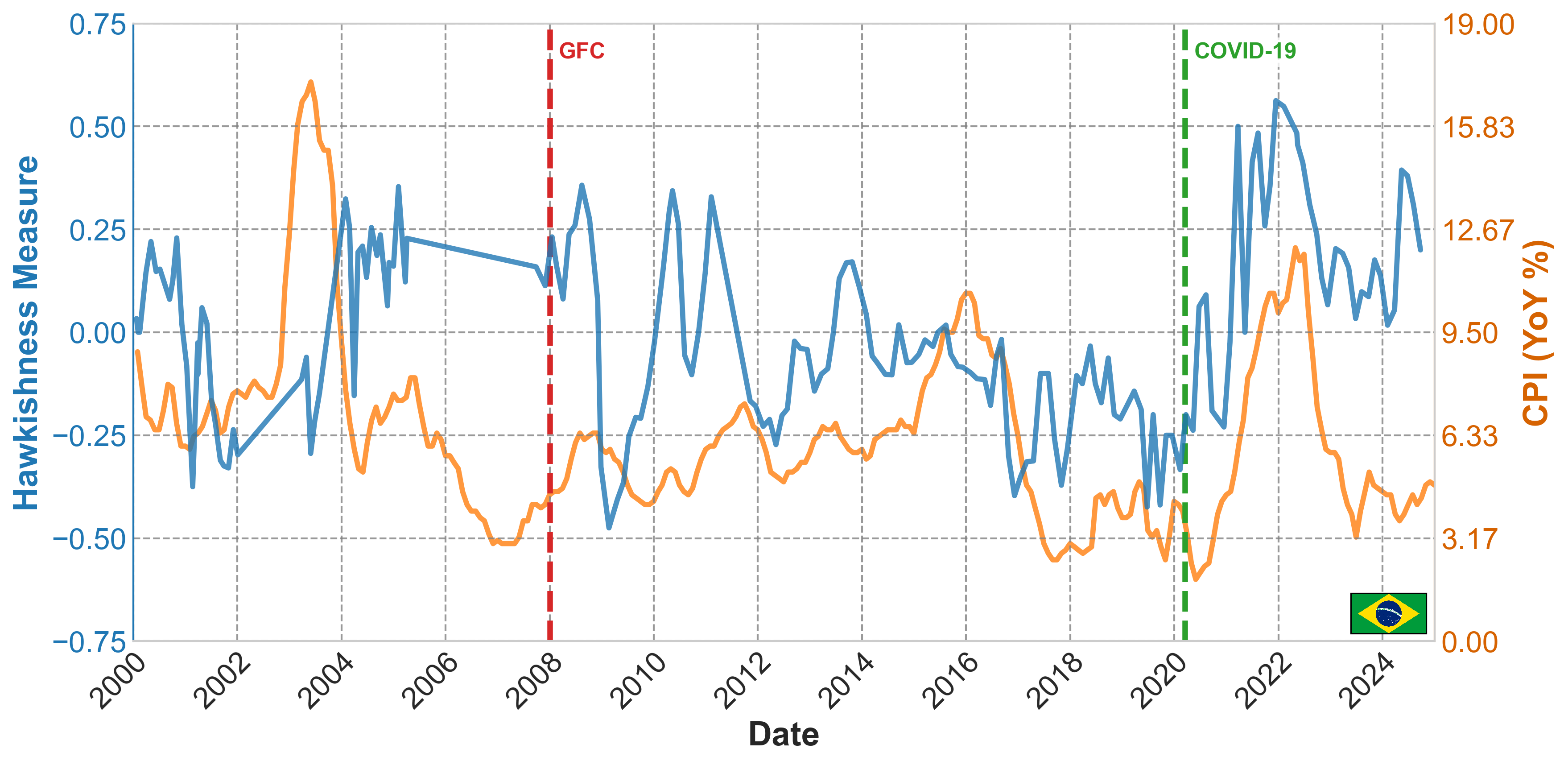}
    \caption{This figure illustrates how the events of the 2008 financial crisis as well as the COVID-19 pandemic affect both the Brazil's inflation index and our hawkishness measure. High levels of hawkishness during these periods reflects central bank's efforts to counteract rising inflation and return it to target levels. Additionally, the figure captures the heightened volatility in both inflation and hawkishness measures as the economy stabilizes following the 1999 financial crisis, during which high domestic debt was amounting to be 40\% of GDP thus, leading to the devaluation of the Brazil real against the U.S. dollar.}
    \label{fig:BCB-INFLATION}
\end{figure}

\begin{figure}
    \centering
    \includegraphics[width=\linewidth]{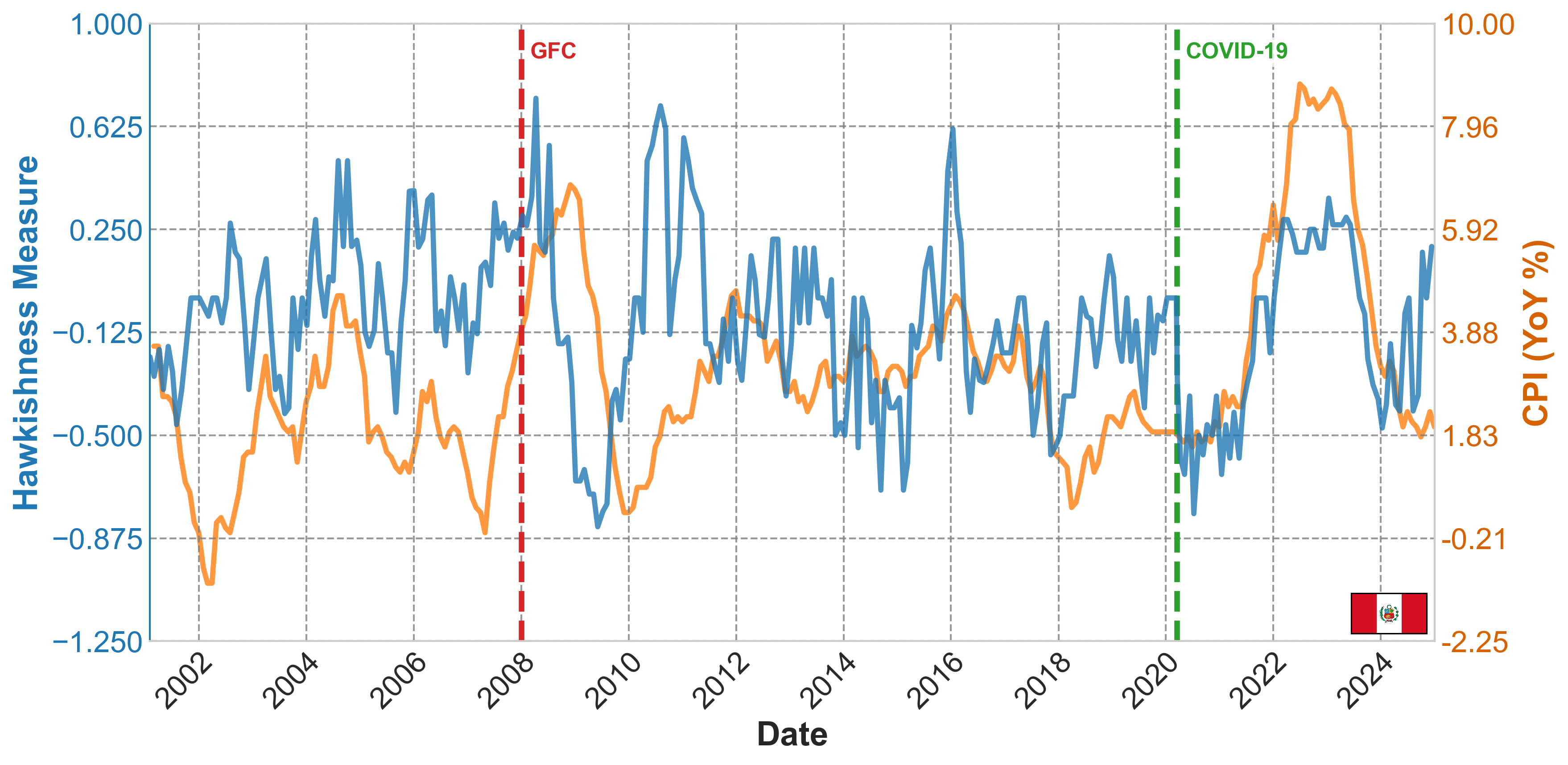}
    \caption{This figure illustrates how the events of the 2008 financial crisis and the 2020 COVID-19 pandemic affect both the Peru's inflation index and our hawkishness measure. BRCP implemented dovish monetary policy during these times to stimulate the economy. Both measures capture the surge in inflation and the corresponding spike in hawkishness caused by disruptions in supply chain and consumer demand, resulting in the BRCP raising rates to promote economic and price stability.}
    \label{fig:BCRP-INFLATION}
\end{figure}

\begin{figure}
    \centering
    \includegraphics[width=\linewidth]{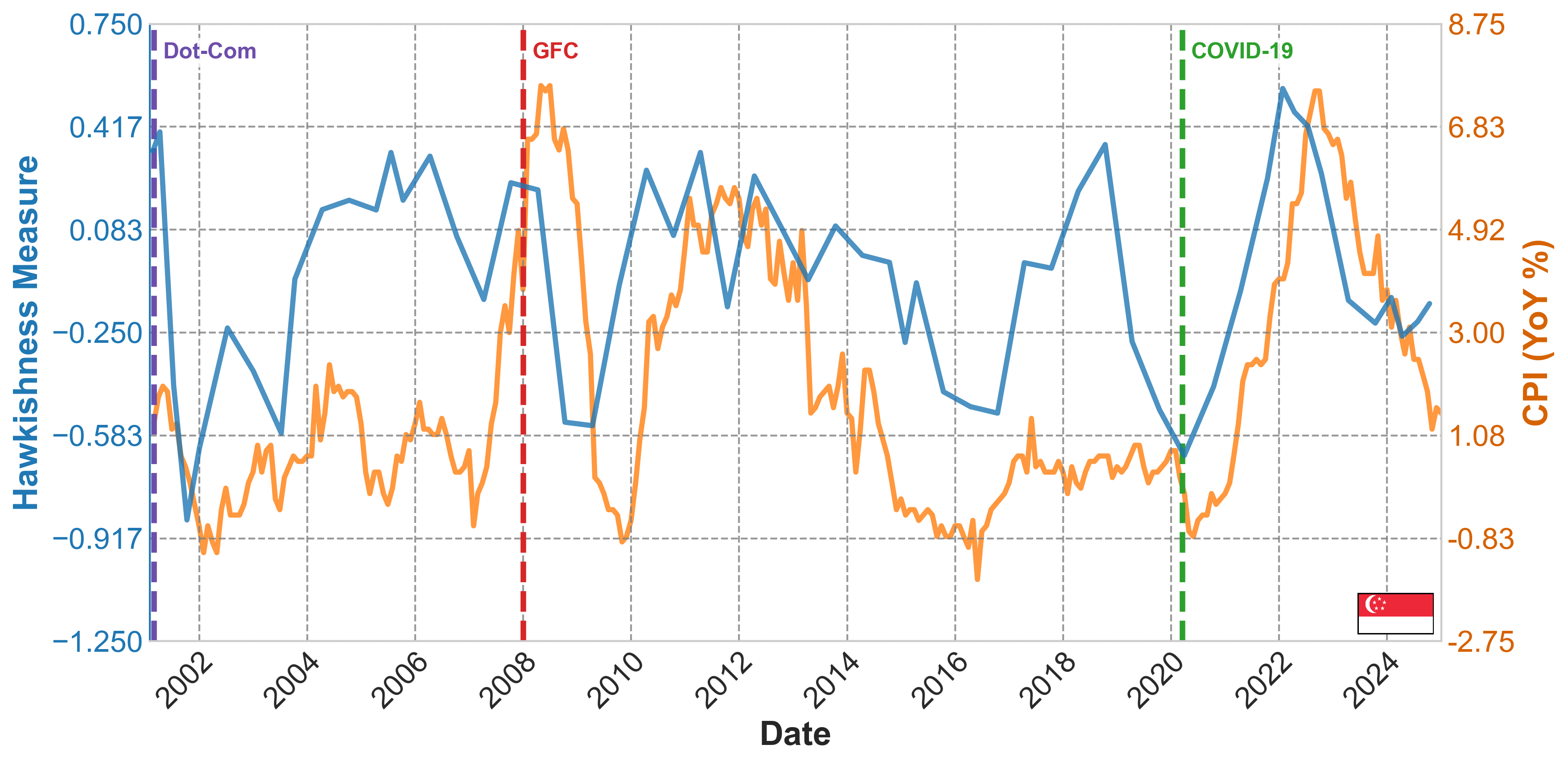}
    \caption{This figure illustrates how the events of the 2001 dot-com bust, 2008 financial crisis, and COVID-19 pandemic affect both the Singapore's inflation index and our hawkishness measure. These periods were marked by dovish monetary policy followed by high hawkishness as the economy contracted, and the MAS worked to stimulate the economy and control inflation after doing so. }
    \label{fig:MAS-INFLATION}
\end{figure}

\begin{figure}
    \centering
    \includegraphics[width=\linewidth]{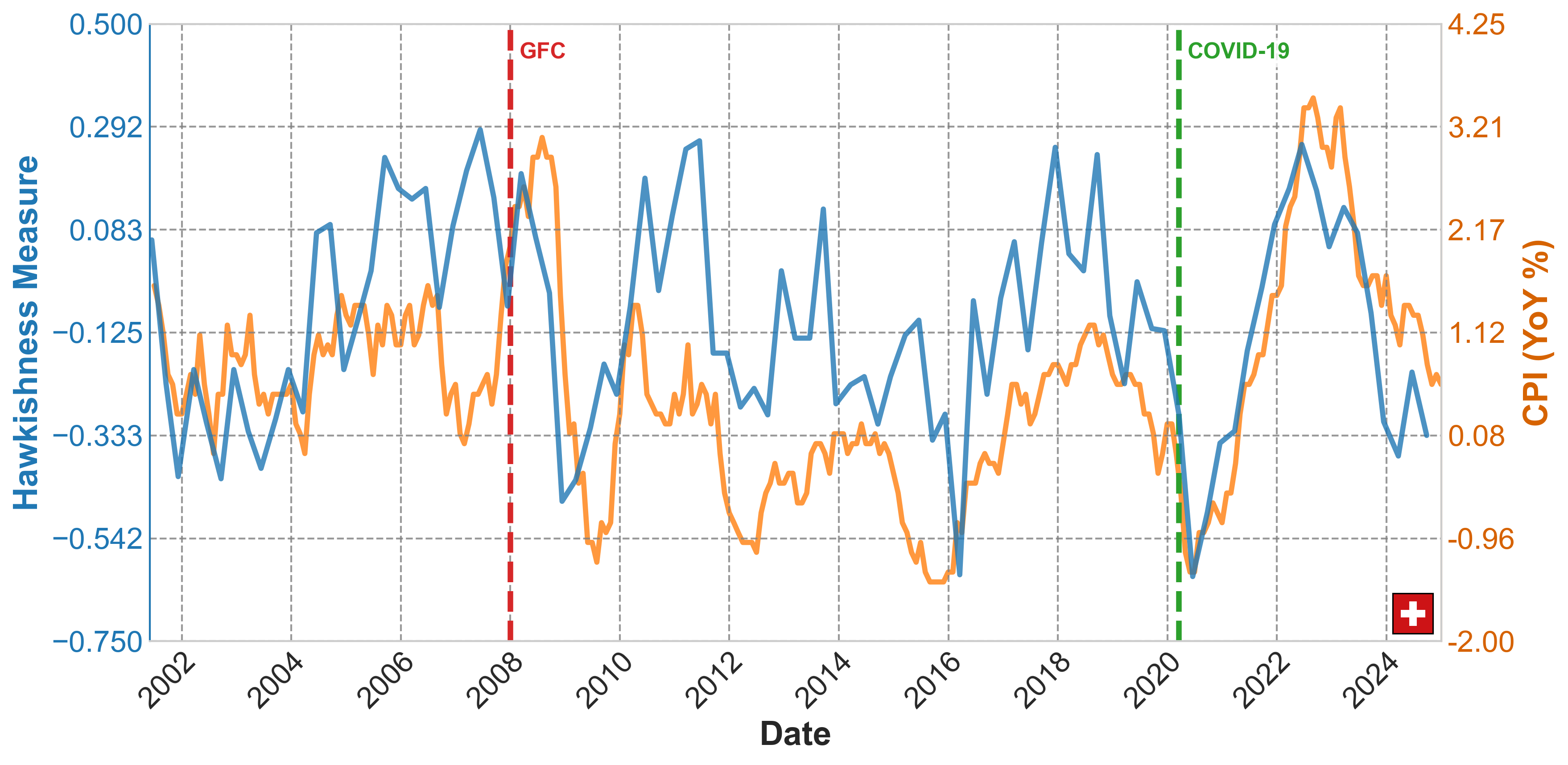}
    \caption{This figure illustrates how the events of the 2008 financial crisis and 2020 COVID-19 pandemic affect both the Switzerland's inflation index and our hawkishness measure. Both periods are marked by extremely low hawkishness due to the SNB trying to stimulate the economy. The figure also captures declining hawkishness in 2011 due to the Eurozone Crisis negatively impacting the exports. Our hawkishness measure also captures high inflation rates post-pandemic that SNB was combating.}
    \label{fig:SNB-INFLATION}
\end{figure}

\begin{figure}
    \centering
    \includegraphics[width=\linewidth]{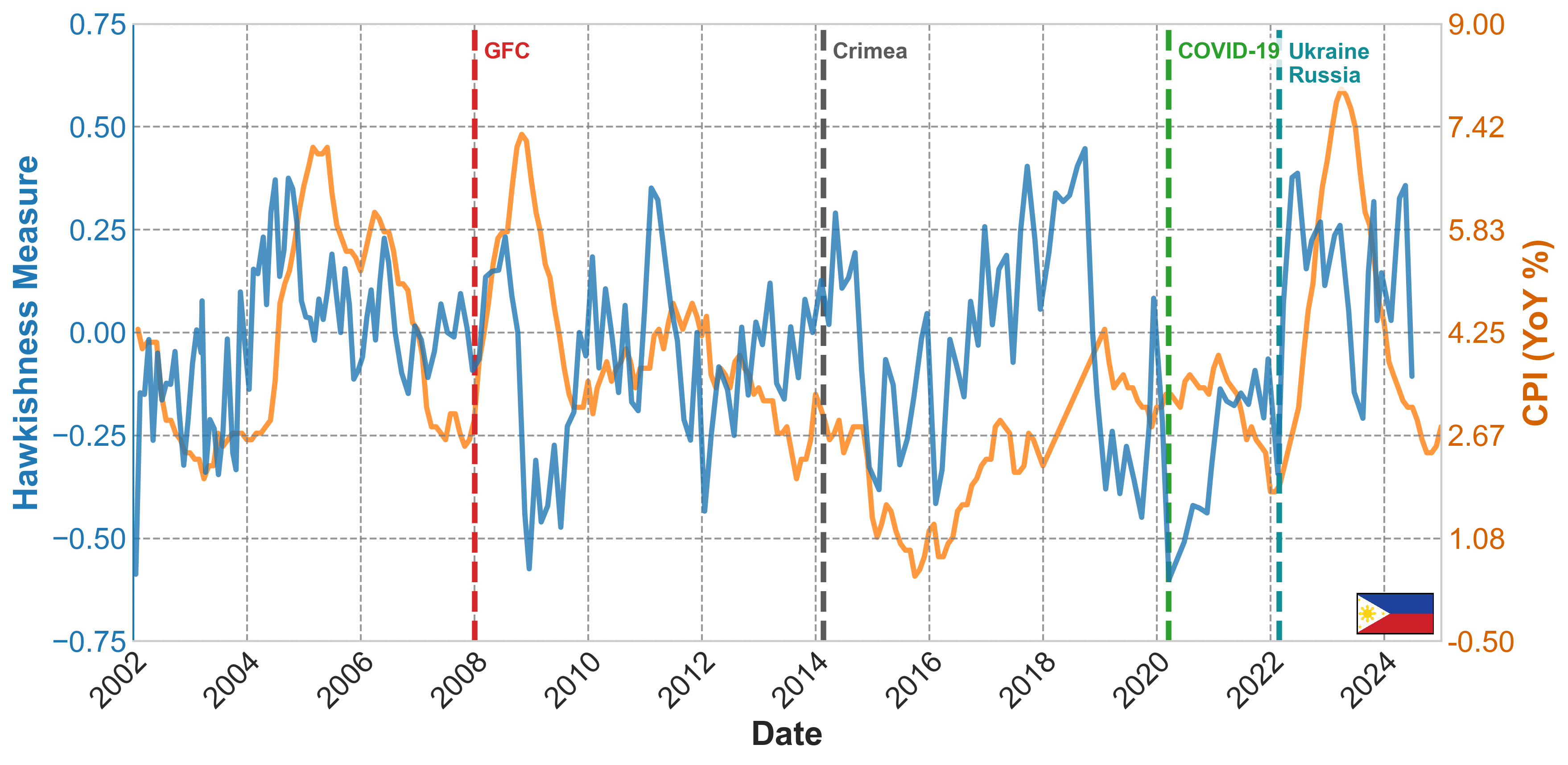}
    \caption{This figure illustrates how the events of the oil shocks in 2008 and 2014 affect both the Philippines's inflation index and our hawkishness measure. Following the oil shock in 2008, the CPI declines in response to increased hawkish monetary policy implementation. Additionally, the figure captures the effects of the 2020 COVID-19 pandemic within the inflation index and our hawkishness measure, resulting in dovish monetary policy to stimulate the economy. Both measures also capture the hawkish monetary policy due to surging food and energy prices from the events of Ukraine-Russia, prompting the BSP to raise rates in response to high inflation. }
    \label{fig:BSP-INFLATION}
\end{figure}

\begin{figure}
    \centering
    \includegraphics[width=\linewidth]{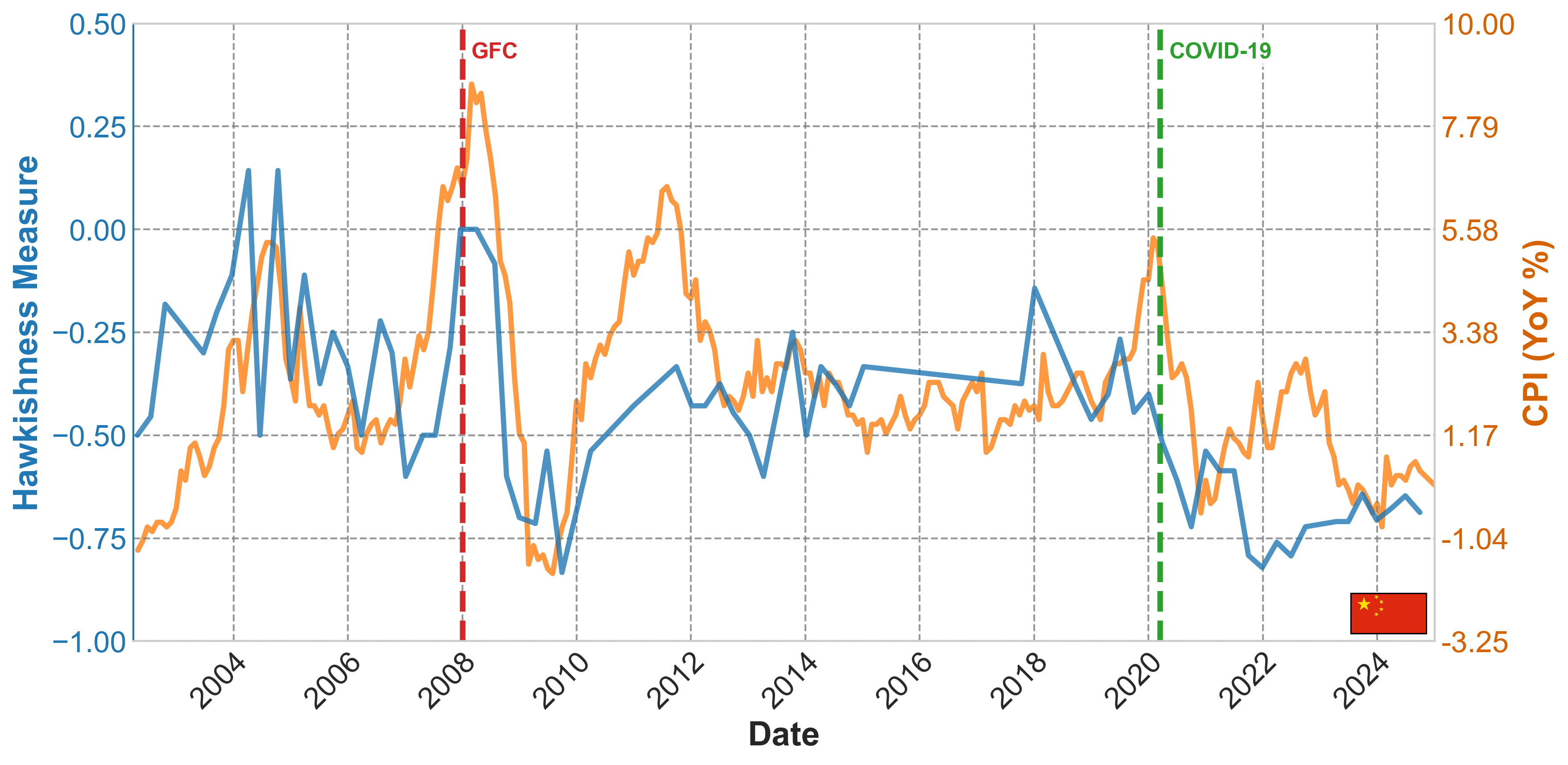}
    \caption{This figure illustrates how the events of the 2008 financial crisis and 2020 COVID-19 pandemic affect both the China's inflation index and our hawkishness measure. Both periods are marked by extremely low hawkishness due to the PBoC working to stimulate the economy.}
    \label{fig:PBoC-INFLATION}
\end{figure}

\begin{figure}
    \centering
    \includegraphics[width=\linewidth]{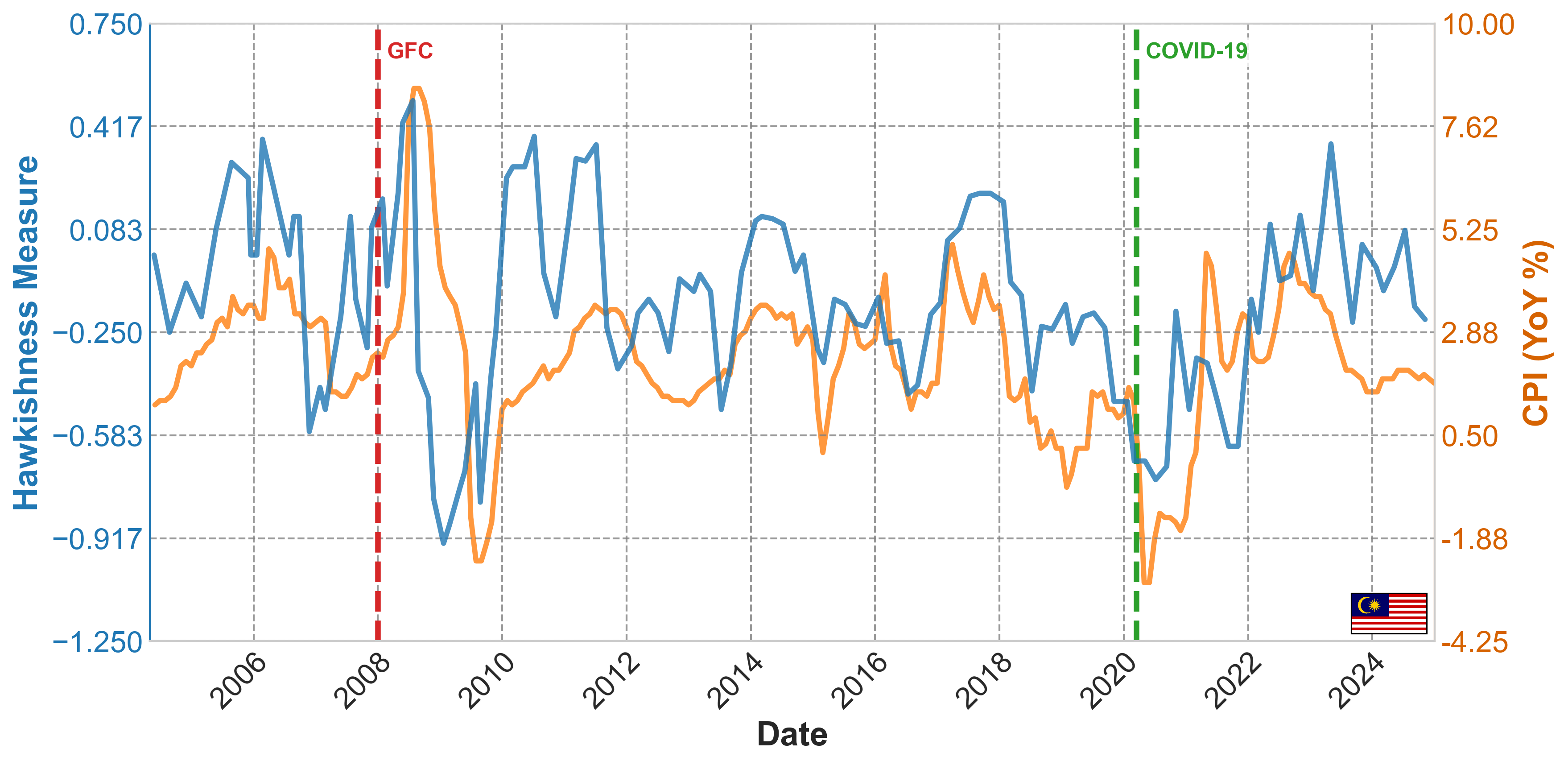}
    \caption{This figure illustrates how the events of the 2008 financial crisis and 2020 COVID-19 pandemic affect both the Malaysia's inflation index and our hawkishness. During these times BNM adopted dovish monetary policy to stimulate the economy. Both measures also capture the high inflation and subsequent hawkish policy caused by the strengthening of the US dollar, the events of Ukraine-Russia, and natural events which disrupted the agriculture sector and food chain.}
    \label{fig:BNM-INFLATION}
\end{figure}

\begin{figure}
    \centering
    \includegraphics[width=\linewidth]{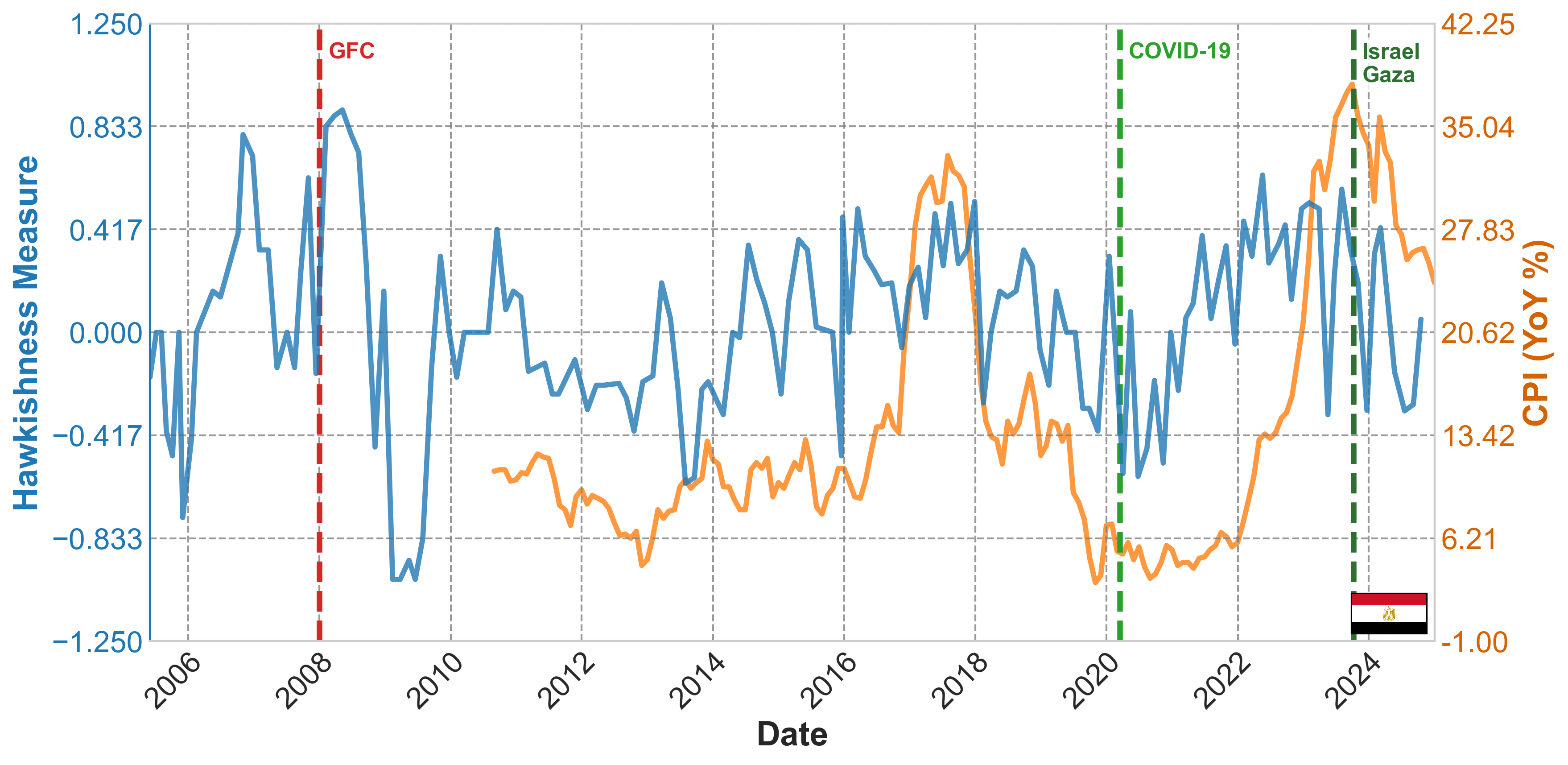}
    \caption{This figure illustrates the effects of the 2020 COVID-19 pandemic within the inflation index and our hawkishness measure during which dovish monetary policy prevailed in order to stimulate the economy. Ukraine-Russia hit the Egyptian economy hard, causing spike in oil and wheat prices and the depreciation of the EGP which led to rising inflation, prompting the CBE is raise rates multiple times.}
    \label{fig:CBE-INFLATION}
\end{figure}

\begin{figure}
    \centering
    \includegraphics[width=\linewidth]{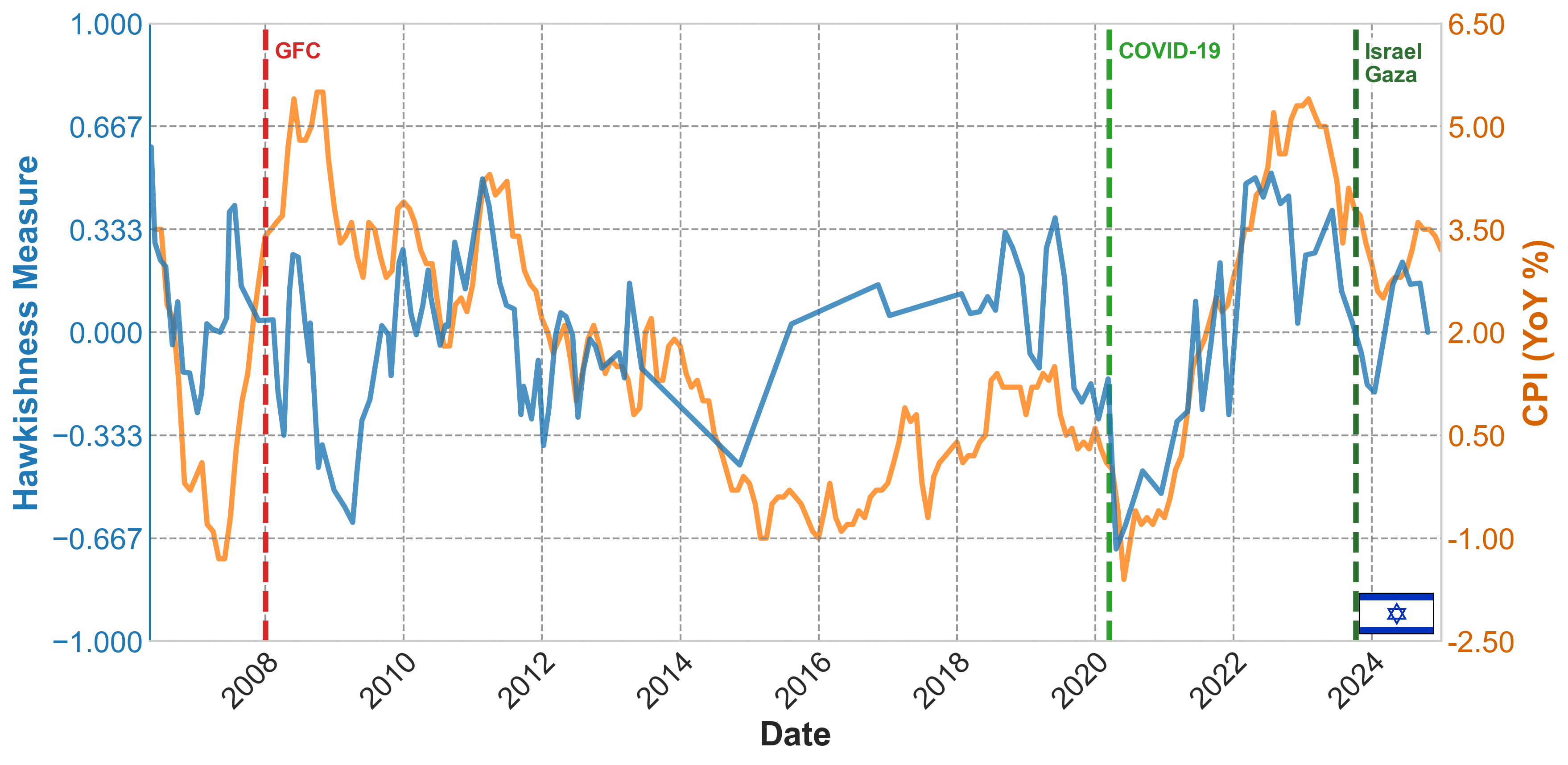}
    \caption{This figure illustrates how the events of the 2020 COVID-19 pandemic and the Israel-Hamas conflict in 2023 affect both the Israel's inflation index and our hawkishness measure. BoI implemented dovish monetary policy, including asset buybacks and rate cuts, to stimulate the economy during these times as major industries such as tourism were impacted.}
    \label{fig:BoI-INFLATION}
\end{figure}

\begin{figure}
    \centering
    \includegraphics[width=\linewidth]{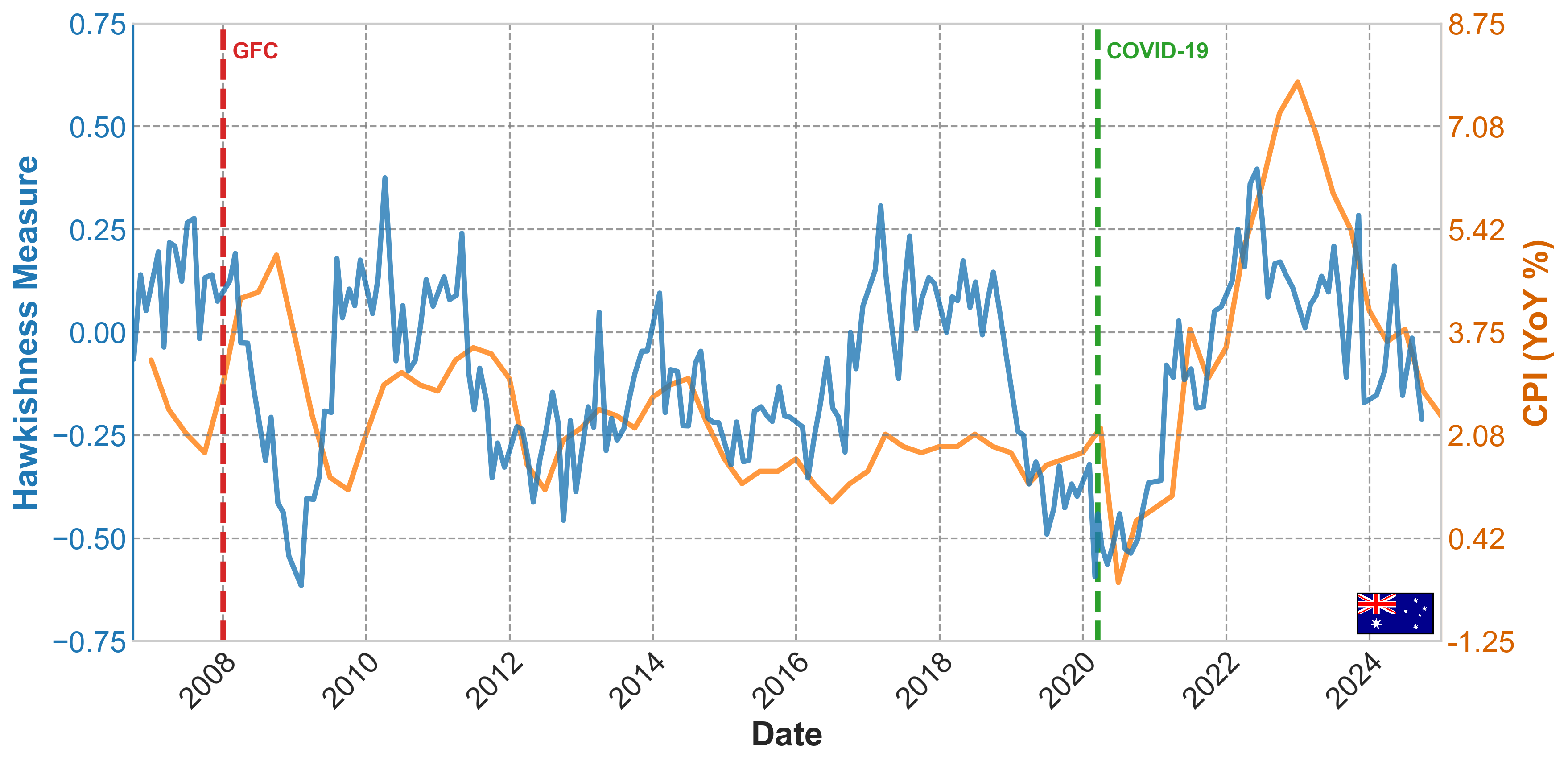}
    \caption{This figure illustrates how the events of the 2008 financial crisis and 2020 COVID-19 pandemic affect both the Australia's inflation index and our hawkishness measure. Both periods are marked by extremely low hawkishness due to the RBA working to stimulate the economy. During the pandemic, the efforts included lowering the cash rate, providing low-cost, long-term funding to banks, and purchasing government bonds. Our measure also captures the rising hawkishness post-pandemic, as the RBA worked to stabilize the economy. }
    \label{fig:RBA-INFLATION}
\end{figure}

\begin{figure}
    \centering
    \includegraphics[width=\linewidth]{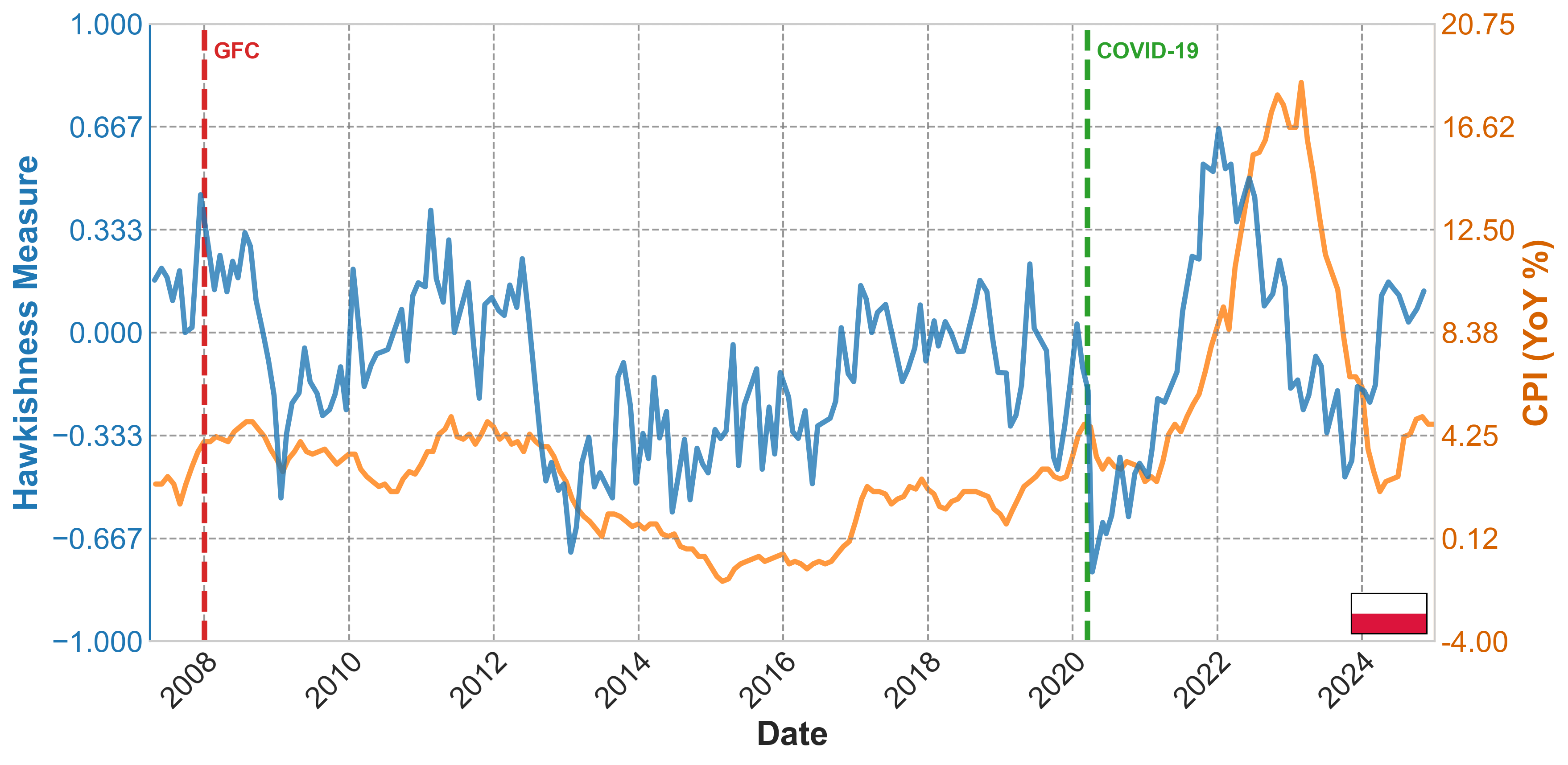}
    \caption{This figure illustrates how the events of the 2008 financial crisis and 2020 COVID-19 pandemic affect both the Poland's inflation index and our hawkishness measure.  Both periods are marked by extremely low hawkishness due to the NBP working to stimulate the economy. Additionally, the figure captures hawkishness measure rising with the inflation index as the NBP raises rates to combat the high inflation level post-pandemic.}
    \label{fig:NBP-INFLATION}
\end{figure}

\begin{figure}
    \centering
    \includegraphics[width=\linewidth]{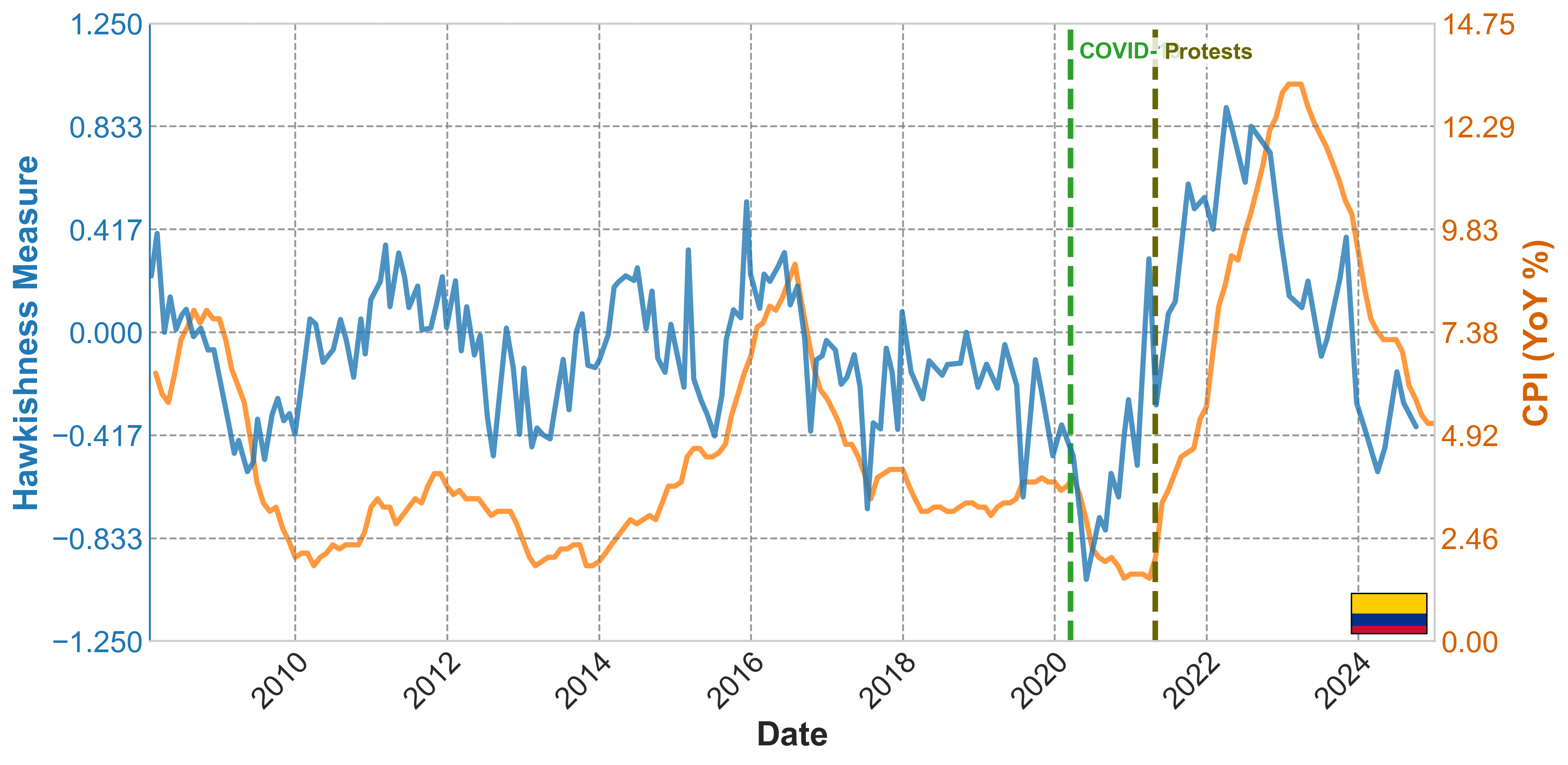}
    \caption{This figure illustrates how the events of the 2020 COVID-19 pandemic affect both the Colombia's inflation index and our hawkishness measure. BanRep implemented dovish monetary policy during these times to stimulate the economy. Both measures also capture the high inflation rate in 2022 caused by supply chain disruptions and high consumer demand, which prompted BanRep to raise rates to stabilize the economy and lower inflation. }
    \label{fig:BanRep-INFLATION}
\end{figure}

\begin{figure}
    \centering
    \includegraphics[width=\linewidth]{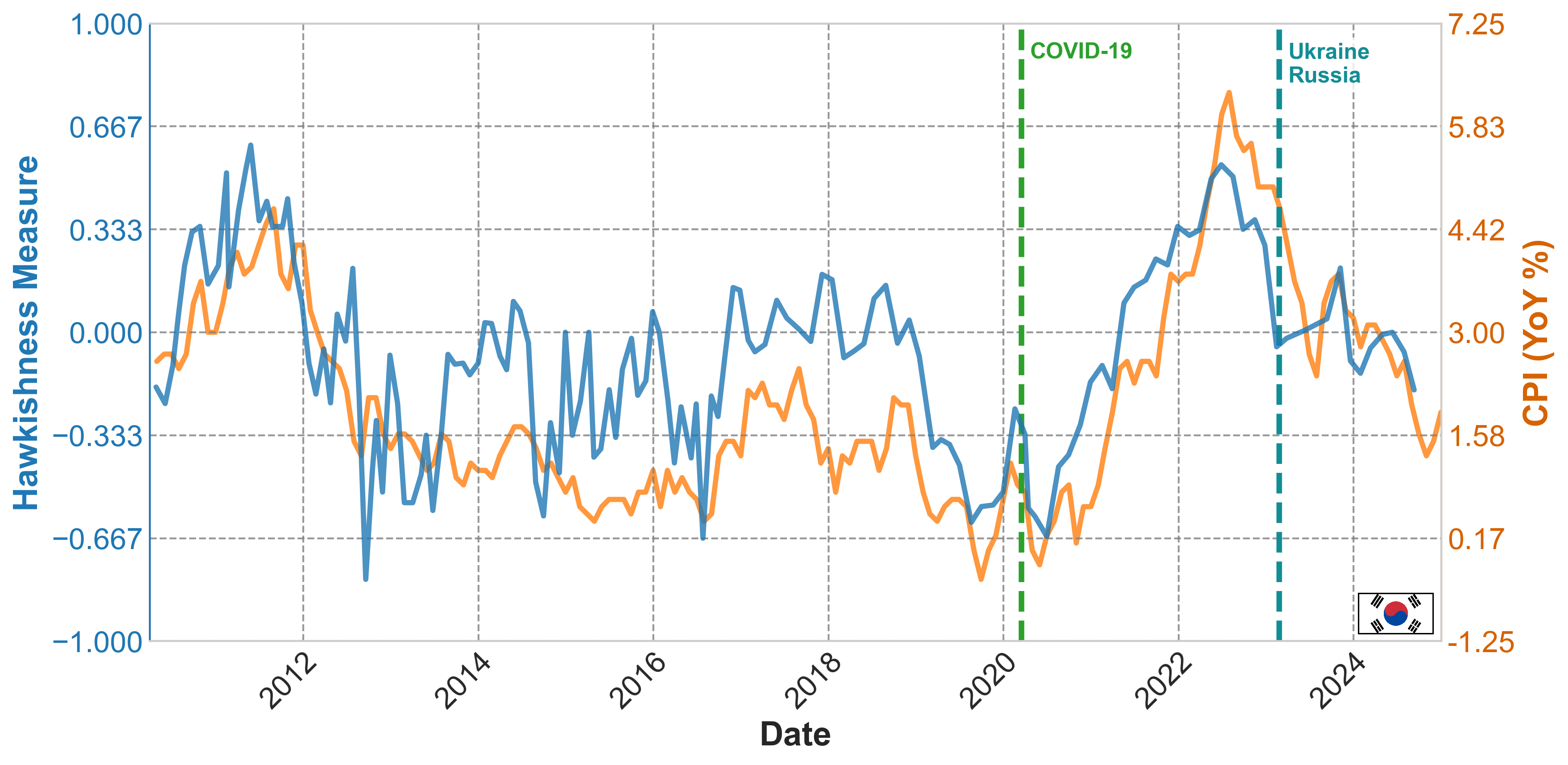}
    \caption{This figure illustrates the effects of the 2020 COVID-19 pandemic on South Korea's inflation index and our hawkishness measure. During this period, the dovish monetary policy was adopted by the central bank in order to stimulate the economy. Both indicators capture the steep incline in inflation in 2022, driven by the post-pandemic supply chain disruptions, increasing commodity prices, and strong consumer demand due to Ukraine-Russia. As a result, the BoK raised the interest rates to stabilize the economy and curb the inflation. Additionally, the effects of spillover from the Eurozone conflict is captured, which caused economic growth to slow, prompting dovish monetary policy.}
    \label{fig:BoK-INFLATION}
\end{figure}

\begin{figure}
    \centering
    \includegraphics[width=\linewidth]{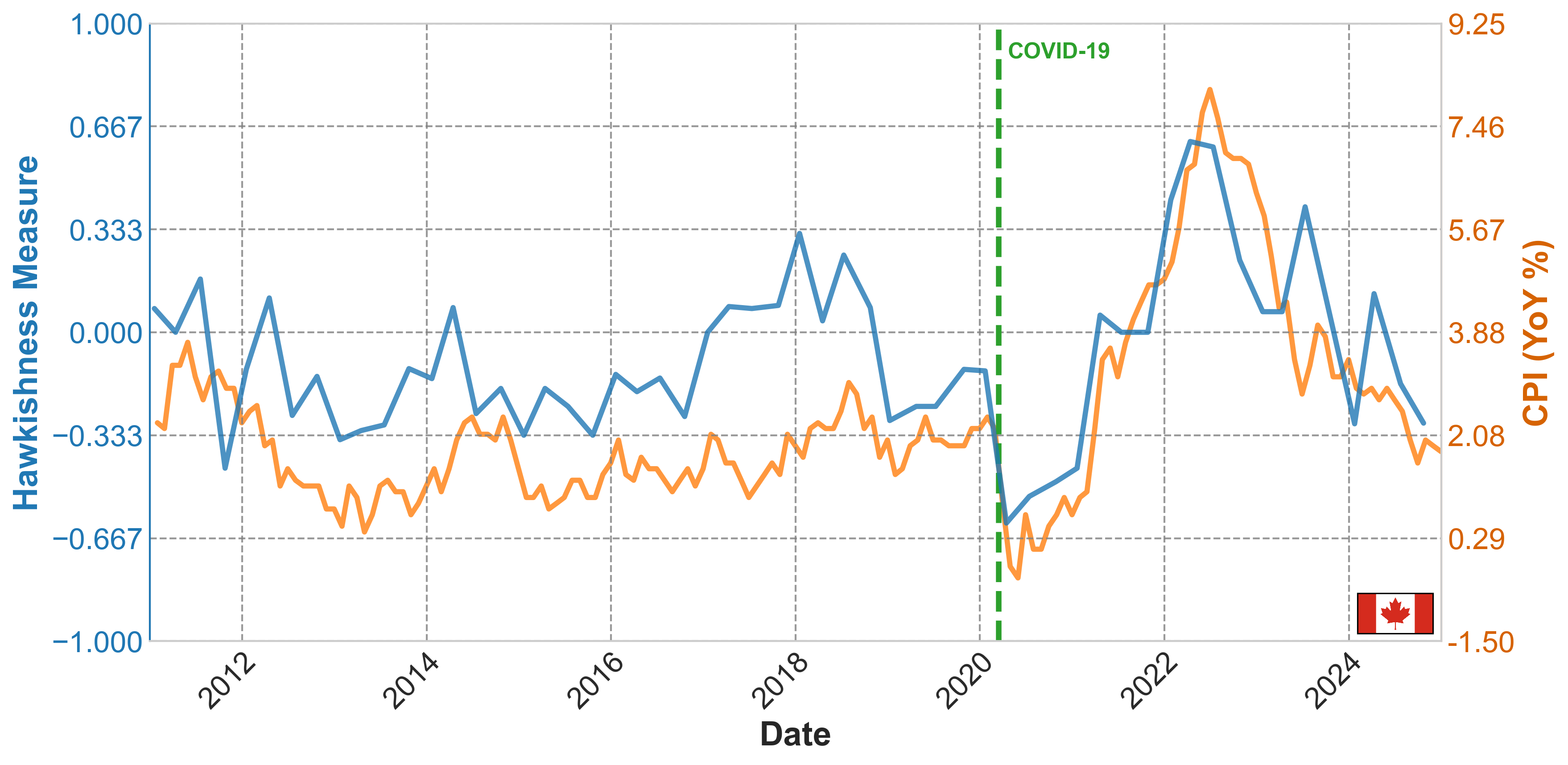}
    \caption{This figure illustrates the effects of the 2020 COVID-19 pandemic on the Canada's inflation index and our hawkishness measure. During this period, BoC implemented dovish monetary policy to stimulate the economy. Additionally, both measures capture the high inflation rate in 2022 caused by supply chain disruptions and high consumer demand, which prompted the BoC to raise rates to stabilize the economy and lower inflation.}
    \label{fig:BoC-INFLATION}
\end{figure}

\begin{figure}
    \centering
    \includegraphics[width=\linewidth]{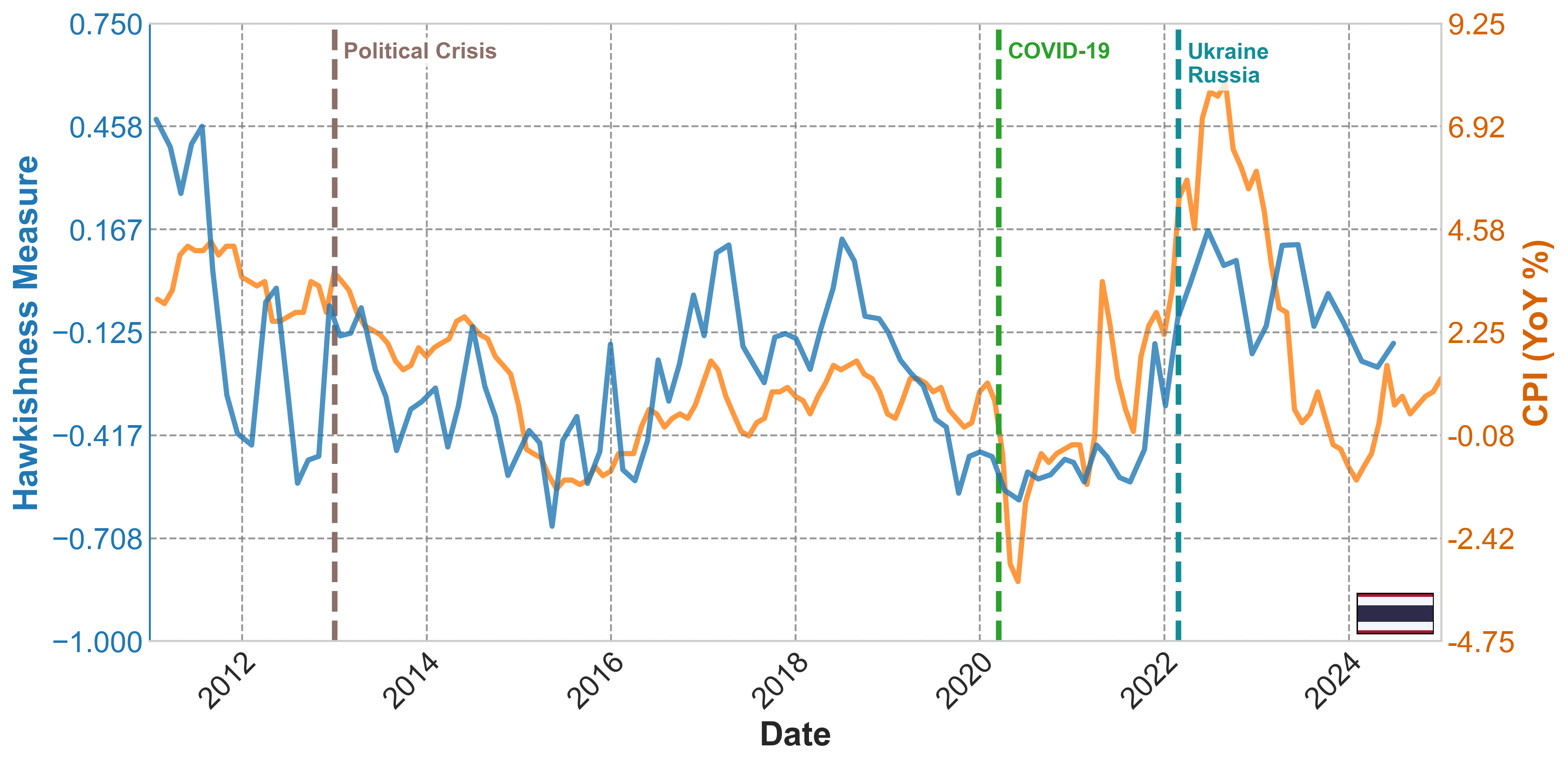}
    \caption{This figure illustrates the effects of the 2013-2014 Thai political crisis and COVID-19 pandemic on Thailand's inflation index and our hawkishness measure. During both periods, BoT adopted dovish monetary policies as the economy contracted. A recovery from the 2014 political unrest is captured by a rise in both the CPI and our hawkishness measure. Additionally, both measures reflect the increase in inflation and subsequent shift to hawkish policy in 2022 due to high energy and commodity prices caused by the events of Ukraine-Russia. }
    \label{fig:BoT-INFLATION}
\end{figure}

\begin{figure}
    \centering
    \includegraphics[width=\linewidth]{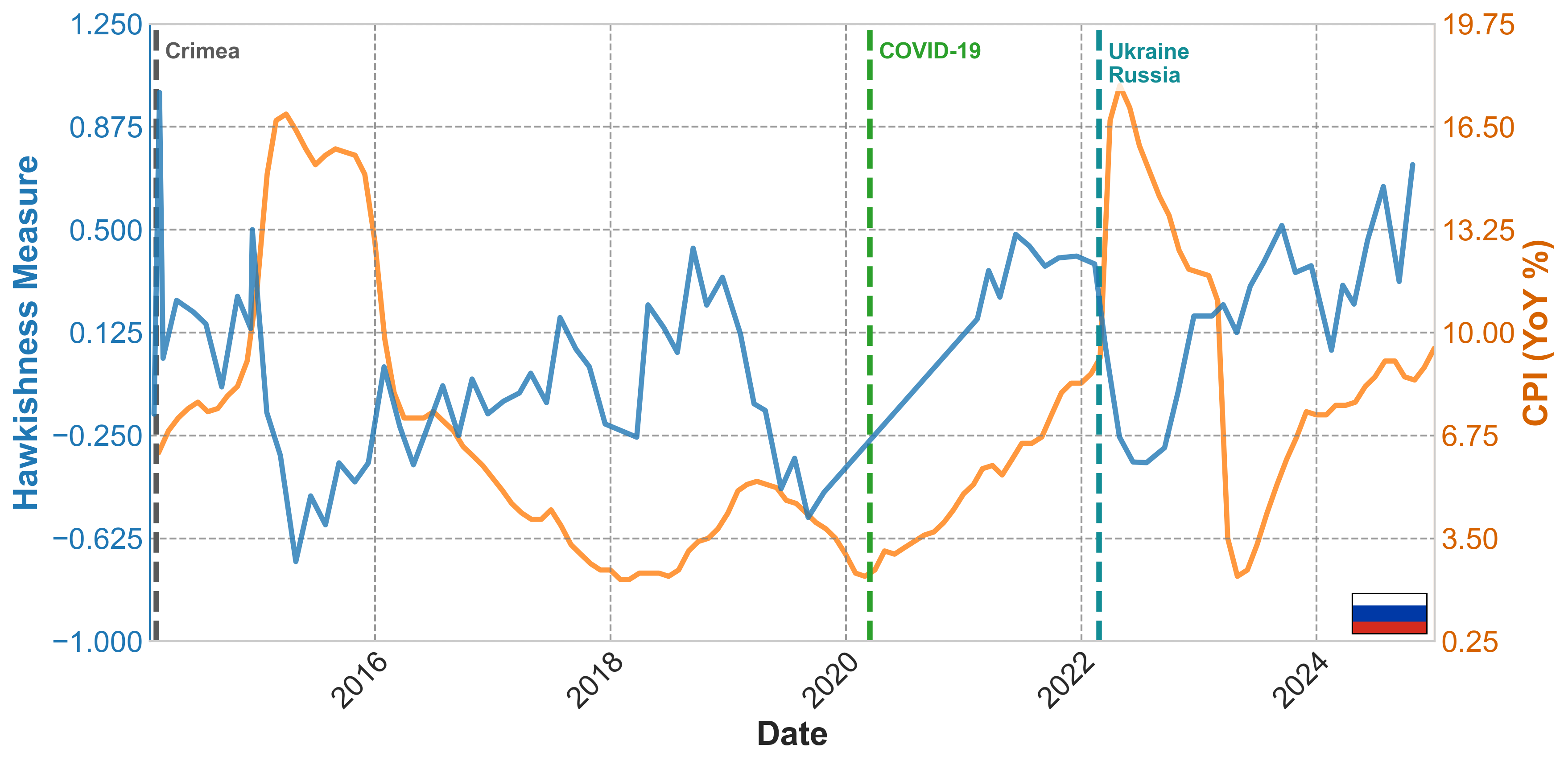}
    \caption{This figure illustrates the effects of major geopolitical and economic events on Russia's inflation index and our hawkishness measure. Elevated hawkishness is observed following the 2014 annexation of Crimea, reflecting CBR's efforts to curb the depreciation of the Russian ruble. The 2020 COVID-19 pandemic prompted dovish monetary policy (lowered interest rates) due to lowered GDP growth and CPI levels to stimulate the economy. The figure also captures the high inflation rate in late 2022 caused by supply chain disruptions due to the pandemic, and rising commodity prices and high consumer demand due to Ukraine-Russia, prompting the central bank to raise rates to stabilize the depreciating ruble.}
    \label{fig:CBR-INFLATION}
\end{figure}

\begin{figure}
    \centering
    \includegraphics[width=\linewidth]{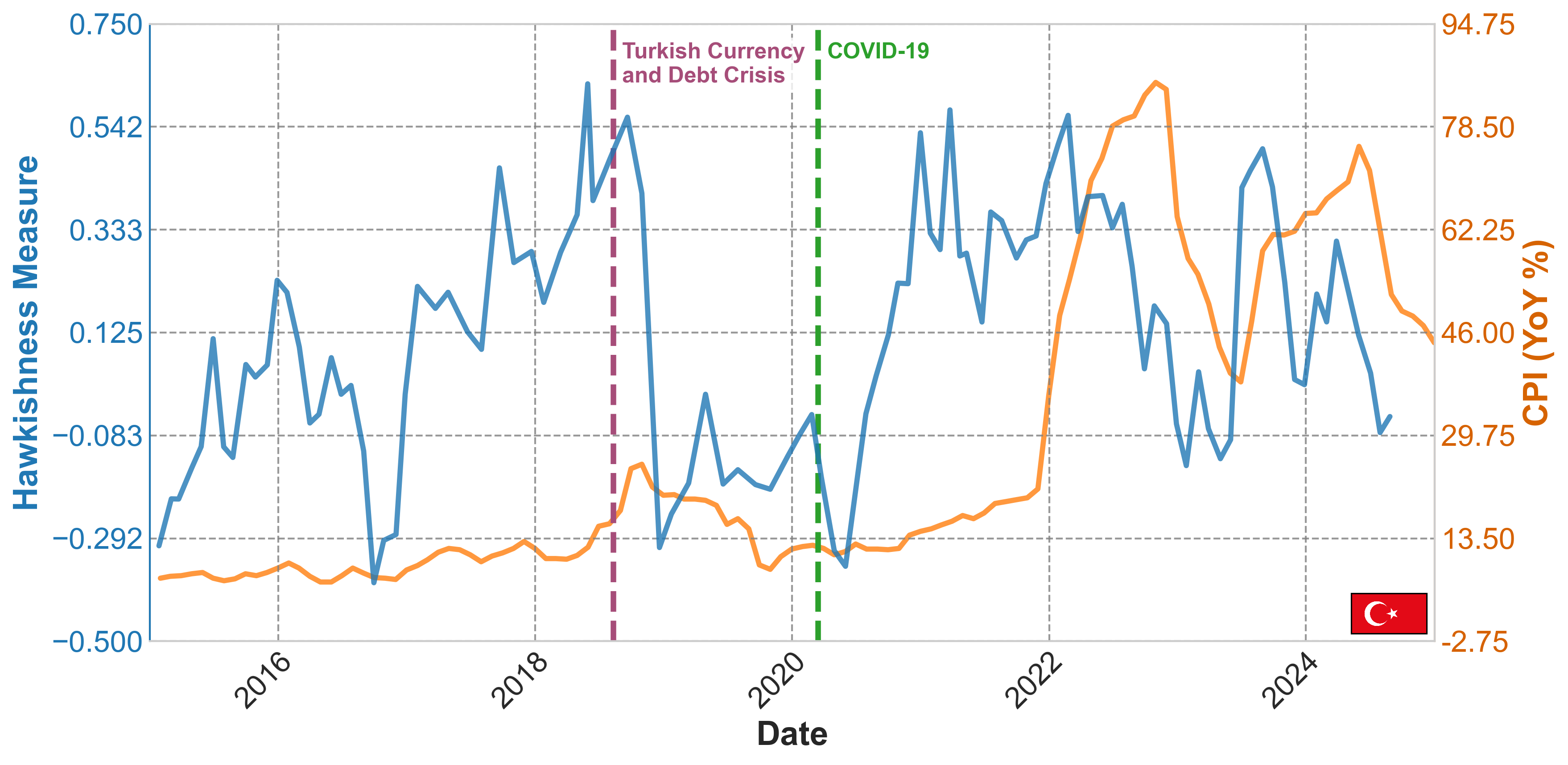}
    \caption{This figure illustrates the effects of the 2018-2019 economic crisis on the Turkey's inflation index and our hawkishness measure. The crisis sparked hawkish monetary policy to combat the Turkish lira depreciation, and then dovish monetary policy to stimulate economic growth to fund the deficit. Additionally, the figure captures the effects of the 2020 COVID-19 pandemic in both measures, and the high inflation rate in late 2022 caused by supply chain disruptions due to the pandemic, resulting in hawkish monetary policy to stabilize the Turkish lira.}
    \label{fig:CBRT-INFLATION}
\end{figure}

\begin{figure}
    \centering
    \includegraphics[width=\linewidth]{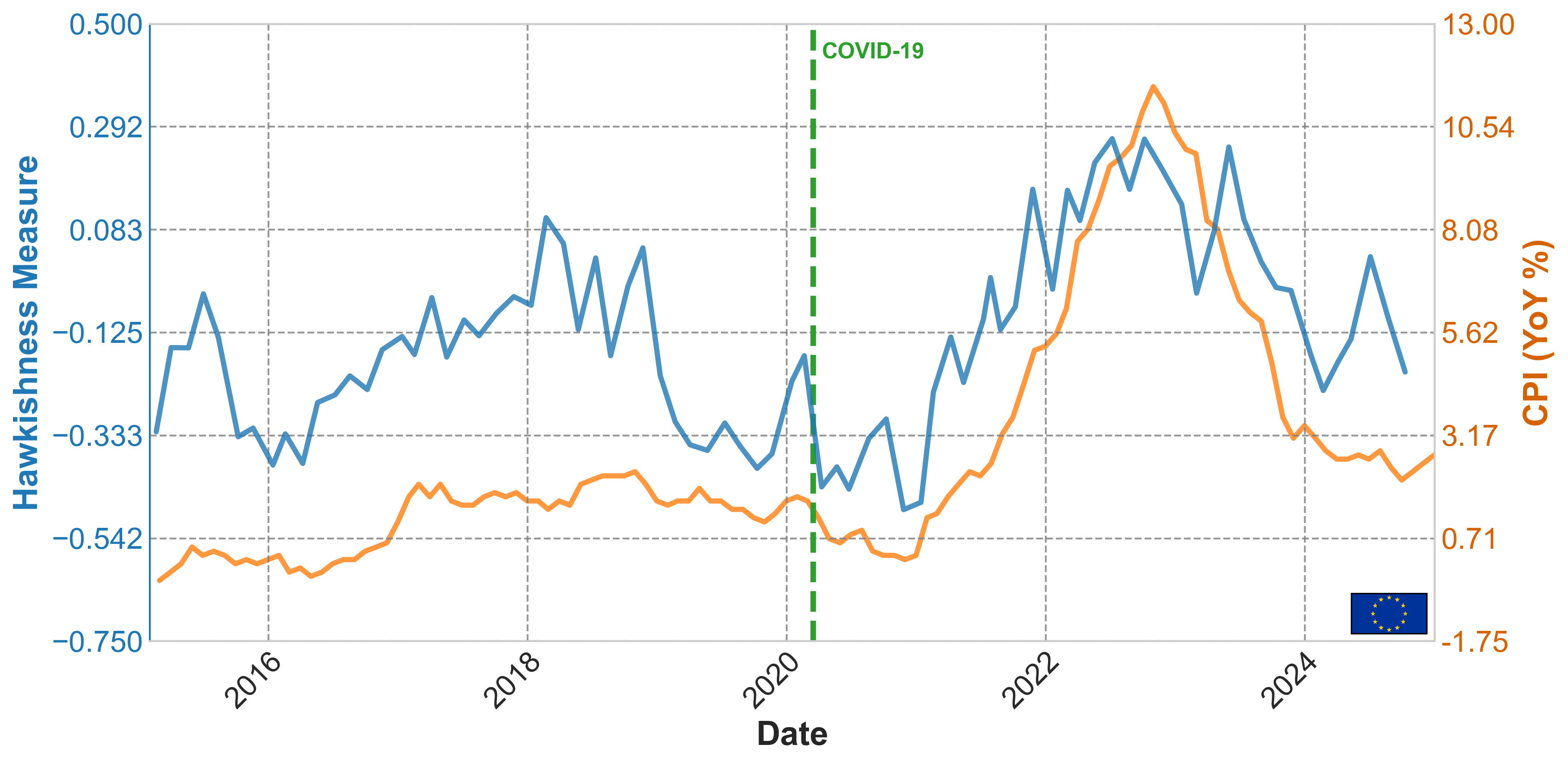}
    \caption{This figure illustrates the effects of the 2020 COVID-19 pandemic on the EU's inflation index and our hawkishness measure. During this period, dovish monetary policy prevailed as the ECB sought to stimulate the economy. Post-pandemic, our hawkishness measure also matches the inflation index when inflation sky-rocketed due to disrupted supply chains and rising demand and the ECB cut rates.}
    \label{fig:ECB-INFLATION}
\end{figure}

\begin{figure}
    \centering
    \includegraphics[width=\linewidth]{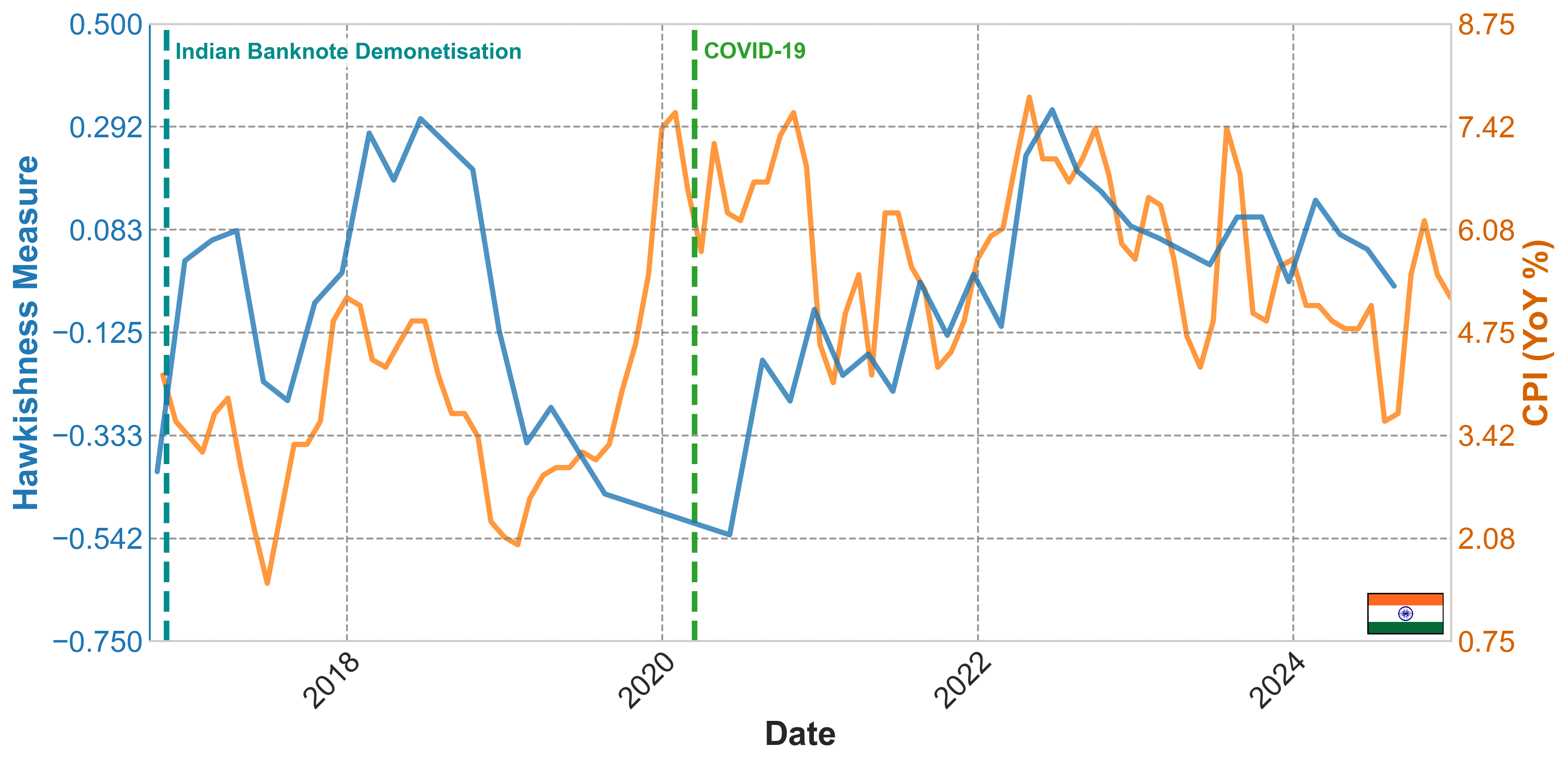}
    \caption{This figure illustrates the effects of the 2020 COVID-19 pandemic on the India's inflation index and our hawkishness measure. The period was marked by volatility, as the monetary policy aimed to support vulnerable households, encourage investment, and maintain price stability. The figure also captures the high hawkishness levels in 2018 reflecting the disinflationary measures implemented to combat liquidity issues following demonetization.}
    \label{fig:RBI-INFLATION}
\end{figure}

\begin{figure}
    \centering
    \includegraphics[width=\linewidth]{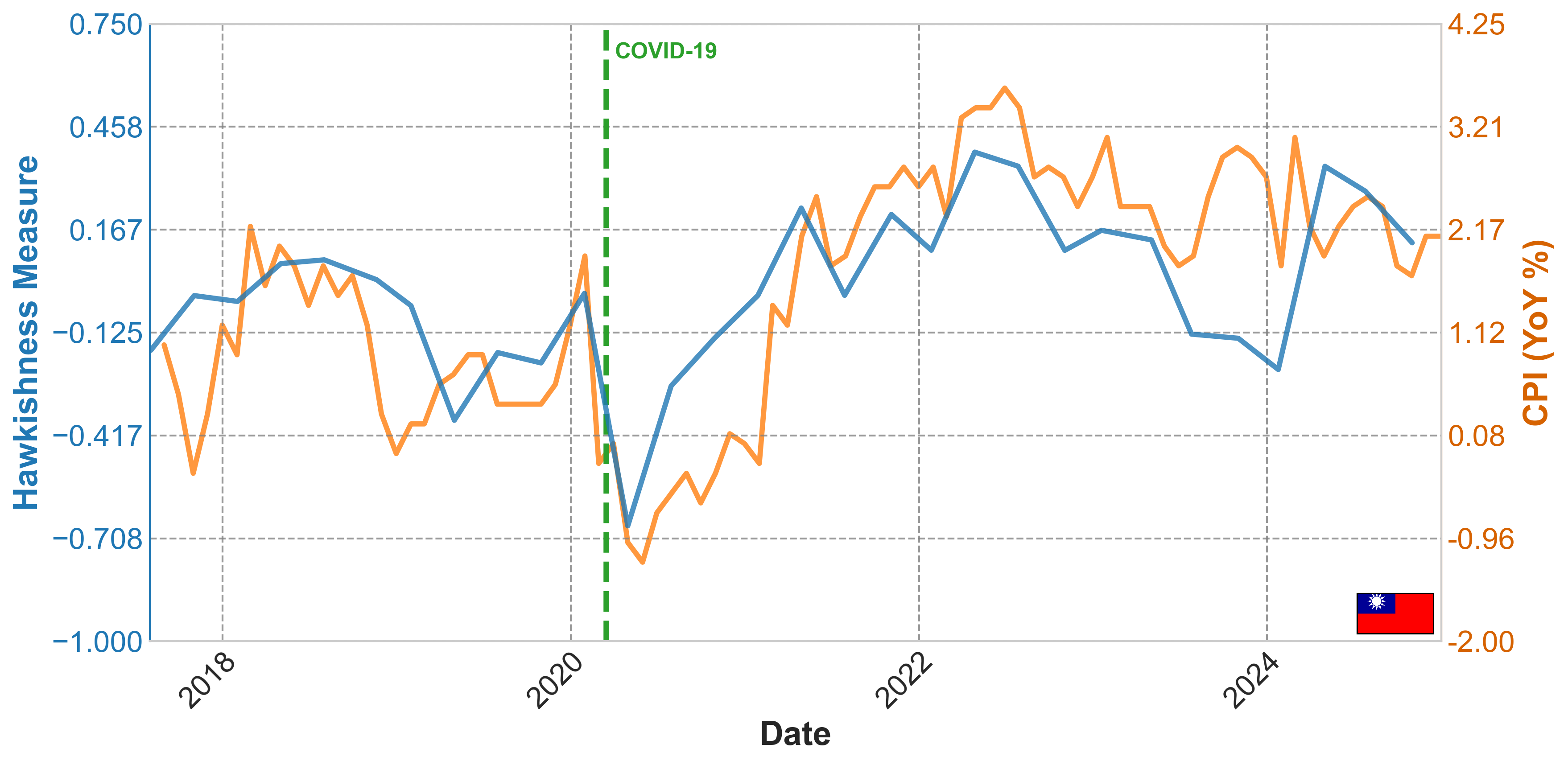}
    \caption{This figure illustrates the effects of the 2020 COVID-19 pandemic on the Taiwan's inflation index and our hawkishness measure. During this period, the central bank implemented dovish monetary policy to stimulate the economy. Additionally, the figure captures the high inflation rate in 2022 caused by supply chain disruptions and high consumer demand, which prompted the central bank to raise rates to stabilize the economy. }
    \label{fig:CBCT-INFLATION}
\end{figure}

\begin{figure}
    \centering
    \includegraphics[width=\linewidth]{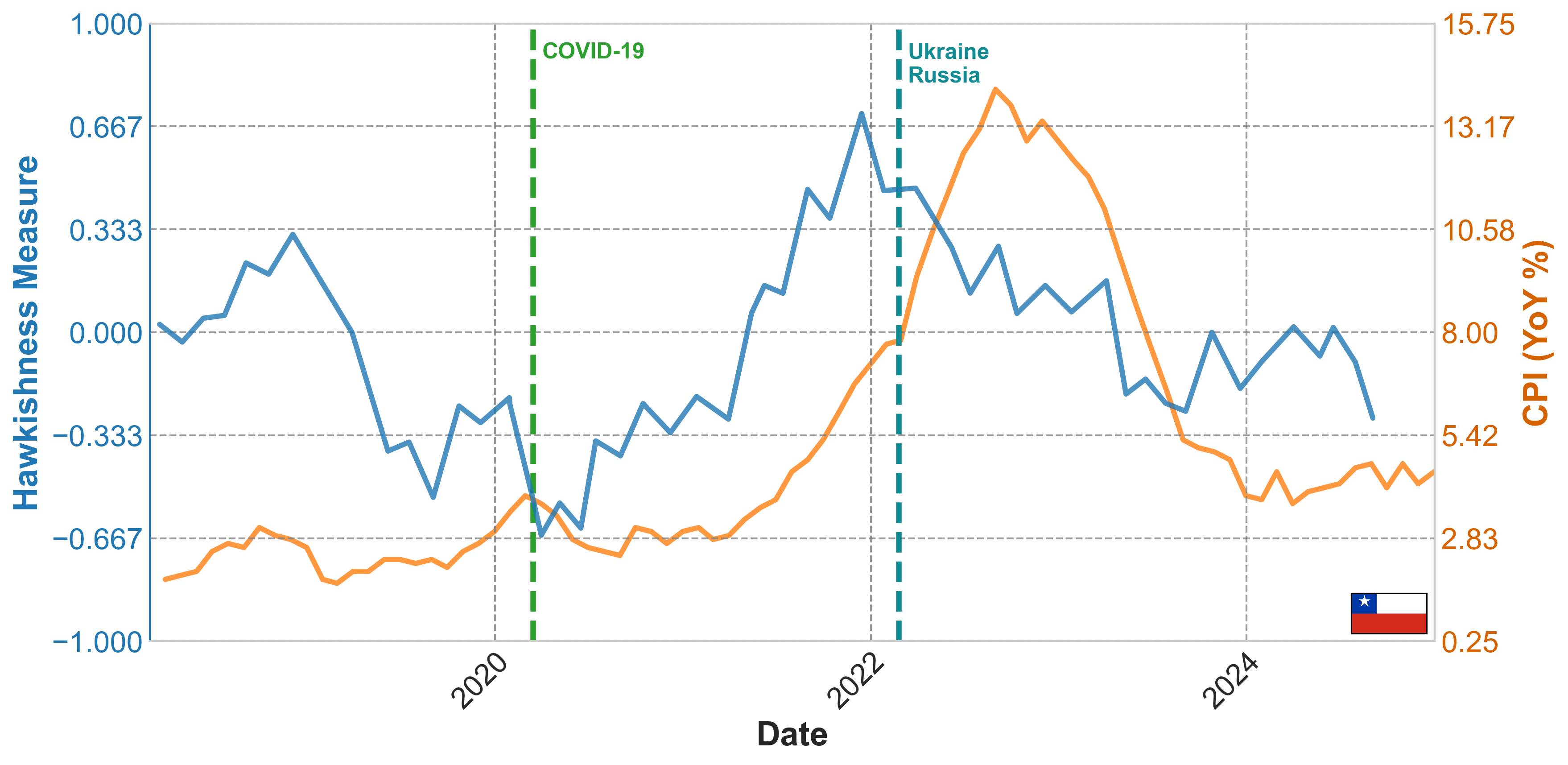}
    \caption{This figure illustrates how both the Chile's inflation index and our hawkishness measure capture the post-pandemic tightening of monetary policy.  With a surge in demand coupled with supply-side bottlenecks, high commodity prices caused by Ukraine-Russia and depreciation of the Peso, the CBoC significantly raised policy rates.}
    \label{fig:CBoC-INFLATION}
\end{figure}

\begin{figure}
    \centering
    \includegraphics[width=\linewidth]{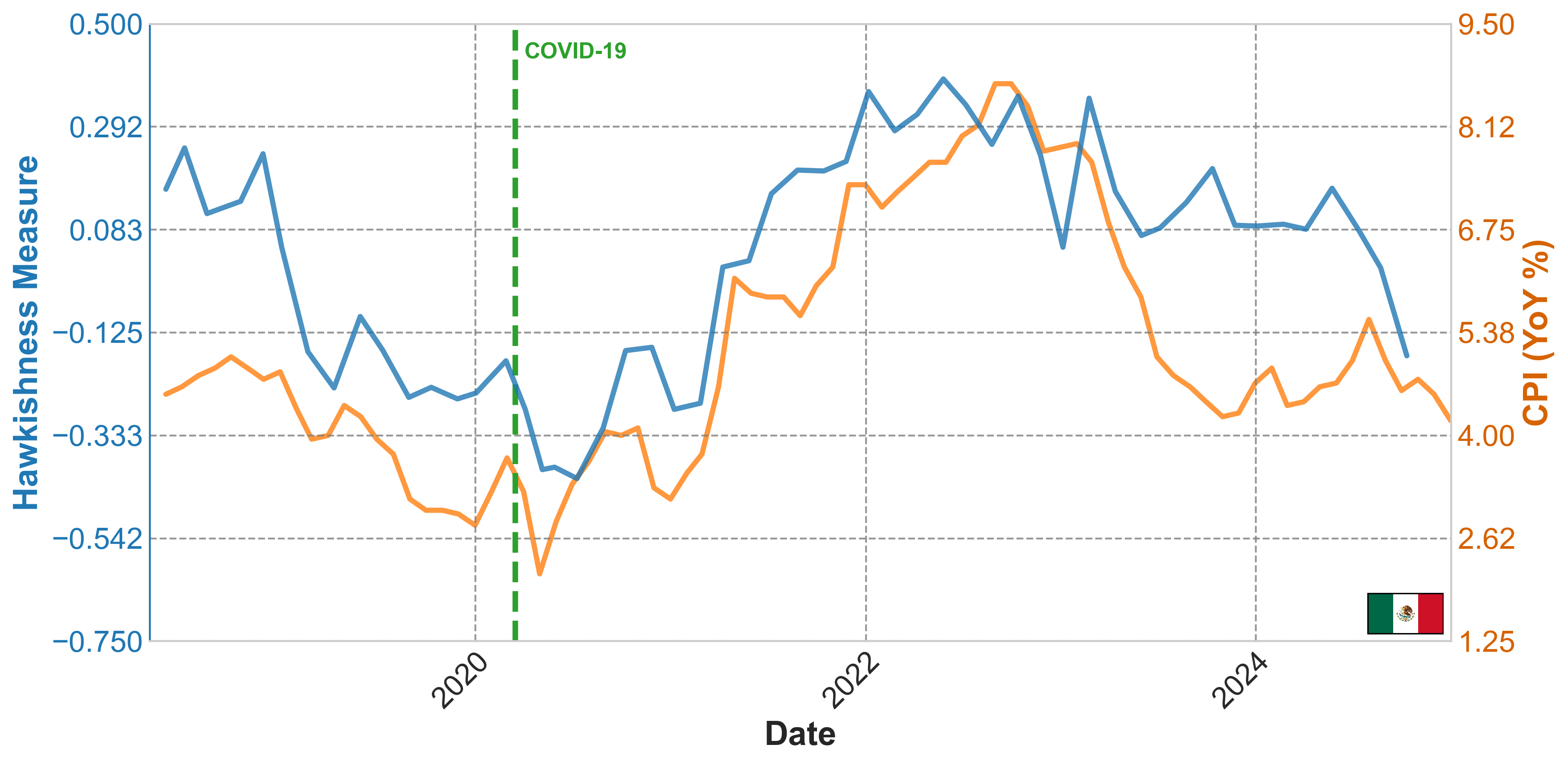}
    \caption{
This figure illustrates the effects of the 2020 COVID-19 pandemic on the Mexico's inflation index and our hawkishness measure. During this period, BoM implemented dovish monetary policy to stimulate the economy. Additionally, the figure captures the high inflation rate in 2022 caused by supply chain disruptions, rising commodity prices, and high consumer demand, which prompted the central bank to raise rates to stabilize the economy and lower inflation.}
    \label{fig:BdeM-INFLATION}
\end{figure}
\clearpage

\subsection{Hindsight 20/20, Foresight Uncertain}
\label{app:Certain time analysis}
\begin{table}[h!]
\caption{Pearson correlation coefficients across all possible pairs of label categories (stance, temporal, certainty) for each bank over all years. Most label pairs exhibit a weak negative correlation. All values within brackets are negative.}
\label{tab:labels correlation}
\begin{tabular}{p{0.13\textwidth} p{0.25\textwidth} p{0.25\textwidth} p{0.25\textwidth}}
\toprule
 & stance-temporal & stance-certainty & temporal-certainty \\
\midrule
\BRAZIL BCB & 0.003541 & 0.015641 & (0.386366) \\
\PERU BCRP & (0.060272) & 0.108143 & (0.414610) \\
\MALAYSIA BNM & (0.063442) & 0.010550 & (0.507737) \\
\PHILIPPINES BSP & 0.002618 & 0.028488 & (0.538671) \\
\COLOMBIA BanRep & 0.019677 & 0.006071 & (0.477708) \\
\CANADA BoC & (0.035004) & 0.023413 & (0.612648) \\
\UK BoE & (0.008751) & 0.062980 & (0.435042) \\
\ISRAEL BoI & (0.087278) & 0.057606 & (0.489864) \\
\JAPAN BoJ & (0.074619) & 0.026308 & (0.388683) \\
\SOUTHKOREA BoK & 0.012551 & 0.010937 & (0.390653) \\
\MEXICO BdeM & 0.083746 & (0.026304) & (0.469378) \\
\THAILAND BoT & (0.069446) & (0.032663) & (0.391205) \\
\TAIWAN CBCT & (0.040279) & 0.063119 & (0.488286) \\
\EGYPT CBE & (0.095191) & 0.077066 & (0.464068) \\
\RUSSIA CBR & (0.029862) & (0.020771) & (0.386689) \\
\TURKEY CBRT & (0.064327) & 0.089533 & (0.478669) \\
\CHILE CBoC & 0.002561 & (0.011903) & (0.305732) \\
\EU ECB & (0.016714) & (0.001914) & (0.351945) \\
\USA FOMC & (0.034457) & 0.056154 & (0.450257) \\
\SINGAPORE MAS & (0.047880) & 0.079492 & (0.595583) \\
\POLAND NBP & 0.005994 & 0.009087 & (0.500110) \\
\CHINA PBoC & (0.283129) & (0.114589) & (0.007756) \\
\AUS RBA & (0.000080) & 0.035070 & (0.465670) \\
\INDIA RBI & (0.017386) & 0.034103 & (0.401618) \\
\SWITZERLAND SNB & 0.006077 & 0.010778 & (0.452553) \\
\midrule \\
\textbf{Average} & (0.035654) & 0.023856 & (0.434060) \\
\bottomrule
\end{tabular}
\end{table}

To analyze the \texttt{Temporal Classification} and \texttt{Uncertainty Estimation} labels, we look at their correlation to see if these two labels are related. As seen in Table \ref{tab:labels correlation}, we observe extremely low values of the Pearson correlation among the stance label with either of the two labels for any of the banks, indicating that \texttt{Stance Detection} is truly independent of both \texttt{Temporal Classification} and \texttt{Uncertainty Estimation}. We notice a weak negative correlation for every single central bank apart from the People's Bank of China. The weak negative correlation implies that there is a high probability that a \texttt{Forward Looking} statement is also \texttt{Uncertain}. Subsequently, \texttt{Certain} statements are likely to be \texttt{Not Forward Looking}.
\label{app:econ_data}

\clearpage
\section{Extended Related Works}
\label{app:extended_related_works}
Research specifically focusing on \texttt{Stance Detection} spans multiple global contexts, including central banks from Israel \citep{KAZINNIK2021308}, Latin America (Colombia, Chile, Peru) \citep{vega2020assessing}, Czech Republic, Hungary, Poland \citep{ROZKRUT2007176}, Japan \citep{oshima2018monetary}, Turkey \citep{iglesias2017emerging}, Mexico \citep{aguilar2022communication}, and India \citep{kumar2024wordsmarketsquantifyingimpact}.While these studies provide valuable local insights, they typically lack comprehensive cross-bank and longitudinal analyses covering significant economic events \citep{ehrmann2020starting, WONG2011432, ARMELIUS2020102116}.

Our dataset uniquely addresses this gap by covering communications from 25 central banks between 1996 to 2024. It provides insights across \texttt{Stance Detection, Uncertainty Estimation}, and \texttt{Temporal Classification} labels, covering critical economic crises. Tables \ref{tab:dataset_comp_table} and \ref{tab:dataset_comp_table_extra} compare our dataset comprehensively with other related datasets, highlighting our dataset's unique breadth and temporal depth. Comparable datasets like \texttt{MONOPOLY}\citep{mathur2022monopoly} and \texttt{EA-MD-QD}\citep{barigozzi2024large}, among others, generally focus on narrower temporal ranges and fewer banks, excluding vital economic crisis periods covered in our analysis. Additionally, our experiments extends beyond benchmarking to human evaluation, Economic Analysis, error analysis, ablation studies on additional banks and on a Congress Committee hearing transcript.

\begin{table}[ht]
\centering
\fontsize{7}{8}\selectfont
\caption{Comparison of existing datasets across year range and geographical coverage and inclusion of four specific additional analytical methods: error analysis, few shot benchmarking, predictive analysis, and systematic data collection. \textit{Error Analysis} is defined as the evaluation of the reasoning of large language models on our specific tasks. \textit{Few Shot Benchmark.} is defined as the inclusion of the task of few shot benchmarking for each of our LLMs and PLMs on our specific tasks. \textit{Predictive Analysis} is defined as the process of using past data to predict future actions, specifically in the context of monetary policy meeting minutes and future policy implementations. \textit{Systematic Data Collection} is defined as following a strict procedure during the collection of monetary policy data in a structured and methodical way.} 
\label{tab:dataset_comp_table_extra}
\begin{tabular}{
    p{0.18\textwidth}  
    p{0.08\textwidth}  
    p{0.13\textwidth}  
    p{0.07\textwidth}  
    p{0.099\textwidth}  
    p{0.08\textwidth}  
    p{0.13\textwidth}  
}
\toprule
\textbf{Paper} & \textbf{Year Range} & \textbf{Geographical Range} & \textbf{Error Analysis} & \textbf{Few Shot Benchmark.} & \textbf{Predictive Analysis} & \textbf{Systematic Data Collection} \\
\midrule
\citet{DVN/23957-2013} & 1995-2012 & 6 emerging economies & $\times$  & $\times$ & \checkmark & $\times$\\
\citet{ARMELIUS2020102116} & 2002-2017 & 23 central banks & $\times$ & $\times$ & \checkmark & $\times$ \\
\citet{mathur2022monopoly} & 2009-2022 & 6 central banks & $\times$ & $\times$ & \checkmark & \checkmark \\
\citet{Kirti2022COVIDPolicies} & 2020 & 74 countries & $\times$ & $\times$ & $\times$ & $\times$ \\
\citet{shah-etal-2023-trillion} & 1996-2022 & United States & \checkmark & $\times$ & \checkmark & \checkmark \\
\citet{Bolhuis2024New} & 2000-2022 & 29 countries & $\times$  & $\times$ & \checkmark & $\times$ \\
\citet{barigozzi2024large} & 2000-$2024^*$ & Euro Area & $\times$  & $\times$ & \checkmark & $\times$\\
\citet{zhang2025camefcausalaugmentedmultimodalityeventdriven} & 2009-2024 & United States & \checkmark & $\times$ & $\times$ & \checkmark \\
\midrule
\textbf{WCB (Ours)} & \textbf{1996-2024} & \textbf{25 central banks} & \textbf{\gcheck} & \textbf{\gcheck} & \textbf{\gcheck} & \textbf{\gcheck} \\
\bottomrule
\end{tabular}
\end{table}

In addition to prior comparison, we summarize below the availability of core reproducibility resources (annotation guides, raw text, and labeled data) across comparable studies in terms of central bank coverage. As shown in Table~\ref{tab:dataset_reproducibility}, existing datasets generally lack publicly available labeled data and annotation documentation, significantly limiting reproducibility and extension, which our WCB dataset addresses.

\begin{table}[ht]
\centering

\caption{Comparison of datasets on reproducibility aspects: availability of annotation guidelines, raw text data, and labeled data.}
\label{tab:dataset_reproducibility}
\begin{tabular}{p{0.23\textwidth} p{0.1\textwidth} p{0.17\textwidth} p{0.17\textwidth} p{0.17\textwidth}}
\toprule
\textbf{Paper} & \textbf{Number of CBs} & \textbf{Annotation Guide} & \textbf{Publicly Available Text Data} & \textbf{Publicly Available Labeled Data} \\
\midrule
Armelious et al. (2020) & 23 & No & No & No \\
Kirti (2022) & 74 & No & Yes & No \\
Bolhuis et al. (2024) & 20 & No & No & No \\
\midrule
\textbf{WCB (Ours)} & \textbf{25} & \textbf{Yes} & \textbf{Yes} & \textbf{Yes} \\
\bottomrule
\end{tabular}
\end{table}
 
\clearpage
\section{Glossary}
\label{app:glossary}

The definitions of common financial terms used in this paper are as follows:

\begin{itemize}

    \item \textbf{Central Bank:} A central bank is a financial institution responsible for the formulation of monetary policy and the regulation of member banks. It typically has privileged control over the production and distribution of money and credit for a nation or a group of nations. \citep{segal2023centralbank}

    \item \textbf{Consumer Price Index (CPI):} CPI is calculated as a weighted average of prices for selected items, with weights reflecting their share in total consumer spending. It is a statistical measure that tracks the average change over time in the prices paid by households for a representative basket of goods and services, and is widely used as a measure for inflation. \citep{fernando2025cpi, hayes2024purchasingpower}

    \item \textbf{Dovish Policy:} A type of policy that favors lower interest rates and quantitative easing to encourage economic growth. \citep{shah-etal-2023-trillion}

    \item \textbf{Harmonized Index of Consumer Prices (HICP):} The Harmonized Index of Consumer Prices (HICP) provides the official measure of consumer price index in the euro area and the EU. \citep{eurostat2024prices}

    \item \textbf{Hawkish Policy:} A type of policy that favors higher interest rates and quantitative tightening to keep inflation in check. \citep{shah-etal-2023-trillion}

    \item \textbf{Inflation:} Inflation is a gradual loss of purchasing power that is reflected in a broad rise in prices for goods and services over time. \citep{fernando2024inflation}

    \item \textbf{Interest Rate:} Interest rate is the amount charged on top of the principal by a lender to a borrower for use of assets, expressed as a percentage or ratio. Central banks use the interest rate as a monetary policy tool. Increasing the cost of borrowing among commercial banks can influence many other interest rates (like mortgage rates, personal loan rates), making it difficult to borrow money. This leads to lower demand for money and can cool the hot economy and vice versa. \citep{banton2025interestrate}

    \item \textbf{Monetary Policy:} Monetary policy is a set of tools used by a nation's central bank to control the overall money supply and promote economic growth, employing strategies such as revising interest rates and changing bank reserve requirements. \citep{segal2023centralbank}

    \item \textbf{Neutral:} In this paper’s context, it refers to a neutral monetary policy that does not favor either hawkishness or dovishness. \citep{shah-etal-2023-trillion}

    \item \textbf{Not Seasonally Adjusted (NSA):} NSA means that the economic data (e.g., GDP) have not been adjusted to remove the effects of predictable seasonal fluctuations. \citep{majaski2022seasonaladjustment}

    \item \textbf{Purchasing Power:} Purchasing power refers to the amount of goods and services one unit of currency can buy. \citep{hayes2024purchasingpower}

\end{itemize}

\clearpage
\section{Central Banks Information, Annotator Disagreement, and Annotation Guides}
\label{app:annotation_guides}
\usubsection{ Federal Open Market Committee}
\begin{center}`
    \textbf{Region: United States of America}
\end{center}
\begin{center}
    \fbox{\includegraphics[width=0.99\textwidth]{resources/flags/Flag_of_the_United_States.png}} 
\end{center}

\begin{center}
    \textbf{Data Collected: 1996-2024}
\end{center}
\vfill
\begin{center}
  \fbox{%
    \parbox{\textwidth}{%
      \begin{center}
      \textbf{Important Links}\\
      \href{https://www.federalreserve.gov/default.htm}{Central Bank Website}\\
       \href{https://huggingface.co/datasets/gtfintechlab/fomc_communication}{Annotated Dataset}\\
     \href{https://huggingface.co/gtfintechlab/model_federal_reserve_system_stance_label}{Stance Label Model} \\
     \href{https://huggingface.co/gtfintechlab/model_federal_reserve_system_time_label}{Time Label Model} \\
     \href{https://huggingface.co/gtfintechlab/model_federal_reserve_system_certain_label}{Certain Label Model} \\
      \end{center}
    }
  }
\end{center}

\newpage

\section*{Monetary Policy Mandate} 
The FOMC is responsible for formulating U.S. monetary policy to achieve its dual mandate: promoting maximum employment and ensuring price stability. 

\textbf{Mandate Objectives:} 
\begin{itemize}
    \item \textbf{Maximum Employment}: Striving for the highest \textit{sustainable} level of employment, recognizing that various factors influence employment levels. 
    \item \textbf{Price Stability}: Aiming for a 2\% annual inflation rate, as measured by the Personal Consumption Expenditures (PCE) Price Index, to maintain purchasing power and economic stability. 
\end{itemize}

\section*{Structure} 
FOMC is integral to the Federal Reserve System, overseeing U.S. monetary policy. 

\textbf{Composition: }

\begin{itemize}
    \item \textbf{Board of Governors:} Seven members appointed by the President and confirmed by the Senate, serving staggered 14-year terms. 
    \item \textbf{Federal Reserve Bank Presidents:} 
    \begin{itemize}
        \item Permanent Member: President of the Federal Reserve Bank of New York. 
        \item Rotating Members: Four of the remaining eleven Reserve Bank presidents, serving one-year terms on a rotating basis. 
    \end{itemize}
\end{itemize}

\textbf{Meeting Structure: } 
\begin{itemize}
    \item \textbf{Frequency:} Eight scheduled meetings annually, approximately every six weeks. 
    \item \textbf{Additional Meetings:} Held as needed to address urgent economic developments. 
\end{itemize}

\section*{Manual Annotation} 

\textbf{Annotators} 
\begin{itemize}
    \item Liam Dolphin
    \item Tiberius Colina
    \item Pranav Aluru
    \item Ahaan Limaye
\end{itemize} 

\textbf{Annotation Agreement} 

The agreement percentage among the pairs of annotators for different labels.
\begin{itemize}
    \item \SD Agreement: 47.8\%
    \item \TC Agreement: 82.5\%
    \item \CE Agreement: 75.1\%
\end{itemize} 

\textbf{Annotation Guide} 

\mptext{fomc}{eight} Economic Status, Dollar Value Change, Energy/House Prices, Foreign Nations, Fed Expectations/Actions/Assets, Money Supply, Key Words/Phrases, corporate bond, and Labor. 

\begin{itemize}
    \item \emph{Economic Status}: A sentence pertaining to the state of the economy, relating to unemployment and inflation.
    \item \emph{Dollar Value Change}: A sentence pertaining to changes such as appreciation or depreciation of value of the United States Dollar on the Foreign Exchange Market.
    \item \emph{Energy/House Prices}: A sentence pertaining to changes in prices of real estate, energy commodities, or energy sector as a whole.
    \item \emph{Foreign Nations}: A sentence pertaining to trade relations between the United States and a foreign economy. If not discussing the United States we label neutral.
    \item \emph{Fed Expectations/Actions/Assets}: A sentence that discusses changes in the Fed yields, bond value, reserves, or any other financial asset value.
    \item \emph{Money Supply}: A sentence that overtly discusses impact to the money supply or changes in demand.
    \item \emph{Key Words/Phrases}: A sentence that contains key word or phrase that would classify it squarely into one of the three label classes, based upon its frequent usage and meaning among particular label classes.
    \item \emph{Corporate Bond}: A sentence that relates to corporate bond issuance.
    \item \emph{Labor}: A sentence that relates to changes in labor productivity.
\end{itemize}

\textbf{Examples: }
\begin{itemize}
    \item ``The Committee then turned to a discussion of the economic and financial outlook, the ranges for the growth of money and debt in 1996, and the implementation of monetary policy over the intermeeting period ahead.''\\
    \textbf{Neutral}: Economic status is only mentioned without diving deeper into increases or decreases in inflation and/or unemployment rate.
    
    \item ``To support the Committee’s decision to raise the target range for the federal funds rate, the Board of Governors voted unanimously to raise the interest rates on required and excess reserve balances to 2.''\\
    \textbf{Hawkish}: The Fed expects higher inflation so they are likely to increase interest rates.
    
    \item ``Labor productivity has continued to rise over recent months, supporting stronger economic growth and reducing inflationary pressures.''\\
    \textbf{Dovish}: The increase in labor productivity can ease inflation concerns without requiring tighter monetary policy.
    
    \item ``The U.S. trade deficit has widened significantly due to increased imports from foreign nations, contributing to upward pressure on inflation.''\\
    \textbf{Hawkish}: A growing trade deficit can fuel inflation, prompting the Fed to consider tightening monetary policy to control price stability.
    
    \item ``Real GDP was anticipated to increase at a rate noticeably below its potential in 2008.''\\
    \textbf{Dovish}: Expected GDP below potential indicates that economy is slowing down, requiring expansionary monetary policy to prevent any drawbacks.

    \item ``It was agreed that the next meeting of the committee would be held on Tuesday-Wednesday, April 26-27, 2016.''\\
    \textbf{Irrelevant}: This sentence solely provides scheduling information and does not offer any insights into a monetary policy stance.
\end{itemize}

\newpage

\begin{longtable}
{p{0.118\textwidth}p{0.183\textwidth}p{0.183\textwidth}p{0.183\textwidth}p{0.183\textwidth}}
\caption{\mptitle{The Federal Open Market Committee}} \\
\toprule
\textbf{Category} & \textbf{Hawkish} & \textbf{Dovish} & \textbf{Neutral} & \textbf{Irrelevant} \\
\midrule
\endfirsthead

\toprule
\textbf{Category} & \textbf{Hawkish} & \textbf{Dovish} & \textbf{Neutral} & \textbf{Irrelevant} \\
\midrule
\endhead

\textbf{Economic Status} & When inflation increases, when unemployment decreases, when economic growth is projected high, when economic output is higher than potential supply/actual output, when economic slack falls. & When inflation decreases, when unemployment increases, when economic growth is projected as low. & When unemployment rate or growth is unchanged, maintained, or sustained. & Sentence is not relevant to monetary policy. \\
\midrule
\textbf{Dollar Value Change} & When the dollar depreciates. & When the dollar appreciates. & N/A & Sentence is not relevant to monetary policy. \\
\midrule
\textbf{Energy/House Prices} & When oil/energy prices increase, when house prices increase. & When oil/energy prices decrease, when house prices decrease. & N/A & Sentence is not relevant to monetary policy. \\
\midrule
\textbf{Foreign Nations} & When the US trade deficit increases. & When the US trade deficit decreases. & When relating to a foreign nation's economic or trade policy. & Sentence is not relevant to monetary policy. \\
\midrule
\textbf{Fed Expectations, Actions, and Assets} & Fed expects high inflation, widening spreads of treasury bonds, increase in treasury security yields, increase in TIPS value, increase bank reserves. & Fed expects subpar inflation, Fed expecting disinflation, narrowing spreads of treasury bonds, decreases in treasury security yields, and reduction of bank reserves. & N/A & Sentence is not relevant to monetary policy. \\
\midrule
\textbf{Money Supply} & Money supply is high, increased demand for goods, low demand for loans. & Money supply is low, M2 increases, increased demand for loans. & N/A & Sentence is not relevant to monetary policy. \\
\midrule
\textbf{Key Words and Phrases} & Indicating a focus on “price stability” and “sustained growth.” & When the stance is ``accommodative,'' indicating a focus on “maximum employment” and “price stability.” & Use of phrases “mixed,” “moderate,” “reaffirmed.” & Sentence is not relevant to monetary policy. \\
\midrule
\textbf{Corporate Bond} & When issuance increases. & When issuance decreases. & N/A & Sentence is not relevant to monetary policy. \\
\midrule
\textbf{Labor} & When productivity or unemployment decreases. & When productivity or unemployment increases. & N/A & Sentence is not relevant to monetary policy. \\
\bottomrule
\label{tb:fomc_mp_stance_guide}
\end{longtable}

\fwcertaintytext{fomc}
\newpage

\begin{table*}
\caption{\fwtitle{The Federal Open Market Committee}}
\vspace{1em}
\begin{tabular}{p{0.3\textwidth}p{0.3\textwidth}p{0.3\textwidth}}
\toprule
\textbf{Label} & \textbf{Description} & \textbf{Example}\\
\midrule
\textbf{Forward Looking} & When it discusses expectations, projections, or anticipations about future economic conditions, policy actions, or resource utilization. & “On the other hand, given the high level of capacity utilization and movements in the exchange rate, there is a danger that higher production costs will increasingly be passed on to prices.” \\
\midrule
\textbf{Not Forward Looking} & When it reflects on recent or past economic data, trends, or events to describe what has already occurred. & “By contrast, exports of goods and services stagnated.” \\
\bottomrule
\end{tabular}
\label{tb:fomc_forward_looking_guide}
\end{table*}

\begin{table*}
\caption{\certaintytitle{The Federal Open Market Committee}}
\vspace{1em}
\begin{tabular}{p{0.3\textwidth}p{0.3\textwidth}p{0.3\textwidth}}
\toprule
\textbf{Label} & \textbf{Description} & \textbf{Example}\\
\midrule
\textbf{Certain} & When an expectation, trend, action, or outcome is definitively stated without ambiguity. Key words include ``remain,'' ``increased,'' ``decreased,'' and ``will continue.''
 & “This is because the utilization of resources is somewhat higher.” \\
\midrule
\textbf{Uncertain} & When it expresses concerns, potential risks, or
need to alter monetary policy if a specific event comes to pass. Key words include ``outlook,'' ``if,'' ``could,'' and ``uncertainty.'' & “These higher prices for imports could impact upon consumer prices.” \\
\bottomrule
\end{tabular}
\label{tb:fomc_certainty_guide}
\end{table*}

\clearpage
\usubsection{People's Bank of China}

\begin{center}
    \textbf{Region: China}
\end{center}
\begin{center}
    \fbox{\includegraphics[width=0.99\textwidth]{resources/flags/Flag_of_China.png}}
\end{center}
\begin{center}
    \textbf{Data Collected: 2003-2024}
\end{center}
\vfill
\begin{center}
  \fbox{%
    \parbox{\textwidth}{%
      \begin{center}
      \textbf{Important Links}\\
      \href{http://www.pbc.gov.cn/en/3688006/index.html}{Central Bank Website}\\
       \href{https://huggingface.co/datasets/gtfintechlab/peoples_bank_of_china}{Annotated Dataset}\\
     \href{https://huggingface.co/gtfintechlab/model_peoples_bank_of_china_stance_label}{Stance Label Model} \\
     \href{https://huggingface.co/gtfintechlab/model_peoples_bank_of_china_time_label}{Time Label Model} \\
     \href{https://huggingface.co/gtfintechlab/model_peoples_bank_of_china_certain_label}{Certain Label Model} \\
      \end{center}
    }
  }
\end{center}

\newpage

\section*{Monetary Policy Mandate} 
The People's Bank of Chines (PBoC) is responsible for determining and implementing the monetary policy of China upon the guidance and advise of the Monetary Policy Committee (MPC). The PBoC also maintains the circulation of the Renminbi and monitors monetary money-laundering suspicions, and also “promot[es] the building up of [the] credit information system”.

\textbf{Mandate Objectives:} 
\begin{itemize}
    \item \textbf{Currency Stability: }Maintain stability of the value of the Renminbi to promote economic growth.
\end{itemize}

\section*{Structure} 
\textbf{Composition: } 
\begin{itemize}
    \item \textbf{Governor}
    \begin{itemize}
        \item Nominated by the Premier of the State Council and decided by the National People's Congress. 
        \item When the National People's Congress is not in session, the Governor is decided by the Standing Committee of the National People's Congress and appointed by the President of the People's Republic of China.
    \end{itemize}
    \item \textbf{Deputy Governors: }There are a number of deputy governors. As of March 2025, there are five. 
\end{itemize}
\textbf{Meeting Structure: } It is unclear when the PBoC meets. However, the MCB, the consultancy body, meets as follows:
\begin{itemize}
\item \textbf{Frequency: }Meetings are on a quarterly basis
\item \textbf{Additional Meetings: }Held if one-third of the members agree or if proposed by the Chairman
\end{itemize}

\section*{Manual Annotation} 

\textbf{Annotators} 

\begin{itemize}
    \item Henry Zhang
    \item Sidharth Subbarao
    \item Harold Huang
    \item Rachel Yuh
\end{itemize}

\textbf{Annotation Agreement} 
The agreement percentage among the pairs of annotators for different labels.
\begin{itemize}
    \item \SD Agreement: 63.2\%
    \item \TC Agreement: 90.7\%
    \item \CE Agreement: 91.5\%
\end{itemize} 
\textbf{Annotation Guide}

\mptext{pboc}{seven}Inflation, Value of the Yuan, Interest Rates, Money Supply, GDP Growth Rate, and Trade Output.
\begin{itemize}
    \item \textit{Inflation: }A sentence about the inflation rate in the economy
    \item \textit{Value of the Yuan: }A sentence about how the value of the Yuan is changing relative to the USD 
    \item \textit{Interest Rates: }A sentence about how interest rates, including the repo rate and loan prime rates are changing. 
    \item \textit{Money Supply: }A sentence pertaining the M2 money supply. PCB may conduct short-term liquidity operations and reverse repo operations to influence liquidity into the market.
    \item \textit{GDP Growth Rate: }A sentence pertaining to changes in GDP or changes in the commodities market or key industry that may impact the GDP
    \item \textit{Trade Output: }A sentence pertaining to China's export or import levels or how macroeconomic factors   may impact foreign trade 
\end{itemize}

\textbf{Examples:}
\begin{itemize}
    \item ``In line with the rules on monetary policy committee of the people's bank of china and with the approval of the state council, presidents of two wholly state-owned commercial banks, Liu Mingkang, Zhang Enzhao and Wu Jinglian will withdraw from the monetary policy committee as their term has expired.''
    \newline\textbf{Irrelevant:} Sentence is not directly related to monetary policy. 
    \item ``Improvement of the managed floating exchange rate regime should be effected, continuously following the principle of “making it a self-initiated, controllable and gradual process.''''
    \newline\textbf{Dovish: } The PBoC is confident in the health of the market is not concerned about the money supply becoming too large hence their desire to focus on floating the exchange rate.
    \item ``It was noted at the meeting that efforts should made to implement the prudent monetary policy, make the financial measures more targeted, flexible and effective, and give more priority to stabilizing the general price level in 2011.''
    \newline\textbf{Hawkish: }The PBoC feels that monetary policy needs to be implemented in order to stabilize prices, which is dovish. 
    \item ``The committee discussed monetary policy and measures to be adopted in the period ahead and concurred that the sound monetary policy should be preserved, interest rates of deposits and loans be maintained stable to ensure steady advancement of market-based interest rate reform.''
    \newline\textbf{Neutral: }The PBoC is emphasizing maintaining the currency policy.

\end{itemize}

\newpage

\begin{longtable}{p{0.118\textwidth}p{0.183\textwidth}p{0.183\textwidth}p{0.183\textwidth}p{0.183\textwidth}}
\caption{\mptitle{People's Bank of China}} \label{tb:pboc_mp_stance_guide}\\
\toprule
\textbf{Category} & \textbf{Hawkish} & \textbf{Dovish} & \textbf{Neutral} & \textbf{Irrelevant} \\
\midrule
\endfirsthead

\toprule
\textbf{Category} & \textbf{Hawkish} & \textbf{Dovish} & \textbf{Neutral} & \textbf{Irrelevant} \\
\midrule
\endhead
\textbf{Inflation} & When inflation rates rise inflationary pressures. & When inflation rates fall. & When inflation rate is stable. & Sentence is not relevant to monetary policy. \\
\midrule
\textbf{Value of the Yuan} & When Yuan depreciates. & When Yuan appreciates. & When Yuan remains stable. & Sentence is not relevant to monetary policy. \\
\midrule
\textbf{Interest Rates} & When the repo rate increases or loan prime rates (LPR) are increased. & When repo rate decreases or when short-term rates are lowered. & When interest rates are maintained at the same level. & Sentence is not relevant to monetary policy. \\
\midrule
\textbf{Money Supply} & When M2 money supply is decreasing. & When PBC restricts money supply. & When money supply remains the same. & Sentence is not relevant to monetary policy. \\
\midrule
\textbf{GDP Growth Rate} & When GDP growth declines. & When GDP growth increases or real economic growth is promoted. & When GDP growth remains constant. & Sentence is not relevant to monetary policy. \\
\midrule
\textbf{Trade Output} & When policy aims to decrease foreign trade. & When there are bottlenecks key industries namely oil, coal, and electricity, or sluggish global economy, or other hindrances to foreign trade. & When there is a focus on maintaining stability in trade. & Sentence is not relevant to monetary policy. \\
\bottomrule
\end{longtable}

\fwcertaintytext{pboc}
\newpage

\begin{table*}
\caption{\fwtitle{People's Bank of China}}
\vspace{1em}
\begin{tabular}{p{0.3\textwidth}p{0.3\textwidth}p{0.3\textwidth}}
\toprule
\textbf{Label} & \textbf{Description} & \textbf{Example}\\
\midrule
\textbf{Forward Looking} & When it discusses future outcomes, key words include “in the future,” “expected,” and “projected.” & ``The credit information system and capital raising system for some should be established.''\\
\midrule
\textbf{Not Forward Looking} & When it references past events, historical data, previous trends, or observed outcomes, and may include key words such as “was,” “in the past,” and “has happened.” & ``Liquidity was adequate in the banking system, growth of money and credit supply rapid and the financial system operated in a stable manner.''\\
\bottomrule
\end{tabular}
\label{tb:pboc_forward_looking_guide}
\end{table*}

\begin{table*}
\caption{\certaintytitle{People's Bank of China}}
\vspace{1em}
\begin{tabular}{p{0.3\textwidth}p{0.3\textwidth}p{0.3\textwidth}}
\toprule
\textbf{Label} & \textbf{Description} & \textbf{Example}\\
\midrule
\textbf{Certain} & When it include indicators based on reliable data and trends, expresses outcomes definitively, includes phrases such as “will,” definitely,” and “certainly.” & ``A mix of new monetary policy tools will be used to keep liquidity adequate at a reasonable level.'' \\
\midrule
\textbf{Uncertain} & When it mentions metrics with high volatility such as global commodity prices or includes qualifying clauses that include hedging words such as ``maybe,'' ``kind of,'' ``sort of,'' ``might,'' or ``possibility.'' & ``While financial support to economic growth must be guaranteed, no negligence should be tolerated in preventing inflation and financial risks.'' \\
\bottomrule
\end{tabular}
\label{tb:pboc_certainty_guide}
\end{table*}

\clearpage
\usubsection{ Bank of Japan}
\begin{center}
    \textbf{Region: Japan}
\end{center}
\begin{center}
    \fbox{\includegraphics[width=0.99\textwidth]{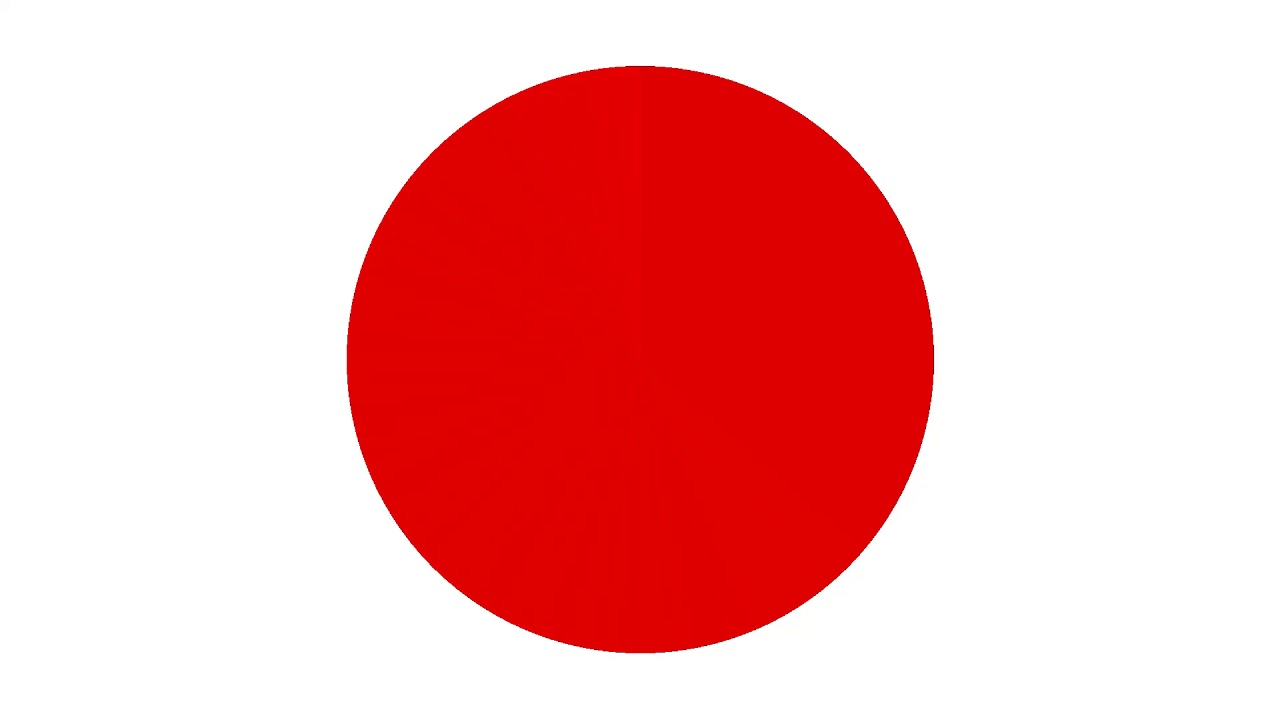}} 
\end{center}
\begin{center}
    \textbf{Data Collected: 1998-2024}
\end{center}
\vfill
\begin{center}
  \fbox{%
    \parbox{\textwidth}{%
      \begin{center}
      \textbf{Important Links}\\
      \href{https://www.boj.or.jp/en/}{Central Bank Website}\\
       \href{https://huggingface.co/datasets/gtfintechlab/bank_of_japan}{Annotated Dataset}\\
     \href{https://huggingface.co/gtfintechlab/model_bank_of_japan_stance_label}{Stance Label Model} \\
     \href{https://huggingface.co/gtfintechlab/model_bank_of_japan_time_label}{Time Label Model} \\
     \href{https://huggingface.co/gtfintechlab/model_bank_of_japan_certain_label}{Certain Label Model} \\
      \end{center}
    }
  }
\end{center}

\newpage

\section*{Monetary Policy Mandate} 
The Bank of Japan implements monetary policy to maintain price stability, which underpins economic activity. It influences interest rates through money market operations to control currency and liquidity. The Policy Board sets the policy direction at Monetary Policy Meetings (MPMs), where economic and financial conditions are reviewed. Decisions are announced immediately, guiding daily market operations to provide or absorb funds as needed.

\textbf{Mandate Objectives:}  
\begin{itemize}  
    \item \textbf{Issuance of Banknotes and Monetary Policy Implementation:} The Bank of Japan is responsible for issuing banknotes and conducting currency and monetary policy operations to ensure effective monetary control.  
    \item \textbf{Financial Stability:} The Bank strives to facilitate the smooth settlement of funds among banks and other financial institutions, thereby contributing to the overall stability of the financial system.  
    \item \textbf{Economic Development:} Through its currency and monetary policy measures, the Bank aims to achieve price stability, fostering the sound and sustainable development of the national economy. The current inflation target is set at 2\%.
\end{itemize}
\section*{Structure} 
The bank's structure consists of the policy board and the bank's officers (Governor, Deputy Governors, Auditors, Executive Directors, and Counsellors.).

\textbf{Composition: }

\begin{itemize}
    \item \textbf{Policy Board:} The Policy Board is established as the Bank's highest decision-making body. The Board determines the guideline for currency and monetary control, sets the basic principles for carrying out the Bank's operations, and oversees the fulfillment of the duties of the Bank's officers, excluding Auditors and Counselors.
    \item \textbf{Bank's Officers:} The number of posts given to the Bank's officers is as follows: the Governor, two Deputy Governors, six Members of the Policy Board, at most three Auditors, at most six Executive Directors, and the Counselors.
\end{itemize}
\textbf{Meeting Structure: } 
 \begin{itemize}
     \item \textbf{Frequency:} Eight scheduled meetings annually, approximately every six weeks. 
     \item \textbf{Additional Meetings:} Held as needed to address urgent economic developments. 
 \end{itemize}
\section*{Manual Annotation} 

\textbf{Annotators} 

\begin{itemize}
    \item Randall Bevan
    \item Alex Chen
    \item Abinav Chari
    \item Aksh Tyagi
\end{itemize}
\textbf{Annotation Agreement} 
The agreement percentage among the pairs of annotators for different labels.
\begin{itemize}
    \item \SD Agreement: 37\%
    \item \TC Agreement: 62.6\%
    \item \CE Agreement: 71.3\%
\end{itemize} 

\textbf{Annotation Guide} 
\mptext{boj}{eleven} Money Supply, Economic Status, Yen Value Change, Energy and Commodity Prices, Foreign Trade, Monetary Policy Stance, Inflation Expectations, Government Bonds and JGB Yields, BoJ Asset Purchases, Labor Market Conditions, and Fiscal Policy Interactions.

\begin{itemize}
    \item \emph{Money Supply}: A sentence that discusses changes in the central bank's control over the money supply of Japan. This includes adjustments in interest rates, quantitative easing policies, and shifts in inflation tolerance, which for Japan, is currently at 2\%. Japan's current quantitative easing policy involves buying \$40 billion worth of Japanese government bonds monthly.
    \item \emph{Economic Status}: A sentence that describes the overall condition of the economy, highlighting trends in inflation, consumer spending, and economic growth. 
    \item \emph{Yen Value Change}: A sentence pertaining to fluctuations in the value of the Yen on the foreign exchange market.
    \item \emph{Energy and Commodity Prices}: A sentence addressing movements in energy and commodity markets or addressing policy that is expected to cause change in energy and commodity markets. For example, lower commodity prices in 2010s allowed the Bank of Japan to keep policies loose without inflation fears.
    \item \emph{Foreign Trade}: A sentence that focuses on foreign trade (e.g., imports and exports) and its impacts on Japan's economy. For example, if Japan has a trade deficit, money would be, on net, leaving the Japanese economy. 
    \item \emph{Monetary Policy Stance (Interest Rates and actions)}: A sentence that outlines the central bank's policy direction through interest rate decisions and asset purchase programs, signaling either a dovish or hawkish stance.
    \item \emph{Inflation Expectations}: A sentence mentions inflation expectations in Japan. Japan's current inflation target is at 2\%.
    \item \emph{Government Bonds and JGB Yields}: A sentence discussing the behavior of government bond yields, including new adjustments in government bond purchases or the yield curve.
    \item \emph{BoJ Asset Purchases (ETFs, REITs, etc.)}: A sentence related to the Bank of Japan's strategy in buying financial assets such as ETFs.
    \item \emph{Labor Market Conditions}: A sentence focused on the labor market, including indicators such as unemployment rates and wage growth.
    \item \emph{Fiscal Policy Interactions}: A sentence that examines the relationships between monetary and fiscal policies, highlighting how changes in, for example,  government spending or taxation affect monetary policy.
\end{itemize}

\textbf{Examples: }
\begin{itemize}
    \item "Some members commented that the growth in demand for some it-related goods was slowing around the world."\\
    \textbf{Dovish}: The subsection in the sentence about slowing demand implies that current economic pressures are subdued, increasing the need for expansionary monetary policy.
    
    \item "With regard to the first factor, one member said that inflation expectations for fiscal 2012 and 2013 had been relatively low at around 0 percent, although medium- to long-term inflation expectations remained stable at around 1.0 percent."\\
    \textbf{Dovish}: The emphasis on very low near-term inflation expectations would allow for expansionary monetary policy without inflation fears.
    
    \item "Some members said that private consumption and fixed asset investment continued to show high growth."\\
    \textbf{Hawkish}: The noted high growth in private consumption and fixed asset investment indicate an overheating economy, which may eventually generate inflationary pressures. This would ultimately require contractionary monetary policy to offset any future inflationary pressures that may arise.
    
    \item "Stock prices in emerging economies had also dropped significantly, due mainly to the rise in inflation."\\
    \textbf{Hawkish}: The sentence indicates that inflationary pressures are affecting asset markets, requiring a tightening stance to counteract the inflation.
    
    \item "The funding conditions of European financial institutions had recently been even more stable."\\
    \textbf{Neutral}: The statement objectively reports that funding conditions are stable without implying any need for either easing or tightening monetary policy.
    
    \item "The representative from the cabinet office made the following remarks."\\
    \textbf{Irrelevant}: This sentence is purely introductory and does not provide any substantive economic or policy-related information for analysis.
\end{itemize}

\newpage

\begin{longtable}{p{0.118\textwidth}p{0.183\textwidth}p{0.183\textwidth}p{0.183\textwidth}p{0.183\textwidth}}
\caption{\mptitle{Bank of Japan}} \\
\toprule
\textbf{Category} & \textbf{Hawkish} & \textbf{Dovish} & \textbf{Neutral} & \textbf{Irrelevant} \\
\midrule
\endfirsthead

\toprule
\textbf{Category} & \textbf{Hawkish} & \textbf{Dovish} & \textbf{Neutral} & \textbf{Irrelevant} \\
\midrule
\endhead

\textbf{Money Supply} & Lower interest rates, quantitative easing via bank purchases, higher inflation tolerance. & Higher interest rates on money, cease quantitative easing (QE), lower inflation tolerance. A 0.25\% interest rate is the highest in Japan currently. & If the interest rates are staying the same. & Sentence is not relevant to monetary policy. \\
\midrule

\textbf{Economic Status} & Liquid economy, increased money flow between businesses and consumers, increasing CPI. & More rigid economy that is shrinking from a state of inflation, less exchange of money between businesses and consumers, decreasing CPI. & If the CPI stays the same or is not affected. & Sentence is not relevant to monetary policy. \\
\midrule

\textbf{Yen Value Change} & When the Yen appreciates (Ex. Allowing the yen to appreciate to reduce import costs). & When the Yen depreciates (Ex. Allowing the yen to weaken to boost exports). & Maintaining a stable Yen exchange rate. & Sentence is not relevant to monetary policy. \\
\midrule

\textbf{Energy and Commodity Prices} & When energy and commodity prices decrease, thus allowing the BoJ to continue their easing measures. & When energy and commodity prices increase. & Stable energy and commodity prices lead the BoJ to make no significant changes. & Sentence is not relevant to monetary policy. \\
\midrule

\textbf{Foreign Trade} & An increase in trade and exports. & A decrease in trade or exports to reduce inflation. & Foreign trade and exports are maintained. & Sentence is not relevant to monetary policy. \\
\midrule

\textbf{Monetary Policy Stance (Interest Rates and actions)} & Expanding asset purchases or lowering interest rates (Ex. Lowering interest rates and purchasing assets to support the economy, such actions took place during COVID-19). & Reducing asset purchases or signaling future interest rate increases (Ex. BoJ reducing its target for JGB purchases or signaling a future interest rate increase to curb inflation). & Keeping monetary policy unchanged, no signs of making interest rate changes or modifying practices of purchasing assets. & Sentence is not relevant to monetary policy. \\
\midrule

\textbf{Inflation Expectations} & High or increasing inflation. & Low or decreasing inflation (target 2\%). & If inflation is unchanged. & Sentence is not relevant to monetary policy. \\
\midrule

\textbf{Government Bonds and JGB Yields} & When BoJ increases bond purchases or targets a 10-year JGB yield near 0\% and indicates readiness to lower the yield further. & When BoJ allows a rise in 10-year JGB yields above 1\% by reducing bond purchases or adjusting yield curve control target further. & 10-year JGB yields remain within the target range (currently 0\% ± 1\%) with no significant change. & Sentence is not relevant to monetary policy. \\
\midrule

\textbf{BoJ Asset Purchases (ETFs, REITs, etc.)} & When BoJ increases asset purchases, injecting liquidity into the market. & When BoJ reduces asset purchases. & No change in the BoJ asset purchase strategy. & Sentence is not relevant to monetary policy. \\
\midrule

\textbf{Labor Market Conditions} & When unemployment rate is high or is increasing or wage growth is low or decreasing, indicating a need for accommodative policy. & When unemployment rate is low or is decreasing, or wage inflation is high or is increasing. & When unemployment rate remains between 2.5\% - 3.5\% with annual wage growth around 1.5\% - 2.5\%. & Sentence is not relevant to monetary policy. \\
\midrule

\textbf{Fiscal Policy Interactions} & Correlated with expansionary fiscal policies (Ex. BoJ gave government stimulus during crisis situations such as 2008 economic crisis and COVID-19 pandemic). & Correlated with contractionary fiscal policies (tax increases, reduced spending). & No connected changes with fiscal policy. & Sentence is not relevant to monetary policy. \\
\bottomrule
\label{tb:boj_mp_stance_guide}
\end{longtable}

\fwcertaintytext{boj}
\newpage

\begin{table*}
\caption{\fwtitle{Bank of Japan}}
\vspace{1em}
\begin{tabular}{p{0.3\textwidth}p{0.3\textwidth}p{0.3\textwidth}}
\toprule
\textbf{Label} & \textbf{Description} & \textbf{Example}\\
\midrule
\textbf{Forward Looking} & When the text mainly uses future tense. Key words include ``will,'' and ``is expected to.'' & ``The Japanese economy was recovering at a moderate pace and was expected to continue to do so.'' \\
\midrule
\textbf{Not Forward Looking} &  When the text is using past or present tense. Key words include ``has,'' and ``did.'' &  ``This member continued, however, that these changes did not reflect an economic downturn.'' \\
\bottomrule
\end{tabular}
\label{tb:boj_forward_looking_guide}
\end{table*}

\begin{table*}
\caption{\certaintytitle{Bank of Japan}}
\vspace{1em}
\begin{tabular}{p{0.3\textwidth}p{0.3\textwidth}p{0.3\textwidth}}
\toprule
\textbf{Label} & \textbf{Description} & \textbf{Example}\\
\midrule
\textbf{Certain} & When the text uses firm language. Key words include ``will,'' ``increased,'' ``decreased.''
 & ``The Japanese economy was on the way to recovery at a moderate pace.'' \\
\midrule
\textbf{Uncertain} & When the text uses cautious language. Key words include ``may'' and ``could.''  & ``Exports were expected to continue to rise against the background of the expansion of overseas economies as a whole.''\\
\bottomrule
\end{tabular}
\label{tb:boj_certainty_guide}
\end{table*}

\clearpage
\usubsection{ Bank of England}
\begin{center}
    \textbf{Region: United Kingdom}
\end{center}
\begin{center}
    \fbox{\includegraphics[width=0.99\textwidth]{resources/flags/Flag_of_the_United_Kingdom.png}} 
\end{center}
\begin{center}
    \textbf{Data Collected: 1999-2024}
\end{center}
\vfill
\begin{center}
  \fbox{%
    \parbox{\textwidth}{%
      \begin{center}
      \textbf{Important Links}\\
      \href{https://www.bankofengland.co.uk}{Central Bank Website}\\
       \href{https://huggingface.co/datasets/gtfintechlab/bank_of_england}{Annotated Dataset}\\
     \href{https://huggingface.co/gtfintechlab/model_bank_of_england_stance_label}{Stance Label Model} \\
     \href{https://huggingface.co/gtfintechlab/model_bank_of_england_time_label}{Time Label Model} \\
     \href{https://huggingface.co/gtfintechlab/model_bank_of_england_certain_label}{Certain Label Model} \\
      \end{center}
    }
  }
\end{center}

\newpage

\section*{Monetary Policy Mandate} 
The Bank of England sets monetary policy to maintain low and stable inflation in the UK.

\textbf{Mandate Objectives:}  
\begin{itemize}
    \item \textbf{Stable Inflation:} The primary objective of monetary policy is to ensure low and stable inflation. This is achieved by targeting a 2\% inflation rate over the medium term, as mandated by the Government.
    \item \textbf{Sustainable Economic Growth:} In addition to inflation control, the Bank supports the Government’s broader economic objective of fostering strong, sustainable, and balanced economic growth.
\end{itemize}

\section*{Structure} 
Monetary policy decisions are made by the Monetary Policy Committee (MPC), which consists of nine members. Each member serves a fixed term, after which they may be reappointed or replaced.
\textbf{Composition: }
\begin{itemize}
    \item \textbf{Monetary Policy Committee (MPC):} The MPC comprises nine members: the Governor, three Deputy Governors (responsible for Monetary Policy, Financial Stability, and Markets \& Banking), the Chief Economist, and four external members appointed by the Chancellor.
    \item \textbf{External Members:} These members are appointed to provide independent perspectives and bring expertise from outside the Bank of England.
    \item \textbf{HM Treasury Representative:} A representative of HM Treasury attends MPC meetings, but does not have voting rights. Their role is to provide insights on fiscal policy and other government economic initiatives while ensuring the Chancellor remains informed about monetary policy decisions.
\end{itemize}
\textbf{Meeting Structure: } 
 The MPC sets and announces policy eight times a year (roughly once every six weeks). In the week leading up to each announcement, the committee meets several times.
\begin{itemize}
    \item \textbf{Pre-MPC Meeting:} Members receive a briefing from staff on the latest economic data and trends. This includes insights on business conditions across the UK from regional agents.
    
    \item \textbf{First Meeting:} Members review and discuss the most recent economic indicators. This meeting typically takes place in the week leading up to the policy announcement.
    
    \item \textbf{Second Meeting:} Members deliberate on potential monetary policy actions. This discussion occurs a few days before the official announcement.
    
    \item \textbf{Final Meeting:} The Governor proposes a policy stance expected to gain majority support, followed by a vote among MPC members. This meeting usually takes place on Wednesday, one day before the decision is announced. Each member has a vote, and the policy with the most votes is selected. In the case of a tie, the Governor casts the deciding vote. Members in the minority express their preferred policy stance.
    
    \item \textbf{MPC Announcement:} The final decision, along with meeting minutes, is published at 12 noon on the day of the announcement, typically a Thursday. Rate announcements are disseminated both on the Bank’s website and directly to market participants.
\end{itemize}
\section*{Manual Annotation} 

\textbf{Annotators} 
\begin{itemize}
    \item Austin Barton
    \item Srimaaye Jegannathan
    \item Andrew Li
    \item Aidan Wu
\end{itemize}

\textbf{Annotation Agreement} 
The agreement percentage among the pairs of annotators for different labels.
\begin{itemize}
    \item \SD Agreement: 59.3\%
    \item \TC Agreement: 77.4\%
    \item \CE Agreement: 68.5\%
\end{itemize} 

\textbf{Annotation Guide}

\mptext{boe}{eight} Economic Status, Pound Value Change, Energy/House Prices, Foreign Nations, BoE Expectations, Actions, Assets, Money Supply, Key Words, Phrases, and Labor.

\begin{itemize}
    \item \emph{Economic Status}: A sentence pertaining to the overall condition of the economy, including indicators such as inflation, unemployment, and growth projections.
    \item \emph{Pound Value Change}: A sentence pertaining to fluctuations in the value of the British Pound on the foreign exchange market, indicating whether it is appreciating or depreciating.
    \item \emph{Energy/House Prices}: A sentence pertaining to variations in the prices of energy commodities and residential properties.
    \item \emph{Foreign Nations}: A sentence pertaining to the economic impacts of trade relations with foreign countries, particularly those affecting the United Kingdom.
    \item \emph{BoE Expectations, Actions, Assets}: A sentence that discusses the Bank of England’s projections, policy actions, and changes in financial assets such as treasury bonds and bank reserves.
    \item \emph{Money Supply}: A sentence that examines changes in the overall availability of money in the economy, including shifts in loan demand and monetary liquidity.
    \item \emph{Key Words, Phrases}: A sentence that includes specific keywords or phrases which clearly indicate a particular monetary policy stance.
    \item \emph{Labor}: A sentence that relates to shifts in labor market conditions, encompassing changes in productivity, wage trends, and job openings.
\end{itemize}

\textbf{Examples: }
\begin{itemize}
    \item ``Recent economic indicators showing rising inflation and falling unemployment have led the Bank of England to signal a tighter monetary policy stance.''\\
    \textbf{Hawkish}: The BOE is concerned that overheating growth will fuel further inflation, prompting interest rate hikes.
    
    \item ``The acceleration of house prices and energy costs is causing upward pressures on consumer demand, which may necessitate a proactive tightening of monetary policy.''\\
    \textbf{Hawkish}: Rising asset prices and energy costs are seen as signals of mounting inflationary pressures.
    
    \item ``Falling inflation paired with a rise in unemployment has convinced policymakers at the Bank of England that a more accommodative stance is needed to support the economy.''\\
    \textbf{Dovish}: Lower inflation and softer labor market conditions suggest that further easing could help sustain economic growth.
    
    \item ``Low money supply growth combined with persistently high loan demand has reinforced the BOE's outlook for maintaining an expansionary monetary policy stance.''\\
    \textbf{Dovish}: These factors indicate subdued economic activity, reducing the likelihood of immediate monetary policy tightening.

    \item ``Total output prices had been flat year on year.''\\
    \textbf{Neutral}: The sentence states that total output prices remained stable and doesn't take a direction on monetary policy.
    
    \item ``The new government had announced its intention to present an updated budget on 8 July.''\\
    \textbf{Irrelevant}: The announcement pertains to a general government schedule and does not contain a monetary policy stance.
\end{itemize}

\newpage
\begin{longtable}{p{0.118\textwidth}p{0.183\textwidth}p{0.183\textwidth}p{0.183\textwidth}p{0.183\textwidth}}
\caption{\mptitle{Bank of England}}     
\label{tb:boe_mp_stance_guide}
\\
\toprule
\textbf{Category} & \textbf{Dovish} & \textbf{Hawkish} & \textbf{Neutral} & \textbf{Irrelevant} \\
\midrule
\endfirsthead

\toprule
\textbf{Category} & \textbf{Dovish} & \textbf{Hawkish} & \textbf{Neutral} & \textbf{Irrelevant} \\
\midrule
\endhead

\textbf{Economic Status} & Inflation decreases, unemployment increases, low projected growth. & Inflation increases, unemployment decreases, high projected growth, demand outgrows supply. & When unemployment rate or growth is unchanged, maintained, or sustained. & Sentence is not relevant to monetary policy. \\
\midrule
\textbf{Pound Value Change} & Pound appreciates. & Pound depreciates. & N/A. & Sentence is not relevant to monetary policy. \\
\midrule
\textbf{Energy and House Prices} & Oil/energy prices decrease, house prices decrease. & Oil/energy prices increase, house prices increase. & N/A. & Sentence is not relevant to monetary policy. \\
\midrule
\textbf{Foreign Nations} & UK trade deficit decreases. & UK trade deficit increases. & Relating to foreign economic policy. & Sentence is not relevant to monetary policy. \\
\midrule
\textbf{BoE Expectations, Actions, Assets} & Expects subpar inflation, disinflation, narrowing spreads of treasury bonds, decreases in treasury security yields, and reduction of bank reserves. & Expects high inflation, widening spreads of treasury bonds, increase in treasury security yields, increase in TIPS value, increase bank reserves. & N/A. & Sentence is not relevant to monetary policy. \\
\midrule
\textbf{Money Supply} & Low money supply, slow M4 growth, increased loan demand. & High money supply, increased demand for goods, low loan demand. & N/A. & Sentence is not relevant to monetary policy. \\
\midrule
\textbf{Key Words and Phrases} & Stance is ``accommodative,'' indicating a focus on ``maximum employment'' and ``price stability.'' & Indicative a focus on ``price stability'' and ``sustained growth.'' & Usage of ``mixed,'' ``moderate,'' ``reaffirmed.'' & Sentence is not relevant to monetary policy. \\
\midrule
\textbf{Labor} & Productivity increases, unemployment decreases, wages increase, job openings increase. & Productivity decreases, unemployment increases, wages decrease, job openings decrease. & N/A. & Sentence is not relevant to monetary policy. \\
\bottomrule
\end{longtable}

\fwcertaintytext{boe}
\newpage

\begin{table*}
\caption{\fwtitle{Bank of England}}
\vspace{1em}
\begin{tabular}{p{0.3\textwidth}p{0.3\textwidth}p{0.3\textwidth}}
\toprule
\textbf{Label} & \textbf{Description} & \textbf{Example}\\
\midrule
\textbf{Forward Looking} & Data or statements that focus on anticipated future economic conditions, forecasts, or expectations, which inform potential policy actions or market movements. & “The upcoming Budget may be expansionary.” \\
\midrule
\textbf{Not Forward Looking} & Data or analysis that reflects past economic performance or trends, used to understand historical conditions. & “Retail sales volumes had fallen sharply in April, in part reflecting weather-related volatility in spending.” \\
\bottomrule
\end{tabular}
\label{tb:boe_forward_looking_guide}
\end{table*}

\begin{table*}
\caption{\certaintytitle{Bank of England}}
\vspace{1em}
\begin{tabular}{p{0.3\textwidth}p{0.3\textwidth}p{0.3\textwidth}}
\toprule
\textbf{Label} & \textbf{Description} & \textbf{Example}\\
\midrule
\textbf{Certain} & Information or outcomes that are highly reliable and predictable, with little room for doubt or uncertainty in economic forecasts or policy guidance. & “That would reduce lending rates for businesses with floating-rate loans linked to LIBOR.” \\
\midrule
\textbf{Uncertain} & Data, forecasts, or situations that involve ambiguity or risk, where the outcome is unpredictable or subject to significant variability, affecting decision-making.  & “It seemed possible that a further broad-based monetary stimulus would on its own be insufficient to transform the outlook for growth.” \\
\bottomrule
\end{tabular}
\label{tb:boe_certainty_guide}
\end{table*}
\clearpage
\usubsection{ Swiss National Bank}
\begin{center}
    \textbf{Region: Switzerland}
\end{center}
\begin{center}
    \fbox{\includegraphics[width=0.99\textwidth]{resources/flags/Flag_of_Switzerland.png}} 
\end{center}
\begin{center}
    \textbf{Data Collected: 2000-2024}
\end{center}
\vfill
\begin{center}
  \fbox{%
    \parbox{\textwidth}{%
      \begin{center}
      \textbf{Important Links}\\
      \href{https://www.snb.ch/en/}{Central Bank Website}\\
       \href{https://huggingface.co/datasets/gtfintechlab/swiss_national_bank}{Annotated Dataset}\\
     \href{https://huggingface.co/gtfintechlab/model_swiss_national_bank_stance_label}{Stance Label Model} \\
     \href{https://huggingface.co/gtfintechlab/model_swiss_national_bank_time_label}{Time Label Model} \\
     \href{https://huggingface.co/gtfintechlab/model_swiss_national_bank_certain_label}{Certain Label Model} \\
      \end{center}
    }
  }
\end{center}

\newpage

\section*{Monetary Policy Mandate} 
According the the Swiss National Banks (SNB) official website, their mandate is to "ensure price stability. In so doing, it shall take due account of economic developments"

\textbf{Mandate Objectives:} 
    \begin{itemize}
    \item \textbf{Price Stability}: Aiming for between 0-2\% annual inflation rate, as measured by the National Consumer Price Index (CPI), to maintain purchasing power while considering the economy's economic situation as a whole.
    \end{itemize}
\section*{Structure} 
\textbf{Composition: }
    \begin{itemize}
        \item \textbf{The Governing Board}: The highest governing body in the SNB made up of the Chairperson, Vice-Chairperson, and one additional member.
        \item \textbf{The Enlarged Governing Board}: 
        \begin{itemize}
            \item \textbf{Permanent Members: }The Governing Board 
            \item \textbf{Alternate Members: }Up to six alternate members (two per governing member) who are recommend by the Bank Council and chosen by the Federal Council, Switzerland's highest governing body.
        \end{itemize}
        \item \textbf{The Bank Council}: Made up of eleven member, six chosen by the Federal Council and five chosen by shareholders
        \begin{itemize}
            \item \textbf{Committees: }Comprises of the Risk Committee, Audit Committee, Compensation Committee, and Nomination Committee. 
        \end{itemize}
        \item \textbf{Shareholders}: Elect five members of the Bank Council and vote on profit distribution.
    \end{itemize}
\textbf{Meeting Structure: } 
\begin{itemize}
    \item \textbf{Frequency: }Quarterly meetings to determine their monetary policy and set interest rates
    \item \textbf{Additional Meetings: }An annual shareholder meeting in April
\end{itemize}

\section*{Manual Annotation} 

\textbf{Annotators} 
\begin{itemize}
    \item Pranav Aluru
    \item Sahasra S. Chava
    \item Arnav Hiray
    \item Dylan Kelly
\end{itemize} 

\textbf{Annotation Agreement} 
The agreement percentage among the pairs of annotators for different labels.
\begin{itemize}
    \item \SD Agreement: 60.3\%
    \item \TC Agreement: 78.0\%
    \item \CE Agreement: 83.4\%
\end{itemize} 

\textbf{Annotation Guide} 
\mptext{snb}{Nine} Interest Rates, Inflation Rate, GDP Growth Forecast, Export Performance, Consumer Demand, Employment Levels, Exchange Rate Movements, Banking Industry, and Tourism. 
\begin{itemize}
    \item \textit{Interest Rates: } A sentence pertaining to changes in interest rates or the Swiss Average Rate Overnight (SARON).
    \item \textit{Inflation Rate: }A sentence pertaining to the inflation rate, consumer price index, or some other inflation-related metric  or references price stability.
    \item \textit{GDP Growth Forecast: }A sentence pertaining to future economic growth expectations.
    \item \textit{Export Performance: }A sentence pertaining to the export level 
    \item \textit{Consumer Demand: }A sentence pertaining to consumer demand and consumer spending level  
    \item \textit{Employment Levels: }A sentence pertaining to employment
    \item \textit{Exchange Rate Movements: }A sentence pertaining to the value of the Swiss Franc.
    \item \textit{Banking Industry: }A sentence pertaining to Switzerland's banking industry.
    \item \textit{Tourism: }A sentence pertaining to the tourism level in Switzerland.
\end{itemize}

\textbf{Examples:}
\begin{itemize}
    \item "The economy has remained relatively robust despite the deterioration in the world economy, but the slowdown is likely to continue over the months to come."\newline\textbf{Dovish: }The economic is projected to slowdown so the SNB will be inclined to promote spending.
    \item "Since the indicators for the demand for labour are still at a high level, the rise in employment is likely to continue in the first part of 2008."\newline\textbf{Hawkish: }The SNB expects employment to increase causing more money to enter into the economy.
    \item "Inflationary pressure from abroad will remain weak."\newline\textbf{Neutral: }This statement does not directly comment on the state of the Swiss economy. 
    \item "Through to the second quarter of 2007, this curve is a little higher than for June."\newline
    \textbf{Irrelevant: }It is unclear what this statement is referring to. 
    
\end{itemize}

\newpage

\begin{longtable}{p{0.118\textwidth}p{0.183\textwidth}p{0.183\textwidth}p{0.183\textwidth}p{0.183\textwidth}}
\caption{\mptitle{Swiss National Bank}} \label{tb:snb_mp_stance_guide}\\ 
\toprule
\textbf{Category} & \textbf{Hawkish} & \textbf{Dovish} & \textbf{Neutral} & \textbf{Irrelevant} \\
\midrule
\endfirsthead

\toprule
\textbf{Category} & \textbf{Hawkish} & \textbf{Dovish} & \textbf{Neutral} & \textbf{Irrelevant} \\
\midrule
\endhead
\textbf{Interest Rates} & When interest rate or SARON rate is high. & When interest rate or SARON rate is low. & When interest rate or SARON rate is stable. & Sentence is not relevant to monetary policy. \\
\midrule
\textbf{Inflation Rate} & When inflation more than the target of 2\%. & When inflation is below the desired level. & When is at the target of 0-2\%. & Sentence is not relevant to monetary policy. \\
\midrule
\textbf{GDP Growth Forecast} & When GDP is increasing. & When GDP is decreasing. & When GDP growth rate is stable. & Sentence is not relevant to monetary policy. \\
\midrule
\textbf{Export Performance} & When exports are performing better than expected and earnings are rising. & When exports are declining and earnings are falling. & When exports are as expected and the rate is sustainable. & Sentence is not relevant to monetary policy. \\
\midrule
\textbf{Consumer Demand} & When consumer demand is higher than expected and inflation is increasing. & When consumer demand is weak and spending is low a. & When the economy is growing at a sustainable rate in line with expectations. & Sentence is not relevant to monetary policy. \\
\midrule
\textbf{Employment Levels} & When unemployment is very low. & When unemployment is high. & Unemployment is at a sustainable level and within projections. & Sentence is not relevant to monetary policy. \\
\midrule
\textbf{Exchange Rate Movements} & When Franc is depreciating. & When currency is appreciating. & When currency value is stable. & Sentence is not relevant to monetary policy. \\
\midrule
\textbf{Banking Industry} & When the banking industry is growing too quickly. & When the global economy is weak and financial activity is low, harming the bank-dependent Swiss economy. & When the banking industry is stable. & Sentence is not relevant to monetary policy. \\
\midrule
\textbf{Tourism} & When the tourism season is very active. & When tourism levels are weak. & When tourism levels are as expected. & Sentence is not relevant to monetary policy. \\
\bottomrule
\end{longtable}

\fwcertaintytext{snb}
\newpage

\begin{table*}
\caption{\fwtitle{Swiss National Bank}}
\vspace{1em}
\begin{tabular}{p{0.3\textwidth}p{0.3\textwidth}p{0.3\textwidth}}
\toprule
\textbf{Label} & \textbf{Description} & \textbf{Example}\\
\midrule
\textbf{Forward Looking} & When the sentence discusses expectations, projections, or anticipations about future economic conditions, policy actions, or resource utilization. & “Although the pace of economic recovery is likely to be moderate for a time, the Committee anticipates a gradual return to higher levels of resource utilization in a context of price stability.” \\
\midrule
\textbf{Not Forward Looking} & When the sentence reflects on recent or past economic data, trends, or events to describe what has already occurred. & “Commodity prices had been mixed recently after trending down earlier.” \\
\bottomrule
\end{tabular}
\label{tb:snb_forward_looking_guide}
\end{table*}

\begin{table*}
\caption{\certaintytitle{Swiss National Bank}}
\vspace{1em}
\begin{tabular}{p{0.3\textwidth}p{0.3\textwidth}p{0.3\textwidth}}
\toprule
\textbf{Label} & \textbf{Description} & \textbf{Example}\\
\midrule
\textbf{Certain} & When the sentence reflects a clear, unanimous, or definitive
decision or expectation about future actions or outcomes without ambiguity. & “To support the Committee’s decision to raise the target range for the federal funds rate, the Board of Governors of the Federal Reserve System voted unanimously to raise the interest rate paid on reserve balances to 0.” \\
\midrule
\textbf{Uncertain} & When the sentence expresses concerns, potential risks, or
varying opinions, often using words like ``could,'' ``may,'' or ``some participants.'' & “Some participants were concerned that inflation could rise as the recovery continued, and some business contacts had reported that producers expected to see an increase in pricing power over time.” \\
\bottomrule
\end{tabular}
\label{tb:snb_certainty_guide}
\end{table*}
\clearpage
\usubsection{Central Bank of Brazil}
\begin{center}
    \textbf{Region: Brazil}
\end{center}
\begin{center}
    \fbox{\includegraphics[width=0.99\textwidth]{resources/flags/Flag_of_Brazil.png}}
\end{center}

\begin{center}
    \textbf{Data Collected: 2000-2024}
\end{center}
\vfill
\begin{center}
  \fbox{%
    \parbox{\textwidth}{%
      \begin{center}
      \textbf{Important Links}\\
      \href{https://www.bcb.gov.br/}{Central Bank Website}\\
       \href{https://huggingface.co/datasets/gtfintechlab/central_bank_of_brazil}{Annotated Dataset}\\
     \href{https://huggingface.co/gtfintechlab/model_central_bank_of_brazil_stance_label}{Stance Label Model} \\
     \href{https://huggingface.co/gtfintechlab/model_central_bank_of_brazil_time_label}{Time Label Model} \\
     \href{https://huggingface.co/gtfintechlab/model_central_bank_of_brazil_certain_label}{Certain Label Model} \\
      \end{center}
    }
  }
\end{center}

\newpage

\section*{Monetary Policy Mandate} 
The main responsibility of the Monetary Policy Committee (Copom) of the Banco Central do Brasil (BCB) is to set the target for the Selic rate (the policy rate).

\textbf{Mandate Objectives:} 
\begin{itemize}
    \item \textbf{Inflation Target}: National Monetary Council (CMN) establishes the annual inflation target in the Brazilian economy, which the BCB's aims to achieve through monetary policy decisions. As of Janurary 2025, the  inflation target is 3\%, with a tolerance of $\pm 1.5\%$
    \item \textbf{Sustainable Economic Growth}: By maintaining a low inflation, the economy can flourish with stable prices, thus, creating employment and enhancing the society's well-being.
\end{itemize}

\section*{Structure}

All members of Copom are selected by the President of the Republic following their approval by the Federal State, subject to the Federal Constitution.

\textbf{Composition: } 
\begin{itemize}
    \item \textbf{Board of Governors:} Nine members appointed by the President and confirmed by the Federal Senate, serving 4-year terms. The Board of Governors consist of
    \begin{itemize}
        \item BCB Governor: Member of Copom who has the casting vote. 
        \item Deputy Governors: Members of the BCB who oversee  many divisions of the bank such as Economic Policy, International Affairs, etc. 
    \end{itemize}
\end{itemize}

\textbf{Meeting Structure: } 

\begin{itemize}
    \item \textbf{Frequency:} Eight regular meetings a year (every 45 days) that take place over two days (two sessions).
\end{itemize}
 
\section*{Manual Annotation} 

\textbf{Annotators} 
\begin{itemize}
    \item Rudra Gopal
    \item Joshua Zhang
    \item Spencer Gosden
    \item Saketh Budideti
\end{itemize}

\textbf{Annotation Agreement} 
The agreement percentage among the pairs of annotators for different labels.
\begin{itemize}
    \item \SD Agreement: 43.4\%
    \item \TC Agreement: 75.6\%
    \item \CE Agreement: 78.4\%
\end{itemize} 

\textbf{Annotation Guide} 
\mptext{bcb}{twelve} Selic Rate, Monetary Policy, Inflation, Aggregate Demand, Food Prices, Industrial Goods Prices, Services Prices, Labor Market, GDP Growth, Exchange Rate, Trade Deficit, and the Cattle Cycle.

\begin{itemize}
    \item \emph{Selic Rate}: The benchmark interest rate set by Brazil's central bank.
    \item \emph{Monetary Policy}: The process by which the central bank controls the money supply, typically through adjustments to interest rates.
    \item \emph{Inflation}: The rate at which the general level of prices for goods and services is rising, reflecting changes in purchasing power.
    \item \emph{Aggregate Demand}: The total demand for goods and services in an economy at a given overall price level and period.
    \item \emph{Food Prices}: The market prices of food products.
    \item \emph{Industrial Goods Prices}: The prices of manufactured goods produced in industrial settings.
    \item \emph{Services Prices}: The prices charged for common services.
    \item \emph{Labor Market}: The interaction between workers and employers regarding employment, wages, and job availability.
    \item \emph{GDP Growth}: The rate of increase in a economy's Gross Domestic Product, indicating economic expansion.
    \item \emph{Exchange Rate}: The value of the Brazilian Real in relation to foreign currencies.
    \item \emph{Trade Deficit}: A situation where a economy's imports exceed its exports.
    \item \emph{Cattle Cycle}: The cyclical fluctuations in beef supply and prices driven by changes in cattle production.
\end{itemize}

\textbf{Examples}
\begin{itemize} 
    \item “Copom unanimously decided to increase the Selic rate by 0.50 p.p. to 11.25\% p.a.”
    \\ \textbf{Hawkish:} Increasing the interest rate is a contractionary move aimed at curbing inflation, indicating that policymakers are concerned about an overheating economy or rising inflation pressures. 
    
    \item “Copom assessed that the scenario—marked by resilient economic activity, labor market pressures, a positive output gap, increased inflation projections, and deanchored expectations—requires a more contractionary monetary policy.”
    \\ \textbf{Hawkish:} The sentences describes a condition (a positive output gap and rising inflation expectations) that justify tightening monetary policy to prevent the economy from overheating. 
    
    \item “All members agreed that it was appropriate to reduce the Selic rate by 0.50 percentage points.”
    \\ \textbf{Dovish:} A reduction in the interest rate is an expansionary action designed to stimulate economic activity. It reflects a concern that the economy might be underperforming, requiring expansionary monetary policy. 
    
    \item “The combination of a robust labor market, expansionary fiscal policy, and vigorous lending to households continues to support consumption and, consequently, aggregate demand.”
    \\ \textbf{Dovish:} The sentence implies that the economy is receiving supportive stimulus from other sectors, reducing the immediate need for tighter monetary policy. Even though it recognizes strong economic activity, it suggests that current measures are sufficient without extra contractionary steps. 
    
    \item “The labor market, which surprised positively throughout 2022, has shown resilience, with a net increase in job creation and relative stability in the unemployment rate.”
    \\ \textbf{Neutral:} The sentence reports economic conditions—namely, stability in the labor market—without suggesting any particular policy action. It serves as an observation of the current economic state rather than a signal for change. 

    \item “However, most of the committee members judged that this interpretation still seems premature and needs further corroboration by data.”
    \\ \textbf{Irrelevant:} This statement mainly expresses skepticism toward an interpretation and does not signal any explicit monetary policy direction or stance.
\end{itemize}

\newpage

\begin{longtable}{p{0.118\textwidth}p{0.183\textwidth}p{0.183\textwidth}p{0.183\textwidth}p{0.183\textwidth}}
\caption{\mptitle{Central Bank of Brazil}} \\
\toprule
\textbf{Category} & \textbf{Hawkish} & \textbf{Dovish} & \textbf{Neutral} & \textbf{Irrelevant} \\
\midrule
\endfirsthead

\toprule
\textbf{Category} & \textbf{Hawkish} & \textbf{Dovish} & \textbf{Neutral} & \textbf{Irrelevant} \\
\midrule
\endhead
\textbf{Selic Rate} & Increase in Selic rate to combat inflation. & Decrease in Selic rate to combat inflation. & Selic rate remains unchanged. & Sentence is not relevant to monetary policy. \\
\midrule
\textbf{Monetary Policy} & Central bank institutes contractionary or tightening monetary policy. & Central bank institutes expansionary or loosens monetary policy. & Monetary policy remains unchanged. & Sentence is not relevant to monetary policy. \\
\midrule
\textbf{Inflation} & Inflation exceeds upper limit of the tolerance band. Increasing inflation, inflationary pressures or expectations of inflation. & Inflation drops below lower limit of tolerance band. Decreasing inflation, inflationary pressures or expectations of inflation. & Inflation is within the central bank’s target range, with inflation and expectations stable. COPOM’s tolerance range for inflation is between 2\% and 5\%, with the target in the middle. & Sentence is not relevant to monetary policy. \\
\midrule
\textbf{Aggregate Demand (AD)} & Increasing aggregate demand. & AD is insufficient or decreasing. & Balanced AD and Aggregate Supply (AS), stable inflation, and steady growth. & Sentence is not relevant to monetary policy. \\
\midrule
\textbf{Food Prices} & Increasing or high food prices. & Decreasing or low food prices. & No change in food prices. & Sentence is not relevant to monetary policy. \\
\midrule
\textbf{Industrial Goods Prices} & Increasing or high price of industrialized goods. & Decreasing or low price of industrialized goods. & No change in price of industrialized goods. & Sentence is not relevant to monetary policy. \\
\midrule
\textbf{Services Prices} & Increasing or high service prices. & Decreasing or low service prices. & No change in price of services. & Sentence is not relevant to monetary policy. \\
\midrule
\textbf{Labor Market} & Low or decreasing unemployment. Demand for workers exceeds available supply. Excessive wage growth. Creation of more jobs. & High or decreasing unemployment. Weak demand for labor; underemployment; stagnant wages. & Full employment without overheating, stable wage growth, and balanced labor supply and demand. Average unemployment is around 10\%. & Sentence is not relevant to monetary policy. \\
\midrule
\textbf{GDP Growth} & GDP grows too quickly, accelerating beyond the economy’s sustainable capacity. & Weak, negative (recession), or insufficient GDP growth to sustain full employment and price stability. & Stable GDP growth without excessive fluctuations. COPOM examines the circumstances and global context; on average, growth is around 1\% quarterly. & Sentence is not relevant to monetary policy. \\
\midrule
\textbf{Exchange Rate (BRL)} & When the BRL depreciates. & When the BRL appreciates. & When a foreign currency appreciates or depreciates. & Sentence is not relevant to monetary policy. \\
\midrule
\textbf{Trade Deficit} & Decrease in exports or increase in imports, leading to a higher trade deficit. & Increase in exports or decrease in imports, leading to a lower trade deficit. & N/A & Sentence is not relevant to monetary policy. \\
\midrule
\textbf{Cattle Cycle} & Supply of beef is low, leading to higher prices. & Supply of beef is high, leading to lower prices. & N/A & Sentence is not relevant to monetary policy. \\
\bottomrule
\label{tb:bcb_mp_stance_guide}
\end{longtable}

\fwcertaintytext{bcb}
\newpage

\begin{table*}
\caption{\fwtitle{Central Bank of Brazil}}
\vspace{1em}
\begin{tabular}{p{0.3\textwidth} p{0.3\textwidth} p{0.3\textwidth}}
\toprule
\textbf{Label} & \textbf{Description} & \textbf{Example} \\
\midrule
\textbf{Forward Looking} & Refers to expectations of or forecasts of metrics. Key words include ``expects,'' ``forecast,'' and ``anticipates.'' & ``Consumer price indices will come under some upward pressure from the prices of rice, milk and dairy products and clothing prices.'' \\
\midrule
\textbf{Not Forward Looking} & Refers to past or ongoing events or values of metrics. Key words include ``previous,'' ``historically,'' and ``recorded.'' & ``Compared to July 2011, the rate decreased 0.9 p.p.'' \\
\bottomrule
\end{tabular}
\label{tb:bcb_forward_looking_guide}
\end{table*}

\begin{table*}
\caption{\certaintytitle{Central Bank of Brazil}}
\vspace{1em}
\begin{tabular}{p{0.3\textwidth} p{0.3\textwidth} p{0.3\textwidth}}
\toprule
\textbf{Label} & \textbf{Description} & \textbf{Example} \\
\midrule
\textbf{Certain} & Utilizes confirmed data or definite terms when discussing a metric. Key words include ``firmly,'' ``will,'' and ``decided.'' & ``In the month, there was no spot intervention in the domestic exchange market.'' \\
\midrule
\textbf{Uncertain} & Utilizes uncertain terms to describe a metric. Key words include 
``suggests,'' ``possibility,'' and ``risk.'' & ``Even though inflation is under control, there are growing uncertainties with respect to how long this trend will continue, despite high productivity gains.'' \\
\bottomrule
\end{tabular}
\label{tb:bcb_certainty_guide}
\end{table*}
\clearpage
\usubsection{ Reserve Bank of India}
\begin{center}
    \textbf{Region: India}
\end{center}
\begin{center}
    \fbox{\includegraphics[width=0.99\textwidth]{resources/flags/Flag_of_India.png}} 
\end{center}

\begin{center}
    \textbf{Data Collected: 2016-2024}
\end{center}
\vfill
\begin{center}
  \fbox{%
    \parbox{\textwidth}{%
      \begin{center}
      \textbf{Important Links}\\
      \href{https://website.rbi.org.in/en/web/rbi}{Central Bank Website}\\
       \href{https://huggingface.co/datasets/gtfintechlab/reserve_bank_of_india}{Annotated Dataset}\\
     \href{https://huggingface.co/gtfintechlab/model_reserve_bank_of_india_stance_label}{Stance Label Model} \\
     \href{https://huggingface.co/gtfintechlab/model_reserve_bank_of_india_time_label}{Time Label Model} \\
     \href{https://huggingface.co/gtfintechlab/model_reserve_bank_of_india_certain_label}{Certain Label Model} \\
      \end{center}
    }
  }
\end{center}

\newpage

\section*{Monetary Policy Mandate}
The Reserve Bank of India (RBI) is responsible for maintaining monetary stability and fostering economic growth.

\textbf{Mandate Objectives:}
\begin{itemize}
    \item \textbf{Inflation Control:} Keeping inflation within a target range of 4\% (+/-2\%), as measured by the Consumer Price Index (CPI).
    \item \textbf{Economic Growth:} Ensuring sufficient credit flow and liquidity to support balanced economic development.
    \item \textbf{Financial Stability:} Managing risks in the financial system to ensure economic resilience.
\end{itemize}

\section*{Structure}
The RBI’s governance framework underpins its role in regulating monetary and financial stability.

\textbf{Composition:}
\begin{itemize}
    \item \textbf{Central Board of Directors:} Led by the RBI Governor, with Deputy Governors and government-appointed members.
    \item \textbf{Monetary Policy Committee (MPC):}
    \begin{itemize}
        \item Six members: Three internal members from the RBI and three external experts.
        \item Responsible for setting policy rates based on inflation and growth forecasts.
    \end{itemize}
\end{itemize}

\textbf{Meeting Structure:}
\begin{itemize}
    \item \textbf{Frequency:} Bi-monthly reviews of economic conditions and policy efficacy.
    \item \textbf{Additional Meetings:} Convened during extraordinary economic developments.
\end{itemize}

\section*{Manual Annotation} 

\textbf{Annotators} 
\begin{itemize}
    \item Priyanshu Mehta
    \item Aryan Garg
    \item Naman Tellakula
    \item Akshar Ravichandran
\end{itemize}

\textbf{Annotation Agreement} 
The agreement percentage among the pairs of annotators for different labels.
\begin{itemize}
    \item \SD Agreement: 50.9\%
    \item \TC Agreement: 69.7\%
    \item \CE Agreement: 77.4\%
\end{itemize} 

\textbf{Annotation Guide}

\mptext{rbi}{fourteen} Inflation, Repo Rate, Reverse Repo Rate, Cash Reserve Ratio, Statutory Liquidity Ratio, GDP Growth Forecast, Monetary Policy Measures, Export Performance, Manufacturing Activity, Consumer Demand, Employment Levels, Commodity Prices, Credit Growth, Exchange Rate:

\begin{itemize}
    \item \emph{Inflation}: A sentence pertaining to the general increase in prices for goods and services; the decrease in the purchasing power of a currency.
    \item \emph{Repo Rate}: A sentence pertaining to the interest rate at which the central bank lends money to commercial banks.
    \item \emph{Reverse Repo Rate}: A sentence pertaining to the rate at which the central bank borrows funds from commercial banks to manage liquidity.
    \item \emph{Cash Reserve Ratio (CRR)}: A sentence pertaining to the percentage of a bank's deposits that must be held in reserve and not lent out.
    \item \emph{Statutory Liquidity Ratio (SLR)}: A sentence pertaining to the minimum percentage of a bank's liabilities that must be held in the form of liquid assets.
    \item \emph{GDP Growth Forecast}: A sentence pertaining to the projected rate of increase in a economy’s Gross Domestic Product, indicating economic expansion.
    \item \emph{Monetary Policy Measures}: A sentence pertaining to actions taken by the central bank to influence the money supply and interest rates.
    \item \emph{Export Performance}: A sentence pertaining to the export activity of a economy in foreign markets.
    \item \emph{Manufacturing Activity}: A sentence pertaining to the level of production and industrial output within the manufacturing sector.
    \item \emph{Consumer Demand}: A sentence pertaining to the overall spending by households on goods and services.
    \item \emph{Employment Levels}: A sentence pertaining to job availability, unemployment, and overall labor market conditions.
    \item \emph{Commodity Prices}: A sentence pertaining to the market prices of primary raw materials and inputs used in production.
    \item \emph{Credit Growth}: A sentence pertaining to the rate at which bank lending increases, reflecting prevailing financial conditions.
    \item \emph{Exchange Rate Movements}: A sentence pertaining to fluctuations in the value of a economy’s currency relative to other currencies.
\end{itemize}

\textbf{Examples}
\begin{itemize}
    \item “With persistently high food inflation, it would be in order to continue with the disinflationary policy stance that we have adopted.” \\
    \textbf{Hawkish:} Increasing the focus on curbing high inflationary pressures indicates a tightening stance aimed at controlling rising prices.
    
    \item “I am, therefore, of the view that a reduction in the policy repo rate by conventional 25 bps will be inadequate.” \\
    \textbf{Hawkish:} This sentence suggests that a minor rate cut is insufficient to address economic challenges, hinting at the need for stronger tightening measures.
    
    \item “There has also been an inching down in the median 3-month and 1-year ahead inflation expectations which is also comforting.” \\
    \textbf{Dovish:} The observation of falling inflation expectations supports a dovish stance, where easing measures are seen as appropriate given the lower inflation outlook.
    
    \item “Export dependent industries such as textiles are not doing well.” \\
    \textbf{Dovish:} Highlighting weak export performance implies that easing measures might be necessary to stimulate growth in export sectors.
    
    \item “Since the Indian middle income consumer is price sensitive, profits have risen.” \\
    \textbf{Neutral:} This sentence provides an observation on market conditions without signaling a clear need for either tightening or easing monetary policy.

    \item “It need not be a concern for the MPC.” \\
    \textbf{Irrelevant:} This statement mentioned the importance of a statement and does not discuss monetary policy or take a monetary policy stance.
\end{itemize}

\begin{longtable}{p{0.118\textwidth}p{0.183\textwidth}p{0.183\textwidth}p{0.183\textwidth}p{0.183\textwidth}}
\caption{\mptitle{Reserve Bank of India}} \label{tb:rbi_mp_stance_guide}
\\
\toprule
\textbf{Category Term} & \textbf{Hawkish} & \textbf{Dovish} & \textbf{Neutral} & \textbf{Irrelevant} \\
\midrule
\endfirsthead

\toprule
\textbf{Category} & \textbf{Hawkish} & \textbf{Dovish} & \textbf{Neutral} & \textbf{Irrelevant} \\
\midrule
\endhead

\textbf{Inflation} & Inflation above target or increasing; rising inflationary pressures prompting tighter policy. & Inflation below target or decreasing; low inflation encouraging easing to support growth. & Inflation at target or stable with no policy change. & Sentence is not relevant to monetary policy. \\
\midrule
\textbf{Repo Rate} & Increasing the repo rate to tighten monetary policy and curb inflation. & Decreasing the repo rate to ease monetary conditions and stimulate economic activity. & Maintaining the repo rate, reflecting balanced conditions. & Sentence is not relevant to monetary policy. \\
\midrule
\textbf{Reverse Repo Rate} & Increasing the reverse repo rate to reduce liquidity by encouraging banks to park funds. & Decreasing the reverse repo rate to boost liquidity by incentivizing lending. & Maintaining the reverse repo rate, indicating steady liquidity management. & Sentence is not relevant to monetary policy. \\
\midrule
\textbf{Cash Reserve Ratio (CRR)} & Increasing CRR to reduce liquidity by requiring banks to hold more reserves. & Decreasing CRR to increase liquidity by allowing more lending. & Maintaining CRR, showing unchanged reserve requirements. & Sentence is not relevant to monetary policy. \\
\midrule
\textbf{Statutory Liquidity Ratio (SLR)} & Increasing SLR to force banks to hold more liquid assets, reducing lending capacity. & Decreasing SLR to allow banks to hold fewer liquid assets, increasing lending capacity. & Maintaining SLR, indicating unchanged liquidity constraints. & Sentence is not relevant to monetary policy. \\
\midrule
\textbf{GDP Growth Forecast} & Projecting lower GDP growth that may signal the need for tighter policy if inflation risks persist. & Projecting higher GDP growth that may lead to easing measures to support expansion. & Projecting stable GDP growth, supporting maintenance of current policy stance. & Sentence is not relevant to monetary policy. \\
\midrule
\textbf{Monetary Policy Measures} & Implementing tightening measures (e.g., raising rates) to combat inflation. & Implementing accommodative measures (e.g., lowering rates) to foster growth. & No significant policy changes, suggesting a continuation of current stance. & Sentence is not relevant to monetary policy. \\
\midrule
\textbf{Exports} & Strong export growth that could overheat the economy and prompt tighter policy. & Weak export performance that may require easing measures to stimulate growth. & Stable export levels with consistent foreign demand, supporting current policy. & Sentence is not relevant to monetary policy. \\
\midrule
\textbf{Manufact-uring Activity} & Rapid expansion that may trigger inflationary pressures, calling for tightening. & Decline in activity suggesting easing measures to revive industrial growth. & Steady, balanced manufacturing output with no policy change indicated. & Sentence is not relevant to monetary policy. \\
\midrule
\textbf{Consumer Demand} & Surging demand that might cause demand-pull inflation and necessitate tightening. & Weak consumer demand indicating low spending, warranting easing measures. & Sustainable, balanced consumer demand with no inflationary impact. & Sentence is not relevant to monetary policy. \\
\midrule
\textbf{Employment} & Very low unemployment that could lead to wage inflation and overheating, prompting tightening. & High unemployment suggesting the need for expansionary policies to stimulate job creation. & Stable employment with balanced growth, indicating no policy shift. & Sentence is not relevant to monetary policy. \\
\midrule
\textbf{Commodity Prices} & Rising prices that can drive cost-push inflation, necessitating tightening measures. & Falling prices reducing inflationary pressures, supporting a dovish stance. & Stable prices contributing to predictable costs, maintaining current policy stance. & Sentence is not relevant to monetary policy. \\
\midrule
\textbf{Credit Growth} & Rapid credit growth that might lead to financial instability and inflation, requiring tightening. & Slow credit growth indicating tight financial conditions, suggesting easing to encourage lending. & Moderate credit growth in line with expansion, indicating no change in policy stance. & Sentence is not relevant to monetary policy. \\
\midrule
\textbf{Exchange Rate Movements} & A depreciating currency increasing import costs and inflation, possibly calling for tightening. & An appreciating currency reducing import costs and inflation, allowing for easing. & Stable exchange rates that keep external pressures in check, maintaining neutrality. & Sentence is not relevant to monetary policy. \\
\bottomrule
\end{longtable}

\fwcertaintytext{rbi}
\newpage

\begin{table*}
\caption{\fwtitle{Reserve Bank of India}}
\vspace{1em}
\centering
\begin{tabular}{p{0.3\textwidth} p{0.3\textwidth} p{0.3\textwidth}}
\toprule
\textbf{Label} & \textbf{Description} & \textbf{Example} \\
\midrule
\textbf{Forward Looking} & 
Future-oriented: describes expectations, projections, or anticipated trends. Key words include: ``Expected to,'' ``Projected,'' ``Forecast,'' and ``Anticipates.'' 
& “The monetary authority expects GDP growth to accelerate next quarter, highlighting its future-oriented perspective.” \\
\midrule
\textbf{Not Forward Looking} & 
Past-oriented: refers to historical events or trends that have already occurred. Key words include ``Has been,'' ``Recorded,'' ``Previous,'' and ``Historically.'' 
& “Historical data shows that inflation remained steady over the past decade, illustrating a retrospective analysis.” \\
\bottomrule
\end{tabular}
\label{tb:rbi_forward_looking_guide}
\end{table*}

\begin{table*}
\caption{\certaintytitle{Reserve Bank of India}}
\vspace{1em}
\centering
\begin{tabular}{p{0.3\textwidth} p{0.3\textwidth} p{0.3\textwidth}}
\toprule
\textbf{Label} & \textbf{Description} & \textbf{Example} \\
\midrule
\textbf{Certain} & 
Indicates definite outcomes with a clear, committed stance. Key words include ``Will,'' ``Is set to,'' ``Confirmed,'' and ``Decided.'' 
& “The board has decided to maintain interest rates at 2.00\%.” \\
\midrule
\textbf{Uncertain} & 
Suggests possibilities or outcomes that are not fully confirmed. Key words include: ``May,'' ``Might,'' ``Could,'' and ``Possibility.'' 
& “The policy rate could be lowered if consumer demand remains weaker than expected.” \\
\bottomrule
\end{tabular}
\label{tb:rbi_certainty_guide}
\end{table*}
\clearpage
\usubsection{European Central Bank}
\begin{center}
    \textbf{Region: European Union}
\end{center}
\begin{center}
    \fbox{\includegraphics[width=0.99\textwidth]{resources/flags/Flag_of_European_Union.png}} 
\end{center}
\begin{center}
    \textbf{Data Collected: 2015-2024}
\end{center}
\vfill
\begin{center}
  \fbox{%
    \parbox{\textwidth}{%
      \begin{center}
      \textbf{Important Links}\\
      \href{https://www.ecb.europa.eu/}{Central Bank Website}\\
       \href{https://huggingface.co/datasets/gtfintechlab/european_central_bank}{Annotated Dataset}\\
     \href{https://huggingface.co/gtfintechlab/model_european_central_bank_stance_label}{Stance Label Model} \\
     \href{https://huggingface.co/gtfintechlab/model_european_central_bank_time_label}{Time Label Model} \\
     \href{https://huggingface.co/gtfintechlab/model_european_central_bank_certain_label}{Certain Label Model} \\
      \end{center}
    }
  }
\end{center}

\newpage

\section*{Monetary Policy Mandate} 

The European Central Bank's core mandate is to maintain price stability of the Euro, the single currency of the Eurozone. It also takes responsibility for overseeing the health of member state central banks, ensuring that risk remains low across the European banking system.

\textbf{Mandate Objectives:} 

\begin{itemize}
    \item \textbf{Price Stability:} The ECB seeks to keep the inflation rate, as measured by the Harmonised Index of Consumer Prices (HICP), as close to 2\% as possible. 
\end{itemize}

The European Union achieves its mandate by manipulating three key interest rates under its control: the Main Refinancing Operations Interest Rate, and the Deposit Facility Interest. It also engages in Open Market Operations via asset purchase programmes.

\section*{Structure} 

The ECB is led by the 26-member Governing Council of the European Central Bank, which makes decisions to raise or lower the three key interest rates under its control.
Rate, and the Marginal Lending Facility Interest Rate. It is also responsible for decisions regarding various quantitative easing policies (asset purchasing programmes) that may be in place.

\textbf{Composition: }

\begin{itemize}
    \item \textbf{Governing Council:} The Governing Council of the ECB consists of six members on the Executive Board, plus the 20 governors of the national central banks of Eurozone countries who follow a rotational voting system.
    \item \textbf{Executive Board:} Six members of the Governing Council form the Executive Board. These members include the President and Vice President of the ECB, and are not aligned with any particular Eurozone member nation.
    \item \textbf{National Central Bank Governors:} The remaining 20 members of the Governing Council consist of the governors of the national central banks of the 20 Eurozone member states.
\end{itemize}

\textbf{Meeting Structure: } 
\begin{itemize}
    \item \textbf{Frequency:} Eight scheduled meetings annually, approximately every six weeks. 
    \item \textbf{Additional Meetings:} Held as needed to address urgent economic developments. 
\end{itemize}

\section*{Manual Annotation} 

\textbf{Annotators} 

\begin{itemize}
    \item Sebastian Jaskowski
    \item Nihal Nizan
    \item Abhishek Pillai
    \item Chaitanya Sri Yetukuri
\end{itemize}

\textbf{Annotation Agreement} 

The agreement percentage among the pairs of annotators for different labels.
\begin{itemize}
    \item \SD Agreement: 52.4\%
    \item \TC Agreement: 74.6\%
    \item \CE Agreement: 82.1\%
\end{itemize}

\textbf{Annotation Guide}

\mptext{ecb}{nine} Economic Status, Euro Value Change, Energy Prices, House Prices, Foreign Nations, ECB Expectations/Actions/Assets, Money Supply, Key Words/Phrases, and Labor.

\begin{itemize} 
    \item \emph{Economic Status}: A sentence pertaining to the general state of the Eurozone economy, incorporating factors such as inflation trends, GDP performance, and overall growth risks. 
    \item \emph{Euro Value Change}: A sentence that addresses variations in the Euro's value relative to other global currencies, whether it appreciates or depreciates. 
    \item \emph{Energy Prices}: A sentence relating to changes in energy costs within the Eurozone, including fluctuations in oil and natural gas prices. 
    \item \emph{House Prices}: A sentence that discusses shifts in the residential real estate market within the Eurozone, particularly changes in housing prices. 
    \item \emph{Foreign Nations}: A sentence concerning trade relations between the European Union and non-Eurozone countries, highlighting changes in trade deficits or external demand for EU goods. 
    \item \emph{ECB Expectations, Actions, and Assets}: A sentence describing the European Central Bank's economic outlook and policy responses, including interest rate adjustments and modifications to asset purchasing programs. 
    \item \emph{Money Supply}: A sentence discussing the availability of monetary resources in the Eurozone, reflecting trends in bank capitalization, changes in the M2 measure, and shifts in loan demand. 
    \item \emph{Key Words and Phrases}: A sentence that includes specific terminology or expressions frequently used by policymakers to indicate a dovish, hawkish, or neutral stance. 
    \item \emph{Labor}: A sentence addressing labor market conditions, including fluctuations in unemployment rates and changes in salaries. 
\end{itemize}

\textbf{Examples}
\begin{itemize}
    \item ``The purchasing managers’ index for manufacturing export orders had fallen into contractionary territory in march owing to the war in Ukraine and lockdowns in China.'' \newline
    \textbf{Dovish}: Slowing economic growth in the manufacturing sector is highlighted, signifying that the central bank may want to lower interest rates to stimulate the economy.
    
    \item ``In addition, options markets had, by and large, priced out low inflation outcomes over the following five years.'' \newline
    \textbf{Hawkish}: Future inflation expectations are high, suggesting that higher inflation may be to come, and the ECB may want to start contracting the money supply.
    
    \item ``Current interest rates are at levels that, maintained for a sufficiently long duration, will make a substantial contribution to this goal.'' \newline
    \textbf{Neutral}: The ECB reiterates that there will be no interest rate change.
    
    \item ``Labour costs had also confirmed continued moderate domestic price pressures.'' \newline
    \textbf{Hawkish}: High labor costs can lead to inflation, which may encourage the ECB to decrease the money supply.
    
    \item ``Data on new export orders pointed to a sharp fall of global trade in the second quarter of the year.'' \newline
    \textbf{Dovish}: This sentence indicates a significant decline in global trade, suggesting economic slowdown that may prompt policymakers to adopt a more dovish monetary policy stance.
    
    \item ``Against this background, it was recalled that the ECB also had secondary objectives.'' \newline
    \textbf{Irrelevant}: This sentence primarily references the ECB’s secondary objectives, which does not directly relate to immediate monetary policy actions.
\end{itemize}

\newpage

\begin{longtable}{p{0.118\textwidth}p{0.183\textwidth}p{0.183\textwidth}p{0.183\textwidth}p{0.183\textwidth}}
\caption{\mptitle{European Central Bank}} \\
\toprule
\textbf{Category} & \textbf{Hawkish} & \textbf{Dovish} & \textbf{Neutral} & \textbf{Irrelevant} \\
\midrule
\endfirsthead

\toprule
\textbf{Category} & \textbf{Hawkish} & \textbf{Dovish} & \textbf{Neutral} & \textbf{Irrelevant} \\
\midrule
\endhead
\textbf{Economic Status} & HICP Inflation increases or is higher than targeted 2\% rate, Inflation risk present on high side, Eurozone GDP and/or Eurozone GDP growth increases. & HICP Inflation decreases or is lower than targeted 2\% rate, Eurozone GDP and/or Eurozone GDP growth decreases, Economic growth risk present on the downside. & Inflation rate remains stable around 2\%, Economic growth remains unchanged. & Sentence is not relevant to monetary policy. \\
\midrule
\textbf{Euro Value Change} & Euro depreciates relative to other global currencies. & Euro appreciates relative to other global currencies. & Minimal or no change in the Euro Value relative to other global currencies. & Sentence is not relevant to monetary policy. \\
\midrule
\textbf{Energy Prices} & Energy prices in the Eurozone, including oil and or natural gas prices, increase. & Energy prices in the Eurozone, including oil and or natural gas prices, decrease. & N/A & Sentence is not relevant to monetary policy. \\
\midrule
\textbf{House Prices} & Housing prices in the Eurozone increase. & Housing price in the Eurozone decrease. & N/A & Sentence is not relevant to monetary policy. \\
\midrule
\textbf{Foreign Nations} & European Union trade deficit increases. & European Union trade deficit decreases, external (to Eurozone) demand for EU goods falls. & When foreign nation outside of the Eurozone or European Union is discussed. & Sentence is not relevant to monetary policy. \\
\midrule
\textbf{ECB Expectations, Actions, and Assets} & Inflation, Economic Growth higher than expected, Increase in key ECB interest rates, Selling off of Securities purchased through Purchasing Programs). & Inflation, Economic Growth lower than expected, Decrease in key ECB interest rates, Increase in or continued exercise of purchasing programs. & Inflation, Economic Growth near expectations, Unchanged ECB interest rates (Deposit Facility, Fixed Rate Tender, Marginal Lending), no change in purchasing program policy. & Sentence is not relevant to monetary policy. \\
\midrule
\textbf{Money Supply} & Money supply is high, High Capitalization of banks, M2 decreases, Low demand for loans. & Money supply is low, Low capitalization for banks, M2 increases, Increased demand for loans. Ex. "Banks reported, on balance, an increase in firms’ demand for loans or drawing of credit lines in the second quarter of 2022". & N/A & Sentence is not relevant to monetary policy. \\
\midrule
\textbf{Key Words and Phrases} & Indicating a focus on ``price stability'' and ``sustained growth.'' & When the stance is ``accommodative,'' a focus on ``maximum employment.'' & Use of phrases like ``mixed,'' ``moderate,'' ``reaffirmed.'' & Sentence is not relevant to monetary policy. \\
\midrule
\textbf{Labor} & When unemployment rate decreases or earnings increase. & When unemployment rate increases or earnings decrease. & Unemployment rate or growth is unchanged. & Sentence is not relevant to monetary policy. \\
\bottomrule
\label{tb:ecb_mp_stance_guide}
\end{longtable}

\fwcertaintytext{ecb}
\newpage

\begin{table*}
\caption{\fwtitle{European Central Bank}}
\vspace{1em}
\begin{tabular}{p{0.3\textwidth}p{0.3\textwidth}p{0.3\textwidth}}
\toprule
\textbf{Label} & \textbf{Description} & \textbf{Example}\\
\midrule
\textbf{Forward Looking} & The statement makes claims or statements about possible or certain future economic conditions. & ``Looking further ahead, in the absence of new disruptions, energy costs were expected to stabilize and supply bottlenecks to ease.'' \\
\midrule
\textbf{Not Forward Looking} & The statement makes claims or statements about current or past economic conditions (regardless of whether these conditions are certain or possible). & ``Clear signs of a recovery in consumption had emerged since may, but spending remained far below pre-lockdown levels.'' \\
\bottomrule
\end{tabular}
\label{tb:ecb_forward_looking_guide}
\end{table*}

\begin{table*}
\caption{\certaintytitle{European Central Bank}}
\vspace{1em}
\begin{tabular}{p{0.3\textwidth}p{0.3\textwidth}p{0.3\textwidth}}
\toprule
\textbf{Label} & \textbf{Description} & \textbf{Example}\\
\midrule
\textbf{Certain} & The speaker/writer is confident in the truth, accuracy, or occurrence of the statement. & ``Annual average hicp inflation for 2014 was 0.4\%.'' \\
\midrule
\textbf{Uncertain} & The speaker/writer expresses doubt, probability, or speculation about the information. & ``Growth could be higher if inflation came down more quickly than expected, if rising real incomes meant that spending increased by more than anticipated, or if the world economy grew more strongly than expected.'' \\
\bottomrule
\end{tabular}
\label{tb:ecb_certainty_guide}
\end{table*}

\clearpage
\usubsection{ Central Bank of the Russian Federation}
\begin{center}
    \textbf{Region: Russia}
\end{center}
\begin{center}
    \fbox{\includegraphics[width=0.99\textwidth]{resources/flags/ru.png}} 
\end{center}
\begin{center}
    \textbf{Data Collected: 2014-2024}
\end{center}
\vfill
\begin{center}
  \fbox{%
    \parbox{\textwidth}{%
      \begin{center}
      \textbf{Important Links}\\
      \href{https://www.cbr.ru/eng/}{Central Bank Website}\\
       \href{https://huggingface.co/datasets/gtfintechlab/central_bank_of_the_russian_federation}{Annotated Dataset}\\
     \href{https://huggingface.co/gtfintechlab/model_central_bank_of_the_russian_federation_stance_label}{Stance Label Model} \\
     \href{https://huggingface.co/gtfintechlab/model_central_bank_of_the_russian_federation_time_label}{Time Label Model} \\
     \href{https://huggingface.co/gtfintechlab/model_central_bank_of_the_russian_federation_certain_label}{Certain Label Model} \\
      \end{center}
    }
  }
\end{center}

\newpage

\section*{Monetary Policy Mandate} 
The CBR is responsible for formulating monetary policy in collaboration with the Russian Federation Government to meet its primary goal of maintaining the price stability of the Ruble. 

\textbf{Mandate Objectives:} 
\begin{itemize}
    \item \textbf{Price Stability}: main priority is to ensure price stability of the Ruble and maintain sustainably low inflation at an annual rate of 4\%, measured by the Consumer Price Index (CPI).
\end{itemize}

\section*{Structure} 

\textbf{Composition: }
\begin{itemize}
    \item \textbf{Governor}: The CBR Governor is appointed by the President of the Russian Federation and serves a four-year term. 
    \item \textbf{Board of Directors}: 14 committee members who are appointed by the State Duma, the President of the Russian Federation, and the CBR Governor for a five‑year period.
\end{itemize}

\textbf{Meeting Structure: } 
\begin{itemize}
    \item \textbf{Frequency}: The CBR holds eight meetings annually split into core and interim meetings.
    \item \textbf{Core Meetings}: Four quarterly meetings where the CBR publishes its medium term macroeconomic forecast, monetary policy report, and a mandatory press release.
    \item \textbf{Interim Meetings}: Four meetings held in-between the core meetings. CBR board members meet with representatives of foreign financial institutions, analysts, and community experts community to answer questions regarding monetary policy and to receive feedback from the audience.
    
\end{itemize}
 
\section*{Manual Annotation} 

\textbf{Annotators} 

\begin{itemize}
    \item Siddhartha Somani
    \item Rohan Bhasin
    \item Dhruv Adha
    \item Andrew DiBiasio
\end{itemize}

\textbf{Annotation Agreement} 
The agreement percentage among the pairs of annotators for different labels.
\begin{itemize}
    \item \SD Agreement: 49.8\%
    \item \TC Agreement: 78.7\%
    \item \CE Agreement: 89.0\%
\end{itemize} 

\textbf{Annotation Guide} 

\mptext{cbr}{thirteen} Economic Status, Ruble Value Change, Oil/Energy Prices, House Prices, Agriculture/Food Prices, Foreign Nations \& Sanctions, Military/Defense, CBR Expectations/Actions/Assets, Inflation Rates (CPI), Money Supply, Key Rate Fluctuations, Key Words/Phrases, and Labor.
\begin{itemize}
    \item \emph{Economic Status}: A sentence pertaining to the state of the economy, relating to inflation levels or economic growth/decay. 
    \item \emph{Ruble Value Change}: A sentence pertaining to the appreciation, depreciation, or maintenance of the Ruble.
    \item \emph{Oil/Energy Prices}: A sentence that discusses fluctuations in Russian crude oil export prices, petroleum development industry performance, natural gas production and distribution, energy commodities, or the oil/energy sector as a whole within the Russian Federation. 
    \item \emph{House Prices}: A sentence pertaining to changes in prices of real estate, domestic/global housing demand, mortgage rates and rental market prices.
    \item \emph{Agriculture/Food Prices}: A sentence pertaining to the performance of the agricultural sector (crop yield, food exports) and/or fluctuations in food prices.
    \item \emph{Foreign Nations \& Sanctions}: A sentence pertaining to international trade relations and/or changes regarding Western sanctions on the Russian Federation and Russian response. 
    \item \emph{Military/Defense}: A sentence pertaining to defense spending and/or its effect on GDP, or about labor shortages exacerbated by wartime recruitment efforts. 
    \item \emph{CBR Expectations/Actions/Assets}: A sentence that discusses changes in the Bank of Russia's key rate, bond yields, foreign exchange reserves, or any other monetary policy instrument or financial asset value.
    \item \emph{Inflation Rates (CPI)}: A sentence regarding the inflation rate as measured by the Consumer Price Index (CPI) and any inflationary/dis-inflationary pressures. 
    \item \emph{Money Supply}: A sentence that overtly discusses impact to the money supply or changes in demand for the Russian Federation.
    \item \emph{Key Rate Fluctuations}: A sentence that describes decisions to manipulate the Key Rate and its corresponding fluctuations.
    \item \emph{Key Words/Phrases}: A sentence containing certain key words that would classify as hawkish, dovish, or neutral, based upon its frequency and sentiment.
    \item \emph{Labor}: A sentence that relates to changes in labor productivity of the Russian Federation.
\end{itemize}

\textbf{Examples:}
\begin{itemize}
    \item ``In February—March 2019, inflation is holding somewhat lower than the Bank of Russia’s expectations.''\\
    \textbf{Dovish}: Low inflation as compared to expectations indicates a raise in price level is needed to meet the target, signaling that quantitative easing should be implemented to promote price growth. 
    
    \item ``Moreover, inflation expectations of households increased in July.''\\
    \textbf{Hawkish}: Inflationary expectations increasing signifies a potential need to implement tighter policies to bring inflation back down.

    \item ``The balance of risks remains skewed towards pro-inflationary risks, especially over a short-term horizon, driven by the VAT increase and price movements in individual food products.''\\
    \textbf{Hawkish}: Risks being aligned with elevated pro-inflationary risks over the short term indicates a need for immediate tightening policy to decrease pro-inflationary pressures.

    \item ``External inflation is stable and does not exert a noticeable influence on domestic prices.''\\
    \textbf{Neutral}: Inflation is stable and therefore no tightening to decrease inflation or easing to increase inflation is needed. 
    
    \item ``While assessing evolving inflation dynamics and economic developments against the forecast, the Bank of Russia admits the possibility of cutting the key rate gradually in coming Q2-Q3.''\\
    \textbf{Dovish}: Cutting the key rate to adjust inflation in line with the target projections is a form of monetary easing.
    
    \item ``On 27 October 2023, the Bank of Russia Board of Directors decided to increase the key rate by 200 basis points to 15.00\% per annum.''\\
    \textbf{Hawkish}: Increasing the key rate is a form of quantitative tightening and indicates a stricter monetary policy stance.
    
    \item ``On 3 February 2017, the Bank of Russia Board of Directors decided to keep the key rate at 10.00\% p.a.''\\
    \textbf{Neutral}: Maintaining the key rate indicates a neutral policy stance as no change is needed. 

    \item ``2 starting from 01.07.2016 auctions were discontinued.''\\
    \textbf{Irrelevant}: This sentence mentions that auctions were discontinued it does not provide any context related to monetary policy or a monetary policy stance.
\end{itemize}

\newpage

\begin{longtable}{p{0.118\textwidth}p{0.183\textwidth}p{0.183\textwidth}p{0.183\textwidth}p{0.183\textwidth}}
\caption{\mptitle{Central Bank of Russia}} \label{tb:cbr_mp_stance_guide}\\
\toprule
\textbf{Category} & \textbf{Hawkish} & \textbf{Dovish} & \textbf{Neutral} & \textbf{Irrelevant} \\
\midrule
\endfirsthead

\toprule
\textbf{Category} & \textbf{Hawkish} & \textbf{Dovish} & \textbf{Neutral} & \textbf{Irrelevant} \\
\midrule
\endhead
\textbf{Economic Status} & When inflation increases/rises above the 4\% target, unemployment decreases, or economic growth is projected high. & When inflation decreases/falls below the 4\% target, unemployment increases, or economic growth is projected low. & When the unemployment rate, growth, inflation, or key rate is unchanged. & Sentence is not relevant to monetary policy. \\ 
\midrule
\textbf{Ruble Value Change} & When the ruble depreciates - CBR may raise rates or intervene in the forex market to support the ruble. & When the ruble appreciates - CBR may intervene to slow appreciation or cut rates to reduce attractiveness of ruble assets. & N/A & Sentence is not relevant to monetary policy. \\ 
\midrule
\textbf{Oil/Energy Prices} & When oil/energy prices increase, when their is strong performance in the oil/energy sector, when oil development and/or exports are high. & When oil/energy prices decrease, when their is poor performance in the oil/energy sector, when oil development and/or exports are low. & N/A & Sentence is not relevant to monetary policy. \\ 
\midrule
\textbf{House Prices} & When housing prices, mortgage rates, housing demand, or rental market prices increase. & When housing prices, mortgage rates, housing demand, or rental market prices decreases. & N/A & Sentence is not relevant to monetary policy. \\ 
\midrule
\textbf{Agriculture and Food Prices} & When the performance of the agricultural sector is high (e.g high exports, large crop yield) or food prices increase. & When the performance of the agricultural sector is low (e.g low exports, low crop yield) or food prices decrease. & N/A & Sentence is not relevant to monetary policy. \\ 
\midrule
\textbf{Foreign Nations \& Sanctions} & When sanctions from other countries tighten, when trading relations worsen, increase in capital inflow, or improving alliances or counter-sanctions that directly challenge Western influence. & When sanctions from other countries ease, when trading relations improve and increase capital outflow, or seeking diplomatic resolutions or negotiations to ease sanctions pressure. & Maintaining existing geopolitical relationships without significant policy shifts and when foreign trade relations remain stable. & Sentence is not relevant to monetary policy. \\ 
\midrule
\textbf{Military and Defense} & Positive effect on GDP from increase in defense spending. & Negative effect on GDP from decreased defense spending. & N/A & Sentence is not relevant to monetary policy. \\ 
\midrule
\textbf{CBR Expectations, Actions, and Assets} & CBR expects higher inflation, tightening reserve requirements. & CBR expects lower inflation, easing reserve requirements. & Expecting balanced growth and ruble/price stability. & Sentence is not relevant to monetary policy. \\ 
\midrule
\textbf{Inflation Rates (CPI)} & CPI rises, signaling inflationary pressure. & CPI (inflation rate) falls, indicating price stability or deflationary pressures. & When CPI remains stable. & Sentence is not relevant to monetary policy. \\ 
\midrule
\textbf{Money Supply} & Tightening money supply, stricter lending policies, targeting lending risks. & Increasing money supply, easing lending conditions, supporting increased liquidity. & N/A & Sentence is not relevant to monetary policy. \\ 
\midrule
\textbf{Key Rate Fluctuations} & Key rate hikes to control inflation and stabilize the ruble. & Key rate cuts stimulate economic activity via monetary easing. & When the key rate is maintained between meetings/terms. & Sentence is not relevant to monetary policy. \\ 
\midrule
\textbf{Key Words and Phrases} & Price stability, target inflation, tightening policies like rate increases or raised reserve requirements. & Economic growth, accommodating, stimulating, supporting demand, consumption, export growth, or raise in GDP. & Use of phrases such as continued, steady, monitoring, and balanced. & Sentence is not relevant to monetary policy. \\ 
\midrule
\textbf{Labor} & When productivity decreases. & When productivity increases. & N/A & Sentence is not relevant to monetary policy. \\ 
\bottomrule
\end{longtable}

\fwcertaintytext{cbr}
\newpage

\begin{table*}
\caption{\fwtitle{Central Bank of Russia}}
\vspace{1em}
\begin{tabular}{p{0.3\textwidth}p{0.3\textwidth}p{0.3\textwidth}}
\toprule
\textbf{Label} & \textbf{Description} & \textbf{Example}\\
\midrule
\textbf{Forward Looking} & Semantics that indicate events or opinions formed based on events predicted to happen in the future such as ``expected'' or ``projected.'' & “The fiscal rule will set off the impact of the oil market conditions on inflation and the domestic economic environment as a whole.” \\
\midrule
\textbf{Not Forward Looking} & Semantics that indicate events or opinions formed based on reliable, known data or trends. & “January 2015 saw a continued fall in real wage growth and a sharp decline in consumer expenditures which exerts a restraining influence on the prices of goods and services.” \\
\bottomrule
\end{tabular}
\label{tb:cbr_forward_looking_guide}
\end{table*}

\begin{table*}
\caption{\certaintytitle{Central Bank of Russia}}
\vspace{1em}
\begin{tabular}{p{0.3\textwidth}p{0.3\textwidth}p{0.3\textwidth}}
\toprule
\textbf{Label} & \textbf{Description} & \textbf{Example}\\
\midrule
\textbf{Certain} & Semantics that indicate events or opinions formed based on reliable, known data or trends. & “The fiscal rule will set off the impact of the oil market conditions on inflation and the domestic economic environment as a whole.” \\
\midrule
\textbf{Uncertain} & Semantics that indicate events or opinions formed based on more volatile or unpredictable factors, like global commodity prices, using words like ``potential,'' ``could,'' ``might,'' or ``may.'' & “The economic recovery may be held back by a slower rollout of vaccination programmes, the spread of new virus strains, and the entailing toughening of restrictions, among other factors.” \\
\bottomrule
\end{tabular}
\label{tb:cbr_certainty_guide}
\end{table*}

\clearpage
\usubsection{ Central Bank of China (Taiwan)}
\begin{center}
    \textbf{Region: Taiwan}
\end{center}
\begin{center}
    \fbox{\includegraphics[width=0.99\textwidth]{resources/flags/Flag_Taiwan.png}} 
\end{center}
\begin{center}
    \textbf{Data Collected: 2017-2024}
\end{center}
\vfill
\begin{center}
  \fbox{%
    \parbox{\textwidth}{%
      \begin{center}
      \textbf{Important Links}\\
      \href{https://www.cbc.gov.tw/}{Central Bank Website}\\
       \href{https://huggingface.co/datasets/gtfintechlab/central_bank_of_china_taiwan}{Annotated Dataset}\\
     \href{https://huggingface.co/gtfintechlab/model_central_bank_of_china_taiwan_stance_label}{Stance Label Model} \\
     \href{https://huggingface.co/gtfintechlab/model_central_bank_of_china_taiwan_time_label}{Time Label Model} \\
     \href{https://huggingface.co/gtfintechlab/model_central_bank_of_china_taiwan_certain_label}{Certain Label Model} \\
      \end{center}
    }
  }
\end{center}

\newpage

\section*{Monetary Policy Mandate} 
The Central Bank of China (CBC) is responsible for maintaining monetary and financial stability in Taiwan while promoting sound economic development.

\textbf{Mandate Objectives:} 
\begin{itemize}
    \item \textbf{Inflation Control:} Maintaining inflation around 2\%, measured by the Consumer Price Index (CPI).
    \item \textbf{Economic Growth:} Supporting sustainable economic growth through efficient monetary policies.
    \item \textbf{Exchange Rate Stability:} Managing the New Taiwan Dollar (NTD) to ensure balanced trade and prevent financial instability.
\end{itemize}

\section*{Structure} 
\textbf{Composition: }
\begin{itemize}
    \item \textbf{Governor:} The CBC Governor oversees the implementation of monetary policies and financial operations.
    \item \textbf{Board of Directors:} Comprising of 15 experts from economics, banking, and finance to set policy priorities.
\end{itemize}

\textbf{Meeting Structure: } 

\begin{itemize}
    \item \textbf{Frequency:} Quarterly meetings to assess economic and monetary conditions.
    \item \textbf{Additional Meetings:} Convened during periods of extraordinary economic developments or financial crises.
\end{itemize}
 
\section*{Manual Annotation} 

\textbf{Annotators} 
\begin{itemize}
    \item Nandini Shukla
    \item Vatsal Dwivedi
    \item Akshar Ravichandran
    \item Siddhant Sukhani
\end{itemize} 

\textbf{Annotation Agreement} 
The agreement percentage among the pairs of annotators for different labels.
\begin{itemize}
    \item \SD Agreement: 56.9\%
    \item \TC Agreement: 83.3\%
    \item \CE Agreement: 84\%
\end{itemize} 

\textbf{Annotation Guide} 

\mptext{cbc}{fourteen} Interest Rates, Monetary Policy Stance, Inflation Control, Economic Growth, Exchange Rate Stability, Historical Context, GDP Growth Forecasts, Inflation Expectations, Export, Manufacturing Activity, Consumer Demand, Commodity Prices, Credit Growth, and Exchange Rate Movements. 

\begin{itemize}
    \item \emph{Interest Rates}: A sentence pertaining to the central bank's set interest rates that influence borrowing costs and overall economic liquidity.
    \item \emph{Monetary Policy Stance}: A sentence describing the direction and approach of the central bank's policy measures, including tightening or easing actions.
    \item \emph{Inflation Control}: A sentence relating to measures aimed at keeping consumer price inflation within target ranges to maintain price stability.
    \item \emph{Economic Growth}: A sentence addressing policies or indicators that reflect the growth of Taiwan's economy.
    \item \emph{Exchange Rate Stability}: A sentence related to the actions taken to maintain a stable value of the currency.
    \item \emph{Historical Context}: A sentence providing reference to past monetary policy actions to historical situations.
    \item \emph{GDP Growth Forecasts}: A sentence discussing projected changes in the Gross Domestic Product, which is often used as an indicator of economic performance.
    \item \emph{Inflation Expectations}: A sentence discussing inflation expectations, including actions or policy that would affect inflation expectations.
    \item \emph{Export}: A sentence pertaining to the performance of a economy’s exports and its implications on the economy.
    \item \emph{Manufacturing Activity}: A sentence discussing manufacturing production, including metrics such as manufacturing output.
    \item \emph{Consumer Demand}: A sentence describing the spending of households and its influence on the broader economy.
    \item \emph{Commodity Prices}: A sentence relating to the fluctuations in prices of commodities such as raw materials and energy.
    \item \emph{Credit Growth}: A sentence discussing the rate at which lending is fluctuating, indicating potential shifts in financial stability.
    \item \emph{Exchange Rate Movements}: A sentence addressing changes in the currency exchange rate.
\end{itemize}

\textbf{Examples: }
\begin{itemize}
    \item ``CBC focuses on price stability under 2\%.''\\
    \textbf{Hawkish}: The sentence signals that the central bank is committed to strict inflation control, implying a hawkish stance.
    
    \item ``During the Asian Financial Crisis, Taiwan protected its currency.''\\
    \textbf{Hawkish}: The sentence highlights a decisive response in a crisis, implying a tight monetary approach.
    
    \item ``Forecast CPI at 1.89\% supports growth.''\\
    \textbf{Dovish}: The sentence suggests a relatively low CPI, indicating that there is leeway for expansionary monetary policy.
    
    \item ``Weak orders for electronics and machinery may warrant supportive monetary steps.''\\
    \textbf{Dovish}: The sentence indicates that easing measures could be necessary to bolster export demand.
    
    \item ``CBC maintained rates at 2.00\% to ensure stability.''\\
    \textbf{Neutral}: The sentence provides a factual statement about unchanged interest rates without directional bias.

    \item ``The bank successively met with a total of 34 domestic banks between august 12 and august 21.''\\
    \textbf{Irrelevant}: The sentence describes a series of routine meetings with domestic banks and does not provide any indication of monetary stance.
\end{itemize}

\newpage

\begin{longtable}{p{0.118\textwidth}p{0.183\textwidth}p{0.183\textwidth}p{0.183\textwidth}p{0.183\textwidth}}
\caption{\mptitle{Central Bank of China}} \label{tb:cbc_mp_stance_guide}\\
\toprule
\textbf{Category} & \textbf{Hawkish} & \textbf{Dovish} & \textbf{Neutral} & \textbf{Irrelevant} \\
\midrule
\endfirsthead

\toprule
\textbf{Category} & \textbf{Hawkish} & \textbf{Dovish} & \textbf{Neutral} & \textbf{Irrelevant} \\
\midrule
\endhead

\textbf{Interest Rates} & High rates to control inflation and reduce liquidity. & Lower rates to encourage borrowing and spending. & Rates remain unchanged. & Sentence is not relevant to monetary policy. \\ 
\midrule
\textbf{Monetary Policy Stance} & Tightening measures such as reduced asset purchases or clear signals of future rate hikes. & Accommodative measures, including rate cuts, quantitative easing, or liquidity injections. & No significant change in policy stance or rates, indicating a wait-and-see approach. & Sentence is not relevant to monetary policy. \\ 
\midrule
\textbf{Inflation Control} & Strong measures to keep CPI well below 2\%. & Tolerance for higher inflation to support economic growth. & Inflation projections stable or within comfortable bounds. & Sentence is not relevant to monetary policy. \\ 
\midrule
\textbf{Economic Growth} & Policies to prevent overheating. & Measures to stimulate growth (e.g., rate cuts). & Balanced approach to growth. & Sentence is not relevant to monetary policy. \\ 
\midrule
\textbf{Exchange Rate Stability} & Aggressive forex interventions to stabilize or strengthen the currency. & Allowing slight depreciation to boost exports; looser approach to FX. & Stable exchange rates are maintained. & Sentence is not relevant to monetary policy. \\ 
\midrule
\textbf{Historical Context} & Tightening responses in past crises. & Accommodative responses in past downturns. & Mentions of periods without major policy shifts. & Sentence is not relevant to monetary policy. \\ 
\midrule
\textbf{GDP Growth Forecasts} & Focus on curbing excessive expansion to mitigate inflation. & Policies that drive higher growth through domestic demand or exports. & Stable projections indicating moderate, sustainable growth. & Sentence is not relevant to monetary policy. \\ 
\midrule
\textbf{Inflation Expectations} & Very strict control near 2\% or lower to avoid price instability. & Willingness to let inflation rise slightly if beneficial for growth. & Projections remain moderate and steady. & Sentence is not relevant to monetary policy. \\ 
\midrule
\textbf{Export} & Robust export growth, risking overheating and trade imbalances. & Declining export demand, suggesting easing measures to support exporters. & Stable export levels with sustainable growth patterns. & Sentence is not relevant to monetary policy. \\ 
\midrule
\textbf{Manufact-uring Activity} & Rapid output potentially causing input shortages or inflation pressures. & Decline in factory output due to weak global demand or disruptions. & Balanced output growth indicating stable demand in key markets. & Sentence is not relevant to monetary policy. \\ 
\midrule
\textbf{Consumer Demand} & Rising domestic consumption pushing inflation beyond targets. & Weak household spending needing policy stimulus. & Stable consumer demand, supporting GDP without driving up inflation. & Sentence is not relevant to monetary policy. \\ 
\midrule
\textbf{Commodity Prices} & Rising global commodity costs (energy, raw materials) pressuring margins. & Falling import prices easing inflation and supporting growth. & Stable commodity prices reducing uncertainty in production and exports. & Sentence is not relevant to monetary policy. \\ 
\midrule
\textbf{Credit Growth} & Rapid credit expansion increasing leverage and financial instability risks. & Slowing credit growth indicating tight conditions, possibly needing easing. & Moderate credit growth aligned with economic needs (e.g., SMEs). & Sentence is not relevant to monetary policy. \\ 
\midrule
\textbf{Exchange Rate Movements} & NTD appreciation, harming exports and requiring tighter interventions. & NTD depreciation, helping exports but risking imported inflation. & Stable rates ensuring balanced trade flows with minimal inflation pressure. & Sentence is not relevant to monetary policy. \\
\bottomrule
\end{longtable}

\fwcertaintytext{cbc}
\newpage




\begin{table*}
    \caption{\fwtitle{Central Bank of China (Taiwan)}}
    \vspace{1em}
    \centering
    \begin{tabular}{p{0.3\textwidth}p{0.3\textwidth}p{0.3\textwidth}}
        \toprule
        \textbf{Category} 
        & \textbf{Definition} 
        & \textbf{Sentence Example} \\
        \midrule
        \textbf{Forward Looking} 
        & Future-oriented: describes expectations, projections, or anticipated trends. Key words include ``expected to,'' ``projected,'' ``forecast,'' and ``anticipates.''
        & “The central bank projects GDP growth of 3\% next year.” \\
        \midrule
        
        \textbf{Not Forward Looking} 
        & Past-oriented: refers to historical events or trends that have already occurred. Key words include ``has been,'' ``recorded,'' ``previous,'' and ``historically.''
        & “Historically, inflation has been below 2\%, reflecting stable price levels in past years.” \\
        \bottomrule
    \end{tabular}
    \label{tb:cbc_forward_looking_guide}
\end{table*}

\begin{table*}
    \caption{\certaintytitle{Central Bank of China}}
    \vspace{1em}
    \centering
    \begin{tabular}{p{0.3\textwidth}p{0.3\textwidth}p{0.3\textwidth}}
        \toprule
        \textbf{Category} 
        & \textbf{Definition} 
        & \textbf{Sentence Example} \\
        \midrule
        \textbf{Certain} 
        & Indicates definite outcomes with a clear, committed stance. Key works include ``will,'' ``is set to,'' ``confirmed,'' and ``decided.''
        & “The central bank will raise the policy rate next month.” \\
        \midrule
        \textbf{Uncertain} 
        & Suggests possibilities or outcomes that are not fully confirmed. Key words include ``may,'' ``might,'' ``could,'' and ``possibility.''
        & “The monetary authority may lower rates if domestic demand softens further.” \\
        \bottomrule
    \end{tabular}
    \label{tb:cbc_certainty_guide}

\end{table*}
\clearpage
\usubsection{ Monetary Authority of Singapore}
\begin{center}
    \textbf{Region: Singapore}
\end{center}
\begin{center}
    \fbox{\includegraphics[width=0.99\textwidth]{resources/flags/Flag_of_Singapore.png}}
\end{center}
\begin{center}
    \textbf{Data Collected: 2013-2024}
\end{center}
\vfill
\begin{center}
  \fbox{%
    \parbox{\textwidth}{%
      \begin{center}
      \textbf{Important Links}\\
      \href{https://www.mas.gov.sg/}{Central Bank Website}\\
       \href{https://huggingface.co/datasets/gtfintechlab/monetary_authority_of_singapore}{Annotated Dataset}\\
     \href{https://huggingface.co/gtfintechlab/model_monetary_authority_of_singapore_stance_label}{Stance Label Model} \\
     \href{https://huggingface.co/gtfintechlab/model_monetary_authority_of_singapore_time_label}{Time Label Model} \\
     \href{https://huggingface.co/gtfintechlab/model_monetary_authority_of_singapore_certain_label}{Certain Label Model} \\
      \end{center}
    }
  }
\end{center}

\newpage

\section*{Monetary Policy Mandate}
The Monetary Authority of Singapore (MAS) is responsible for maintaining monetary and financial stability in Singapore while fostering economic growth.

\textbf{Mandate Objectives:}
\begin{itemize}
    \item \textbf{Exchange Rate Policy:} Managing the Singapore dollar (SGD) for price stability and economic growth.
    \item \textbf{Inflation Control:} Controlling inflationary pressures using exchange rate tools.
    \item \textbf{Financial Stability:} Ensuring the resilience of Singapore's financial system.
\end{itemize}

\section*{Structure}
\textbf{Composition:}
\begin{itemize}
    \item \textbf{Managing Director:} Oversees policy implementation and financial operations.
    \item \textbf{Board of Directors:} Composed of economic and financial experts setting strategic policy directions.
\end{itemize}

\textbf{Meeting Structure:}
\begin{itemize}
    \item \textbf{Frequency:} Regular policy reviews to assess macroeconomic conditions.
    \item \textbf{Additional Meetings:} Held during significant economic events.
\end{itemize}

\section*{Manual Annotation}

\textbf{Annotators} 

\begin{itemize}
    \item Marc-Alain Adjahi
    \item Jessica Ly
    \item Vivek Atmuri
    \item Giovanni Hsu
\end{itemize}

\textbf{Annotation Agreement}
The agreement percentage among the pairs of annotators for different labels.
\begin{itemize}
    \item \SD Agreement: 47.5\%
    \item \TC Agreement: 91.9\%
    \item \CE Agreement: 79\%
\end{itemize} 
\textbf{Annotation Guide}

\mptext{mas}{six} Exchange rate policy, Inflation control, Employment, Trade, Foreign labor and Talent attraction, and Global Supply Chain Hub.

\begin{itemize}
    \item \emph{Exchange Rate Policy}: A sentence pertaining to changes such as the appreciation or depreciation of the value of the Singapore dollar (SGD).
    \item \emph{Inflation Control}: A sentence pertaining to adjusting the inflationary pressure by weakening or strengthening the SGD.
    \item \emph{Employment}: A sentence pertaining to changes in the employment sectors.
    \item \emph{Trade}: A sentence pertaining to trade relations between Singapore and a foreign economy. If not discussing Singapore, we label irrelevant.
    \item \emph{Foreign Labor and Talent Attraction}: A sentence that discusses changes in immigration policies to increase or limit foreign talent.
    \item \emph{Global Supply Chain Hub}: A sentence that discusses Singapore's involvement in the global supply chain.
\end{itemize}

\textbf{Examples: }
\begin{itemize}
    \item ``To combat rising inflation, the central bank may need to steepen the exchange rate band.''\\
    \textbf{Hawkish}: Increasing the exchange rate band indicates a tighter monetary policy stance to maintain price stability.
    
    \item ``A flattening of the band would provide greater flexibility during economic downturns.''\\
    \textbf{Dovish}: Flattening the exchange rate band indicates a more accommodative stance to support the economy.

    \item ``The current monetary policy remains effective without any proposed changes.''\\
    \textbf{Hawkish}: There are no changes made to the monetary policy and remains neutral.
    
    \item ``External inflationary pressures are likely to stay muted for the rest of 2003.''\\
    \textbf{Dovish}: The statement suggests that inflationary pressures are not expected to worsen, thereby creating room for expansionary monetary policy.

    \item ``MAS will be shifting the schedule of its semi-annual cycle from January/July to April/October.''\\
    \textbf{Irrelevant}: This sentence relates to an operational or scheduling change and does not mention monetary policy decisions or stances.

\end{itemize}

\newpage

\begin{longtable}{p{0.118\textwidth}p{0.183\textwidth}p{0.183\textwidth}p{0.183\textwidth}p{0.183\textwidth}}
\caption{\mptitle{Monetary Authority of Singapore}} \label{tb:mas_mp_stance_guide} \\
\toprule
\textbf{Category} & \textbf{Hawkish} & \textbf{Dovish} & \textbf{Neutral} & \textbf{Irrelevant} \\
\midrule
\endfirsthead

\toprule
\textbf{Category} & \textbf{Hawkish} & \textbf{Dovish} & \textbf{Neutral} & \textbf{Irrelevant} \\
\midrule
\endhead
\textbf{Exchange Rate Policy} & In favor of or advocating for the appreciation of the Singapore dollar (SGD). & In favor of or advocating for the depreciation or weakening of the Singapore dollar (SGD). & Identifying economic trends but not suggestive of action to the currency. & Sentence is not relevant to monetary policy. \\
\midrule
\textbf{Inflation Control} & Favoring a stronger SGD to reduce inflationary pressures. & Tending towards weakening or maintaining the SGD due to the reduction of inflationary pressures. & Discusses inflation as a whole and mentions statistics but don’t take a stance on it. & Sentence is not relevant to monetary policy. \\
\midrule
\textbf{Employment} & Lack of employment due to the strengthening of the SGD due to other concerns. & Favoring a weaker SGD to boost export-driven employment sectors. & Discusses employment metrics without suggesting action either way. & Sentence is not relevant to monetary policy. \\
\midrule
\textbf{Trade} & Favoring stricter trade policies and protective measures to guard domestically produced items against international competition. & Favoring open markets and more relaxed policies to boost exports and secure Singapore’s role as a trade hub. & Discusses previous trade strategies or options for future opportunities. & Sentence is not relevant to monetary policy. \\
\midrule
\textbf{Foreign Labor Policy} & Sentences discuss the need for stricter immigration policies in order to keep local jobs and limit foreign labor. & Suggesting the allowing of an increase in foreign talent in the local markets to foster innovation and economic growth. & Discusses the maintaining of immigration policies and SGD or mentioning local vs. foreign labor forces. & Sentence is not relevant to monetary policy. \\
\midrule
\textbf{Global Supply Chain} & Suggesting the mitigation of global supply chain disruptions by focusing on more local sourcing. & Sentences implies the increase in involvement in global supply chain in order to enhance Singapore’s position as a hub and to invite investment. & Maintaining Signapore’s policy and involvement in the global supply chain due to the lack of concerns of disruptions or happiness with Singapore’s place as the hub. & Sentence is not relevant to monetary policy. \\
\bottomrule
\end{longtable}

\fwcertaintytext{mas}
\newpage

\begin{table*}
\caption{\fwtitle{Monetary Authority of Singapore}}
\vspace{1em}
\begin{tabular}{p{0.3\textwidth}p{0.3\textwidth}p{0.3\textwidth}}
\toprule
\textbf{Label} & \textbf{Description} & \textbf{Example}\\
\midrule
\textbf{Forward Looking} & The sentence is written in future tense or is suggestive of an action MAS will take. It could also mention a solution to a problem in the past or mention a modification to previous policies. & “To combat rising inflation, the central bank will steepen the exchange rate band.” \\
\midrule
\textbf{Not Forward Looking} & The sentence states facts about a previous event or mentions action that is currently being taken. It could also discuss the pros and cons of a previous policy or reminisce on what could have been done differently.  & “Previously, the flattening of the band provided greater flexibility during economic downturns.” \\
\bottomrule
\end{tabular}
\label{tb:mas_forward_looking_guide}
\end{table*}

\begin{table*}
\caption{\certaintytitle{Monetary Authority of Singapore}}
\vspace{1em}
\begin{tabular}{p{0.3\textwidth}p{0.3\textwidth}p{0.3\textwidth}}
\toprule
\textbf{Label} & \textbf{Description} & \textbf{Example}\\
\midrule
\textbf{Certain} & The sentence mentions an event that will happen, an event that is very likely to happen, or mentions an event that happened in the past. & “A flattening of the band will very likely provide greater flexibility during economic downturns.” \\
\midrule
\textbf{Uncertain} & The sentence mentions an event that should happen, an event that will likely happen, an event that is unsure to hapen, or an event that happened in the past, but is speculative of the reason. & “A flattening of the band should provide greater flexibility during economic downturns.” \\
\bottomrule
\end{tabular}
\label{tb:mas_certainty_guide}
\end{table*}
\clearpage
\usubsection{ Bank of Korea}
\begin{center}
    \textbf{Region: South Korea}
\end{center}
\begin{center}
    \fbox{\includegraphics[width=0.99\textwidth]{resources/flags/Flag_of_South_Korea.png}} 
\end{center}
\begin{center}
    \textbf{Data Collected: 2010-2024}
\end{center}
\vfill
\begin{center}
  \fbox{%
    \parbox{\textwidth}{%
      \begin{center}
      \textbf{Important Links}\\
      \href{https://www.BoK.or.kr/eng/main/main.do}{Central Bank Website}\\
       \href{https://huggingface.co/datasets/gtfintechlab/bank_of_korea}{Annotated Dataset}\\
     \href{https://huggingface.co/gtfintechlab/model_bank_of_korea_stance_label}{Stance Label Model} \\
     \href{https://huggingface.co/gtfintechlab/model_bank_of_korea_time_label}{Time Label Model} \\
     \href{https://huggingface.co/gtfintechlab/model_bank_of_korea_certain_label}{Certain Label Model} \\
      \end{center}
    }
  }
\end{center}

\newpage

\section*{Monetary Policy Mandate} 
The Bank of Korea (BoK) serves as South Korea’s central bank, responsible for formulating and implementing monetary policy, maintaining financial stability, and issuing the national currency, the South Korean won (KRW). In addition to its policy making functions, the BoK provides banking services to the government and financial institutions, oversees payment systems, and manages the economy’s foreign exchange reserves. It also conducts research on macroeconomic and financial issues to support policy decisions.

\textbf{Mandate Objectives:}
The functions and responsibilities of the BoK are established by various legislative frameworks, primarily the Bank of Korea Act, which defines its authority, powers, and obligations.
\begin{itemize}
    \item \textbf{Price Stability:} Maintaining price stability as the foundation of sustainable economic growth.
    \item \textbf{Financial Stability:} Ensuring the soundness of the financial system to prevent systemic risks.
    \item \textbf{Economic Growth Support:} Contributing to balanced economic development and employment stability.
    \item \textbf{Foreign Exchange Management:} Managing foreign reserves and exchange rate policies to maintain external stability.
\end{itemize}

\section*{Structure}
The Bank of Korea (BoK) Monetary Policy Board is responsible for formulating and implementing monetary policy. Before each meeting, BoK staff prepare detailed reports on domestic and international economic conditions, financial market developments, and policy recommendations. Senior officials present these analyses during meetings to inform decision-making.

\textbf{Composition:}
\begin{itemize}
    \item \textbf{Executive Members:} The Monetary Policy Board consists of the following key officials:
    \begin{itemize}
        \item The Governor (Chair): serves a four-year term and may be reappointed for a single consecutive term
        \item The Senior Deputy Governor: serves a three-year term and may be reappointed for a single consecutive term
        \item The Deputy Governors: serves a three-year term and may be reappointed for a single consecutive term
    \end{itemize}
    \item \textbf{Non-Executive Members:} Additional members are appointed by the President of South Korea upon the recommendation of the government and financial sector authorities.
\end{itemize}

\textbf{Meeting Structure: } 
\begin{itemize}
    \item \textbf{Frequency:} Eight times a year, following the release of key economic indicators such as inflation reports, GDP growth figures, and financial stability assessments. Additionally, the Board convenes regularly
    on the second and fourth Thursdays of each month to discuss various economic and financial issues. 
    \item \textbf{Additional Meetings: } Held if deemed necessary by the Chairman or upon request by at least two members.
\end{itemize}
 
\section*{Manual Annotation} 

\textbf{Annotators} 

\begin{itemize}
    \item Soungmin (Min) Lee
    \item Sarah Chae
    \item Eric Kim
    \item Woojin Kang
\end{itemize} 

\textbf{Annotation Agreement} 
The agreement percentage among the pairs of annotators for different labels.
\begin{itemize}
    \item \SD Agreement: 48.0\%
    \item \TC Agreement: 72.9\%
    \item \CE Agreement: 65.6\%
\end{itemize} 
\textbf{Annotation Guide}

\mptext{bok}{14}: Economic Status, Keywords/Phrases, KRW Value Change, Housing Debt/Loan, House Prices, Foreign Nations, BoK Expectations/Actions/Assets, Money Supply, Labor, Interest rate gap between South Korea and United States, US Dollar Value, Foreign Exchange (FX) Market, COVID-19, and Energy Prices.

\begin{itemize}
    \item \textit{Economic Status}: A sentence pertaining to the state of the economy, relating to unemployment and inflation.
    \item \textit{Keywords/Phrases}: A sentence that contains key word or phrase that would classify it squarely into one of the three label classes, based upon its frequent usage and meaning among particular label classes.
    \item \textit{KRW Value Change}: A sentence pertaining to changes such as appreciation or depreciation of value of the South Korea Won on the Foreign Exchange Market.
    \item \textit{Housing Debt/Loan}: A sentence that relates to changes in debt in households.
    \item \textit{House Prices}: A sentence pertaining to changes in prices of real estate.
    \item \textit{Foreign Nations}: A sentence pertaining to trade relations between South Korea and a foreign economy. If not discussing South Korea we label neutral.
    \item \textit{BoK Expectations/Actions/Assets}: A sentence that discusses changes in BoK yields, bond value, reserves, or any other financial asset value.
    \item \textit{Money Supply}: A sentence that overtly discusses impact to the money supply or changes in demand.
    \item \textit{Labor}: A sentence that relates to changes in labor productivity.
    \item \textit{Interest rate gap between South Korea and United States}: A sentence pertaining to changes such as increase or decrease of the interest rate gap between the United States and South Korea.
    \item \textit{US Dollar Value}: A sentence about fluctuation of the dollar’s value which can influence monetary policy in a dovish or hawkish direction.
    \item \textit{FX Market}: A sentence about stability’s changes in the foreign exchange market.
    \item \textit{COVID-19}: A sentence indicating whether the government is tightening or easing COVID policy.
    \item \textit{Energy Prices}: A sentence pertaining to changes in prices of energy commodities or the energy sector as a whole.
\end{itemize}

\textbf{Examples: }
\begin{itemize}
    \item ``Concerning prices, the member presented the view that a depreciation of the won could further fuel inflationary pressures.''\\
    \textbf{Hawkish}: This statement warns that a depreciation of the won may lead to increased inflation, suggesting the need for a tighter monetary policy.
    
    \item ``Some of the members emphasized that it would be necessary to prepare against growing instability of the external sector due to the worsening global financial conditions including the rising won-dollar exchange rate, the narrowing current account surplus, and a possible foreign capital outflow owing to the widening gap between domestic and overseas interest rates.''\\
    \textbf{Hawkish}: This sentence reflects a hawkish stance by highlighting external financial instability and recommending proactive measures to counter potential inflationary pressures.
    
    \item ``Finally, in terms of financial stability, the member assessed that growth in household lending and housing prices had slowed since the second half of last year, but continued caution about the risk of a buildup of financial imbalances was required.''\\
    \textbf{Dovish}: This statement is dovish as it underscores caution and a preference for maintaining accommodative measures despite slowing growth in household lending and housing prices.
    
    \item ``In light of the narrowing interest rate gap between South Korea and the United States, some members suggested that the Bank of Korea may consider keeping interest rates low to support domestic economic recovery and encourage borrowing.''\\
    \textbf{Dovish}: This sentence advocates for lower interest rates to stimulate economic recovery, which is an expansionary monetary policy stance.

    \item ``The member, however, pointed out that the price path had not been showing a commensurate change.''\\
    \textbf{Neutral}: This sentence notes that the price path has not changed significantly without implying any particular monetary policy stance.
    
    \item ``The member thus emphasized the need to pay attention to this issue.''\\
    \textbf{Irrelevant}: This sentence is irrelevant because it does not provide enough context related to economic policy or market conditions relevant to the Bank of Korea.
\end{itemize}

\newpage

\begin{longtable}{p{0.118\textwidth}p{0.183\textwidth}p{0.183\textwidth}p{0.183\textwidth}p{0.183\textwidth}}
\caption{Bank of Korea Annotation Guide} \\
\toprule
\textbf{Category} & \textbf{Hawkish} & \textbf{Dovish} & \textbf{Neutral} & \textbf{Irrelevant} \\
\midrule
\endfirsthead

\toprule
\textbf{Category} & \textbf{Hawkish} & \textbf{Dovish} & \textbf{Neutral} & \textbf{Irrelevant} \\
\midrule
\endhead
\textbf{Economic Status} & When inflation decreases, when unemployment increases, when economic growth is projected as low. & When inflation increases, when unemployment decreases when economic growth is projected high when economic output is higher than potential supply/actual output when economic slack falls. & When unemployment rate or growth is unchanged, maintained, or sustained. & Sentence is not relevant to monetary policy. \\ 
\midrule
\textbf{KRW Value Change} & When KRW value appreciates. & When KRW value depreciates. & N/A & Sentence is not relevant to monetary policy. \\
\midrule
\textbf{US Dollar Value} & When USD is moderate or weakening. & When USD becomes stronger, as this indicates potential depreciation of KRW. & N/A & Sentence is not relevant to monetary policy. \\
\midrule
\textbf{Interest rate gap between South Korea and United States} & When US interest rate is aligned with South Korea’s, US not interfering with South Korea’s interest rates. & When US interest rate is higher than South Korea’s, potentially pressuring South Korea to raise interest rates. & N/A & Sentence is not relevant to monetary policy. \\
\midrule
\textbf{FX market} & Stable FX market. & Unstable FX market. & N/A & Sentence is not relevant to monetary policy. \\
\midrule
\textbf{Housing Debt/Loan} & When public or private debt level decreases, signaling more flexibility for borrowing and fiscal stimulus. & When public or private debt level decreases, signaling more flexibility for borrowing and fiscal stimulus. & N/A & Sentence is not relevant to monetary policy. \\
\midrule
\textbf{Housing Prices} & When house prices decrease or expected to decrease. & When house prices increase or expected to increase. & N/A & Sentence is not relevant to monetary policy. \\
\midrule
\textbf{Foreign \newline Nations} & When the Korean trade deficit decreases. & When the Korean trade deficit increases. & When relating to a foreign nation’s economic or trade policy. & Sentence is not relevant to monetary policy. \\
\midrule
\textbf{BoK Expectations/Actions/\newline Assets} & BoK expect supbar inflation, BoK expecting disinflation, narrowing spreads of treasury bonds, decreases in treasury security yields, and reduction of bank reserves. & When BoK signals the need for tightening, raising interest rates, or increasing reserves to curb inflation. & N/A & Sentence is not relevant to monetary policy. \\
\midrule
\textbf{Keywords/\newline Phrases} & When the stance is "accommodative", indicating a focus on “maximum employment” and “price stability.” & Indicating a focus on “price stability” and “sustained growth.” & Use of phrases “mixed”, “moderate”, “reaffirmed.” & Sentence is not relevant to monetary policy. \\
\midrule
\textbf{Money \newline Supply} & Money supply is low, M2 increases, increased demand for loans. & Money supply is high, increased demand for goods, low demand for loans. & N/A & Sentence is not relevant to monetary policy. \\
\midrule
\textbf{Labor} & When productivity increases, solid labor market. & When productivity decreases, unstable labor market. & N/A & Sentence is not relevant to monetary policy. \\
\midrule
\textbf{COVID-19} & Easing COVID-19 regulation: boost tourism industry and increased productivity. & Increased COVID-19 regulation: detrimental for tourism industry and decreased productivity. & N/A & Sentence is not relevant to monetary policy. \\
\midrule
\textbf{Energy} & When oil/energy prices decrease. & When oil/energy prices increase. & N/A & Sentence is not relevant to monetary policy. \\
\bottomrule
\label{tb:bok_mp_stance_guide}
\end{longtable}

\fwcertaintytext{bok}
\pagebreak

\begin{table*}
\caption{\fwtitle{Bank of Korea}} \vspace{1em}
\begin{tabular}{p{0.3\textwidth}p{0.3\textwidth}p{0.3\textwidth}}
\toprule
\textbf{Label} & \textbf{Description} & \textbf{Example} \\
\midrule
\textbf{Forward Looking} & When sentence contains words such as “expected”, “anticipated,” “projected,” “will,” or “forecasted.” & “Budget 2025 measures will provide additional support to growth.” \\
\midrule
\textbf{Not Forward Looking} & When sentence contains phrase like “had been,” “previously,” or “last year.” & “Domestic headline inflation moderated in September to 1.8.” \\
\bottomrule
\end{tabular}
\label{tb:bok_forward_looking_guide}
\end{table*}

\begin{table*}
\caption{\certaintytitle{Bank of Korea}} \vspace{1em}
\begin{tabular}{p{0.3\textwidth}p{0.3\textwidth}p{0.3\textwidth}}
\toprule
\textbf{Label} & \textbf{Description} & \textbf{Example} \\
\midrule
\textbf{Certain} & When sentence contains phrase like “will,” “must,” “certainly,” or “definitely.” & “The monetary policy statement will be released at 6 p.m. on the same day as the MPC meeting.” \\
\midrule
\textbf{Uncertain} & When sentence contains phrase like “might,” “could,” “may,” “possibly,” “likely,” or “uncertain.” & “These developments may increase the risk to the outlook for inflation.” \\
\bottomrule
\end{tabular}
\label{tb:bok_certainty_guide}
\end{table*}

\clearpage
\usubsection{Reserve Bank of Australia}
\begin{center}
    \textbf{Region: Australia}
\end{center}
\begin{center}
    \fbox{\includegraphics[width=0.99\textwidth]{resources/flags/aus-flag1.png}} 
\end{center}
\begin{center}
    \textbf{Data Collected: 2006-2024}
\end{center}
\vfill
\begin{center}
  \fbox{%
    \parbox{\textwidth}{%
      \begin{center}
      \textbf{Important Links}\\
      \href{https://www.rba.gov.au/}{Central Bank Website}\\
       \href{https://huggingface.co/datasets/gtfintechlab/reserve_bank_of_australia}{Annotated Dataset}\\
     \href{https://huggingface.co/gtfintechlab/model_reserve_bank_of_australia_stance_label}{Stance Label Model} \\
     \href{https://huggingface.co/gtfintechlab/model_reserve_bank_of_australia_time_label}{Time Label Model} \\
     \href{https://huggingface.co/gtfintechlab/model_reserve_bank_of_australia_certain_label}{Certain Label Model} \\
      \end{center}
    }
  }
\end{center}

\newpage

\section*{Monetary Policy Mandate} 
The Reserve Bank of Australia (RBA) serves as Australia's central bank, responsible for conducting monetary policy, maintaining financial stability, and issuing the national currency. Beyond its policy-making role, the RBA provides select banking and registry services to Australian government agencies, overseas central banks, and official institutions. Additionally, it manages Australia's gold and foreign exchange reserves.

The functions and responsibilities of the RBA are defined by various legislative frameworks, primarily the \textit{Reserve Bank Act 1959}, which establishes its statutory authority, powers, and obligations.

\textbf{Mandate Objectives}
The RBA's primary objectives, as outlined in the \textit{Reserve Bank Act 1959}, include:

\begin{itemize}
    \item \textbf{Currency Stability:} Ensuring the stability of the Australian dollar.
    \item \textbf{Full Employment:} Promoting full employment within Australia.
    \item \textbf{Economic Prosperity:} Supporting the economic welfare and prosperity of the Australian people.
\end{itemize}

\section*{Structure}
The Reserve Bank Board is responsible for formulating monetary policy. Before each meeting, RBA staff prepare comprehensive reports on domestic and international economic conditions, financial markets, and policy recommendations. Senior staff attend and present during these meetings.
\begin{itemize}
    \item \textbf{Composition}
\begin{itemize}
    \item \textbf{Executive Members:} The Board comprises three \textit{ex officio} members:
    \begin{itemize}
        \item The Governor (Chair)
        \item The Deputy Governor (Deputy Chair)
        \item The Secretary to the Treasury
    \end{itemize}
    The Governor and Deputy Governor serve terms of up to seven years and may be reappointed.
    \item \textbf{Non-Executive Members:} Six non-executive members are appointed by the Treasurer for terms of up to five years.
\end{itemize}
    \item \textbf{Meeting Structure:}
    The Board convenes eight times per year, following the release of key economic indicators, such as inflation and economic activity reports. Meetings take place at the RBA's Head Office in Sydney.

    \item \textbf{Quorum and Decision-Making:} A quorum requires five members, with the Governor or, in their absence, the Deputy Governor presiding. Decisions are made by majority vote, with the Chair holding a casting vote if necessary. Minutes of monetary policy meetings are published two weeks after each meeting. The Governor and Deputy Governor do not partake in discussions concerning their terms of employment.

    \item \textbf{Meeting Schedule:} Meetings commence on a Monday afternoon and conclude the following day. The outcome is announced at 2:30 PM on the second day. Following each meeting, the Governor holds a media conference to explain the policy decision.
\end{itemize}

\textbf{Meeting Structure:}
The Board convenes eight times per year, following the release of key economic indicators, such as inflation and economic activity reports. Meetings take place at the RBA's Head Office in Sydney.

\textbf{Quorum and Decision-Making:} A quorum requires five members, with the Governor or, in their absence, the Deputy Governor presiding. Decisions are made by majority vote, with the Chair holding a casting vote if necessary. Minutes of monetary policy meetings are published two weeks after each meeting. The Governor and Deputy Governor do not partake in discussions concerning their terms of employment.

\textbf{Meeting Schedule:} Meetings commence on a Monday afternoon and conclude the following day. The outcome is announced at 2:30 PM on the second day. Following each meeting, the Governor holds a media conference to explain the policy decision.

\section*{Manual Annotation} 

\textbf{Annotators} 

\begin{itemize}
    \item Eric Van Ness
    \item Ankit Agrawal
    \item Lana Duke
    \item Nithil Balaji
\end{itemize}

\textbf{Annotation Agreement} 
The agreement percentage among the pairs of annotators for different labels.
\begin{itemize}
    \item \SD Agreement: 63.3\%
    \item \TC Agreement: 88.8\%
    \item \CE Agreement: 83.2\%
\end{itemize} 

\textbf{Annotation Guide} 

\mptext{rba}{eleven} Economic Status, Government Spending, Key Words and Phrases, Dollar Value Change, Energy, Commodity, House Prices, RBA Expectations, Actions, Assets, Foreign Nations, Money Supply, Labor, Consumer Sentiment, Wages.

\textbf{Definitions:}
\begin{itemize}
    \item \textit{Economic Status}: A sentence discussing the current state of inflation, unemployment, or overall economic growth, and how these indicators influence monetary policy decisions.

    \item \textit{Government Spending}: A sentence pertaining to levels of government spending, such as enacting expansionary or contractionary fiscal policy.

    \item \textit{Key Words and Phrases}: A sentence that contains key word or phrase that would classify squarely into one of the three label classes, based upon its frequent usage and meaning among particular label classes.

    \item \textit{Dollar Value Change}: A sentence pertaining to changes such as appreciation or depreciation of value of the Australian Dollar on the Foreign Exchange Market.

    \item \textit{Energy/Commodity/House Prices}: A sentence pertaining to Energy via changes in prices of energy commodities or the energy sector as a whole, Commodities via precious metals or agricultural goods, or House Prices via single-family homes or the real estate sector as a whole.

    \item \textit{RBA Expectations/Actions/Assets}: A sentence that discusses changes in the cash rate, bond value, reserves, or any other financial asset value.

    \item \textit{Foreign Nations}: A sentence pertaining to trade relations between Australia and a foreign economy. If not discussing Australia we label neutral.

    \item \textit{Money Supply}: A sentence that overtly discusses impact to the money supply or changes in demand.

    \item \textit{Labor}: A sentence that relates to changes in labor productivity.

    \item \textit{Consumer Sentiment}: A sentence reflecting the general confidence of households regarding future economic conditions.

    \item \textit{Wages}: A sentence highlighting wage growth or decline, which impacts inflation and consumer spending
\end{itemize}

\textbf{Examples: }
\begin{itemize}
    \item ``The AONIA remained below target despite a recent policy change.''\\
    \textbf{Dovish}: Indicates that lower-than-target overnight lending rates suggest an easing stance.
    
    \item ``The RBA’s stance has been accommodative toward recent inflationary pressures.''\\
    \textbf{Dovish}: Suggests that the central bank is prioritizing economic support over tightening measures.
    
    \item ``Inflation has proved to be persistent and a challenge, staying above the inflation target even with recent efforts.''\\
    \textbf{Hawkish}: Implies that ongoing high inflation may necessitate a tightening of monetary policy.
    
    \item ``Import trade restrictions for China, one of Australia’s key trading partners could lead to supply shortages and higher prices, consequentially increasing inflationary pressures.''\\
    \textbf{Hawkish}: Highlights how external trade barriers can drive inflation upward, justifying stricter policy.
    
    \item ``After consideration, the cash rate target will remain unchanged as of October 2024.''\\
    \textbf{Neutral}: Reflects a decision to maintain the current policy without alteration.

    \item ``Members recognized that the calibration of this guidance was not precise or straightforward.''\\
    \textbf{Irrelevant}: Discusses the clarity and formulation of policy guidance rather than actual economic indicators or monetary actions.
\end{itemize}

\newpage

\begin{longtable}{p{0.118\textwidth}p{0.183\textwidth}p{0.183\textwidth}p{0.183\textwidth}p{0.183\textwidth}}
\caption{\mptitle{Reserve Bank of Australia}} \label{tb:rba_mp_stance_guide} \\

\toprule
\textbf{Category} & \textbf{Dovish} & \textbf{Hawkish} & \textbf{Neutral} & \textbf{Irrelevant} \\
\midrule
\endfirsthead

\toprule
\textbf{Category} & \textbf{Dovish} & \textbf{Hawkish} & \textbf{Neutral} & \textbf{Irrelevant} \\
\midrule
\endhead

\textbf{Economic Status} & 
When inflation decreases, when economic growth is predicted to be low, when cash rate (the goal overnight lending rate) target will be lowered, when inflation is below the inflation target, when the AONIA (the overnight lending rate) is below the cash rate target. & 
When inflation increases, when economic growth is predicted to be high, when cash rate target will be raised, when inflation is above the inflation target, when the AONIA (aka cash rate) is above the cash rate target. & 
When inflation remains unchanged, when the cash rate target remains unchanged, when inflation is at the inflation target, when the cash rate is at the cash rate target. & Sentence is not relevant to monetary policy. \\
\midrule

\textbf{Government Spending} & 
When government spending decreases. & 
When government spending increases. & N/A & Sentence is not relevant to monetary policy. \\
\midrule

\textbf{Key Words and Phrases} & 
When the stance is “accommodative,” indicating a focus on “maximum employment” and “price stability.” & 
Indicating a focus on “price stability” and “sustained growth.” & 
Use of phrases “mixed,” “moderate,” “reaffirmed.” & Sentence is not relevant to monetary policy. \\
\midrule

\textbf{Dollar Value Change} & 
When the Australian Dollar appreciates. & 
When the Australian Dollar depreciates. & N/A & Sentence is not relevant to monetary policy. \\
\midrule

\textbf{Energy, Commodity, House Prices} & 
When energy, commodities, or home prices decrease. & 
When energy, commodities, or home prices increase. & N/A & Sentence is not relevant to monetary policy. \\
\midrule

\textbf{RBA Expectations, Actions, Assets} & 
RBA expects subpar inflation, RBA expecting disinflation, narrowing spreads of treasury bonds, decreases in treasury security yields, and reduction of bank reserves. & 
RBA expects high inflation, widening spreads of treasury bonds, increase in treasury security yields, increase bank reserves. & N/A & Sentence is not relevant to monetary policy. \\
\midrule

\textbf{Foreign Nations} & 
When Australia’s trade deficit decreases or there are positive developments in trade relations, especially with top trading partners such as China, Japan, US, and the EU. & 
When Australia’s trade deficit increases or trade relations worsen, especially concerning top trading partners such as China, Japan, US, and the EU. & 
When relating to a foreign nation’s policy with no effect. & Sentence is not relevant to monetary policy. \\
\midrule

\textbf{Money Supply} & 
Money supply is low, M2 increases, increased demand for loans. & 
Money supply is high, increased demand for goods, low demand for loans. & N/A & Sentence is not relevant to monetary policy. \\
\midrule

\textbf{Labor} & 
When productivity increases, when unemployment increases. & 
When productivity decreases, when unemployment decreases. & N/A & Sentence is not relevant to monetary policy. \\
\midrule

\textbf{Consumer Sentiment} & 
Negative Consumer Sentiment. & 
Positive Consumer Sentiment. & N/A & Sentence is not relevant to monetary policy. \\
\midrule

\textbf{Wages} & 
Wages decrease. & 
Wages increase. & N/A & Sentence is not relevant to monetary policy. \\
\bottomrule
\end{longtable}

\fwcertaintytext{rba}

\newpage

\begin{table*}
\caption{\fwtitle{Reserve Bank of Australia}}
\vspace{1em}
\begin{tabular}{p{0.3\textwidth}p{0.3\textwidth}p{0.3\textwidth}}
\toprule
\textbf{Label} & \textbf{Description} & \textbf{Example} \\
\midrule
\textbf{Forward Looking} & Statements that anticipate or project future economic conditions, expectations, or policy actions. Key words include “outlook,” “expect,” “forecast,” “prospects,” and “likely.” & “The outlook for the labor market remains positive.” \\
\midrule
\textbf{Not Forward Looking} & Statements that describe past economic data, trends, or events. Key words include “has been,” “stable,” “last year,” and “past.” & “The economic outlook has been stable for the past year.” \\
\bottomrule
\end{tabular}
\label{tb:rba_forward_looking_guide}
\end{table*}

\begin{table*}
\caption{\certaintytitle{Reserve Bank of Australia}}
\vspace{1em}
\begin{tabular}{p{0.3\textwidth}p{0.3\textwidth}p{0.3\textwidth}}
\toprule
\textbf{Label} & \textbf{Description} & \textbf{Example} \\
\midrule
\textbf{Certain} & Statements that are definitively stated with clarity and without ambiguity. Key words include “guaranteed,” “remain,” “determined,” “decided,” “mandated,” “confirmed,” and “ensured.” & “The cash rate will remain at 4\% until inflation targets are met.” \\
\midrule
\textbf{Uncertain} & Statements that express potential risks, conditions, or future possibilities. Key words include “potential,” “may,” “could,” “unclear,” “contingent,” “likely,” and “possibly.” & “The RBA may consider further tightening if inflationary pressures persist.” \\
\bottomrule
\end{tabular}
\label{tb:rba_certainty_guide}
\end{table*}
\clearpage
\usubsection{ Bank of Israel}
\begin{center}
    \textbf{Region: Israel}
\end{center}
\begin{center}
    \fbox{\includegraphics[width=0.99\textwidth]{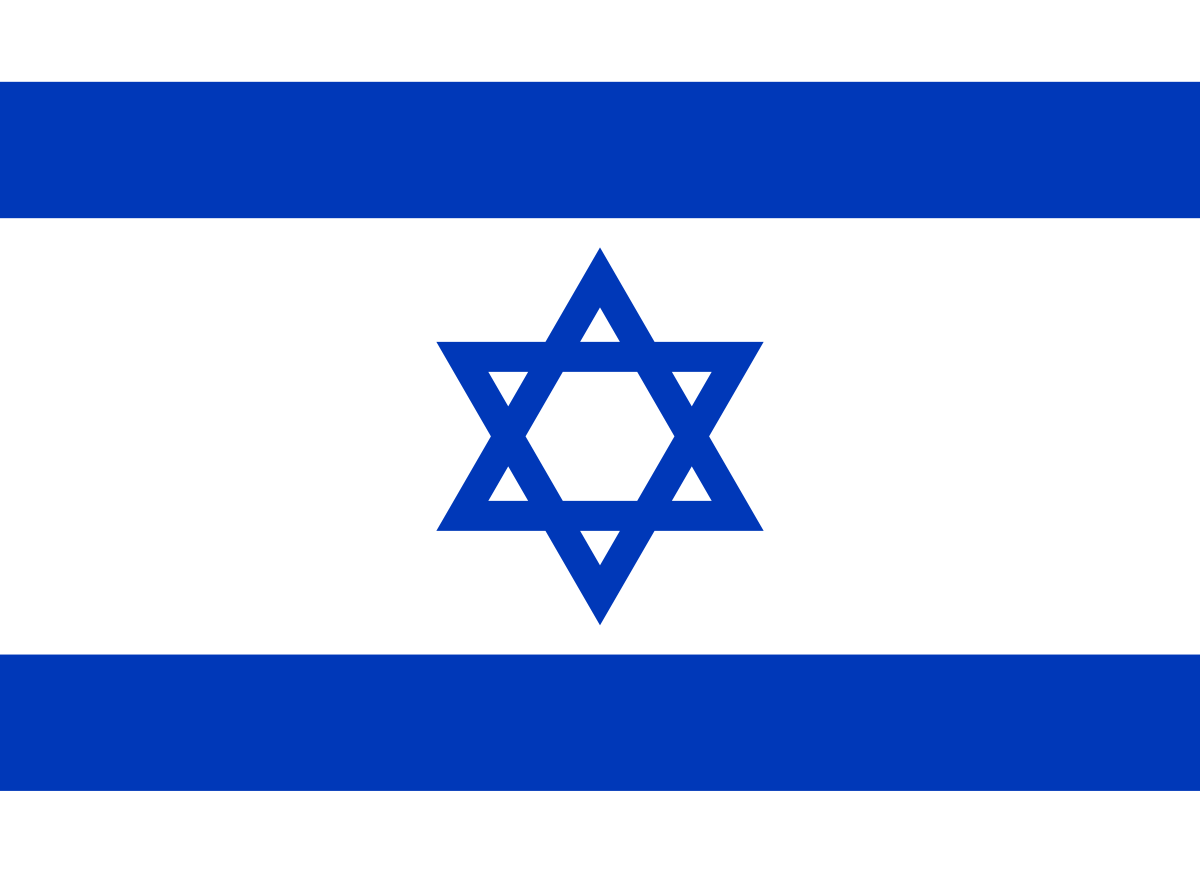}} 
\end{center}
\begin{center}
    \textbf{Data Collected: 2006-2024}
\end{center}
\vfill
\begin{center}
  \fbox{%
    \parbox{\textwidth}{%
      \begin{center}
      \textbf{Important Links}\\
      \href{https://www.boi.org.il/en/}{Central Bank Website}\\
       \href{https://huggingface.co/datasets/gtfintechlab/bank_of_israel}{Annotated Dataset}\\
     \href{https://huggingface.co/gtfintechlab/model_bank_of_israel_stance_label}{Stance Label Model} \\
     \href{https://huggingface.co/gtfintechlab/model_bank_of_israel_time_label}{Time Label Model} \\
     \href{https://huggingface.co/gtfintechlab/model_bank_of_israel_certain_label}{Certain Label Model} \\
      \end{center}
    }
  }
\end{center}

\newpage

\section*{Monetary Policy Mandate} 
The Bank of Israel's main function is to protect the value of the local currency - in other words, to maintain price stability. Maintaining the value of money is important for economic stability and strength, and for creating the conditions necessary for continued growth of output and employment.

\textbf{Mandate Objectives:} 
\begin{itemize}
    \item \textbf{Price Stability:} The Bank of Israel's main function is to protect the value of the local currency.
    \item \textbf{Economic Growth:} Additional functions are to support the attainment of the other goals of the government's economic policy, especially growth, employment, and the narrowing of social gaps, as well as the stability and proper activity of the financial system – all these provided that price stability is not undermined in the long run.
\end{itemize}
\section*{Structure} 

\textbf{Composition: }
\begin{itemize}
    \item \textbf{Officials:} The Governor, who serves as chairperson of the Committee, the Deputy Governor, and an additional Bank employee who is appointed by the Governor. 
    \item \textbf{Public Representatives:} The other three members of the Committee are representatives of the public.
\end{itemize}
\textbf{Meeting Structure: } 
The Monetary Committee meets eight times a year to discuss the current economic outlook and makes rate decisions based on the current price stability and economic growth.
\section*{Manual Annotation} 

\textbf{Annotators} 

\begin{itemize}
    \item Tyson Brown
    \item Matthew Lu
    \item Gilbert Mao
    \item Daniel Wu
\end{itemize}

\textbf{Annotation Agreement} 
The agreement percentage among the pairs of annotators for different labels.
\begin{itemize}
    \item \SD Agreement: 38.7\%
    \item \TC Agreement: 78.2\%
    \item \CE Agreement: 65.6\%
\end{itemize} 

\textbf{Annotation Guide} 

\mptext{boi}{10} Interest Rate, Capital Market Inflation Rate, Total Assets, Gross Domestic Product (GDP), Unemployment Rate, Job Vacancy Rate, NIS to Euro (or Dollar), Inflation, Housing Market, War.

\textbf{Important Factors}
\begin{itemize}
    \item \textit{Interest Rate:} The Bank of Israel has adjusted its benchmark interest rate, influencing the cost of borrowing and impacting economic activity by either stimulating or cooling down demand.
    \item \textit{Capital Market Inflation Rate:} Inflation in Israel's financial markets has been reflected in fluctuations in asset prices, with stock and bond markets responding to macroeconomic conditions and central bank policies.
    \item \textit{Total Assets:} The Bank of Israel's balance sheet has expanded as foreign reserves increased, injecting liquidity into the financial system and affecting overall economic stability.
    \item \textit{GDP:} Israel’s GDP growth has been influenced by the central bank’s monetary policies, balancing inflation control with efforts to sustain economic expansion.
    \item \textit{Unemployment Rate:} The unemployment rate has shifted due to monetary policy measures aimed at either job creation or inflation containment, affecting labor market dynamics.
    \item \textit{Job Vacancy Rate:} Changes in the job vacancy rate indicate shifts in labor demand, reflecting the impact of central bank policies on hiring trends and wage pressures.
    \item \textit{NIS to Euro (or Dollar):} The exchange rate of the Israeli shekel against the Euro and the U.S. dollar has fluctuated, influencing import costs, export competitiveness, and inflation levels.
    \item \textit{Inflation:} The Bank of Israel has implemented monetary policies to stabilize inflation, ensuring that price levels remain within the target range for economic stability.
    \item \textit{Housing Market:} Interest rate adjustments have directly affected mortgage rates, impacting housing affordability and real estate market activity.
    \item \textit{War:} Ongoing geopolitical conflicts have introduced economic uncertainty, affecting market confidence, trade flows, and the central bank’s policy responses.
\end{itemize}

\textbf{Examples: }
\begin{itemize}
    \item ``The Monetary Committee decided to leave the interest rate unchanged, but sees a real possibility of having to raise the interest rate in future decisions.''\\
    \textbf{Hawkish}: This sentence implies that despite the current stance, future policy may tighten, reflecting caution against inflation.
    
    \item ``The Bank of Israel increased bond purchases and foreign currency reserves.''\\
    \textbf{Hawkish}: This example indicates a proactive measure to curb potential overheating of the economy.
    
    \item ``The Committee members agreed that if necessary, the Committee will take additional steps to make monetary policy even more accommodative.''\\
    \textbf{Dovish}: The sentence suggests a readiness to ease monetary conditions in order to support economic growth.
    
    \item ``The Committee decided to sell bonds and securities to reduce the money supply.''\\
    \textbf{Dovish}: By selling bonds, the bank is expanding its balance sheet, a typical sign of an easing or accommodative policy stance.
    
    \item ``It was decided to keep the interest rate unchanged at 0.1 percent.''\\
    \textbf{Neutral}: This sentence reflects a steady policy stance with no movement towards tightening or easing.

    \item ``Other participants in the discussion are the directors of the research and market operations departments, and economists from various departments who prepare and present the material for discussion.''\\
    \textbf{Irrelevant}: This sentence discusses the participants in a discussion rather than focusing on monetary policy or actions that influence monetary policy decisions.
\end{itemize}

\newpage

\begin{longtable}{p{0.118\textwidth}p{0.183\textwidth}p{0.183\textwidth}p{0.183\textwidth}p{0.183\textwidth}}
\caption{\mptitle{Bank of Israel}} \label{tb:boi_mp_stance_guide} \\

\toprule
\textbf{Category} & \textbf{Dovish} & \textbf{Hawkish} & \textbf{Neutral} & \textbf{Irrelevant} \\
\midrule
\endfirsthead

\toprule
\textbf{Category} & \textbf{Dovish} & \textbf{Hawkish} & \textbf{Neutral} & \textbf{Irrelevant} \\
\midrule
\endhead

\textbf{Interest Rate} & 
When interest rates are too high, less money to spend, reducing economic growth. & 
When interest rates are too low, lots of borrowing, reduces value of money. & 
When interest rates remain the same. & 
Sentence is not relevant to monetary policy. \\
\midrule

\textbf{Capital Market Inflation Rate} & 
Likely to accept higher inflation for lower rates. & 
Aim to keep inflation low with tighter policies. & 
Stable inflation without policy shifts. & 
Sentence is not relevant to monetary policy. \\
\midrule

\textbf{Total Assets} & 
Expanding balance sheets, such as selling bonds. & 
Reducing balance sheet to protect against overheating. & 
Maintaining a stable balance sheet. & 
Sentence is not relevant to monetary policy. \\
\midrule

\textbf{GDP} & 
Shrinking GDP indicating need for stimulus. & 
Rising GDP increases inflation/employment. & 
GDP remains steady. & 
Sentence is not relevant to monetary policy. \\
\midrule

\textbf{Unemploy-ment Rate} & 
Focus on lowering unemployment. & 
Willing to tolerate higher unemployment due to inflation. & 
Balancing unemployment and inflation. & 
Sentence is not relevant to monetary policy. \\
\midrule

\textbf{Job Vacancy Rate} & 
Stimulating economy to fill vacancies. & 
High vacancies as wage inflation signal; tighter stance. & 
Monitored but not policy-shaping. & 
Sentence is not relevant to monetary policy. \\
\midrule

\textbf{NIS to Euro (or Dollar)} & 
High NIS makes exports expensive, lowering trade. & 
Higher interest rates to make NIS more attractive. & 
Inflation at target, balanced economy. & 
Sentence is not relevant to monetary policy. \\
\midrule

\textbf{Housing Market} & 
Lower interest rates make mortgages more affordable. & 
Higher rates discourage borrowing, cool prices. & 
Inflation and rates at target levels. & 
Sentence is not relevant to monetary policy. \\
\midrule

\textbf{War} & 
Lower risk perception, lower rates encouraged. & 
Higher risk perception, tighter policy. & 
War impact unchanged. & 
Sentence is not relevant to monetary policy. \\
\bottomrule
\end{longtable}

\fwcertaintytext{boi}

\begin{table*}
\caption{\fwtitle{Bank of Israel}}
\vspace{1em}
\begin{tabular}{p{0.3\textwidth}p{0.3\textwidth}p{0.3\textwidth}}
\toprule
\textbf{Label} & \textbf{Description} & \textbf{Example} \\
\midrule
\textbf{Forward Looking} & Discusses expectations, projections, or anticipations about future actions or events. Key words include "future", "expect", "project", "anticipate". & “The Monetary Committee decided to leave the interest rate unchanged, but sees a real possibility of having to raise it in future decisions.” \\
\midrule
\textbf{Not Forward Looking} & Reflects on recent or past economic data, trends, or events to describe historical or current events. Key words include "past", "historical", "analyzed", "occurred". & “The interest rate can be left at its current level as inflation is expected to converge within the target range beginning in another year.” \\
\bottomrule
\end{tabular}
\label{tb:boi_forward_looking_guide}
\end{table*}

\begin{table*}
\caption{\certaintytitle{Bank of Israel}}
\vspace{1em}
\begin{tabular}{p{0.3\textwidth}p{0.3\textwidth}p{0.3\textwidth}}
\toprule
\textbf{Label} & \textbf{Description} & \textbf{Example} \\
\midrule
\textbf{Certain} & An expectation, trend, action, or outcome that is clearly stated. Key words include "remain", "increased", "decreased", "will continue". & “As expected, the central bank kept the interest rate unchanged at 0.5 percent this month.” \\
\midrule
\textbf{Uncertain} & An unclear or unexplained event or action. Key words include "if", "could", "outlook", "uncertainty". & “An open discussion on monetary policy follows, which ends with a vote on the level of the interest rate.” \\
\bottomrule
\end{tabular}
\label{tb:boi_certainty_guide}
\end{table*}
\clearpage
\usubsection{Bank of Canada}
\begin{center}
    \textbf{Region: Canada}
\end{center}
\begin{center}
    \fbox{\includegraphics[width=0.99\textwidth]{resources/flags/Flag_of_Canada.png}}
\end{center}
\begin{center}
    \textbf{Data Collected: 2005-2024}
\end{center}
\vfill
\begin{center}
  \fbox{%
    \parbox{\textwidth}{%
      \begin{center}
      \textbf{Important Links}\\
      \href{https://www.bankofcanada.ca/}{Central Bank Website}\\
       \href{https://huggingface.co/datasets/gtfintechlab/bank_of_canada}{Annotated Dataset}\\
     \href{https://huggingface.co/gtfintechlab/model_bank_of_canada_stance_label}{Stance Label Model} \\
     \href{https://huggingface.co/gtfintechlab/model_bank_of_canada_time_label}{Time Label Model} \\
     \href{https://huggingface.co/gtfintechlab/model_bank_of_canada_certain_label}{Certain Label Model} \\
      \end{center}
    }
  }
\end{center}

\newpage

\section*{Monetary Policy Mandate} 
The goal of Canada’s monetary policy is to promote the economic and financial well-being of Canadians.

\textbf{Mandate Objectives:} 
\begin{itemize}
    \item \textbf{Controlled Inflation Rate:} Aiming for a low and stable inflation rate at $2\%$ to allow Canadians to make spending and investment decisions confidently. 
\end{itemize}

\section*{Structure}
\textbf{Composition: }\\
\begin{itemize}
    \item \textbf{Governing Council:} The Governing Council is the policy-making body of the Bank. It consists of the Governor, Senior Deputy Governor and the Deputy Governors.
    \item \textbf{Monetary Policy Review Committee:} Composed of Governing Council members and select senior Bank leaders that discuss the challenges and merits of different courses of action.
    \item \textbf{Economics Departments:} Four economics departments, which produce the forecasts and other analyses that feed into the process
\end{itemize}
\textbf{Meeting Structure: } 
 \begin{itemize}
     \item \textbf{Frequency:} Eight times per year on fixed dates to decide the policy interest rate.
 \end{itemize}
\section*{Manual Annotation} 

\textbf{Annotators} 

\begin{itemize}
    \item Erik Larson
    \item Anish Kanduri
    \item Ryan Sheppard
    \item Ojas Kalra
\end{itemize}

\textbf{Annotation Agreement} 
The agreement percentage among the pairs of annotators for different labels.
\begin{itemize}
    \item \SD Agreement: 78\%
    \item \TC Agreement: 84.8\%
    \item \CE Agreement: 80.4\%
\end{itemize} 

\textbf{Annotation Guide} 
\mptext{boc}{eight} Inflation, Employment, Economic Growth, Interest Rates, Monetary Stimulus, Exchange Rates, Commodity Prices, and Fiscal Policy Interaction.

\begin{itemize} 
    \item \emph{Inflation}: A sentence pertaining to the evolution of price levels, encompassing trends of changing inflation rates or pressures on inflation. 
    \item \emph{Employment}: A sentence pertaining to labor market conditions, focusing on unemployment rates, job creation trends, and workforce strength. 
    \item \emph{Economic Growth}: A sentence pertaining to the pace and direction of economic expansion or contraction, including implications for stimulus during slowdowns or overheating during rapid growth. 
    \item \emph{Interest Rates}: A sentence pertaining to interest rate levels, discussing policy decisions aimed at modifying interest rates to address the needs of the economy. 
    \item \emph{Monetary Stimulus}: A sentence pertaining to actions that expand liquidity, such as increased asset purchases. 
    \item \emph{Exchange Rates}: A sentence pertaining to the movement of a nation's currency value or their economic implications. 
    \item \emph{Commodity Prices}: A sentence pertaining to shifts in the prices of Canada's commodities. 
    \item \emph{Fiscal Policy Interaction}: A sentence pertaining to the interactions between government fiscal measures (such as spending and taxation) and monetary policy decisions.
\end{itemize}

\textbf{Examples: } 
\begin{itemize} 
    \item ``Inflation is expected to exceed 3\%, requiring tightening.''\\ 
    \textbf{Hawkish}: This sentence implies that, due to rising inflation above the target, the Bank of Canada may adopt tightening policies to control inflationary pressures.
    
    \item ``Unemployment has fallen significantly, prompting concerns about wage inflation.''\\
    \textbf{Hawkish}: This example indicates that a tight labor market may lead to wage pressures, suggesting that monetary policy could tighten to prevent an overheating economy.
    
    \item ``The central bank cut rates to support the slowing economy.''\\
    \textbf{Dovish}: This sentence states that the Bank of Canada took an accommodative stance to stimulate economic activity during economic slowdowns.
    
    \item ``Weak economic activity signals the need for further monetary stimulus.''\\
    \textbf{Dovish}: This example reflects the view that subdued growth conditions warrant more easing to boost economic performance.
    
    \item ``Interest rates remained steady this month.''\\
    \textbf{Neutral}: This sentence portrays a situation where the policy remains unchanged, reflecting stability without leaning towards tightening or easing.

    \item ``We will take decisions one meeting at a time.'' \\
    \textbf{Irrelevant}: This sentence discusses the way the Bank of Canada plans to make monetary policy decisions, however, it does not explicitly discuss any monetary policy action.
\end{itemize}

\newpage

\begin{longtable}{p{0.118\textwidth}p{0.183\textwidth}p{0.183\textwidth}p{0.183\textwidth}p{0.183\textwidth}}
\caption{Bank of Canada Annotation Guide} \label{tb:boc_mp_stance_guide} \\
\toprule
\textbf{Category} & \textbf{Dovish} & \textbf{Hawkish} & \textbf{Neutral} & \textbf{Irrelevant} \\
\midrule
\endfirsthead

\toprule
\textbf{Category} & \textbf{Dovish} & \textbf{Hawkish} & \textbf{Neutral} & \textbf{Irrelevant} \\
\midrule
\endhead

\textbf{Inflation} & 
Below target or deflation concerns leading to easing. & 
Above target or rising inflation, prompting tightening. & 
Descriptive with no policy implication. & 
Sentence is not relevant to monetary policy. \\
\midrule

\textbf{Employment} & 
High unemployment, weak labor market prompting easing. & 
Low unemployment, risk of overheating, tightening bias. & 
Labor description without policy implication. & 
Sentence is not relevant to monetary policy. \\
\midrule

\textbf{Economic Growth} & 
Slow growth or recession risk, stimulus needed. & 
Strong growth, risk of overheating. & 
Growth data with no implied action. & 
Sentence is not relevant to monetary policy. \\
\midrule

\textbf{Interest Rates} & 
Lowering or keeping rates low to support economy. & 
Raising rates to counter inflation. & 
Rate level mentioned without bias. & 
Sentence is not relevant to monetary policy. \\
\midrule

\textbf{Monetary Stimulus} & 
Increasing liquidity or asset purchases. & 
Reducing purchases or liquidity, signaling tightening. & 
Describes programs neutrally. & 
Sentence is not relevant to monetary policy. \\
\midrule

\textbf{Exchange Rates} & 
Depreciation prompts easing to support growth. & 
Appreciation prompts concern about overheating. & 
Neutral exchange rate statement. & 
Sentence is not relevant to monetary policy. \\
\midrule

\textbf{Commodity Prices} & 
Falling prices suggest weak demand, needing easing. & 
Rising prices suggest inflation, tightening needed. & 
Price movement with no policy stance. & 
Sentence is not relevant to monetary policy. \\
\midrule

\textbf{Fiscal Policy Interaction} & 
Monetary policy needed to support lagging fiscal stimulus. & 
Strong fiscal stimulus might require tightening. & 
Neutral fiscal commentary. & 
Sentence is not relevant to monetary policy. \\
\bottomrule
\end{longtable}

\fwcertaintytext{boc} 
\newpage

\begin{table*}
\caption{\fwtitle{Bank of Canada}}
\vspace{1em}
\begin{tabular}{p{0.3\textwidth}p{0.3\textwidth}p{0.3\textwidth}}
\toprule
\textbf{Label} & \textbf{Description} & \textbf{Example} \\
\midrule
\textbf{Forward Looking} & Sentences that predict future economic conditions, policy actions, or outcomes. Key phrases include ``expected to,'' ``forecast,'' ``anticipated to occur.'' & ``Government spending is not projected to contribute to growth in 2012.'' \\
\midrule
\textbf{Not Forward Looking} & Sentences referencing past or present economic data, actions, or outcomes. These include descriptions of historical economic data, past policy actions, or statement discussing recent trends. & ``Recent data show the Canadian economy is returning to potential growth.'' \\
\bottomrule
\end{tabular}
\label{tb:boc_forward_looking_guide}
\end{table*}

\begin{table*}
\caption{\certaintytitle{Bank fo Canada}}
\vspace{1em}
\begin{tabular}{p{0.3\textwidth}p{0.3\textwidth}p{0.3\textwidth}}
\toprule
\textbf{Label} & \textbf{Description} & \textbf{Example} \\
\midrule
\textbf{Certain} & Sentences presenting definitive statements or outcomes with high confidence. Key phrases include ``is,'' ``has been,'' ``will be.'' & ``Inflation expectations remain well-anchored.'' \\
\midrule
\textbf{Uncertain} & Sentences suggesting variability, unpredictability, or tentative outcomes. Key phrases include ``could,'' ``might,'' ``expected,'' ``may.'' & ``Economic growth might slow next quarter depending on global market conditions.'' \\
\bottomrule
\end{tabular}
\label{tb:boc_certainty_guide}
\end{table*}
\clearpage
\usubsection{ Bank of Mexico}
\begin{center}
    \textbf{Region: Mexico}
\end{center}
\begin{center}
    \fbox{\includegraphics[width=0.99\textwidth]{resources/flags/Flag_of_Mexico.png}} 
\end{center}
\begin{center}
    \textbf{Data Collected: 2018-2024}
\end{center}
\vfill
\begin{center}
  \fbox{%
    \parbox{\textwidth}{%
      \begin{center}
      \textbf{Important Links}\\
      \href{https://www.banxico.org.mx/indexen.html}{Central Bank Website}\\
       \href{https://huggingface.co/datasets/gtfintechlab/bank_of_mexico}{Annotated Dataset}\\
     \href{https://huggingface.co/gtfintechlab/model_bank_of_mexico_stance_label}{Stance Label Model} \\
     \href{https://huggingface.co/gtfintechlab/model_bank_of_mexico_time_label}{Time Label Model} \\
     \href{https://huggingface.co/gtfintechlab/model_bank_of_mexico_certain_label}{Certain Label Model} \\
      \end{center}
    }
  }
\end{center}

\newpage

\section*{Monetary Policy Mandate} 
The Banco de Mexico aims to control the value of Mexico's currency and improve the wellbeing of its citizens.

\textbf{Mandate Objectives:}  
\begin{itemize}
    \item \textbf{Stable currency rate:} The main mandate of the Banco de Mexico is to maintain a stable value for the value of Mexico's currency in the long term.
\end{itemize}
\section*{Structure} 
\textbf{Composition: }
\begin{itemize}
    \item \textbf{Governor: } There is one governor on the Board, appointed by the President and confirmed by the Senate. The Governor has a six-year term, starting on the first of January of the fourth year of each presidential term. 
    \item \textbf{Deputy Governors: } There are four deputy governors in the board, appointed by the President and confirmed by the Senate. The Deputy Governors have an eight-year period. To be replaced or reappointed every two years, on the first, third, and fifth years of each presidential term.
\end{itemize}
\textbf{Meeting Structure: } 
 \begin{itemize}
     \item \textbf{Frequency:}  The Board of Governors meets eight times annually on prefixed dates.
 \end{itemize}
\section*{Manual Annotation} 

\textbf{Annotators} 
\begin{itemize}
    \item Arnav Tripathi
    \item Chittebbayi Penugonda
    \item Christian Lindler
    \item Surya Subramanian
\end{itemize}

\textbf{Annotation Agreement} 

The agreement percentage among the pairs of annotators for different labels.
\begin{itemize}
    \item \SD Agreement: 57.5\%
    \item \TC Agreement: 77.8\%
    \item \CE Agreement: 63.3\%
\end{itemize} 
\textbf{Annotation Guide}

\mptext{bom}{7} Peso Value Change, Trade Activity, Labor, Key Words, Economic Growth,  Market Sentiment, Energy/House Prices.

\begin{itemize}
    \item \textit{Peso Value Change:}  
    Sentences corresponding to peso value changes such as inflation and/or foreign market appreciation/depreciation. A weaker peso may be seen as dovish, especially when Banco de México is attempting to support exports by making Mexican goods cheaper in foreign markets. A stronger peso helps control inflation by making imports cheaper, but it can hurt export competitiveness.

    \item \textit{Trade Activity:}  
    Sentences pertaining to the import/export ratio—an increase in one or decrease of another. Banco de México favors policies that stimulate exports, especially given Mexico’s strong manufacturing sector and trade agreements (such as USMCA). A focus on export growth could support job creation and help maintain balance in current account deficits.

    \item \textit{Labor:}  
    Sentences mentioning statistics about the increase or decrease in productivity. If there are concerns about declining productivity—especially in key industries in Mexico like oil, manufacturing, or agriculture—Banco de México might lean hawkish, signaling the need to cool off the economy to address inefficiencies.

    \item \textit{Key Words:}  
    Sentences that have keywords that make a clear, strong indication towards classifications.


    \item \textit{Economic Growth:}  
    Sentences conveying the performance in economic growth the bank wishes to see in the near future. Given Mexico’s economic structure, fostering growth in sectors like manufacturing, tourism, and remittances can be seen as a dovish move aimed at reducing poverty and unemployment.


    \item \textit{Market Sentiment:}  
    Sentences mentioning what the outlook is for the market and how investors are acting.

    \item \textit{Energy/House Prices:}  
    Sentences that mention anything about housing price levels and energy price levels. Given Mexico’s reliance on oil and gas revenues, a dovish outlook might stem from efforts to lower energy prices domestically, encouraging consumption and reducing inflationary pressure on households.


\end{itemize}

\textbf{Examples:}

\begin{itemize}
    \item ``Another member noted that non-core inflation has been below forecasts.''\\ 
    \textbf{Dovish}: This sentence states that non-core inflation is lower than expected, possibly opening up opportunities for expansionary monetary policy.
    
    \item ``However, another member specified that breakeven inflation still remains at levels above 4\% for all terms.''\\ 
    \textbf{Hawkish}: This sentence highlights that breakeven inflation exceeds 4\% across all terms, hinting at potential future tightening to counter inflation concerns.
    
    \item ``In this context, some members noted that the confidence of economic agents decreased.''\\ 
    \textbf{Dovish}: This sentence suggests that a decline in economic confidence might encourage a more supportive monetary policy approach to improve confidence.

    \item ``Another member stated that the dynamics within the components of inflation has led to an upward adjustment of forecasts.''\\ 
    \textbf{Hawkish}: This sentence asserts that specific dynamics have already resulted in a revision upward of forecasts, requiring contractionary monetary policy to address inflation concerns.
    
    \item ``Some members considered that the economy's cyclical position have not changed significantly since the previous monetary policy decision.''\\ 
    \textbf{Neutral}: This sentence indicates that the economy’s cyclical stance remains stable, implying no immediate change in policy direction.
    
    \item ``One member added that the bank of Canada is also preparing to do so.''\\ 
    \textbf{Irrelevant}: This sentence refers to potential preparatory actions by the Bank of Canada without directly addressing a monetary policy decision or stance.
\end{itemize}

\newpage

\begin{longtable}{p{0.118\textwidth}p{0.183\textwidth}p{0.183\textwidth}p{0.183\textwidth}p{0.183\textwidth}}
\caption{\mptitle{Bank of Mexico}} \label{tb:bom_mp_stance_guide} \\
\toprule
\textbf{Category} & \textbf{Dovish} & \textbf{Hawkish} & \textbf{Neutral} & \textbf{Irrelevant} \\
\midrule
\endfirsthead

\toprule
\textbf{Category} & \textbf{Dovish} & \textbf{Hawkish} & \textbf{Neutral} & \textbf{Irrelevant} \\
\midrule
\endhead
\textbf{Peso Value Change} & Peso appreciates, leading to lower inflation. & Peso depreciates, leading to higher inflation. & N/A & Sentence is not relevant to monetary policy. \\
\midrule
\textbf{Trade Activity} & Favors decreased imports and increased exports, supporting domestic growth. & Favors increased imports and decreased exports, raising dependency on foreign markets. & N/A & Sentence is not relevant to monetary policy. \\
\midrule
\textbf{Labor} & Favors increased labor productivity, improving economic output. & Favors lower productivity, signaling economic slowdown. & Labor market remains stable, no notable changes. & Sentence is not relevant to monetary policy. \\
\midrule
\textbf{Key Words} & Words like “Accommodative”, “Expansionary”, “Easing”, indicating support for stimulus. & Words like “Tighter”, “Restrictive”, “Balance”, indicating a shift toward tightening. & Words like “Mixed”, “Moderate”, signaling a neutral stance. & Sentence is not relevant to monetary policy. \\
\midrule
\textbf{Economic Growth} & Supports expansion, higher consumer spending, and investment. & Advocates policies leading to economic slowdown or controlled growth. & Economy remains balanced, no significant shifts in output. & Sentence is not relevant to monetary policy. \\
\midrule
\textbf{Market Sentiment} & Bullish outlook, rising investor confidence in the economy. & Bearish outlook, increased caution among investors. & Neutral market stance, adopting a wait-and-see approach. & Sentence is not relevant to monetary policy. \\
\midrule
\textbf{Energy/ House Prices} & Rising property values and lower energy costs, benefiting consumers. & Housing market slowdown, rising energy costs, increasing financial burden. & N/A & Sentence is not relevant to monetary policy. \\
\bottomrule
\end{longtable}

\fwcertaintytext{bom}

\newpage
\begin{table*}
\caption{\fwtitle{Bank of Mexico}}
\vspace{1em}
\begin{tabular}{p{0.3\textwidth}p{0.3\textwidth}p{0.3\textwidth}}
\toprule
\textbf{Label} & \textbf{Description} & \textbf{Example}\\
\midrule
\textbf{Forward Looking} & Focusing on future. Key words include “expect,” “anticipate,” and “project.” & “We expect inflation to decline over the next quarter.”\\
\midrule
\textbf{Not Forward Looking} & Focusing on past. Key words include “remained” and “over past X years/months.” & “Inflation remained stable over the past year.”\\
\bottomrule
\end{tabular}
\label{tb:bom_forward_looking_guide}
\end{table*}

\begin{table*}
\caption{\certaintytitle{Bank of Mexico}}
\vspace{1em}
\begin{tabular}{p{0.3\textwidth}p{0.3\textwidth}p{0.3\textwidth}}
\toprule
\textbf{Label} & \textbf{Description} & \textbf{Example}\\
\midrule
\textbf{Certain} & Sentiment is sure. Key words include “reported,” “show,” and “according to.” & “Reports confirm that economic growth accelerated this quarter.”\\
\midrule
\textbf{Uncertain} & Sentiment is unsure. Key words include “likely,” “estimate,” and “projected.” & “It is estimated that inflation may rise in the next period.”\\
\bottomrule
\end{tabular}
\label{tb:bom_certainty_guide}
\end{table*}

\clearpage
\usubsection{ Narodowy Bank Polski}
\begin{center}
    \textbf{Region: Poland}
\end{center}
\begin{center}
    \fbox{\includegraphics[width=0.99\textwidth]{resources/flags/Flag_of_Poland.png}}
\end{center}

\begin{center}
    \textbf{Data Collected: 2007-2024}
\end{center}
\vfill
\begin{center}
  \fbox{%
    \parbox{\textwidth}{%
      \begin{center}
      \textbf{Important Links}\\
      \href{https://nbp.pl/en/}{Central Bank Website}\\
       \href{https://huggingface.co/datasets/gtfintechlab/national_bank_of_poland}{Annotated Dataset}\\
     \href{https://huggingface.co/gtfintechlab/model_national_bank_of_poland_stance_label}{Stance Label Model} \\
     \href{https://huggingface.co/gtfintechlab/model_national_bank_of_poland_stance_label}{Time Label Model} \\
     \href{https://huggingface.co/gtfintechlab/model_national_bank_of_poland_certain_label}{Certain Label Model} \\
      \end{center}
    }
  }
\end{center}

\newpage

\section*{Monetary Policy Mandate} 
The MPC is responsible for formulating Poland's monetary policy to achieve the primary object of maintaining price stability.

\textbf{Mandate Objectives:} 
\begin{itemize}
    \item \textbf{Price Stability}: Aiming for a 2.5\% annual inflation rate, as measured by annual change in the consumer price index, with a symmetric band for deviations of $\pm$1\% for the medium term. 
\end{itemize}

\section*{Structure} 
The MPC is integral to the National Bank of Poland as they set the economy's monetary policy by influencing interest rates and the money supply.

\textbf{Composition: }

\begin{itemize}
    \item \textbf{Chairperson:} The Governor of the NBP. They are appointed by the Sejm (the lower house of Poland's parliament) and serve a 6-year term.
    \item \textbf{Committee Members}: Nine members appointed in equal proportions by the President, Senate, and the Sejm. They serve a 6-year term.
\end{itemize}

\textbf{Meeting Structure: } 
\begin{itemize}
    \item \textbf{Frequency:} The MPC typically meets once a month, with each meeting lasting two days. 
    \item \textbf{Non-decision-making Meetings:} The MPC also holds non-decision-making meetings to analyze the economy and explore policy options.
    \item \textbf{Additional Meetings:} Held as needed to address urgent economic developments. 
\end{itemize}

\section*{Manual Annotation} 

\textbf{Annotators} 
\begin{itemize}
    \item Saketh Budideti
    \item Huzaifa Pardawala
    \item Harsit Mittal
    \item Rachel Yuh
\end{itemize}

\textbf{Annotation Agreement} 
The agreement percentage among the pairs of annotators for different labels.
\begin{itemize}
    \item \SD Agreement: 75.8\%
    \item \TC Agreement: 86\%
    \item \CE Agreement: 81.3\%
\end{itemize} 

\textbf{Annotation Guide} 
\mptext{nbp}{13}Interest Rates, Inflation Rate, GDP Growth, Unemployment Rate, Required Reserve Ratio, Money Supply, Foreign Exchange Reserves, Private Consumption, Prices, Lending, Domestic Manufacturing, Gold Holdings, Net Exports.

\begin{itemize}
    \item \emph{Interest Rates}: A sentence pertaining to various interest rates used by the NBP. This includes, but is not limited to, the reference rate, Lombard rate, deposit rate, discount rate, and inflation rate. 
    \item \emph{Inflation Rate}: A sentence pertaining to the inflation rate (the change in the consumer price index).
    \item \emph{GDP Growth}: A sentence pertaining to the GDP growth of Poland.
    \item \emph{Unemployment Rate}: A sentence pertaining to the unemployment rate of Poland.
    \item \emph{Required Reserve Ratio}: A sentence pertaining to the required reserve ratio, that is, the minimum percentage of reservable liabilities that depository institutions (e.g. banks) must hold at any given time.
    \item \emph{Money Supply}: A sentence that discusses the impact to money supply or changes in demand.
    \item \emph{Foreign Exchange Reserves}: A sentence pertaining to foreign assets controlled by the NBP.
    \item \emph{Private Consumption}: A sentence pertaining to private consumption in Poland.
    \item \emph{Prices}: A sentence pertaining to a change in prices of goods and services in Poland.
    \item \emph{Lending}: A sentence that discusses changes in lending behaviors of actors such as corporations.
    \item \emph{Domestic Manufacturing}: A sentence discusses the growth or output of the manufacturing industry in Poland.
    \item \emph{Gold Holdings}: A sentence pertaining to the gold holdings of the NBP.
    \item \emph{Net Exports}: A sentence discussion the net exports of Poland.
\end{itemize}

\textbf{Examples: }
\begin{itemize}
    \item ``The Council judged that the current level of the NBP interest rates was conducive to meeting the NBP inflation target in the medium term''\\
    \textbf{Neutral}: Interested rates are expected to meeting the inflation target, signaling that no change should be made.
    
    \item ``Some Council members pointed to a continually high rate of wage growth and to a very fast drop in unemployment, facilitating a further strong rise in wages.''\\
    \textbf{Hawkish}: High wage growth and low unemployment indicate wage-driven inflation, requiring tighter monetary policy to combat.

    \item ``Some Council members stressed that in the event of significant price growth that would jeopardize meeting the inflation target in the medium term, it might be justified to consider an increase in the NBP interest rates in the coming quarters.''\\
    \textbf{Hawkish}: The emphasis on potential interest rates increases to address price growth signals commitment to tighten monetary policy amid inflation risks.
    
    \item ``These meeting participants also assessed that the acceleration of interest rate increases might not reduce inflation expectations of households, which are highly adaptive in nature.''\\
    \textbf{Dovish}: Participants questions the effectiveness of rate hikes in lowering inflation expectations, suggesting a looser monetary policy stance.

    \item ``In effect, it was judged that the annual GDP growth in 2022 Q4 slowed down markedly for another consecutive quarter.''\\
    \textbf{Dovish}: Slowing GDP growth indicates a need for easing to stimulate economic momentum.

    \item ``The discussion at the meeting focused on the outlook for economic growth abroad and in Poland, fiscal policy, zloty exchange rate developments and the situation in the credit market and the banking sector.''\\
    \textbf{Irrelevant}: The sentence discusses the topics of the meeting, indicating no monetary policy stance.
\end{itemize}

\newpage

\begin{longtable}{p{0.118\textwidth}p{0.183\textwidth}p{0.183\textwidth}p{0.183\textwidth}p{0.183\textwidth}}
\caption{\mptitle{Narodowy Bank Polski}} \label{tb:nbp_mp_stance_guide} \\
\toprule
\textbf{Category} & \textbf{Hawkish} & \textbf{Dovish} & \textbf{Neutral} & \textbf{Irrelevant} \\
\midrule
\endfirsthead

\toprule
\textbf{Category} & \textbf{Hawkish} & \textbf{Dovish} & \textbf{Neutral} & \textbf{Irrelevant} \\
\midrule
\endhead
\textbf{Interest Rates} & When interest rates decrease. & When interest rates increase. & When interest rates are unchanged, maintained, or sustained. & Sentence is not relevant to monetary policy. \\
\midrule
\textbf{Inflation Rate} & When inflation is below target. & When inflation is above target. & When inflation is at target. & Sentence is not relevant to monetary policy. \\
\midrule
\textbf{GDP Growth} & Slowing GDP growth. & When GDP growth is at a sustained increase above potential. & Stable GDP growth. & Sentence is not relevant to monetary policy. \\
\midrule
\textbf{Unemploy-ment Rate} & Increase in unemployment rates. & Decrease in unemployment rates. & Stable unemployment rates. & Sentence is not relevant to monetary policy. \\
\midrule
\textbf{Required Reserve Ratio} & Decrease in required reserve ratio. & Increase in required reserve ratio. & Maintaining the required reserve ratio. & Sentence is not relevant to monetary policy. \\
\midrule
\textbf{Money Supply} & Low money supply or increased demand for loans. & Money supply is high, increased demand for goods, low demand for loans. & Money supply is stable. & Sentence is not relevant to monetary policy. \\
\midrule
\textbf{Foreign Exchange Reserves} & Increasing reserves, indicating boosting market confidence. & Decreasing reserves. & Maintained reserves. & Sentence is not relevant to monetary policy. \\
\midrule
\textbf{Private Consumption} & Decrease in consumption due to weak consumer sentiment. & Increase in consumption due to rising wages or government spending. & Sustained private consumption. & Sentence is not relevant to monetary policy. \\
\midrule
\textbf{Prices} & Falling prices indicating reduced inflation. & Rising prices indicating increased inflation. & Maintaining existing price levels. & Sentence is not relevant to monetary policy. \\
\midrule
\textbf{Corporate Lending} & Decline in lending, low investment. & Rapid increase in corporate borrowing. & Maintaining existing lending levels. & Sentence is not relevant to monetary policy. \\
\midrule
\textbf{Domestic Manufacturing} & Weak domestic manufacturing output or growth. & Significant domestic manufacturing output or growth. & Stable domestic manufacturing output and growth. & Sentence is not relevant to monetary policy. \\
\midrule
\textbf{Gold Holdings} & Falling gold holdings indicating flexibility in monetary policy. & Rising gold holdings indicating tighter monetary policy. & Maintained gold holdings. & Sentence is not relevant to monetary policy. \\
\midrule
\textbf{Net Exports} & Low net exports indicative of low demand or uncompetitive industries. & High net exports, indicating higher risk of inflation. & Maintained net export levels. & Sentence is not relevant to monetary policy. \\
\bottomrule
\end{longtable}

\fwcertaintytext{nbp}
\newpage
\begin{table*}
\caption{\fwtitle{Narodowy Bank Polski}}
\vspace{1em}

\begin{tabular}{p{0.3\textwidth}p{0.3\textwidth}p{0.3\textwidth}}
\toprule
\textbf{Label} & \textbf{Description} & \textbf{Example}\\
\midrule
\textbf{Forward Looking} & text describing events anticipated to occur in the future. Sentences often utilize language such as “expected” or “projected.” & “The Council underscored that data incoming in 2023 Q3 signalled a further deterioration in economic conditions in the euro area, including in Germany.” \\
\midrule
\textbf{Not Forward Looking} & text referring to past or recent economic policy, data, trends, or events. & “In this context, it was noted that corporate lending, including for investments, had rebounded.” \\
\bottomrule
\end{tabular}
\label{tb:nbp_forward_looking_guide}
\end{table*}

\begin{table*}
\caption{\certaintytitle{Narodowy Bank Polski}}
\vspace{1em}

\begin{tabular}{p{0.3\textwidth}p{0.3\textwidth}p{0.3\textwidth}}
\toprule
\textbf{Label} & \textbf{Description} & \textbf{Example}\\
\midrule
\textbf{Certain} & The sentence reflects a clear, firm, or definitive
decision or expectation about actions or outcomes without ambiguity. & “The Council decided to increase the NBP interest rates to the following levels reference rate to 4.75, lombard rate to 6.25, deposit rate to 3.25, rediscount rate to 5.00.” \\
\midrule
\textbf{Uncertain} & The sentence expresses potential risks, varying opinions, or uncertainties about events, outcomes, or actions, often using words like “could,” “might,” or “may.” & “These meeting participants also assessed that the acceleration of interest rate increases might not reduce inflation expectations of households, which are highly adaptive in nature.” \\
\bottomrule
\end{tabular}
\label{tb:nbp_certainty_guide}
\end{table*}

\clearpage
\usubsection{ Central Bank of Turkey}
\begin{center}
    \textbf{Region: Turkey}
\end{center}
\begin{center}
    \fbox{\includegraphics[width=0.99\textwidth]{resources/flags/Flag_of_Turkey.png}} 
\end{center}
\begin{center}
    \textbf{Data Collected: 2015-2024}
\end{center}
\vfill
\begin{center}
  \fbox{%
    \parbox{\textwidth}{%
      \begin{center}
      \textbf{Important Links}\\
      \href{http://www.tcmb.gov.tr/}{Central Bank Website}\\
       \href{https://huggingface.co/datasets/gtfintechlab/central_bank_republic_of_turkey}{Annotated Dataset}\\
     \href{https://huggingface.co/gtfintechlab/model_central_bank_republic_of_turkey_stance_label}{Stance Label Model} \\
     \href{https://huggingface.co/gtfintechlab/model_central_bank_republic_of_turkey_time_label}{Time Label Model} \\
     \href{https://huggingface.co/gtfintechlab/model_central_bank_republic_of_turkey_certain_label}{Certain Label Model} \\
      \end{center}
    }
  }
\end{center}

\newpage

\section*{Monetary Policy Mandate} 
The Central Republic Bank of Turkey (CBRT) is responsible for formulating Türkiye’s monetary policy to achieve the primary objective of maintaining price stability.

\textbf{Mandate Objectives:} 
\begin{itemize}
    \item \textbf{Price Stability}: The primary goal of the CBRT is to achieve and maintain price stability, using an inflation targeting regime to guide monetary policy.
    \item \textbf{Inflation Targeting}: Since 2006, the CBRT has implemented an explicit inflation targeting framework, setting annual inflation targets in coordination with the government, with an uncertainty band to account for fluctuations.
    \item \textbf{Financial Stability}: Post-2010, the CBRT enhanced its monetary policy to also safeguard financial stability, introducing tools like required reserves and the Reserve Options Mechanism to mitigate external shocks.
\end{itemize}

\section*{Structure} 
The Monetary Policy Committee (MPC) is responsible for setting the economy’s monetary policy by determining interest rates and guiding liquidity management, making it key to the CRBT.

\textbf{Composition: }

\begin{itemize}
    \item \textbf{Governor:} The Governor of the CBRT, who also serves as the chair of the MPC. The Governor is appointed by the President of Türkiye.
    \item \textbf{Committee Members:} The MPC consists of the Governor, the Deputy Governors of the CBRT, and additional members appointed by the President.
\end{itemize}

\textbf{Meeting Structure: } 

\begin{itemize}
    \item \textbf{Frequency}: At least eight times a year according to a pre-announced schedule.
    \item \textbf{Regular Reports}: The CBRT publishes a Monthly Price Developments Report following inflation data releases and a Financial Stability Report analyzing the financial system.
    \item \textbf{Non decision-making meetings}: The committee holds discussions on economic developments and policy options outside of formal decision-making sessions.
\end{itemize}

\section*{Manual Annotation} 

\textbf{Annotators} 
\begin{itemize}
    \item Veer Guda
    \item Tarun Mandapati
    \item Tamas Mester
    \item Alexander Wang
\end{itemize}

\textbf{Annotation Agreement} 
The agreement percentage among the pairs of annotators for different labels.
\begin{itemize}
    \item \SD Agreement: 54.2\%
    \item \TC Agreement: 88.2\%
    \item \CE Agreement: 89.5\%
\end{itemize} 
\textbf{Annotation Guide} 

\mptext{cbrt}{6} Economic Status, Currency Exchange Rate (Turkish Lira/TRY), Inflation Rates, Interest Rates, Government Debt \& Fiscal Policy, and Foreign Trade Balance.

\begin{itemize}
    \item \textit{Economic Status}: Represents the overall condition of the Turkish economy, including inflation, unemployment, and growth trends.
    \item \textit{Currency Exchange Rate (Turkish Lira - TRY)}: Fluctuations in the value of the Turkish lira relative to other currencies, influencing trade and inflation.
    \item \textit{Inflation Rates}: Measures the rate of price increases for goods and services in Turkey, impacting purchasing power and policy decisions.
    \item \textit{Interest Rate}s: The CBRT's benchmark policy rate, affecting borrowing costs, investment, and overall economic activity.
    \item \textit{Government Debt \& Fiscal Policy}: Represents Turkey’s government borrowing levels and fiscal management, influencing monetary policy choices.
    \item \textit{Foreign Trade Balance}: Measures Turkey’s exports and imports, with trade surpluses or deficits impacting currency strength and policy decisions.
\end{itemize}

\textbf{Examples:}
\begin{itemize}
    \item ``Thus, the unemployment rate declined by 0.4 points compared to previous month and realized as 9.6\%.''\\ 
    \textbf{Hawkish}: The sentence reports a decline in unemployment, potentially justifying a more hawkish approach to curb inflation.

    \item ``Thus, the rise in automobile prices amounted to 34.01\% in the first six months of 2023.''\\ 
    \textbf{Hawkish}: This sentence highlights a significant increase in automobile prices, suggesting potential inflationary pressures that may prompt a tighter monetary policy stance.
    
    \item ``The committee has therefore decided to keep the interest rates at current levels.''\\ 
    \textbf{Neutral}: The sentence simply states that interest rates will remain unchanged, reflecting no immediate bias toward tightening or easing policy.
    
    \item ``Annual inflation remained flat in rents, but decreased in other subgroups.''\\ 
    \textbf{Dovish}: With inflation showing stability in rents and declines elsewhere, the overall price pressure appears subdued, which may support expansionary monetary policy.
    
    \item ``In sum, economic activity is expected to recede in the third quarter.''\\ 
    \textbf{Dovish}: This forward Looking forecast of a slowdown in economic activity points to the possibility of an easing monetary stance to promote economic growth.
    
    \item ``Impact analyses of the CBRT's regulations are conducted for all components of the framework by assessing their impact on inflation, interest rates, exchange rates, reserves, expectations, securities, and financial stability with a holistic approach.''\\ 
    \textbf{Irrelevant}: The sentence focuses on a comprehensive evaluation process of regulatory impacts without providing any specific insight into monetary policy stance.
\end{itemize}

\newpage

\begin{longtable}{p{0.118\textwidth}p{0.183\textwidth}p{0.183\textwidth}p{0.183\textwidth}p{0.183\textwidth}}
\caption{\mptitle{Central Bank of Turkey}}
\label{tb:cbrt_mp_stance_guide} \\
\toprule
\textbf{Category} & \textbf{Dovish} & \textbf{Hawkish} & \textbf{Neutral} & \textbf{Irrelevant} \\
\midrule
\endfirsthead
\toprule
\textbf{Category} & \textbf{Dovish} & \textbf{Hawkish} & \textbf{Neutral} & \textbf{Irrelevant} \\
\midrule
\endhead
\textbf{Economic Status} & When inflation is decreasing, unemployment is rising, or economic growth is weak. & When inflation is increasing, unemployment is low, and economic growth is strong. & When economic indicators remain stable. & Sentence is not relevant to monetary policy. \\
\midrule
\textbf{Currency Exchange Rate (Turkish Lira/TRY)} & When the lira depreciates significantly, increasing import costs and inflationary pressure. & When the lira appreciates, reducing import costs and signaling potential monetary tightening. & When the exchange rate remains stable. & Sentence is not relevant to monetary policy. \\
\midrule
\textbf{Inflation Rate} & When inflation is slowing down, suggesting room for rate cuts. & When inflation is rising, requiring tighter monetary policy. & When inflation remains within the target range. & Sentence is not relevant to monetary policy. \\
\midrule
\textbf{Money Supply} & Low money supply, any increase in M2, demand for loans is higher. & High money supply, demand for goods increase, demand for loans decrease. & N/A. & Sentence is not relevant to monetary policy. \\
\midrule
\textbf{Government Debt} & When government debt increases significantly, prompting expectations of monetary stimulus to manage borrowing costs. & When debt decreases, indicating fiscal discipline and allowing tighter monetary policy. & When debt levels remain stable. & Sentence is not relevant to monetary policy. \\
\midrule
\textbf{Foreign Trade Balance} & When the trade deficit widens, indicating higher imports than exports, potentially justifying a weaker lira and looser monetary policy. & When the trade deficit narrows or a surplus is recorded, strengthening the lira and allowing for tighter policy. & When the trade balance remains relatively unchanged. & Sentence is not relevant to monetary policy. \\
\midrule
\textbf{Key Words and Phrases} & When the sentiment is growth oriented, e.g., "accommodative," "maximum employment," "price stability," "heightened consumption," "low inflation." & When sentiment is restrictive, indicating a focus on "price stability" and "sustained growth." & Use of neutral terms, e.g., "mixed," "moderate," "reaffirmed." & Sentence is not relevant to monetary policy. \\
\bottomrule
\end{longtable}

\fwcertaintytext{cbrt}

\newpage

\begin{table*}
\caption{\fwtitle{Central Bank of Turkey}}
\vspace{1em}
\begin{tabular}{p{0.3\textwidth}p{0.3\textwidth}p{0.3\textwidth}}
\toprule
\textbf{Label} & \textbf{Description} & \textbf{Example}\\
\midrule
\textbf{Forward Looking} & Statements describing anticipated economic developments, monetary policy actions, or financial trends. Often includes phrases like “expected,” “projected,” or “forecasted.” & “The CBRT expects inflation to moderate in the second half of the year as the impact of recent monetary tightening measures takes hold.” \\
\midrule
\textbf{Not Forward Looking} & Statements referring to past or current economic trends, monetary decisions, or historical data points. & “Inflation in the previous quarter rose to 70.3\%, largely driven by food and energy price hikes.” \\
\bottomrule
\end{tabular}
\label{tb:cbrt_forward_looking_guide}
\end{table*}

\begin{table*}
\caption{\certaintytitle{Central Bank of Turkey}}
\vspace{1em}
\begin{tabular}{p{0.3\textwidth}p{0.3\textwidth}p{0.3\textwidth}}
\toprule
\textbf{Label} & \textbf{Description} & \textbf{Example}\\
\midrule
\textbf{Certain} & The sentence expresses a firm, definitive decision or expectation about monetary policy, economic trends, or financial markets. Often avoids speculative language. & “The CBRT has decided to raise the policy rate from 42.5\% to 45\% to counter inflationary pressures.” \\
\midrule
\textbf{Uncertain} & The sentence reflects potential risks, varying opinions, or uncertainties about future monetary policy or economic conditions, using words such as “may,” “could,” or “might.” & “The Monetary Policy Committee noted that further rate hikes might be necessary if inflation does not show signs of stabilizing.” \\
\bottomrule
\end{tabular}
\label{tb:cbrt_certainty_guide}
\end{table*}

\clearpage
\usubsection{ Bank of Thailand}
\begin{center}
    \textbf{Region: Thailand}
\end{center}
\begin{center}
    \fbox{\includegraphics[width=0.99\textwidth]{resources/flags/Flag_of_Thailand.png}}
\end{center}
\begin{center}
    \textbf{Data Collected: 2011-2024}
\end{center}
\vfill
\begin{center}
  \fbox{%
    \parbox{\textwidth}{%
      \begin{center}
      \textbf{Important Links}\\
      \href{https://www.bot.or.th/en/home.html}{Central Bank Website}\\
       \href{https://huggingface.co/datasets/gtfintechlab/bank_of_thailand}{Annotated Dataset}\\
     \href{https://huggingface.co/gtfintechlab/model_bank_of_thailand_stance_label}{Stance Label Model} \\
     \href{https://huggingface.co/gtfintechlab/model_bank_of_thailand_time_label}{Time Label Model} \\
     \href{https://huggingface.co/gtfintechlab/model_bank_of_thailand_certain_label}{Certain Label Model} \\
      \end{center}
    }
  }
\end{center}

\newpage

\section*{Monetary Policy Mandate} 
The Bank of Thailand (BoT) aims to foster a stable, sustainable and inclusive macroeconomic and financial environment in order to allow Thai citizens to sustainably achieve higher quality of life over the long term.

\textbf{Mandate Objectives:} 
\begin{itemize}
    \item \textbf{Quality of Life:} BoT focuses heavily on the quality of life of the Thai citizens.
    \item \textbf{Stable financial environment:} The bank aims to have a sustainable, stable and inclusive macroeconomic and financial environment.
\end{itemize}
\section*{Structure} 
The Bank of Thailand has the following meeting composition and structure:

\textbf{Composition: }
The MPC will continue to comprise of 7 members, as in the present arrangement.  The BOT Governor will act as Chairman, given the Governor's responsibility in implementing the MPC's decisions.  The Governor also acts as a linkage with the other policy committees of the BOT.  Two deputy Governors of the BOT and four external members form the remaining membership of the MPC, who aim to provide a balance of views, as intended in the new BOT Act.  External members have a fixed term of 3 years, and cannot serve more than two consecutive terms.
\begin{itemize}
    \item \textbf{Internal Members:} There will be 3 internal members. The BOT Governor will act as Chairman, given the Governor's responsibility in implementing the MPC's decisions.  The Governor also acts as a linkage with the other policy committees of the BOT.  Two deputy Governors of the BOT are the other internal members.\\
    \item \textbf{External Members} Four external members form the remaining membership of the MPC, who aim to provide a balance of views, as intended in the new BOT Act.  External members have a fixed term of 3 years, and cannot serve more than two consecutive terms.
\end{itemize}
\textbf{Meeting Structure: } 
 \begin{itemize}
     \item \textbf{Frequency:} Six scheduled meetings annually.
     \item \textbf{Additional Meetings:} Held as needed.
 \end{itemize}
\section*{Manual Annotation} 

\textbf{Annotators} 
\begin{itemize}
    \item Eric Liao
    \item Tyler Huang
    \item Kenny Cao
    \item Christopher Traylor
\end{itemize}

\textbf{Annotation Agreement} 
The agreement percentage among the pairs of annotators for different labels.
\begin{itemize}
    \item \SD Agreement: 35.5\%
    \item \TC Agreement: 59.1\%
    \item \CE Agreement: 60.7\%
\end{itemize} 

\textbf{Annotation Guide}

\mptext{bot}{9} Inflation, Economic Growth, Copper Prices, Peso Exchange Rate, Foreign Reserves, External Demand (Trade Partners), Key Words/Phrases, Labor Market, and Energy Prices. 
\begin{itemize}
    \item \emph{Economic Status}: A sentence pertaining to the state of the economy, relating to factors such as unemployment and inflation.
    \item \emph{Thai Baht Value Change}: A sentence pertaining to the appreciation or depreciation of the Thai Baht in the foreign exchange market.
    \item \emph{Energy/House Prices}: A sentence pertaining to fluctuations in energy sectors or housing prices, indicative of broader economic conditions.
    \item \emph{Foreign Policy}: A sentence pertaining to trade relations and economic interactions between Thailand and its international partners. If not directly related, the sentence is labeled neutral.
    \item \emph{MPC Expectations/Actions/ Assets}: A sentence that discusses the MPC's projections, actions, or asset valuations, influencing monetary policy decisions.
    \item \emph{Money Supply}: A sentence that overtly discusses changes in the money supply or shifts in the demand for loans.
    \item \emph{Key Words/Phrases}: A sentence that encapsulates specific key words or phrases which classify monetary policy stances based on their conventional usage.
    \item \emph{Labor Market}: A sentence pertaining to changes in labor market performance and productivity that may affect monetary policy.
\end{itemize}

\textbf{Examples:}
\begin{itemize}
    \item ``The latest risk-adjusted inflation forecast for 2024 eased to 3.9 percent from 4.2 percent in the previous meeting in December.''\\ 
    \textbf{Dovish}: The sentence reports a decline in inflation forecast, which reduces the urgency for tightening of monetary policy and suggest less inflationary pressure.

    \item ``Headline inflation increased in June and July driven mainly by rising transport costs such as higher domestic petroleum prices and tricycle fare hikes.''\\ 
    \textbf{Hawkish}: This sentence highlights an increase in headline inflation, suggesting potential inflationary pressures that may prompt a tighter monetary policy stance.
    
    \item ``The highlights of the discussions on the 1st December 2011 meeting were approved by the monetary board during its regular meeting held on 15th Decemeber 2011.''\\ 
    \textbf{Irrelevant}: The sentence is simply procedural and does not provide any information about the monetary policy stance.
    
    \item ``Nonetheless, inflation forecasts for 2016-2017 stayed close to the midpoint of the target range.''\\ 
    \textbf{Neutral}: The sentence indicates that the inflation forecasts remained near the midpoint of the target range, it implies that inflation is stable but does not convey any information on need for monetary tightening or easing. 
    
    \item ``However, the moderation in growth in Q1 2014, particularly in the US, as well as tighter financial conditions and geopolitical tensions in the Middle East and Russia, could dampen the overall momentum for the rest of 2014.''\\ 
    \textbf{Dovish}: This sentence indicates some downside risks to economic growth, which hints at potential easing measures.
\end{itemize}

\newpage

\begin{longtable}{p{0.118\textwidth}p{0.183\textwidth}p{0.183\textwidth}p{0.183\textwidth}p{0.183\textwidth}}
\caption{\mptitle{Bank of Thailand}} \label{tb:bot_mp_stance_guide} \\
\toprule
\textbf{Category} & \textbf{Dovish} & \textbf{Hawkish} & \textbf{Neutral} & \textbf{Irrelevant} \\
\midrule
\endfirsthead

\toprule
\textbf{Category} & \textbf{Dovish} & \textbf{Hawkish} & \textbf{Neutral} & \textbf{Irrelevant} \\
\midrule
\endhead

\textbf{Economic Status} & 
When inflation decreases, When unemployment increases, when economic growth is projected as low. & 
When inflation increases, when unemployment decreases when economic growth is projected high when economic output is higher than potential supply/actual output when economic slack falls. & 
When unemployment rate or growth is unchanged, maintained, or sustained. & 
Sentence is not relevant to monetary policy. \\
\midrule

\textbf{Thai Baht Value Change} & 
When the Thai Baht appreciates. & 
When the Thai Baht depreciates. & 
When there is no real change in Thai Baht worth. & 
Sentence is not relevant to monetary policy. \\
\midrule

\textbf{Energy/ House Prices} & 
When oil/energy sectors dwindle and prices decrease, when housing prices decrease because this can indicate a smaller consumer output, and with a dovish economy there will be a larger consumer output. & 
When oil/energy sectors are augmented/prices increase, when housing prices increase because this can indicate a large consumer output, and the economy could benefit from a hawkish policy as it will help stabilize prices. & 
When there is market uncertainty in the oil/energy sectors, when housing prices have an uncertain outlook as this means there is not clear indicators on how consumer spending is. & 
Sentence is not relevant to monetary policy. \\
\midrule

\textbf{Foreign Policy} & 
When partner economics are starting to expand, Thailand’s monetary policy will be more free as Thailand’s economic output throughput is directly proportional to its trading partners. & 
When partner economies are starting to fall or become highly volatile, Thailand should be more hawkish as the partner trading profitability is uncertain. & 
When describing specifics on an unrelated foreign nation’s economic or trade policy as this is likely to have little effect on the monetary policy control for Thailand or uncertainty in partners economies. & 
Sentence is not relevant to monetary policy. \\
\midrule

\textbf{MPC Expectations/ Actions/ Assets} & 
MPC expects subpar inflation, disinflation, narrowing of spread in Thai government bonds/lower yields, low debt serviceability because in these cases, it is in best interest to allow a freer monetary policy to boost spending/market growth. & 
MPC expects high inflation, widening of spread in Thai government bonds/higher yields, high debt serviceability because in these cases, the market has some room in which the monetary policy may be more strict without a huge negative effect. & 
MPC’s expectations of future economic status remainings uncertain because there is not much indicator for which way the economic status should sway. & 
Sentence is not relevant to monetary policy. \\
\midrule

\textbf{Money Supply} & 
Money supply is low, M2 measure increases, increased demand for loans. & 
Money supply is high, increased demand for goods, low demand for loans. & 
N/A. & 
Sentence is not relevant to monetary policy. \\
\midrule

\textbf{Key Words and Phrases} & 
When the stance is "accommodative", indicating a focus on “maximum employment” and “price stability”. & 
Indicating a focus on “price stability” and “sustained growth”. & 
Use of phrases “mixed”, “moderate”, “reaffirmed. & 
Sentence is not relevant to monetary policy. \\
\midrule

\textbf{Labor Market} & 
When labor markets see improvements, Thailand’s policy is more dovish because it leads to lower inflationary pressure to support economic growth and employment. & 
When labor markets see a decline, Thailand’s policy can be more hawkish because it can increase inflationary pressures due to more limited supplies. & 
When labor markets don’t see a change in the productivity of labor markets, there isn’t a change in the monetary policy. & 
Sentence is not relevant to monetary policy. \\
\bottomrule
\end{longtable}

\fwcertaintytext{bot}
\newpage

\begin{table*}
\caption{\fwtitle{Bank of Thailand}}
\vspace{1em}
\begin{tabular}{p{0.3\textwidth}p{0.3\textwidth}p{0.3\textwidth}}
\toprule
\textbf{Label} & \textbf{Description} & \textbf{Example} \\
\midrule
\textbf{Forward Looking} & Sentences that reflect projections or expectations about future economic conditions. They highlight anticipated trends such as inflation moderation or predicted changes in unemployment. & ``Inflation is anticipated to moderate in the coming year.'' \\
\midrule
\textbf{Not Forward Looking} & Sentences that refer to past economic performance or recent data points, such as prior inflation rates or past GDP growth. & ``Last quarter's GDP growth was lower than expected.'' \\
\bottomrule
\end{tabular}
\label{tb:bot_forward_looking_guide}
\end{table*}

\begin{table*}
\caption{\certaintytitle{Bank of Thailand}}
\vspace{1em}
\begin{tabular}{p{0.3\textwidth}p{0.3\textwidth}p{0.3\textwidth}}
\toprule
\textbf{Label} & \textbf{Description} & \textbf{Example} \\
\midrule
\textbf{Certain} & Sentences presenting definitive statements with high confidence. These include clear projections on economic growth or unambiguous stances on monetary policy. & ``The central bank will raise interest rates in the next meeting.'' \\
\midrule
\textbf{Uncertain} & Sentences that incorporate words or phrases implying doubt or unpredictability, such as “may,” “could,” or “is expected to.” & ``Economic growth might slow amid global economic uncertainties.'' \\
\bottomrule
\end{tabular}
\label{tb:bot_certainty_guide}
\end{table*}

\clearpage
\usubsection{Central Bank of Egypt}
\begin{center}
    \textbf{Region: Egypt}
\end{center}
\begin{center}
    \fbox{\includegraphics[width=0.99\textwidth]{resources/flags/Flag_of_Egypt.png}}
\end{center}
\begin{center}
    \textbf{Data Collected: 2012-2024}
\end{center}
\vfill
\begin{center}
  \fbox{%
    \parbox{\textwidth}{%
      \begin{center}
      \textbf{Important Links}\\
      \href{https://www.cbe.org.eg/en/}{Central Bank Website}\\
       \href{https://huggingface.co/datasets/gtfintechlab/central_bank_of_egypt}{Annotated Dataset}\\
     \href{https://huggingface.co/gtfintechlab/model_central_bank_of_egypt_stance_label}{Stance Label Model} \\
     \href{https://huggingface.co/gtfintechlab/model_central_bank_of_egypt_time_label}{Time Label Model} \\
     \href{https://huggingface.co/gtfintechlab/model_central_bank_of_egypt_certain_label}{Certain Label Model} \\
      \end{center}
    }
  }
\end{center}

\newpage

\section*{Monetary Policy Mandate} 
The Monetary Policy Committee (MPC) oversees the Central Bank of Egypt and is responsible for setting price targets, overnight interbank rate, and other policy rates. \newline

\textbf{Mandate Objectives:} 
\begin{itemize}
    \item \textbf{Minimize the Inflation Gap}: Transitioning the
economy to a “flexible inflation-targeting framework” by setting several inflation targets to transition from the current inflation level to medium-term level inflation. Goal was to achieve 7\% ($\pm$ 2\%) by Q4 2024 $5\% (\pm 2\%)$ by Q4 2026. Currently core inflation is 23.220\% y/y. 
    \item \textbf{Minimize the Output Gap}: Striving the maximize output with respect of full capacity maximization while minimizing macroeconomic volatility and maintaining price stability 
\end{itemize}
\section*{Structure} 
\textbf{Composition: }The MPC consists of a Governor, two deputy Governors, six members, and three non-executive Board Members, selected by the Board of Directors.

\textbf{Meeting Structure: } 
\begin{itemize}
    \item \textbf{Frequency}: Eight times a year, the schedule of which is announced at the beginning of the year
    \item \textbf{Additional Meetings}: Held as necessary in extraordinary circumstances
\end{itemize}

\section*{Manual Annotation} 

\textbf{Annotators} 

\begin{itemize}
    \item Riya Bhadani
    \item Ameen Agbaria
    \item Jaden Hamer
    \item Pranay Maddireddy
\end{itemize}

\textbf{Annotation Agreement} 
The agreement percentage among the pairs of annotators for different labels.
\begin{itemize}
    \item \SD Agreement: 52.4\%
    \item \TC Agreement: 81.9\%
    \item \CE Agreement: 88\%
\end{itemize} 
\textbf{Annotation Guide} 

\mptext{cbe}{12} exchange rate, inflation, real GDP, foreign investment, employment, foreign debt, interest rates, poverty, tourism, oil and natural gas, remittance payments, and Suez Canal. Each sentence is labeled as hawkish, dovish, neutral, or irrelevant.
\begin{itemize}
    \item \emph{Exchange rate}: Sentence pertaining to the value of Egyptian Pound (EGP) against United States Dollar (USD).
    \item \emph{Inflation}: A sentence pertaining past or current inflation levels, changes in inflation, or target inflation.
    \item \emph{Real GDP}: A sentence about changes to Egypt's economic output or target GDP.
    \item \emph{Employment}: A sentence about changes to the employment levels in Egypt or target employment levels.
    \item \emph{Foreign Investment}: A sentence about the level of foreign assets Egypt has or changes to foreign assets.
    \item \emph{Foreign Debt}: A sentence regarding the level of foreign debt in Egypt or changes to foreign debt.
    \item \emph{Interest Rates}: A sentence pertaining to past or current interest rate levels, changes to interest rate or target interest rate.
    \item \emph{Poverty}: A sentence pertaining to current poverty levels or changes to poverty levels.
    \item \emph{Tourism}: A sentence pertaining to current tourism revenue or foreseeable or past changes to tourism revenue.
    \item \emph{Oil and Natural Gas}: A sentence pertaining to changes in oil or natural gas prices, supply, demand, or exports, or a sentence regarding the affect on Egypt's economy due to global oil prices or global oil demand. Note oil and natural gas includes crude oil, petroleum, and petroleum derivatives.
    \item \emph{Remittance Payments}: A sentence pertaining to changes in level of remittance payments as a function of GDP.
    \item \emph{Suez Canal}: A sentence regarding the number of or change to the number of cargo ships passing through the Suez Canal, or geopolitical factors that may affect Suez Canal revenue.
\end{itemize}

\textbf{Examples}: 
\begin{itemize}
    \item ``With the gradual easing of previous shocks, inflationary pressures continued to subside, as annual headline and core inflation edged downward for the fifth consecutive month to reach 25.7 percent and 24.4 percent in July 2024, respectively.''\\
    \textbf{Dovish}: The inflation rate has consistently been decreasing suggesting that the MPC will decrease interest rates. 
    \item ``The slight increase in the remittances of the Egyptians working abroad by 1.6 percent to register US \$ 31.9 billion.''\\
    \textbf{Hawkish}: Increase in remittance payments means larger money supply in Egypt and more economic activity, increase the money supply in the market. Therefore, MPC would increase interest rates. 
    \item ``Globally, the expansion of economic activity continued to weaken, financial conditions eased, and trade tensions continued to weigh on the outlook.''\\
    \textbf{Neutral}: Both hawkish and dovish sentiment are expressed. Weakened economic growth and trade tensions indicates decreasing interest rates but easing financial conditions suggests that may not be necessary.
    \item ``The MPC is watching these developments and is weighing the possible effects of the above on inflation in the near future.''\\
    \textbf{Irrelevant}: There is missing information that makes the Stance unclear.
\end{itemize}

\newpage

\begin{longtable}{p{0.118\textwidth}p{0.183\textwidth}p{0.183\textwidth}p{0.183\textwidth}p{0.183\textwidth}}
\caption{\mptitle{Central Bank of Egypt}} \\
\toprule
\textbf{Category} & \textbf{Hawkish} & \textbf{Dovish} & \textbf{Neutral} & \textbf{Irrelevant} \\
\midrule
\endfirsthead

\toprule
\textbf{Category} & \textbf{Hawkish} & \textbf{Dovish} & \textbf{Neutral} & \textbf{Irrelevant} \\
\midrule
\endhead

\textbf{Exchange rate} & When exchange rate of EGP against USD increases. & When exchange rate of EGP against USD decreases. & When exchange rate remains stable. & Sentence is not relevant to monetary policy. \\
\midrule
\textbf{Inflation} & When inflation or CPI increases or when mention of  maintaining a flexible inflation target, "price stability over the medium term", mandate, structural reform, or some adjacent topic. & When inflation or CPI decreases. & When inflation remains unchanged or changes minimally. & Sentence is not relevant to monetary policy. \\
\midrule
\textbf{Real GDP} & When real GDP growth is strong. & When real GDP growth rate in Egypt is sluggish. & When GDP remains steady, stable, or unchanged. & Sentence is not relevant to monetary policy. \\
\midrule
\textbf{Employment} & When unemployment levels decrease. & When unemployment levels increase. & No or negligible changes to unemployment levels. & Sentence is not relevant to monetary policy. \\
\midrule
\textbf{Foreign Investment} & When there is an increase in foreign assets in Egypt. & When there is a decrease in foreign assets in Egypt. & When amount of foreign assets remain unchanged. & Sentence is not relevant to monetary policy. \\
\midrule
\textbf{Foreign Debt} & When Egypt reduces its foreign debt or secures debt relief, improving fiscal balance. & When Egypt's foreign debt increases, raising repayment pressure and fiscal instability. & When foreign debt levels remain manageable with regular servicing or foreign debt levels remain unchanged. & Sentence is not relevant to monetary policy. \\
\midrule
\textbf{Interest Rates} & When interest rates increase. & When interest rates decrease. & When interest rates remain unchanged. & Sentence is not relevant to monetary policy. \\
\midrule
\textbf{Poverty} & When poverty rate decreases indicating economic recovery or raising inflation. & While inflation might stabilize, economic growth slows, increasing poverty levels. & When poverty rate remains unchanged. & Sentence is not relevant to monetary policy. \\
\midrule
\textbf{Tourism} & When tourism revenue increases. & When tourism revenue declines. & When tourism revenue remains steady. & Sentence is not relevant to monetary policy. \\
\midrule
\textbf{Oil and Natural Gas} & When global oil/gas prices surge, leading to higher energy export revenues but potentially higher domestic energy costs. & When global oil/gas prices drop, lowering export revenues but reducing domestic fuel costs. & When oil/gas prices and exports remain stable, providing steady revenue without major shocks. & Sentence is not relevant to monetary policy. \\
\midrule
\textbf{Remittance payments} & When remittance payments decrease as a percentage of GDP. & When remittance payments increase as a percentage of GDP. & When remittance payments remain unchanged. & Sentence is not relevant to monetary policy. \\
\midrule
\textbf{Suez Canal} & When the number of cargo ships increases, boosting Suez Canal revenue. & When the number of cargo ships declines, affecting Suez Canal revenue, usually due to a war in the region or blockage in the Red Sea. & When the number of cargo ships passing through the canal doesn’t change. & Sentence is not relevant to monetary policy. \\
\bottomrule
\label{tb:cbe_mp_stance_guide}
\end{longtable}

\fwcertaintytext{cbe}
\newpage

\begin{table*}
\caption{\fwtitle{Central Bank of Egypt}}
\vspace{1em}
\begin{tabular}{p{0.3\textwidth}p{0.3\textwidth}p{0.3\textwidth}}
\toprule
\textbf{Label} & \textbf{Description} & \textbf{Example} \\
\midrule
\textbf{Forward Looking} & When the sentence discusses future events, economic outlook, possible changes to policy positions, and may use keywords such as “forecast,” “expected,” “judges,” or “future.” & “The Committee reiterates that the path of future policy rates remains a function of inflation expectations.” \\
\midrule
\textbf{Not Forward Looking} & When the sentence discusses a past or present economic event, data, or policy position, and may use keywords such as “driven by,” “reflections,” “decided,” “stabilized.” & “Labor market data show that the unemployment rate narrowed to 12.5 percent in the quarter ending June 2016 after peaking at 13.4 percent in the quarter ending December 2013, supported by lower real unit labor costs relative to productivity.” \\
\bottomrule
\end{tabular}
\label{tb:cbe_forward_looking_guide}
\end{table*}

\begin{table*}
\caption{\certaintytitle{Central Bank of Egypt}}
\vspace{1em}
\begin{tabular}{p{0.3\textwidth}p{0.3\textwidth}p{0.3\textwidth}}
\toprule
\textbf{Label} & \textbf{Description} & \textbf{Example} \\
\midrule
\textbf{Certain} & When the sentence expresses certainty about a future event or outcome, it may use words such as “declined,” “increased,” “decided,” “revised,” or “data show.” & “The overnight deposit and the overnight lending rates were kept at 9.5 percent and 12.5 percent, respectively.” \\
\midrule
\textbf{Uncertain} & When the statement expresses ambiguity about future events or causes of a past event, it may use words such as “expected,” “uncertain,” “implies,” “projected.” & “However, economic activity is expected to grow at a slower rate than previously projected, given the uncertainty and negative spillovers from the global scene.” \\
\bottomrule
\end{tabular}
\label{tb:cbe_certainty_guide}
\end{table*}
\clearpage
\usubsection{ Bank Negara Malaysia}
\begin{center}
    \textbf{Region: Malaysia}
\end{center}
\begin{center}
    \fbox{\includegraphics[width=0.99\textwidth]{resources/flags/Flag_of_Malaysia.png}} 
\end{center}
\begin{center}
    \textbf{Data Collected: 2004-2024}
\end{center}
\vfill
\begin{center}
  \fbox{%
    \parbox{\textwidth}{%
      \begin{center}
      \textbf{Important Links}\\
      \href{https://www.bnm.gov.my/}{Central Bank Website}\\
       \href{https://huggingface.co/datasets/gtfintechlab/bank_negara_malaysia}{Annotated Dataset}\\
     \href{https://huggingface.co/gtfintechlab/model_bank_negara_malaysia_stance_label}{Stance Label Model} \\
     \href{https://huggingface.co/gtfintechlab/model_bank_negara_malaysia_time_label}{Time Label Model} \\
     \href{https://huggingface.co/gtfintechlab/model_bank_negara_malaysia_certain_label}{Certain Label Model} \\
      \end{center}
    }
  }
\end{center}

\newpage

\section*{Monetary Policy Mandate} 
The Monetary Policy Committee (MPC) of Bank Negara Malaysia focuses on price stability and economic growth through a combination of low interest rates and inclusive financial sector. \\ \\
\textbf{Mandate Objectives:} 
\begin{itemize}
    \item \textbf{Price Stability}: Price stability is achieved by maintaining the Overnight Policy Rate (OPR) below 3\%. OPR influences the interest rates of other areas such as consumer and business loans.
    \item \textbf{Economic Growth}: Economic growth is supported through financial inclusion and providing ample credit. The bank makes sure every financial sectors have access to financial services for financial inclusion.
\end{itemize}

\section*{Structure} 
\textbf{Composition: }
\begin{itemize}
    \item \textbf{Monetary Policy Committee (MPC)}: The MPC regulates the monetary policies for the Bank Negara Malaysia. The MPC is a small group which includes the Governor, Deputy Governors, and three to seven other members. These members include external members selected by the Minister of Finance.
    \item \textbf{Financial Stability Executive Committee (FSEC)}: The FSEC maintains financial stability by overseeing and deciding on policies related to financial stability. The committee includes seven members that are mostly independent from the Bank.

\end{itemize}

\textbf{Meeting Structure: } 
\begin{itemize}
    \item \textbf{Frequency}: The MPC meets a minimum of six times a year.
    \item \textbf{Additional Meetings}: The Monetary Policy Working Group decides whether there should be more meetings for the MPC.
\end{itemize}

\section*{Manual Annotation} 

\textbf{Annotators} 

\begin{itemize}
    \item Eric Kim
    \item Aiden Chiang
    \item Rudra Gopal
    \item Spencer Gosden
\end{itemize}

\textbf{Annotation Agreement} 
The agreement percentage among the pairs of annotators for different labels.
\begin{itemize}
    \item \SD Agreement: 45.8\%
    \item \TC Agreement: 82.8\%
    \item \CE Agreement: 75.9\%
\end{itemize} 

\textbf{Annotation Guide}

\mptext{bnm}{thirteen} Economic status, Malaysian ringgit Value Change, Household loan, Foreign nations, BNM Expectations/Actions/Assets, Money Supply, Labor, Interest rate gap between Malaysia and United States, COVID, Energy/Housing Prices, and 1997 Asian Financial Crisis.

\begin{itemize}
    \item \emph{Economic Status}: A sentence pertaining to the state of the economy, relating to unemployment and inflation
    \item \emph{Malaysian ringgit Change}: A sentence pertaining to changes such as appreciation or depreciation of value of the Malaysian ringgit on the Foreign Exchange Market
    \item \emph{Household loan}: A sentence pertaining to changes in public or private debt levels.
    \item \emph{Foreign Nations}: A sentence pertaining to trade relations between Malaysia and a foreign economy. If not discussing Malaysia we label neutral.
    \item \emph{BNM Expectations/Actions/Assets}: A sentence that discusses changes in the BNM yields, bond value, reserves, or any other financial asset value.
    \item \emph{Money Supply}: A sentence that overtly discusses impact to the money supply or changes in demand.
    \item \emph{Labor}: A sentence that relates to changes in labor productivity.
    \item \emph{Interest rate gap between Malaysia and United States}: A sentence pertaining to whether the US interest rate is lower or higher compared to Malaysia's interest rate.
    \item \emph{Energy/Housing Prices}: A sentence pertaining to changes in prices of real estate, energy commodities, or energy sector as a whole.
    \item \emph{1997 Asian Financial Crisis}: A sentence pertaining to when the financial crisis caused deflationary pressures or capital outflows.
\end{itemize}

\textbf{Examples: }
\begin{itemize}
    \item ``At the Monetary Policy Committee (MPC) meeting today, Bank Negara Malaysia decided to leave the Overnight Policy Rate (OPR) unchanged at 3.50 percent.''\\
    \textbf{Neutral}: Maintaining existing interest rates indicates no change in monetary policy.
    
    \item ``Notwithstanding the stronger aggregate demand and higher global energy prices, the domestic inflation rate, as measured by the Consumer Price Index, remained low at 1.2\% in the second quarter. The underlying inflation in the economy is expected to remain low.''\\
    \textbf{Dovish}: Low inflation rates indicate room for economic growth and an accommodative monetary policy.

    \item ``Over the course of the COVID-19 crisis, the OPR was reduced by a cumulative 125 basis points to a historic low of 1.75\% to provide support to the economy. The unprecedented conditions that necessitated such actions have since abated. With the domestic growth on a firmer footing, the MPC decided to begin reducing the degree of monetary accommodation. This will be done in a measured and gradual manner, ensuring that monetary policy remains accommodative to support a sustainable economic growth in an environment of price stability.''\\
    \textbf{Hawkish}: Reducing monetary accommodation signals a shift towards a tighter economic policy.
    
    \item ``At the current OPR level, the monetary policy stance remains supportive of the economy and is consistent with the current assessment of inflation and growth prospects.''\\
    \textbf{Hawkish}: A monetary policy stance that supports growth of the economy indicates an tighter approach.
    
    \item ``Headline inflation is projected to average between 2.2\% - 3.2\% in 2022. Given the improvement in economic activity amid lingering cost pressures, underlying inflation, as measured by core inflation, is expected to trend higher to average between 2.0\% - 3.0\% in 2022.''\\
    \textbf{Hawkish}: A rise in inflation and cost pressure indicates a tighter monetary policy stance to maintain price stability.

    \item ``The MPC will continue to carefully evaluate the global and domestic economic and financial developments and their implications on the overall outlook for inflation and growth of the Malaysian economy.''\\
    \textbf{Neutral}: There is no mention of a policy shift since the MPC is solely observing the situation.
\end{itemize}

\begin{longtable}{p{0.118\textwidth}p{0.183\textwidth}p{0.183\textwidth}p{0.183\textwidth}p{0.183\textwidth}}
\caption{\mptitle{Bank Negara Malaysia}} \label{tb:bnm_mp_stance_guide} \\
\toprule
\textbf{Category} & \textbf{Dovish} & \textbf{Hawkish} & \textbf{Neutral} & \textbf{Irrelevant} \\
\midrule
\endfirsthead

\toprule
\textbf{Category} & \textbf{Dovish} & \textbf{Hawkish} & \textbf{Neutral} & \textbf{Irrelevant} \\
\midrule
\endhead

\textbf{Economic Status} & When inflation drops below BNM's 1\% target, unemployment increases, or economic growth is projected as low. & When inflation exceeds BNM's 3\% target, unemployment decreases, economic growth is projected as high, economic output surpasses potential supply, or economic slack declines. & When unemployment rate or growth is unchanged, maintained, or sustained. & Sentence is not relevant to monetary policy. \\
\midrule
\textbf{Malaysian Ringgit Value Change} & When the ringgit appreciates beyond BNM targets and inflation is low (below 1.5\%), signaling a need to boost growth. & When the ringgit depreciates beyond BNM targets and inflation is high (above 3\%), indicating a need for economic cooling. & When inflation is between 1.5\% and 3\%, and the ringgit's value changes at a healthy rate. & Sentence is not relevant to monetary policy. \\
\midrule
\textbf{Household Loan} & When public or private debt levels rise, leading to stricter fiscal policies to manage borrowing (household debt-to-GDP ratio exceeds 84\%). & When public or private debt levels decrease, allowing for more borrowing flexibility and fiscal stimulus (household debt-to-GDP ratio drops below 84\%). & N/A & Sentence is not relevant to monetary policy. \\
\midrule
\textbf{Foreign Nations} & When Malaysia's trade deficit increases with the U.S., Singapore, or Japan, signaling more exports than imports and ringgit appreciation. & When Malaysia's trade deficit rises with China or falls with the U.S., Singapore, or Japan, indicating ringgit depreciation. & When referring to foreign nations' economic or trade policies. & Sentence is not relevant to monetary policy. \\
\midrule
\textbf{BNM Expectations, Actions, and Assets} & When BNM expects subpar inflation, predicts disinflation, observes narrowing treasury bond spreads, declining treasury yields, or reduces bank reserves. & When BNM signals tightening, raises interest rates, or increases reserves to curb inflation. & When BNM believes current economic conditions are satisfactory or will self-correct. & Sentence is not relevant to monetary policy. \\
\midrule
\textbf{Money Supply} & When money supply is low, M2 growth rate falls below 4\%, and loan demand increases. & When money supply is high, M2 growth rate exceeds 6\%, demand for goods rises, and loan demand is low. & N/A & Sentence is not relevant to monetary policy. \\
\midrule
\textbf{Labor} & When productivity increases, solid labor market, strong semiconductor, electronics, and tourism industries, growing wages and consumer spending. & When productivity decreases, tourism is down, exports decreasing, downturn in global technology sector. & N/A & Sentence is not relevant to monetary policy. \\
\midrule
\textbf{Interest rate gap between Malaysia and United States} & When US interest rate is aligned with Malaysia’s, US not interfering with Malaysia’s interest rates. & When US interest rate is higher than Malaysia, potentially pressuring Malaysia to raise interest rates. & N/A & Sentence is not relevant to monetary policy. \\
\midrule
\textbf{COVID} & When easing Covid regulation: boost tourism industry and increased productivity. & When increased Covid: regulation detrimental for tourism industry and decreased productivity. & N/A & Sentence is not relevant to monetary policy. \\
\midrule
\textbf{Energy and Housing Prices} & When oil/energy prices decrease, when house prices decrease or expected to decrease. & When oil/energy prices increase, when house prices increase or expected to increase. & N/A & Sentence is not relevant to monetary policy. \\
\midrule
\textbf{1997 Asian Financial Crisis} & When the financial crisis causes deflationary pressures, prompting the central bank to lower interest rates to stimulate demand and growth. & When the financial crisis causes capital outflows, and the central bank raises interest rates to defend the currency and attract foreign capital. & N/A & Sentence is not relevant to monetary policy. \\
\bottomrule
\end{longtable}

\fwcertaintytext{bnm}
\clearpage

\begin{table*}
\caption{\fwtitle{Bank Negara Malaysia}}
\vspace{1em}
\begin{tabular}{p{0.3\textwidth} p{0.3\textwidth} p{0.3\textwidth}}
\toprule
\textbf{Label} & \textbf{Description} & \textbf{Example} \\
\midrule
\textbf{Forward Looking} & When a sentence contains phrases such as “expected,” “anticipated,” “projected,” “will,” or “forecasted.”  & “Budget 2025 measures will provide additional support to growth.” \\
\midrule
\textbf{Not Forward Looking} & When a sentence contains phrases such as “had been,” “previously,” or “last year.”  & “Domestic headline inflation moderated in September to 1.8.” \\
\bottomrule
\end{tabular}
\label{tb:bnm_forward_looking_guide}
\end{table*}

\begin{table*}
\caption{\certaintytitle{Bank Negara Malaysia}}
\vspace{1em}
\begin{tabular}{p{0.3\textwidth}p{0.3\textwidth}p{0.3\textwidth}}
\toprule
\textbf{Label} & \textbf{Description} & \textbf{Example} \\
\midrule
\textbf{Certain} & When a sentence contains phrases such as “will,” “must,” “certainly,” or “definitely.”  & “The monetary policy statement will be released at 6 p.m. on the same day as the MPC meeting.” \\
\midrule
\textbf{Uncertain} & When a sentence contains phrases such as “might,” “could,” “may,” “possibly,” “likely,” or “uncertain.”  & “These developments may increase the risk to the outlook for inflation.” \\
\bottomrule
\end{tabular}
\label{tb:bnm_certainty_guide}
\end{table*}

\clearpage
\usubsection{ Central Bank of the Philippines}
\begin{center}
    \textbf{Region: Philippines}
\end{center}
\begin{center}
    \fbox{\includegraphics[width=0.99\textwidth]{resources/flags/Flag_of_Philippines.png}}
\end{center}
\begin{center}
    \textbf{Data Collected: 2002-2024}
\end{center}
\vfill
\begin{center}
  \fbox{%
    \parbox{\textwidth}{%
      \begin{center}
      \textbf{Important Links}\\
      \href{https://www.bsp.gov.ph/SitePages/Default.aspx}{Central Bank Website}\\
       \href{https://huggingface.co/datasets/gtfintechlab/central_bank_of_the_philippines}{Annotated Dataset}\\
     \href{https://huggingface.co/gtfintechlab/model_central_bank_of_the_philippines_stance_label}{Stance Label Model} \\
     \href{https://huggingface.co/gtfintechlab/model_central_bank_of_the_philippines_time_label}{Time Label Model} \\
     \href{https://huggingface.co/gtfintechlab/model_central_bank_of_the_philippines_certain_label}{Certain Label Model} \\
      \end{center}
    }
  }
\end{center}

\newpage

\section*{Monetary Policy Mandate} 
\textbf{Mandate Objectives:} The BSP's primary objective is to promote price stability that fosters balanced and sustainable growth of the economy. 

\section*{Structure} 
\textbf{Composition: }The BSP's Monetary Board contains seven members, who are all appointed by the President of the Philippines. The seven members are:
\begin{itemize}
     \item The Governor of the BSP, who also serves as the Chairman of the Monetary Board for a six-year term.
    \item A member of the Cabinet who serves a six-year term.
    \item Five full-time members, three of whom serve a six-year term, and the other three serve a three-year term.
\end{itemize}

\textbf{Meeting Structure: } 
\begin{itemize}
    \item \textbf{Frequency:} Six scheduled meetings annually (bi-monthly). The number of meetings decreased from seven starting in 2025.
    \item \textbf{Additional Meetings:} The Governor of the BSP or two other board members may call a meeting whenever necessary to address urgent economic situations. 
\end{itemize}
 
\section*{Manual Annotation} 

\textbf{Annotators} 

\begin{itemize}
    \item Vedant Nahar
    \item Saloni Jain
    \item Saketh Budideti
    \item Jiashen Zhang
\end{itemize}
\textbf{Annotation Agreement} 
The agreement percentage among the pairs of annotators for different labels.
\begin{itemize}
    \item \SD Agreement: 53.5\%
    \item \TC Agreement: 88.1\%
    \item \CE Agreement: 80.7\%
\end{itemize} 

\textbf{Annotation Guide} 

\mptext{bsp}{five} Policy, Environment, Society, Market, and Trade.

\begin{itemize}
    \item \emph{Policy}: Decisions implemented by the government or its institutions to regulate, direct, or influence national affairs. Policies often impact laws and public services.
    \item \emph{Environment}: Decisions related to environmental sustainability, energy consumption, or natural resources.
    \item \emph{Society}: Decisions aimed at addressing the well-being of society. Policies often address income inequality, cost-of-living, and social programs.
    \item \emph{Market}: Decisions that influence financial markets, including changes in interest rates, bond yields, and liquidity management.
    \item \emph{Trade}: Decisions that impact trade, including imports, exports, exchange rates, and trade balances.
\end{itemize}

\textbf{Examples: }
\begin{itemize}
    \item ``Maintain the BSP’s policy interest rates at 7.5 percent for the overnight RRP (borrowing) rate and 9.75 percent for the overnight RP (lending) rate; b.''\\
    \textbf{Neutral}: Maintaining existing interest rates indicates no change in monetary policy.
    
    \item ``The Monetary Board (MB) decided to: a) Increase the BSP’s key policy interest rate by 50 basis points to 4.50 percent for the overnight RRP (borrowing) facility, effective 28 September 2018; and, b) Adjust the interest rates on the overnight deposit and overnight lending facilities accordingly.''\\
    \textbf{Hawkish}: Increase in interest rate and Reverse Repurchase Rate (RRP) signals contractionary monetary policy. 

    \item ``In August, the peso depreciated against the US dollar amid renewed global recession concerns as a result of protracted trade conflict between the US and China.''\\
    \textbf{Hawkish}: A depreciating peso in addition to recession concerns is likely to drive inflation higher, creating a need for contractionary monetary policy to ensure the stability of the peso and the economy.
    
    \item ``Moreover, unemployment rose to 10.2 percent in the fourth quarter of 2002 as against 9.8 percent in the same period a year ago.''\\
    \textbf{Dovish}: A sharp increase in the unemployment rate signals monetary easing to stimulate the economy. 

    \item ``Compared to the previous policy meeting, baseline inflation forecast for 2013 has been revised downward due primarily to lower oil prices.''\\
    \textbf{Dovish}: Lower inflation forecasts indicate room for monetary easing without the risks of high inflation.

    \item ``The highlights of the discussions during the 9 November 2017 monetary policy meeting were approved by the Monetary Board during its regular meeting held on 28 November 2017.''\\
    \textbf{Irrelevant}: The sentence mentions the approval of a past discussion with no mention of a monetary policy stance.
\end{itemize}

\newpage

\begin{longtable}{p{0.118\textwidth}p{0.183\textwidth}p{0.183\textwidth}p{0.183\textwidth}p{0.183\textwidth}}
\caption{Annotation Guide} \label{tb:bsp_mp_stance_guide} \\
\toprule
\textbf{Category} & \textbf{Dovish} & \textbf{Hawkish} & \textbf{Neutral} & \textbf{Irrelevant} \\
\midrule
\endfirsthead

\toprule
\textbf{Category} & \textbf{Dovish} & \textbf{Hawkish} & \textbf{Neutral} & \textbf{Irrelevant} \\
\midrule
\endhead
\textbf{Policy} & When government spending decreases or financial regulations are tightened. & When government spending increases or financial regulations are loosened. & When policy statements maintain current spending levels or regulations. & Sentence is not relevant to monetary policy. \\
\midrule
\textbf{Environment} & Energy prices fall, new price controls are implemented on fossil fuels, added subsidies for renewable energy, adverse effects of el niño. & Energy prices rise, production costs rise, or subsidies are removed. & When statements maintain current prices, confirm existing policy (e.g. subsidies, price controls) without signaling change. & Sentence is not relevant to monetary policy. \\
\midrule
\textbf{Society} & When there is higher taxes, reduced subsidies, reduced social programs, or reduced minimum wages. & When there are lower taxes, increased subsidies, new social programs, or increases in minimum wages. & When societal measures are discussed with no clear indication of change to fields such as taxes, minimum wage, subsidies, or social programs. & Sentence is not relevant to monetary policy. \\
\midrule
\textbf{Market} & When there is a reduction in interest rates, increase in the money supply, higher demand for loans, narrowing bond spread, higher bond yields. & Where there is an increase in interest rates, decrease in money supply, lower demand for loans, expanding bond spread, lower bond yields. & When there is no change in interest rates, money supply, or demand for loans. & Sentence is not relevant to monetary policy. \\
\midrule
\textbf{Trade} & Philippine peso appreciates, increased trade deficit, decreased exports. & Philippine peso depreciates, increased trade surplus, increased imports. & Maintained trade activity and currency valuation. & Sentence is not relevant to monetary policy. \\
\midrule
\bottomrule
\end{longtable}

\fwcertaintytext{bsp}
\newpage

\begin{table*}
\caption{\fwtitle{Bangko Sentral ng Pilipinas}}
\begin{tabular}{p{0.3\textwidth}p{0.3\textwidth}p{0.3\textwidth}}
\toprule
\textbf{Label} & \textbf{Description} & \textbf{Example}\\
\midrule
\textbf{Forward Looking} & text describing events anticipated to occur in the future. Sentences often utilize language such as “expected” or “projected.” & “These developments form the economic backdrop for the latest WEO projections by the IMF, which point to a modest recovery in 2017.” \\
\midrule
\textbf{Not Forward Looking} & text referring to past or recent economic policy, data, trends, or events. & “Compared to the survey two quarters ago, fewer banks have tightened lending standards over the past six months based on the Fed’s Senior Loan Officer Survey.” \\
\bottomrule
\end{tabular}
\label{tb:bsp_forward_looking_guide}
\end{table*}

\begin{table*}
\caption{\certaintytitle{Bangko Sentral ng Pilipinas}}
\begin{tabular}{p{0.3\textwidth}p{0.3\textwidth}p{0.3\textwidth}}
\toprule
\textbf{Label} & \textbf{Description} & \textbf{Example}\\
\midrule
\textbf{Certain} & The sentence reflects a clear, firm, or definitive
decision or expectation about actions or outcomes without ambiguity. & “Corporate bond issuances also increased to P119 billion, higher by 20   percent compared to the level a year ago.” \\
\midrule
\textbf{Uncertain} & The sentence expresses potential risks, varying opinions, or uncertainties about events, outcomes, or actions, often using words like “could,” “might,” or “may.” & “However, a number of risks remain, notably the expected slower growth trajectory of the US economy, a disorderly unwinding of global imbalances, and a potential increase in financial market volatility.” \\
\bottomrule
\end{tabular}
\label{tb:bsp_certainty_guide}
\end{table*}

\clearpage
\usubsection{Central Bank of Chile}
\begin{center}
    \textbf{Region: Chile}
\end{center}
\begin{center}
    \fbox{\includegraphics[width=0.99\textwidth]{resources/flags/Chile.png}}
\end{center}
\begin{center}
    \textbf{Data Collected: 2018-2024}
\end{center}
\vfill
\begin{center}
  \fbox{%
    \parbox{\textwidth}{%
      \begin{center}
      \textbf{Important Links}\\
      \href{https://www.bcentral.cl/en/home}{Central Bank Website}\\
       \href{https://huggingface.co/datasets/gtfintechlab/central_bank_of_chile}{Annotated Dataset}\\
     \href{https://huggingface.co/gtfintechlab/model_central_bank_of_chile_stance_label}{Stance Label Model} \\
     \href{https://huggingface.co/gtfintechlab/model_central_bank_of_chile_time_label}{Time Label Model} \\
     \href{https://huggingface.co/gtfintechlab/model_central_bank_of_chile_certain_label}{Certain Label Model} \\
      \end{center}
    }
  }
\end{center}

\newpage

\section*{Monetary Policy Mandate} 
The primary mandate of the Bank of Chile is to control the inflation levels in the economy.

\textbf{Mandate Objectives:} 

\begin{itemize}
    \item \textbf{Price Stability:} The Central Bank of Chile's primary goal is to ensure stability of the currency.This implies keeping inflation low and stable over time. This translates to keeping the Consumer Price Inflation at around 3\% with a $\pm 1$\% tolerance over a two year horizon.
\end{itemize}

\section*{Structure} 
\textbf{Composition: }
\begin{itemize}
    \item \textbf{Board of the Central Bank of Chile:} The Board is made up of five members, appointed by the President of the Republic, upon prior approval by the Senate. Each member serves for ten years. At the same time, one of them serves as the Governor for a term of five years.
    \item \textbf{Additional Members:} Bank managers and analysts also participate in the meetings,  along with a representative of the Ministry of Finance, all of them without voting rights.
\end{itemize}
\textbf{Meeting Structure: } 
 \begin{itemize}
     \item \textbf{Frequency and Length:} There are 8 meetings that occur every year. Four of these meetings are held on the business day prior to the publication of the Monetary Policy Report and they take place on one day. On the other four occasions, the meeting lasts one and a half days.
 \end{itemize}
\section*{Manual Annotation} 

\textbf{Annotators} 

\begin{itemize}
    \item Sagnik Nandi
    \item Rutwik Routu
    \item Ankith Keshireddy
    \item Dylan Patrick Kelly
\end{itemize}

\textbf{Annotation Agreement} 
The agreement percentage among the pairs of annotators for different labels.
\begin{itemize}
    \item \SD Agreement: 55.5\%
    \item \TC Agreement: 69.6\%
    \item \CE Agreement: 75.5\%
\end{itemize} 

\textbf{Annotation Guide} 

\mptext{cboc}{nine} Inflation, Economic Growth, Copper Prices, Peso Exchange Rate, Foreign Reserves, External Demand (Trade Partners), Key Words/Phrases, Labor Market, Energy Prices. 

\begin{itemize}
    \item \emph{Inflation}: A sentence relating to the monetary policy decision of the Central Bank of Chile.
    
    \item \emph{Economic Growth}: A sentence pertaining to the state of the economy, relating to unemployment, investment, and inflation.
    
    \item \emph{Copper Prices}: A sentence relating to changes in copper prices, since Chile is the largest copper exporter and therefore, is an economic indicator for Chile.
    
    \item \emph{Peso Exchange Rate}: A sentence pertaining to changes in the peso, such as appreciation or depreciation of the value of the Peso on the Foreign Exchange Market.
    
    \item \emph{Foreign Reserves}: A sentence reflecting the central bank’s stance on liquidity and currency stabilization.
    
    \item \emph{External Demand}: Since Chile’s economy is export-dependent, a sentence pertaining to global demand becomes a critical factor.
    
    \item \emph{Key Words/Phrases}: A sentence that contains a key word or phrase that would classify it squarely into one of the three label classes, based upon its frequent usage and meaning among particular label classes.
    
    \item \emph{Labor Market}: A sentence that relates to changes in labor productivity.
    
    \item \emph{Energy Prices}: Lower energy prices reduce inflation and production costs, while higher energy prices increase them, prompting different monetary responses from the Central Bank.
\end{itemize}

\textbf{Examples: }
\begin{itemize}
    \item ``Meanwhile, the longer-term rates had risen less than their external counterparts, in a context in which the local risk indicators remained contained.''\\
    \textbf{Neutral}: This statement only describes the market conditions without indicating a tightening or easing policy stance. 
    
    \item ``The median of the Financial Traders Survey (FTS) remained around 3.5\% for the third month in a row, while 90\% of the respondents expected inflation to be higher than 3\%.''\\
    \textbf{Hawkish}: The fact that 90\% of the respondents believed that the inflation was above 3\% suggests that there are some inflationary pressures.
    
    \item ``In this context, a major challenge would be to maintain the impulse of monetary and fiscal policy, as well as the ability to safeguard financial stability, until the economy was able to achieve a self-sustained growth rate, reducing the gaps and frictions that still remained.''\\
    \textbf{Dovish}: The statement suggests a goal to maintain stability which suggests a more accommodative approach.
    
    \item ``All five board members agreed that, from the analysis of the background information submitted in the preparation of the March MP report, it could be concluded that there was still no evidence of consolidation of said inflationary convergence.''\\
    \textbf{Hawkish}: In this sentence, it was evident from the board members' agreement that the inflation has not yet stabilized, signaling that inflation is too high or volatile. Thus, potentially requiring tightening monetary policy.
    
    \item ``There were also differences across multiple sources of information.''\\
    \textbf{Irrelevant}: This sentence points out that differences exist across information sources but is ultimately carry any monetary policy stance.
\end{itemize}

\clearpage

\begin{longtable}{p{0.118\textwidth}p{0.183\textwidth}p{0.183\textwidth}p{0.183\textwidth}p{0.183\textwidth}}
\caption{\mptitle{Central Bank of Chile}} \label{tb:cboc_mp_stance_guide} \\
\toprule
\textbf{Category} & \textbf{Hawkish} & \textbf{Dovish} & \textbf{Neutral} & \textbf{Irrelevant} \\
\midrule
\endfirsthead

\toprule
\textbf{Category} & \textbf{Hawkish} & \textbf{Dovish} & \textbf{Neutral} & \textbf{Irrelevant} \\
\midrule
\endhead

\textbf{Inflation} & Sentences indicating inflation significantly below the 2\% target suggest concerns over deflation or weak demand. & Sentences where inflation exceeds 4\% highlight the necessity for tighter monetary policies to curb inflationary pressures. & Sentences indicating inflation rates within the 2.5\% to 3.5\% ranges suggest stable price levels with no immediate need for policy adjustments. & Sentence is not relevant to monetary policy. \\
\midrule

\textbf{Economic Growth} & Descriptions of GDP growth below 2\% or unemployment over 8\% indicate the need for economic stimulus. & Descriptions of GDP growth above 5\% and unemployment below 4\% suggest an overheating economy requiring cooling measures. & Sentences indicating GDP growth between 3\% and 4\%, and unemployment rates from 4\% to 6\%, represent balanced economic conditions. & Sentence is not relevant to monetary policy. \\
\midrule

\textbf{Copper Prices} & Statements that copper prices are below approximately \$7,000/ton indicate reduced export revenues and potential economic strain. & References to copper prices exceeding \$10,000/ton suggest potential inflationary pressures, possibly necessitating tighter monetary policies. & Descriptions of copper prices within \$7,500 to \$9,500/ton indicate stable conditions conducive to current policy settings. & Sentence is not relevant to monetary policy. \\
\midrule

\textbf{Peso Exchange Rate} & Any mention of the peso depreciating more than 5\% suggests potential harm to export competitiveness, which might call for central bank intervention. & Mentions of the peso depreciating more than 5\%-10\% in a short period, pointing to possible inflation spikes and the need for a tighter monetary stance. & Statements indicating minor fluctuations (less than 2\% per month) suggest a stable exchange rate. & Sentence is not relevant to monetary policy. \\
\midrule

\textbf{Foreign Reserves} & Rapid increases in reserves (more than 5\% per month) can indicate an intention to inject liquidity into the economy. & Decreases in reserves (more than 5\%) can suggest a tightening of monetary liquidity. & Moderate changes (within 2\% month-over-month) in reserves indicate stability. & Sentence is not relevant to monetary policy. \\
\midrule

\textbf{External Demand (Trade Partners)} & Sentences indicating a weakening in external demand, especially from major trade partners like China, suggest economic downturns. & Increased global demand for Chilean exports, as mentioned in statements, indicates robust economic health. & Stable external demand levels imply a steady economic environment. & Sentence is not relevant to monetary policy. \\
\midrule

\textbf{Key Words and Phrases} & Sentences focusing on “stimulating growth,” “accommodative monetary policy,” or “reducing unemployment” are categorized as dovish. & Phrases like “inflation control,” “price stability,” or “reducing liquidity” are considered hawkish. & Neutral terms include “balanced,” “steady,” or “in line with expectations.” & Sentence is not relevant to monetary policy. \\
\midrule

\textbf{Labor Market} & References to high unemployment rates (over 8\%) and stagnant wage growth (less than 2\%) indicate a need for supportive monetary policies. & Low unemployment (less than 4\%) and rising wages (over 5\%) suggest an overheating labor market that may need cooling measures. & Descriptions of unemployment rates between 4\%-6\% and wage growth between 3\%-4\% indicate a healthy labor market. & Sentence is not relevant to monetary policy. \\
\midrule

\textbf{Energy Prices} & Statements of energy prices declining by more than 10\% often suggest decreasing inflationary pressures and reduced production costs. & Descriptions of energy prices increasing by more than 10\% indicate rising inflation pressures, potentially requiring tighter monetary measures. & Sentences noting energy price changes within a 1\%-2\% range month-over-month suggest price stability. & Sentence is not relevant to monetary policy. \\
\bottomrule
\end{longtable}

\fwcertaintytext{cboc}
\clearpage

\begin{table*}
\caption{\fwtitle{Central Bank of Chile}}
\vspace{1em}
\begin{tabular}{p{0.3\textwidth}p{0.3\textwidth}p{0.3\textwidth}}
\toprule
\textbf{Label} & \textbf{Description} & \textbf{Example}\\
\midrule
\textbf{Forward Looking} & Sentences that contain predictions or anticipations about
future events or trends. & “However, the latest indicators showed a rebound in confidence and outlook, anticipating a faster recovery in the second half of the year, supported by progress in the vaccination processes, the high savings accumulated by households, and the fiscal stimulus plans.” \\
\midrule
\textbf{Not Forward-looking} & Sentences that discuss current or past events without reference to future expectations. & “Household expectations had also improved, although their performance and that of other indicators more closely related to private consumption augured a slow rebound in this expenditure component.” \\
\bottomrule
\end{tabular}
\label{tb:cboc_forward_looking_guide}
\end{table*}

\begin{table*}
\caption{\certaintytitle{Central Bank of Chile}}
\vspace{1em}
\begin{tabular}{p{0.3\textwidth}p{0.3\textwidth}p{0.3\textwidth}}
\toprule
\textbf{Label} & \textbf{Description} & \textbf{Example}\\
\midrule
\textbf{Certain} & Sentences that state facts or outcomes with certainty, without speculative language. & “Headline inflation had reached 13.7\% annually in September, slightly less than in August, while core inflation -- the CPI minus volatiles -- had risen to 11.1\% annually.” \\
\midrule
\textbf{Uncertain} & Sentences that include words indicating probability, speculation, or uncertainty such as ``likely,'' ``could,'' or ``might.'' & “It was also noted that the external risks had implications for inflation and monetary policy in Chile that could differ in both the short and the long term.” \\
\bottomrule
\end{tabular}
\label{tb:cboc_certainty_guide}
\end{table*}
\clearpage
\usubsection{ Central Reserve Bank of Peru}
\begin{center}
    \textbf{Region: Peru}
\end{center}
\begin{center}
    \fbox{\includegraphics[width=0.99\textwidth]{resources/flags/Flag_of_Peru.png}}
\end{center}

\begin{center}
    \textbf{Data Collected: 2001-2024}
\end{center}
\vfill
\begin{center}
  \fbox{%
    \parbox{\textwidth}{%
      \begin{center}
      \textbf{Important Links}\\
      \href{https://www.bcrp.gob.pe/en/}{Central Bank Website}\\
       \href{https://huggingface.co/datasets/gtfintechlab/central_reserve_bank_of_peru}{Annotated Dataset}\\
     \href{https://huggingface.co/gtfintechlab/model_central_reserve_bank_of_peru_stance_label}{Stance Label Model} \\
     \href{https://huggingface.co/gtfintechlab/model_central_reserve_bank_of_peru_time_label}{Time Label Model} \\
     \href{https://huggingface.co/gtfintechlab/model_central_reserve_bank_of_peru_certain_label}{Certain Label Model} \\
      \end{center}
    }
  }
\end{center}

\newpage

\section*{Monetary Policy Mandate} 
The BCRP (Banco Central de Reserva del Peru) is tasked with regulating the amount of money, manage international reserves, issue notes and coins, and periodically report on the economy's finances

\textbf{Mandate Objectives:} 
\begin{itemize}
    \item \textbf {Price Stability}: Maintain price stability via inflation targeting 2\% per year with $\pm$ 1\%.
    \item \textbf {Interbank Interest Rate}: Either injecting or sterilizing liquidity depending on whether the reference rate is facing upward or downward pressure.
\end{itemize}
\section*{Structure} 
BCRP's was established by the constitution of Peru and serves to maintain monetary stability in the economy.

\textbf{Composition: }
\begin{itemize}
    \item \textbf{The Board of Directors}: They have the highest authority of the central bank. The board is responsible for determining the policies needed to fulfill the bank's mission and also is in charge of general activities. The Executive and Legislative branches of Peru appoint three members each to the board. 
    \item \textbf{General Manager}: The General Manager is in charge of the technical and administrative processes.
\end{itemize}

\textbf{Meeting Structure: } 
\begin{itemize}
    \item \textbf{Frequency}: The Board of Directors meet once a month or 12 times a year at dates published at the start of the year.
    \item \textbf{Purpose}: The meetings general discuss what actions need to be taken in order to meet the bank goals. The decisions made during the meeting are immediately announced to the press afterward.
    \item \textbf{Method}: Monetary policy actions that can be taken include changing the reference rate depending on inflationary or deflationary pressures.
\end{itemize}

\section*{Manual Annotation} 

\textbf{Annotators} 

\begin{itemize}
    \item Meghaj Tarte
    \item Min Lee
    \item Saketh Budideti
    \item Sloka Chava
\end{itemize}

\textbf{Annotation Agreement} 
The agreement percentage among the pairs of annotators for different labels.
\begin{itemize}
    \item \SD Agreement: 56.6\%
    \item \TC Agreement: 82.2\%
    \item \CE Agreement: 87.6\%
\end{itemize} 

\textbf{Annotation Guide} 

\mptext{bcrp}{twelve} Economic Status, Sol Value Change, Housing Debt/Loan, Foreign Nations, BCRP Expectations/Actions/Assets, Money Supply, Keywords/Phrases, Dollar Value, FX Market, COVID, Energy/Housing Prices, and Weather.

\begin{itemize}
    \item \emph{Economic Status}: A sentence pertaining to the state of the economy, relating to unemployment and inflation.
    \item \emph{Sol Value Change}: A sentence pertaining to changes such as appreciation or depreciation of value of the Peruvian sol on the Foreign Exchange Market
    \item \emph{Housing Debt/Loan}:  A sentence that relates to changes in debt in households.
    \item \emph{Foreign Nations}: A sentence pertaining to trade relations between Peru and a foreign economy. If not discussing Peru we label neutral. 
    \item \emph{BCRP Expectations/Actions/Assets}: A sentence that discusses changes in CRBP yields, bond value, reserves, or any other financial asset value. 
    \item \emph{Money Supply}:  A sentence that overtly discusses impact to the money supply or changes in demand. 
    \item \emph{Keyword/Phrases}: A sentence that contains key word or phrase that would classify it squarely into one of the three label classes, based upon its frequent usage and meaning among particular label classes
    \item \emph{Dollar value}:  A sentence about fluctuation of the dollar’s value which can influence monetary policy in a dovish or hawkish direction.
    \item \emph{Foreign Exchange (FX) Market}: A sentence about stability’s changes in the foreign exchange market
    \item \emph{COVID}:  A sentence indicating whether the government is tightening or easing COVID policy
    \item \emph{Energy/Housing prices}: A sentence pertaining to changes in prices of real estate, energy commodities, or energy sector as a whole. 
    \item \emph{Weather}: A sentence pertaining to how drastic changes to weather affect the economy.
\end{itemize}

\textbf{Examples: }
\begin{itemize}
    \item ``In light of the narrowing interest rate gap between Peru and the United States, some members suggested that the Central Reserve Bank of Peru may consider keeping interest rates low to support domestic economic recovery and encourage borrowing.''\\
    \textbf{Dovish}: This statement supports economic recovery via lowering interest rates, thus, reflecting a dovish stance.

    \item ``This measure is mainly preventive in a context of strong dynamism of domestic demand, a situation in which withdrawing monetary stimulus is advisable in order to maintain inflation within the target range.''\\
    \textbf{Hawkish}: This sentence suggests 'withdrawing monetary stimulus', which implies tightening monetary policy. 

    \item ``Therefore, between May 2022 and March 2024, the minimum legal reserve requirement rate in local currency was kept at 6 percent in the context of a progressive reduction of the monetary easing linked to the COVID-19 epidemic. This rate was lowered to 5.5 percent in April of this year in order to increase the amount of money that may be lent out and to supplement monetary easing.''\\
    \textbf{Dovish}: This statement indicates a dovish stance as it suggests monetary easing and  lowering of the legal reserve requirement rate to increase the money available for lending. 

     \item ``This higher level is explained by the new methodology that is being used to calculate this indicator since September 29, and now includes bonds involving a longer maturity in the basket''\\
    \textbf{Neutral}: This statement is purely informational, describing the change in how an indicator is calculated without any discussion of policy implications.
     \\

    \item ``The board will approve the monetary program for the next month on its session of November 5.''\\
    \textbf{Irrelevant}: This statement is purely administrative, providing details about a scheduled board session without indicating any monetary policy stance.
\end{itemize}

\newpage

\begin{longtable}{p{0.118\textwidth}p{0.183\textwidth}p{0.183\textwidth}p{0.183\textwidth}p{0.183\textwidth}}
\caption{\mptitle{Central Reserve Bank of Peru}} \label{tb:bcrp_mp_stance_guide} \\
\toprule
\textbf{Category} & \textbf{Hawkish} & \textbf{Dovish} & \textbf{Neutral} & \textbf{Irrelevant} \\
\midrule
\endfirsthead

\toprule
\textbf{Category} & \textbf{Hawkish} & \textbf{Dovish} & \textbf{Neutral} & \textbf{Irrelevant} \\
\midrule
\endhead
\textbf{Economic Status} & When inflation decreases, when unemployment increases, or when economic growth is projected as low. & When inflation increases, unemployment decreases, or economic growth is projected as high. & When unemployment rate or growth is unchanged, maintained, or sustained. & Sentence is not relevant to monetary policy. \\ 
\midrule
\textbf{Sol Value Change} & When Sol value appreciates. & When Sol value depreciates. & N/A & Sentence is not relevant to monetary policy. \\
\midrule
\textbf{Housing Debt/Loan} & When public or private debt levels decrease, signaling more flexibility for borrowing and fiscal stimulus. & When public or private debt levels increase, prompting strict fiscal policy to limit borrowing and manage rising debt. & N/A & Sentence is not relevant to monetary policy. \\
\midrule
\textbf{Foreign Nations} & When the Peruvian trade deficit decreases. & When the Peruvian trade deficit increases. & When relating to a foreign nation’s economic or trade policy. & Sentence is not relevant to monetary policy. \\
\midrule
\textbf{BCRP Expectations, Actions, and Assets} & When BCRP expects subpar inflation, disinflation, narrowing spreads of treasury bonds, decreases in treasury security yields, or reductions in bank reserves. & When BCRP signals the need for tightening, raising interest rates, or increasing reserves to curb inflation. & N/A & Sentence is not relevant to monetary policy. \\
\midrule
\textbf{Money Supply} & When the money supply is low, M2 increases, or demand for loans increases. & When the money supply is high, demand for goods increases, or demand for loans decreases. & N/A & Sentence is not relevant to monetary policy. \\
\midrule
\textbf{Keywords and Phrases} & When the stance is "accommodative," indicating a focus on "maximum employment" and "price stability." & Indicating a focus on "price stability" and "sustained growth." & Use of phrases such as “mixed,” “moderate,” or “reaffirmed.” & Sentence is not relevant to monetary policy. \\
\midrule
\textbf{Dollar Value} & When USD is moderate or weakening. & When USD strengthens, indicating potential depreciation of Sol. & N/A & Sentence is not relevant to monetary policy. \\
\midrule
\textbf{FX Market} & Stable FX market, leading to less government involvement. & Unstable FX market, leading to more government involvement. & N/A & Sentence is not relevant to monetary policy. \\
\midrule
\textbf{COVID} & Easing COVID regulations boosts the tourism industry and increases productivity. & Increased COVID regulations harm the tourism industry and decrease productivity. & N/A & Sentence is not relevant to monetary policy. \\
\midrule
\textbf{Energy and Housing Prices} & When oil/energy prices decrease or house prices decrease or are expected to decrease. & When oil/energy prices increase or house prices increase or are expected to increase. & N/A & Sentence is not relevant to monetary policy. \\
\midrule
\textbf{Weather} & Favorable weather conditions benefit agriculture, fishing, and manufacturing sectors. & Unfavorable weather conditions, such as droughts, harm agriculture, fishing, and manufacturing sectors. & N/A & Sentence is not relevant to monetary policy. \\
\bottomrule
\end{longtable}

\fwcertaintytext{bcrp}
\clearpage

\begin{table*}
\caption{\fwtitle{Central Reserve Bank of Peru}}
\vspace{1em}
\begin{tabular}{p{0.3\textwidth}p{0.3\textwidth}p{0.3\textwidth}}
\toprule
\textbf{Label} & \textbf{Description} & \textbf{Example}\\
\midrule
\textbf{Forward Looking} & Events that are referencing future actions or future data projections. Sentences using language such as ``expected,'' ``projected, ''coming,`` and ``next.'' & ``The BCRP expects to drop interest rates in the next coming meetings.''\\
\midrule
\textbf{Not Forward Looking} & Events that are referencing past actions or past data. Sentences using language such as ``was,'' ``stayed,'' ``maintained,'' and ``kept.'' & ``The BCRP decreased the reference rate by 25 bps.'' \\
\bottomrule
\end{tabular}
\label{tb:bcrp_forward_looking_guide}
\end{table*}

\begin{table*}
\caption{\certaintytitle{Central Reserve Bank of Peru}}
\vspace{1em}
\begin{tabular}{p{0.3\textwidth}p{0.3\textwidth}p{0.3\textwidth}}
\toprule
\textbf{Label} & \textbf{Description} & \textbf{Example}\\
\midrule
\textbf{Certain} & The sentence is very clear about the stance it is taking or it is reflecting about past actions which are certain as they have already been done. Key words include ``certain,'' ``evidently,'' and ``absolutely.'' & ``The BCRP will not change the reference rate at this time.''\\
\midrule
\textbf{Uncertain} & The sentence is unsure about the stance it is taking or its speculating about the future which can be hard to predict. Key words include ``unsure,'' ``might,'' ``may,'' ``expects,'' and ``hopes.'' & ``The BCRP may increase rates at the next meeting.'' \\
\bottomrule
\end{tabular}
\label{tb:bcrp_certainty_guide}
\end{table*}
\clearpage
\usubsection{Bank of the Republic}
\begin{center}
    \textbf{Region: Colombia}
\end{center}
\begin{center}
    \fbox{\includegraphics[width=0.99\textwidth]{resources/flags/Flag_of_Colombia.png}}
\end{center}
\begin{center}
    \textbf{Data Collected: 2014-2024}
\end{center}
\vfill
\begin{center}
  \fbox{%
    \parbox{\textwidth}{%
      \begin{center}
      \textbf{Important Links}\\
      \href{http://www.banrep.gov.co/}{Central Bank Website}\\
       \href{https://huggingface.co/datasets/gtfintechlab/bank_of_the_republic_colombia}{Annotated Dataset}\\
     \href{https://huggingface.co/gtfintechlab/model_bank_of_the_republic_colombia_stance_label}{Stance Label Model} \\
     \href{https://huggingface.co/gtfintechlab/model_bank_of_the_republic_colombia_time_label}{Time Label Model} \\
     \href{https://huggingface.co/gtfintechlab/model_bank_of_the_republic_colombia_certain_label}{Certain Label Model} \\
      \end{center}
    }
  }
\end{center}

\newpage

\section*{Monetary Policy Mandate} 

The Banco de la Rep\'ublica (BanRep), in addition to being responsible for maintaining the functioning of payment systems and managing cultural activities, aims to formulate monetary policy to achieve its mandate.

\textbf{Mandate Objectives:}
\begin{itemize}
    \item \textbf{Maintaining purchasing power} Aims to maintain a low and stable annual inflation target of 3\%, measured by the CPI.
    \item \textbf{Economic Growth:} Strives for economic output to be as close to its maximum sustainable potential value as possible. 
    \item \textbf{Financial Stability:} Promote the ability of the Colombian economy to allocate resources and mitigate financial risks properly.
\end{itemize}

\section*{Structure} 
\textbf{Composition: }
\begin{itemize}
    \item Seven directors, including the Minister of Finance, the Governor of the Bank (who is elected by the Board), and five members appointed by the President, each serving four-year renewable terms. Two board members are replaced every presidential term. 
\end{itemize}

\textbf{Meeting Structure: } 

\begin{itemize}
    \item \textbf{Meeting Requirements}: 
    \begin{itemize}
        \item At least five members must be in attendance, one of whom is the the Minister of Finance.
        \item Decisions require a majority vote of at least four members, and national loans require a unanimous vote.
    \end{itemize}
    \item \textbf{Frequency}: Monthly for a total of 12 meetings a year. However, BanRep generally does not make decisions in May, August, or November.
\end{itemize}
 
\section*{Manual Annotation} 

\textbf{Annotators} 

\begin{itemize}
    \item Joshua Zhang
    \item Saksham Purbey
    \item Adhish D Rajan
    \item Anir Vetukuri
\end{itemize}

\textbf{Annotation Agreement} 
The agreement percentage among the pairs of annotators for different labels.
\begin{itemize}
    \item \SD Agreement: 57.3\%
    \item \TC Agreement: 80.4\%
    \item \CE Agreement: 79.9\%
\end{itemize} 

\textbf{Annotation Guide} 

\mptext{banrep}{eight} Inflation, Employment, Growth, Peso Value Change, Trade, BanRep Interest Rate, Energy, and Climate.

\begin{itemize}
    \item \emph{Inflation}: How much inflation is present in the Colombian Economy, measured by CPI, and whether or not this value increases.
    \item \emph{Employment}: The rate of unemployment in Colombia and its change over time.
    \item \emph{Growth}: The rate of GDP growth in Colombia's economy, and any changes to its value.
    \item \emph{Peso Value Change}: The appreciation, depreciation, or maintenance of the Colombian Peso.
    \item \emph{Trade}: The amount of imports and exports going in, and any changes in the amounts and distribution of these goods.
    \item \emph{BanRep Interest Rate}: The current interest rate set by the Central Bank of Colombia and any decisions surrounding it.
    \item \emph{Energy}: The availability and price of energy in Colombia
    \item \emph{Climate}: Notable climate conditions impacting Colombia, such as El Ni\~no.
\end{itemize}

\textbf{Examples: }
\begin{itemize}
    \item ``They noted that the concern of a significant impact on food or power costs by the El Ni\~no phenomenon has been waning.''\\
    \textbf{Dovish}: Poor climate conditions like El Ni\~no put inflationary pressures on agricultural and energy sectors. With these effects improving, there is less inflation and more room to cut rates to stimulate the economy.
    
    \item ``They cautioned against sudden and unexpected interest rate cuts, which could fuel inflation expectations and currency depreciation, potentially jeopardizing the approximation to the inflation target and necessitating reversals with significant credibility costs.''\\
    \textbf{Hawkish}: The BanRep committee members state their hesitancy toward cutting rates, indicating a hawkish outlook.
    
    \item ``In the case of electricity rates, they emphasize that the risk of power outages has dissipated as hydro generation has been increasing and new solar plants are coming online.'' \\
    \textbf{Dovish}: Better power infrastructure is less likely to put inflationary pressures on energy prices, reducing the need to increase rates.

    \item ``This trend persists because essential items like rents, education, and personal services, among others, tend to adjust in line with observed inflation and wage increments.'' \\
    \textbf{Neutral}: Makes a neutral remark regarding observed trends surrounding inflation, but does not provide a reason to either cut or raise rates.

    \item ``A regular meeting of the board of directors of Banco de la Republica was held in the city of Bogota on may 23, 2008.''\\
    \textbf{Irrelevant}: This sentence is a statement about an administrative meeting and does not contain any monetary policy stance.
\end{itemize}

\newpage

\begin{longtable}{p{0.118\textwidth}p{0.183\textwidth}p{0.183\textwidth}p{0.183\textwidth}p{0.183\textwidth}}
\caption{\mptitle{Bank of the Republic}} \\
\toprule
\textbf{Category} & \textbf{Dovish} & \textbf{Hawkish} & \textbf{Neutral} & \textbf{Irrelevant} \\
\midrule
\endfirsthead

\toprule
\textbf{Category} & \textbf{Dovish} & \textbf{Hawkish} & \textbf{Neutral} & \textbf{Irrelevant} \\
\midrule
\endhead
\textbf{Inflation} & When inflation is declining or is below the 3\% target. & When inflation is rising or is above the 3\% target. & When inflation is close to the target and unchanged. & Sentence is not relevant to monetary policy. \\
\midrule
\textbf{Employment} & Low/decreasing employment, deterioration of job quality, or sectors lobbying for rate cuts to improve the labor market. & High/increasing employment or improvements in the labor market. & Stable unemployment rates. & Sentence is not relevant to monetary policy. \\
\midrule
\textbf{Economic Growth} & When economic growth is currently low or expected to be low in the future. & When economic growth is currently too high to be sustainable, or expected to be as such. & When economic growth is stable without causing economic overheating. & Sentence is not relevant to monetary policy. \\
\midrule
\textbf{Peso Value Change} & When the Peso appreciates. & When the Peso depreciates. & N/A & Sentence is not relevant to monetary policy. \\
\midrule
\textbf{Trade} & More exports or fewer imports, leading to a lower trade deficit. & Less exports or more imports, leading to a higher trade deficit. & N/A & Sentence is not relevant to monetary policy. \\
\midrule
\textbf{BanRep Interest Rate} & When BanRep decides to or is planning on lowering the interest rate. & When BanRep decides to or is planning on raising the interest rate. & When BanRep leaves interest rate unchanged. & Sentence is not relevant to monetary policy. \\
\midrule
\textbf{Energy} & Falling energy prices, higher energy availability. & Rising energy prices, lower energy availability. & N/A & Sentence is not relevant to monetary policy. \\
\midrule
\textbf{Climate} & When climate in the region improves, reducing agricultural risks leading to inflation (for examples, food supply and prices). & When poor climate and weather conditions increase food prices or reduce agricultural output, leading to inflation. & N/A & Sentence is not relevant to monetary policy. \\
\midrule
\textbf{Investment} & Lower levels of investment or delays in investment. & Higher levels of investment. & N/A & Sentence is not relevant to monetary policy. \\
\midrule
\textbf{Security Prices} & Higher prices of public debt. & Lower prices of public debt. & N/A & Sentence is not relevant to monetary policy. \\
\midrule
\textbf{Credit} & Decrease in credit or low levels of credit growth. & When credit growth is high. & N/A & Sentence is not relevant to monetary policy. \\
\bottomrule
\label{tb:banrep_mp_stance_guide}
\end{longtable}

\fwcertaintytext{banrep}
\newpage

\begin{table*}
\caption{\fwtitle{Bank of the Republic}}
\vspace{1em}
\begin{tabular}{p{0.3\textwidth}p{0.3\textwidth}p{0.3\textwidth}}
\toprule
\textbf{Label} & \textbf{Description} & \textbf{Example}\\
\midrule
\textbf{Forward Looking} & Discusses projections or expectations of future values of metrics. Uses words like ``expects,'' ``forecasts,'' or ``anticipates.'' & ``Economic activity indicators suggest that GDP would remain on a recovery path during the second quarter, with varying performances among sectors.'' \\
\midrule
\textbf{Not Forward Looking} & Discusses historical or ongoing data/events. Uses words like ``previous,'' ``historic'' or ``recorded.'' & ``The directors agree that the restrictive monetary policy stance has fulfilled its objective of reducing excess spending in the economy.'' \\
\bottomrule
\end{tabular}
\label{tb:banrep_forward_looking_guide}
\end{table*}

\begin{table*}
\caption{\certaintytitle{Bank of the Republic}}
\vspace{1em}
\begin{tabular}{p{0.3\textwidth}p{0.3\textwidth}p{0.3\textwidth}}
\toprule
\textbf{Label} & \textbf{Description} & \textbf{Example}\\
\midrule
\textbf{Certain} & Refers to concrete decisions, historical facts/data, or projections for the future with definite and unambiguous diction. Uses words like ``firmly,'' ``will,'' or ``decided.'' & ``he Board of Directors reached a unanimous decision to lower interest rates to support economic recovery and address inflation, which is effectively decreasing and approaching the target.''\\
\midrule
\textbf{Uncertain} & Refers to concrete decisions, historical facts/data, or projections for the future with ambiguous diction and a lack of definite terms. Uses words like ``may,'' ``possible,'' or ``suggested.'' & ``They cautioned against sudden and unexpected interest rate cuts, which could fuel inflation expectations and currency depreciation, potentially jeopardizing the approximation to the inflation target and necessitating reversals with significant credibility costs.'' \\
\bottomrule
\end{tabular}
\label{tb:banrep_certainty_guide}
\end{table*}
\clearpage
\usubsection{Czech National Bank}

\label{app:czech_annot_guide}
\begin{center}
    \textbf{Region: Czech Republic}
\end{center}
\begin{center}
    \fbox{\includegraphics[width=0.99\textwidth]{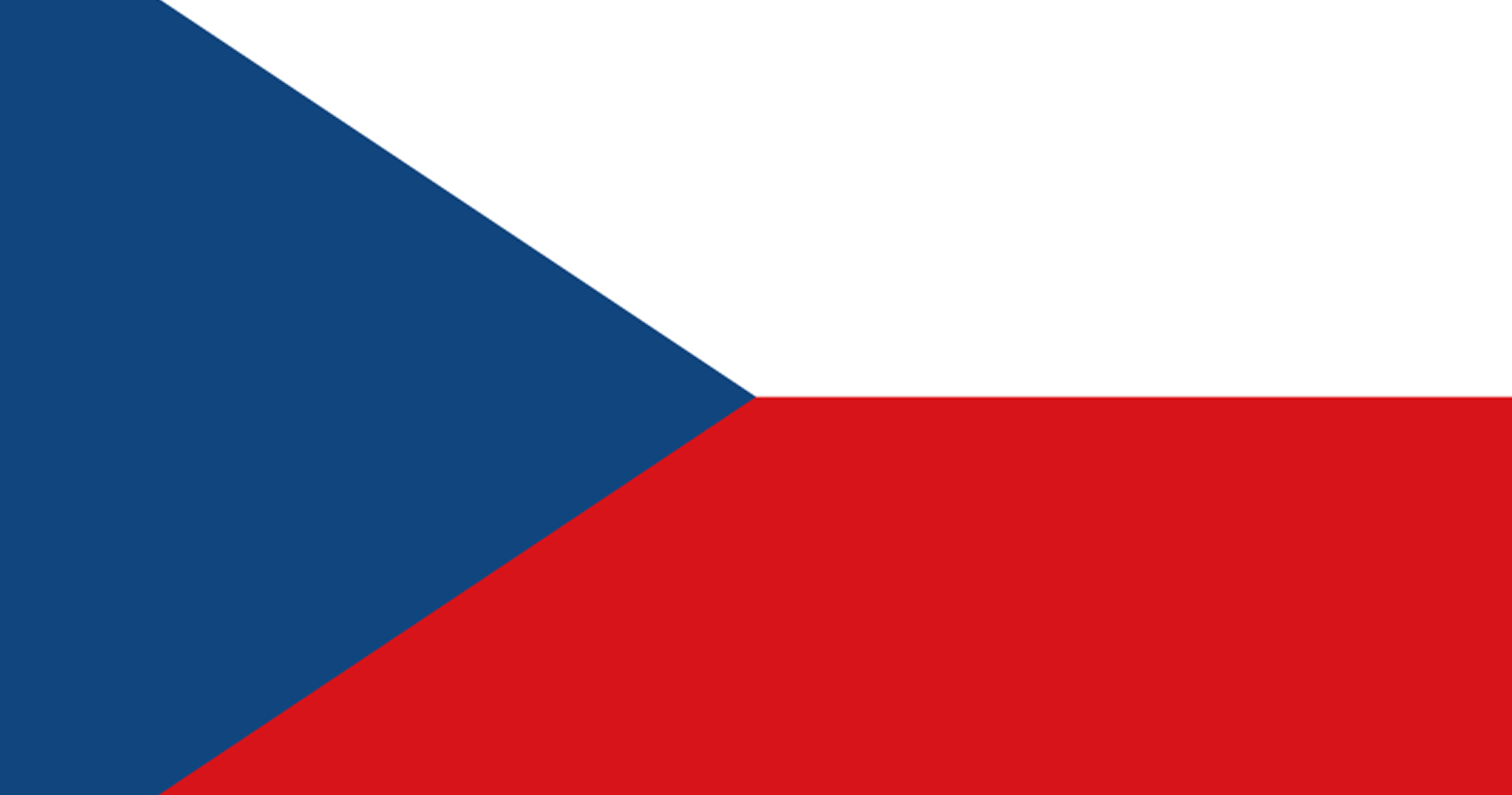}}
\end{center}
\begin{center}
    \textbf{Data Collected: 2013-2024}
\end{center}
\vfill
\begin{center}
  \fbox{%
    \parbox{\textwidth}{%
      \begin{center}
      \textbf{Important Links}\\
      \href{https://www.cnb.cz/}{Central Bank Website}\\
       \href{https://huggingface.co/datasets/gtfintechlab/czech_national_bank}{Annotated Dataset}\\
      \end{center}
    }
  }
\end{center}

\newpage

\section*{Monetary Policy Mandate}
The Czech National Bank (CNB) is responsible for maintaining monetary and financial stability in the Czech Republic while promoting sound economic development.

\textbf{Mandate Objectives:}
\begin{itemize}
    \item \textbf{Inflation Control:} Managing inflation within the CNB's target range, which is currently announced as 2\%.
    \item \textbf{Currency management:} Issuing legal tender and managing the Czech money supply.
    \item \textbf{Price and Financial Stability:} Ensuring currency stability and the stability of the broader Czech financial system through effective monetary policy decisions.
\end{itemize}

\section*{Structure}
\textbf{Composition:}
\begin{itemize}
    \item \textbf{Governor:} Represents the Czech National Bank externally and chairs central bank meetings.
    \item \textbf{Deputy Governors:} Two deputy governors that primarily oversee departments (Monetary department; Financial Markets and Resolution Department), but can also serve as the acting governor if necessary.
    \item \textbf{Other members:} Four other board members that manage other departments within the CNB.
\end{itemize}

\textbf{Meeting Structure:}
\begin{itemize}
    \item \textbf{Frequency:} Regular meetings are held eight times annually to assess economic and monetary conditions.
    \item \textbf{Additional Meetings:} Held during extraordinary economic developments.
\end{itemize}

\section*{Manual Annotation}

\textbf{Annotators} 

\begin{itemize}
    \item Marc-Alain Adjahi
\end{itemize}

\textbf{Annotation Guide}
\mptext{cnb}{six} Economic Status, Currency Stability, Inflation Rates, Interest Rates, External Shocks, and Wage Dynamics.
\begin{itemize}
    \item \emph{Economic Status}: A sentence pertaining to the overall performance of the economy, particularly focusing on fluctuations in inflation, unemployment, and economic growth.
    \item \emph{Currency Stability}: A sentence addressing changes in the exchange rate of the Czech koruna, reflecting its appreciation or depreciation and its impact on import costs and monetary conditions.
    \item \emph{Inflation Rates}: A sentence focusing on trends in inflation, especially in relation to the CNB's target range, and indicating potential monetary policy actions in response to deviations.
    \item \emph{Interest Rates}: A sentence relating to adjustments in the CNB's interest rate policy, highlighting decisions to raise, lower, or maintain borrowing costs as a means to influence economic activity.
    \item \emph{External Shocks}: A sentence examining the impact of external economic factors, such as shifts in international demand or economic conditions abroad, that could influence domestic stability and policy decisions.
    \item \emph{Wage Dynamics}: A sentence describing changes in wage growth relative to productivity, assessing their effects on domestic demand, inflation pressures, and the overall balance of the economy.
\end{itemize}

\textbf{Examples: }
\begin{itemize}
    \item ``Inflation would be curbed by the strengthening of the Czech koruna vis-à-vis the euro, pulling import prices down and tightening monetary conditions in the economy.''\\
    \textbf{Dovish}: This example illustrates a scenario where easing conditions support a dovish stance.
    
    \item ``At the close of the meeting, the Board decided unanimously to lower the CNB two-week repo rate by 0.50 percentage points to 4.75\%.''\\
    \textbf{Dovish}: The decision to lower rates is indicative of a dovish monetary policy approach.
    
    \item ``The risks presented by the forecast had accumulated on the inflation side, with a rise in money supply and fiscal deficit potentially sending a stronger demand impulse than anticipated.''\\
    \textbf{Hawkish}: This sentence demonstrates concerns over overheating, suggesting a hawkish response.
    
    \item ``At the close of the meeting, the Board decided unanimously to raise the CNB two-week repo rate by 0.25 percentage points to 5.25\%.''\\
    \textbf{Hawkish}: The rate increase reflects a move towards tighter policy, characteristic of a hawkish stance.
    
    \item ``The actual data did not confirm a substantial weakening, and money supply growth continued to reach a figure of about 12\% year-on-year, supporting a stable rise in household consumption.''\\
    \textbf{Neutral}: The stable economic indicators give rise to a neutral policy outlook.

    \item ``The next bank board meeting on monetary and economic development will take place on 28 January 1999.''\\
    \textbf{Irrelevant}: This sentence solely provides scheduling information and does not offer any insights into monetary policy.
\end{itemize}

\newpage

\begin{table}[h!]
    \centering
    \small
    \caption{\mptitle{Czech National Bank}}
    \begin{tabular}    {p{0.118\textwidth}p{0.183\textwidth}p{0.183\textwidth}p{0.183\textwidth}p{0.183\textwidth}}
    
\toprule
\textbf{Category} & \textbf{Hawkish} & \textbf{Dovish} & \textbf{Neutral} & \textbf{Irrelevant} \\
\midrule

    \textbf{Economic Status} & When inflation increases, unemployment decreases, or economic growth is projected high, indicating a potential overheating of the economy. & When inflation decreases, unemployment increases, or economic growth is projected as low. & When economic indicators remain stable. & Sentence is not relevant to monetary policy. \\
    \midrule
    \textbf{Currency Stability} & When the Czech koruna appreciates, reducing import costs, signaling a stronger economy, and potential tightening of policy. & When the Czech koruna depreciates, increasing import costs, potentially signaling a need for a looser monetary policy to stabilize the economy. & When the koruna exchange rate remains relatively stable over a period, without sharp fluctuations. & Sentence is not relevant to monetary policy. \\
    \midrule
    \textbf{Inflation Rates} & When inflation rises above the CNB's target range, requiring tighter policy to anchor expectations. & When inflation trends downward, providing room for looser monetary policy to support growth. & When inflation rates remain stable within a target range. & Sentence is not relevant to monetary policy. \\
    \midrule
    \textbf{Interest Rates} & When interest rates are raised, signaling the CNB's intent to curb inflation by making borrowing more expensive. & When interest rates are lowered, indicating the CNB is encouraging borrowing and investment by reducing the cost of credit. & When interest rates remain unchanged from previous levels. & Sentence is not relevant to monetary policy. \\
    \midrule
    \textbf{External Shocks} & When external demand surges, driven by recovery in key markets, potentially requiring tighter policy to avoid overheating. & When external shocks, such as a downturn in Germany or an EU recession, weaken Czech exports, calling for monetary easing to stabilize growth. & When external economic conditions remain stable, causing no immediate impact. & Sentence is not relevant to monetary policy. \\
    \midrule
    \textbf{Wage Dynamics} & When wage growth accelerates beyond productivity levels, risking inflation and higher costs for businesses. & When wage growth slows, signaling weakening domestic demand and easing inflationary pressures. & When wage growth aligns with productivity, posing no immediate risk to inflation or demand. & Sentence is not relevant to monetary policy. \\
    \bottomrule
    \label{tb:cnb_mp_stance_guide}
    \end{tabular}
\end{table}

\fwcertaintytext{cnb}
\newpage

\begin{table*}
\caption{\fwtitle{Czech National Bank}}
\vspace{1em}
\begin{tabular}{p{0.3\textwidth}p{0.3\textwidth}p{0.3\textwidth}}
\toprule
\textbf{Label} & \textbf{Description} & \textbf{Example}\\
\midrule
\textbf{Forward Looking} & Projections of economic data and the geopolitical landscape. Comments about what is expected to come in future meetings or in future events. & “The forecast expected headline inflation to increase this year from its current very low but positive level.” \\
\midrule
\textbf{Not Forward Looking} & Reflections of previous economic data and the geopolitical landscape. Comments about events that have already happened in the past. & “Average wage growth had risen to 2.2 in 2015 q1, slightly above the current forecast.” \\
\bottomrule
\end{tabular}
\label{tb:cnb_forward_looking_guide}
\end{table*}

\begin{table*}
\caption{\certaintytitle{Czech National Bank}}
\vspace{1em}
\begin{tabular}{p{0.3\textwidth}p{0.3\textwidth}p{0.3\textwidth}}
\toprule
\textbf{Label} & \textbf{Description} & \textbf{Example}\\
\midrule
\textbf{Certain} & References data points or events that have already happened in the past. If the sentence references the future, it uses that reflect certainty, such as ``will.'' & “The labour market was meanwhile still assessed as strongly overheated.” \\
\midrule
\textbf{Uncertain} & Projections of the future that do not reflect certainty. Key words include ``may'' and ``should.'' & “In light of recent events, the still unclear form of Brexit could be added to this uncertainty.” \\
\bottomrule
\end{tabular}
\label{tb:cnb_certainty_guide}
\end{table*}

\end{document}